\documentclass{article} 
\usepackage[preprint]{neurips_2025}

\usepackage{xcolor}

\usepackage{amsmath,amsfonts,bm}

\def\eqref#1{equation~\ref{#1}}

\def\1{\bm{1}}

\DeclareMathAlphabet{\mathsfit}{\encodingdefault}{\sfdefault}{m}{sl}
\SetMathAlphabet{\mathsfit}{bold}{\encodingdefault}{\sfdefault}{bx}{n}

\newcommand{\E}{\mathbb{E}}

\definecolor{mydarkblue}{rgb}{0,0.08,0.45}
\usepackage[colorlinks=true,
    linkcolor=mydarkblue,
    citecolor=mydarkblue,
    filecolor=mydarkblue,
    urlcolor=mydarkblue]{hyperref}  
\usepackage{url}
\usepackage{subcaption}
\usepackage{graphicx}
\usepackage{wrapfig}
\usepackage{booktabs}
\usepackage{multirow}

\usepackage{amsfonts}
\usepackage{amsmath}
\usepackage{amsthm}
\usepackage{bbm}

\usepackage{tablefootnote}

\usepackage{alphalph}

\title{Bias Analysis in Unconditional Image \\
Generative Models} 

\author{Xiaofeng Zhang$^{1}$, Michelle Lin$^{1}$, Simon Lacoste-Julien$^{1,2}$, Aaron Courville$^{1}$ and Yash Goyal$^{2}$ \\
$^{1}$Mila, Université de Montréal $^{2}$Samsung AI Lab, Montréal \\
}

\begin{document}

\maketitle

\begin{abstract}
    The widespread adoption of generative AI models has raised growing concerns about representational harm and potential discriminatory outcomes. Yet, despite growing literature on this topic, the mechanisms by which bias emerges—especially in unconditional generation—remain disentangled.
    We define the \textit{bias of an attribute} as the difference between the probability of its presence in the observed distribution and its expected proportion in an ideal reference distribution. 
    In our analysis, we train a set of unconditional image generative models and adopt a commonly used bias evaluation framework to study bias shift between training and generated distributions. 
    Our experiments reveal that the detected attribute shifts are small. We find that the attribute shifts are sensitive to the attribute classifier used to label generated images in the evaluation framework, particularly when its decision boundaries fall in high-density regions. 
    Our empirical analysis indicates that this classifier sensitivity is often observed in attributes values that lie on a spectrum, as opposed to exhibiting a binary nature.
    This highlights the need for more representative labeling practices, understanding the shortcomings through greater scrutiny of evaluation frameworks, and recognizing the socially complex nature of attributes when evaluating bias.
\end{abstract}

\section{Introduction}

Generative AI models have achieved realistic generation qualities for various modalities including text~\citep{llama,gpt4}, image~\citep{dalle2,stablediffusion22,stablediffusion3}, audio~\citep{audiogen}, and video~\citep{videodiffusion,makevideo}. They are consequently employed for commercial uses and are available to every internet user across the world. The widespread use of these high-performing models, along with the potential social biases embedded in their generation, increase the risk of representational harm. Examples within image generation include  \cite{PetaPixel2023} and \cite{WPAIbias2023} that report racial and gender biases in popular text-to-image (T2I) systems including DALL-E~\citep{dalle}, Stable Diffusion~\citep{stablediffusion22} and Midjourney (\url{https://www.midjourney.com}).


There is growing literature in studying these models and investigating how these models manifest biases~\citep{dalleval,stereotype,stablebias,FairDiffusion,DBLP:journals/corr/abs-2308-00755}. Some of the works focus on the presence of bias in generated samples~\citep{dalleval,stereotype,stablebias}, while a few others compare how bias is differently presented (\emph{bias shift}) in dataset and generation~\citep{FairDiffusion,DBLP:journals/corr/abs-2308-00755,diffusionfaceanalysis}. When it comes to the comparison between the dataset and the generation, evidence from studies has been inconclusive. ~\cite{FairDiffusion} report that images generated by Stable Diffusion~\citep{stablediffusion22} show cases of bias and even bias amplification compared to the training data (LAION-5B)~\citep{laion}.
On the other hand, ~\cite{DBLP:journals/corr/abs-2308-00755}  discover that \emph{bias shift} can be mainly attributed to discrepancies between training captions and model prompts. 

As such, our work investigates the question: \textit{What is the behaviour of the bias shift between the training and generation distributions?} We focus on the unconditional image generative models, specifically, diffusion process~\cite{ddpm,scorematching19} and generative adversarial network~\cite{gan,biggan}. This focus is motivated by the need to exclude potential confounding factors in the generative process such as conditional classifier-free guidance (CFG)~\cite{cfgguidance}. Since CFG reduces diversity in the generated outputs~\cite{cfgguidance,bvg}, we hypothesize that this mode centric behavior introduced by the CFG will affect bias. 
As such, this work represents a step toward isolating and better understanding the contributions of the unconditional generator itself to observed bias shifts. Thus, in our experiments, we focus on \emph{unconditional pixel-level} image generative models \emph{without any guidance during training or inference}.

We use image datasets that has multiple attributes to conduct empirical study. In our context, we formalize the bias shift problem as the attribute frequency shift from training to generation. Although bias in discriminative models has been extensively studied~\cite{Wang2022ABR}, it is a common approach to use trained classifiers to label generated samples while evaluating generative models~\cite{dalleval,stablebias,finetuneforfairness,ullmfairness}.

We train unconditional generative models, including diffusion models and GANs on CelebA~\cite{celebadataset} and DeepFashion~\cite{deepfashiondataset}. Then, we use the aforementioned framework to evaluate the attribute bias shift. To ensure a consistent and comparable estimation of attribute labels, we use the trained classification model to label both the evaluation set and generation set, and compute the attribute bias shift based on the model predicted labels.

\begin{figure}[!tbp]
    \centering
    \begin{subfigure}[h]{0.22\textwidth}
        \includegraphics[width=\linewidth]{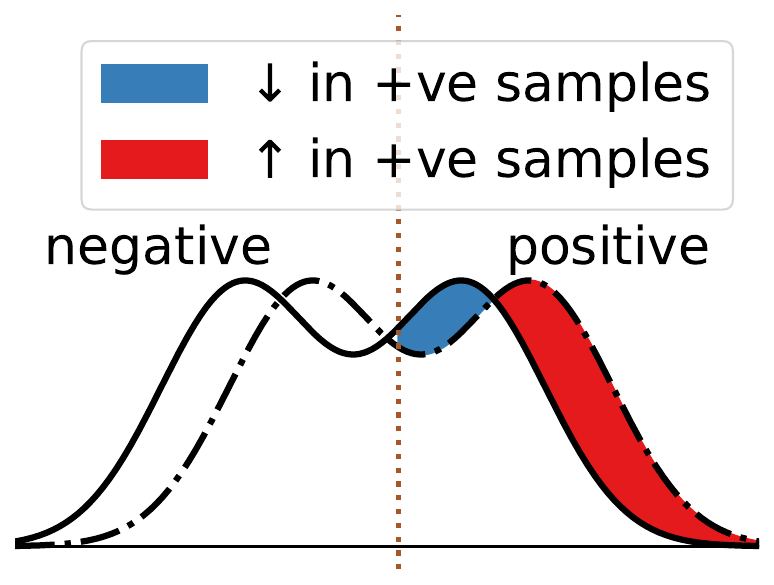}
        \caption{\textit{spectrum-based}}
        \label{fig:bihigh}
    \end{subfigure}
    \begin{subfigure}[h]{0.22\textwidth}
        \includegraphics[width=\linewidth]{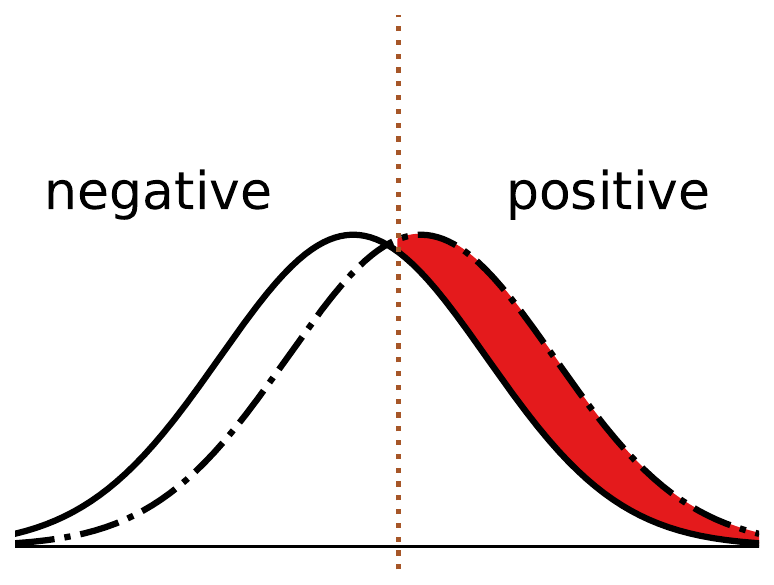}
         \caption{\textit{spectrum-based}}
         \label{fig:unihigh}
    \end{subfigure}
    \begin{subfigure}[h]{0.22\textwidth}
        \includegraphics[width=\linewidth]{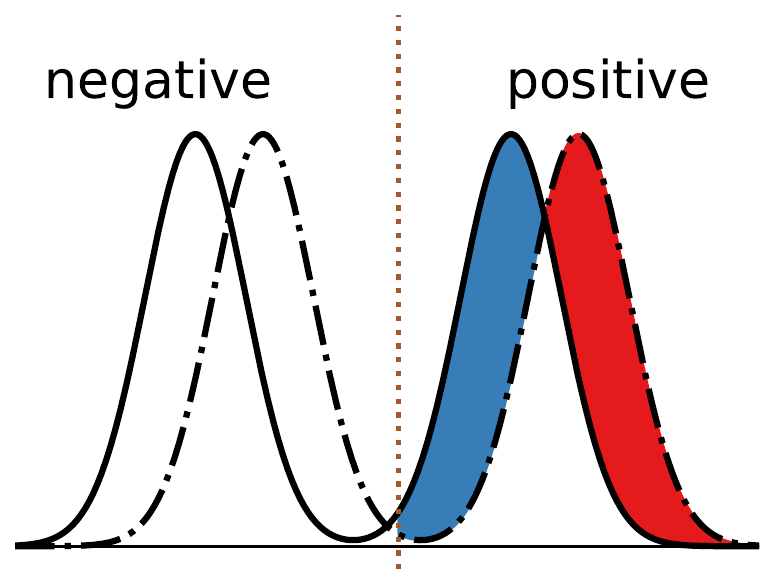}
         \caption{\textit{non-spectrum-based}}
         \label{fig:bilow}
    \end{subfigure}
    \begin{subfigure}[h]{0.25\textwidth}
        \includegraphics[width=\linewidth]{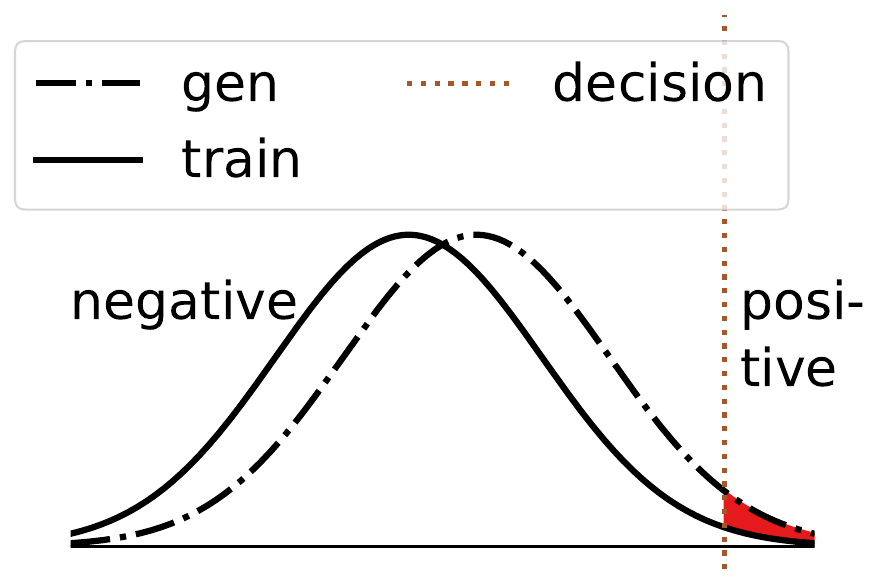}
         \caption{\textit{non-spectrum-based}}
         \label{fig:unilow}
    \end{subfigure}
    \caption{\textbf{Illustrations depicting bias shift.} The plots represent the distributions of samples with respect to the likelihood of an attribute (solid for training data, dashed for generation). 
    The decision boundary (brown) binarizes the likelihood into positive and negative classes.
    In each subfigure, the generation distribution is translated from the training. Bias shift is the difference between red and blue areas. When the boundary falls in a low-density region (Figs.~\ref{fig:bilow} and \ref{fig:unilow}), the bias shifts  tend to be small, and vice versa (Figs.~\ref{fig:bihigh} and \ref{fig:unihigh}). Mathematical proof is in Appendix~\ref{app:translationproof}. Detailed discussion is in Section~\ref{sec:cls_revisiting} with distributions obtained from real datasets. }
    \label{fig:cartoon}
    \vspace{-1em}
\end{figure}

Our analysis yield the following findings: 1) Attribute bias shift occurs between training and generation distributions for unconditional image generative models. The magnitude of attribute bias shift is correlated with the \textit{spectrum-based} nature of the attribute. See Figure~\ref{fig:cartoon} for an intuitive understanding. 2) BigGAN models~\citep{biggan} have larger attribute bias shifts compared to diffusion models, despite having similar image generation metrics (e.g., FID, etc.). 3)  Classifier decision boundary density is a strong predictor of shift sensitivity.




\section{Related Works}
\textbf{Bias in Image Generation. \ \ }
Previous studies focus on social biases in image generation, often concluding that these models ~\citep{FairDiffusion,dalleval} fail to reflect biases as observed in U.S. labor statistics~\citep{stablebias,stereotype}. Different studies select various public models and develop their own evaluation benchmarks. For example, \cite{dalleval} investigate minDALL-E~\citep{mindalle}, Karlo~\citep{karlo}, and Stable Diffusion~\citep{stablediffusion22} v1.4, while \cite{stablebias} use Stable Diffusion v1.4, v2, and Dall-E 2~\citep{dalle2}. 
However, there are different components in T2I models that may contribute to generative biases. We focus on the unconditional image generative model that directly outputs the generations, without text conditioning or guidance.

\textbf{Bias Shift between Train and Generation. \ \ }
Few studies attempt to compare bias between the generation and training distributions. These efforts often rely on publicly available Stable Diffusion models, comparing generated outputs with the LAION-5B training set~\citep{laion}, a large-scale dataset lacking explicit attribute labels. Given a text prompt, ~\citep{FairDiffusion} select a subset of LAION-5B based on pre-trained image-prompt similarity, then compare the bias between this subset and the images generated using the same prompt. In contrast, \cite{DBLP:journals/corr/abs-2308-00755} select subsets based on keywords in image captions, which may overlook relevant images. 
To avoid this large-scale dataset search and subset comparison, we train generative models using datasets with labeled attributes, ensuring reliable bias estimation across both the training and generation.

\textbf{Bias-related Attribute Label Prediction. \ \ }
To calculate bias in generation, the generated images need to be assigned attribute labels, which is non-trivial in the case of unconditional generation. 
Some studies~\citep{stereotype} infer the labels in the representation space of self-supervised learning models, for example, CLIP~\citep{clip21}. 
Some methods use pre-trained vision language models and conduct zero-shot text generation. \citep{dalleval} use BLIP-2~\cite{blip2} and get the label through visual question answering (VQA). \cite{stablebias} use BLIP
with VQA task and ViT~\citep{vit}
with image captioning task. However, pre-trained models introduce their own biases~\citep{foundationModel,clipbias}, rendering the predicted labels unreliable for accurate bias evaluation. Some approaches~\citep{FairDiffusion} train an attribute classifier on other available supervised learning datasets. In our case, we train the classifier on the same dataset used for analysis, resulting in more accurate predictions.



\section{Bias Evaluation Method}

\subsection{Bias Definition}

We aim to study the bias shift from training to generation for unconditional generative models.
Bias is a complex concept, including algorithmic bias, dataset bias, and disparate model performance across different categories, etc.~\cite{Wang2022ABR,Verma2018FairnessDE,Pagano2023BiasAU}.
In this work, we define our scope and the definition of bias very specifically: following previous works that study bias in T2I systems~\cite{stablebias}, bias is encoded in the T2I systems if the attribute proportion in the generation is different from that in the real world statistics.
More precisely, we define bias for an attribute as the difference between the probability of its presence in the observed distribution and its expected proportion in an ideal reference distribution. 
For a discussion of the scope, broader impacts, and limitations of these definitions, see Section~\ref{sec:conclusion}.


Considering a set of binary attributes\footnote{The use of binary attributes can be extended to $K$-way attributes by binarizing the $K$-way attributes as $K$ $1$-vs-all binary attributes.} $\mathcal{C}$ for which we want to study bias, each image in the dataset is annotated for every attribute. Given an attribute $C \in \mathcal{C}$, 
we denote the ideal probability for attribute $C$ in the reference distribution as $P^{\text{ideal}}(C)$ , depending on the context.
We denote the probability of this attribute in the data distribution as $P^{\text{data}}(C)$. We can use either $P^{\text{train}}(C)$ or $P^{\text{val}}(C)$ as an estimation for $P^{\text{data}}(C)$ and compare with the reference probability to determine the degree of bias. For example, we define the bias of the data distribution relative to $P^{\text{ideal}}(C)$ as 
\begin{equation}
    B^{\text{data}} (C) = P^{\text{data}}(C) - P^{\text{ideal}}(C) .
\end{equation}
To get the bias on the generation set, we need to calculate the proportion for this attribute in the generation set $P^{\text{gen}}(C)$. We can then measure the bias in the generation 
\begin{equation}
    B^{\text{gen}} (C) = P^{\text{gen}}(C) - P^{\text{ideal}}(C) .
\end{equation}




\begin{figure}[!tbp]
    \centering
    \includegraphics[width=0.85\linewidth]{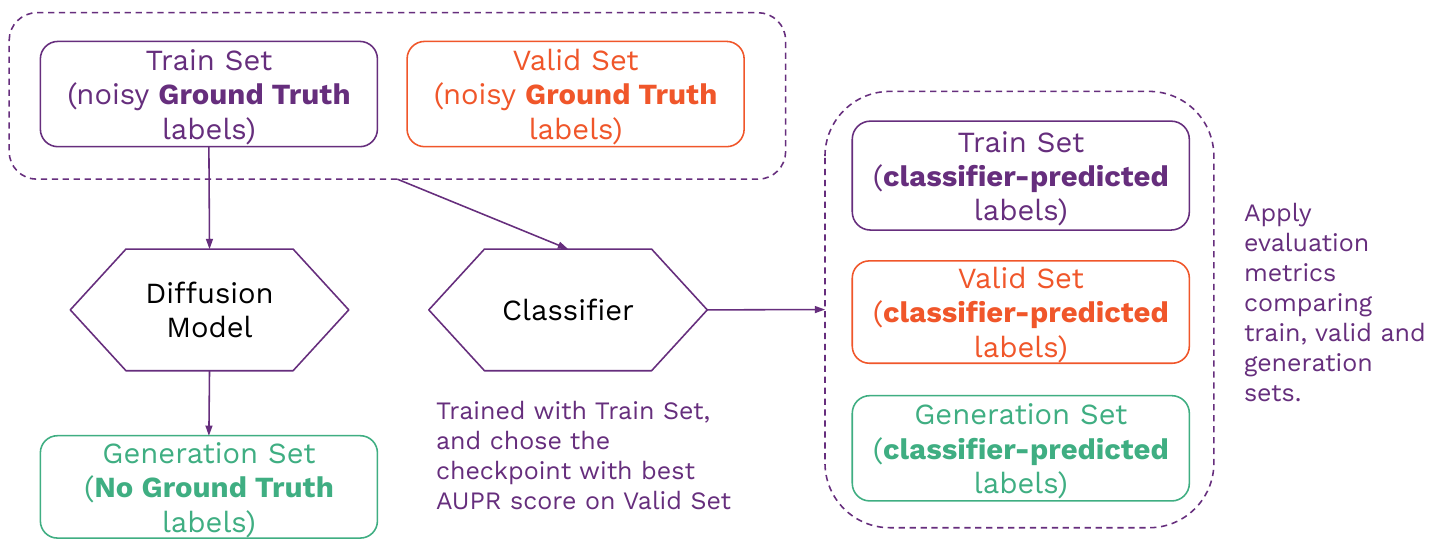}
    \caption{\textbf{Bias shift evaluation framework.} Unconditional generative models are trained on the training set. The pre-trained classifier is fine-tuned on the training set and validated on the validation set using ground truth labels and is then used to classify training, validation, and generation sets. The bias shift evaluation metrics are calculated based on the \textbf{classifier-predicted labels}.}
    \label{fig:pipeline}
    \vspace{-1em}
\end{figure}

\subsection{Bias Shift Evaluation Framework}
\label{sec:framework}
Fig.~\ref{fig:pipeline} illustrates the bias shift evaluation framework that is adopted in our study.
We train image generative models for unconditional image generation only using images from the training set, without feeding ground truth labels into the models. We generate 10,000 images for each checkpoint during training. To calculate the proportion for each attribute in the generation distribution, we require attribute labels for the generated images. Previous works rely on either large pre-trained models~\cite{stablebias,dalleval} or specific classifiers~\cite{FairDiffusion,finetuneforfairness} to get attribute labels for generated images. We apply a trained classifier, developed using the training and validation sets with ground truth labels, to the generated images to obtain classifier-predicted attribute labels.

Although these works acknowledge the biases that attribute prediction models may exhibit, there are no good alternatives for automatic attribute labeling. Similarly, the trained classifier that we use in our framework inevitably introduces errors. However, in our research scope which is to study the bias shift from training to generation, we can use the same classifier to classify both the dataset and the generation set to keep the classification protocol the same.
In addition, we use $P^{val}(C)$ to estimate the probability of attribute $C$ in the data distribution, as the classifier may overfit to the training set.
By adopting these techniques, we aim to minimize the potential error introduced by the classifier in our attribute bias (frequency) shift evaluation framework for generative models.

Given a binary attribute $C \in \mathcal{C}$, we can therefore define 
\textbf{attribute bias shift} between generation and training data as 
\begin{equation}
    B_\text{shift} (C) = |B^\text{gen} (C) - B^\text{data} (C)| = |P^\text{gen}_{\texttt{cls}}(C) - P^\text{val}_{\texttt{cls}}(C)| .
\end{equation}
The subscript \texttt{cls} stands for using classifier-predicted labels. In attribute bias shift, the expected probability for the attribute $C$ in the ideal reference distribution $P^\text{ideal}(C)$ is canceled out. Attribute bias shift remains the same regardless which ideal bias reference we select. Thus, attribute bias shift in our empirical study is evaluated as the attribute frequency shift from training to generation. If attribute frequency shift is close to 0, then the generation distribution and the training distribution exhibit the same level of attribute bias for the given attribute. 

\textbf{Attribute bias shift} evaluates changes in bias between data and generation distribution for each attribute considered in the study. To provide an overall understanding of the magnitude of bias shift across all attributes, we propose to use the average of attribute bias shift across all the attributes considered in the dataset. 
\textbf{Average attribute bias shift (ABS)}\footnote{We emphasize the bias shift is for certain attribute in the main text to avoid confusion with social bias w.r.t. protected attributes (e.g., gender, race, etc.). We omit ``attribute'' in the abbreviation.} evaluates the overall bias shift magnitude across all attributes considered between the training and the generated data distributions. This value represents the absolute difference between probabilities and is expressed as a percentage. We define this metric as 
\begin{equation}
    \text{ABS} = \E_{C \in \mathcal{C}} B_\text{shift}(C) .
\end{equation}

Averaging the bias across all attributes is not appropriate for all cases, potentially leading to masking of biases and loss of nuance, etc. However, in our case, the bias shift for each attribute is always positive, and the attributes considered in our study are generally non-protected, such as color, shape, and texture. Therefore, we use the average of bias shifts across all attributes to provide a quantitative analysis of the overall bias shift phenomenon.

\section{Experiments}
\label{sec:exp}

\begin{figure}[t]
    \centering
    \begin{subfigure}[ht]{0.29\textwidth}
    \centering
        \includegraphics[width=0.98\textwidth]{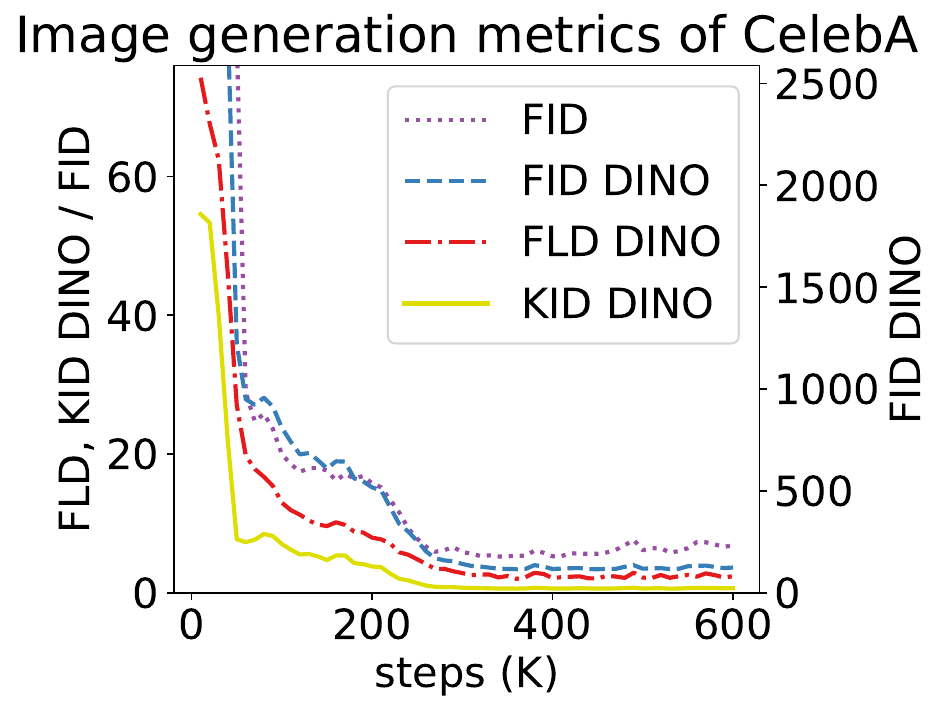}
        \caption{\tiny FID, KID, and FLD of CelebA}
        \label{fig:fldfid_celeba}
    \end{subfigure}
    \begin{subfigure}[ht]{0.34\textwidth}
    \centering
        \includegraphics[width=0.98\textwidth]{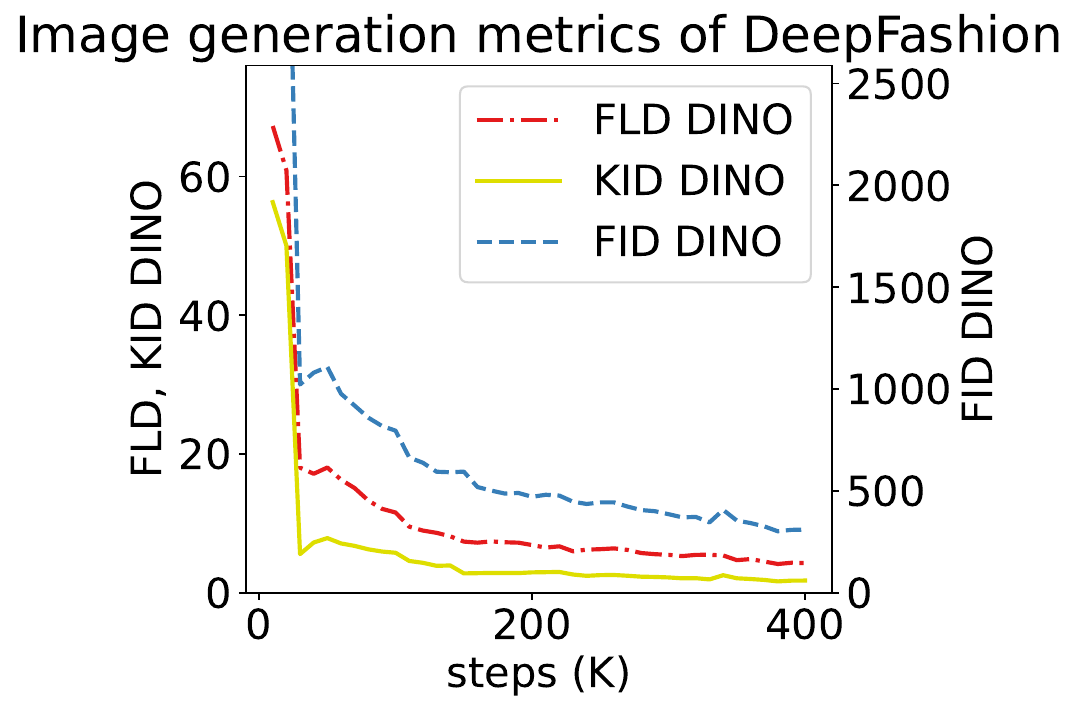}
        \caption{\tiny FID, KID, and FLD of DeepFashion}
        \label{fig:fldfid_deepfashion}
    \end{subfigure}
    \begin{subfigure}[ht]{0.30\textwidth}
    \centering
        \includegraphics[width=0.98\textwidth]{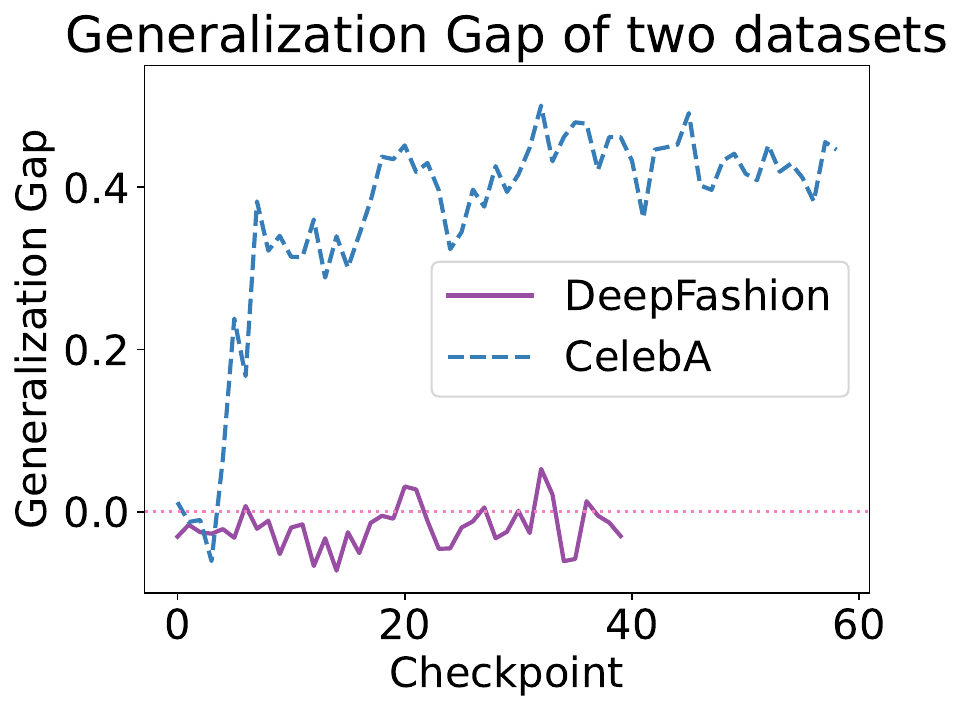}
        \caption{ \tiny 
        Generalization gap}
        \label{fig:gengap_celeba}
    \end{subfigure}
    \caption{\textbf{Evaluation metrics for image generation throughout training.} In \ref{fig:fldfid_celeba} and \ref{fig:fldfid_deepfashion}, FID, KID, and FLD values converge to small values showing the good quality of generated images and good coverage of modes of the training distribution. In \ref{fig:gengap_celeba}, the positive or slightly negative generalization gaps indicate that the trained models do not have severe memorization issues.}
    \label{fig:overallmetrics}
\end{figure}

\subsection{Experimental setup}
\textbf{Datasets. \ \ }
We apply our proposed bias evaluation framework to two real image datasets -- CelebA~\citep{celebadataset} and DeepFashion~\citep{deepfashiondataset}.
CelebA~\citep{celebadataset} is a large-scale dataset with more than 200,000 celebrity facial images, each labeled with 40 binary attributes. It covers a wide range of facial features, from details (e.g., earrings, pointy nose) to outlines (e.g., hair color, gender, age).
DeepFashion~\citep{deepfashiondataset} is a clothes dataset with over 800,000 diverse fashion images.
We use a subset with 26 fine-grained attribute annotations to train the classifier. We then study the attribute bias shift over these fine-grained attributes. For both datasets, we follow the training/validation/test set split from the official release. More details about these datasets are in Appendix~\ref{app:datasets}.

\textbf{Backbone models in the framework. \ \ }
We follow the setup from~\cite{adm21} to train unconditional ablated diffusion models (ADMs). We train models of varying sizes by adjusting the number of channels in the U-Net~\citep{unet} bottleneck layer (32 for tiny, 64 for small, and 256 for large), with proportional changes in each layer. In the following sections, we report the results of the large diffusion model if the model size is not otherwise specified. We generate 10,000 images per checkpoint using 100 inference steps across training.
We use a ResNeXt-50-based~\citep{resnext17} image classifier and a Swin Transformer-based~\cite{swint} image classifier. The results presented in the main paper are obtained using the ResNeXt-50-based classifier, and the corresponding results using the Swin Transformer-based classifier are provided in the appendix.
For comparison with a GAN model, we train a BigGAN~\citep{biggan} model using the recommended settings.
Implementation details are in Appendix~\ref{app:hyperparams}.

\begin{figure}[!tbp]
\centering
    \begin{subfigure}{0.28\textwidth}
        \centering
        \includegraphics[width=\textwidth]{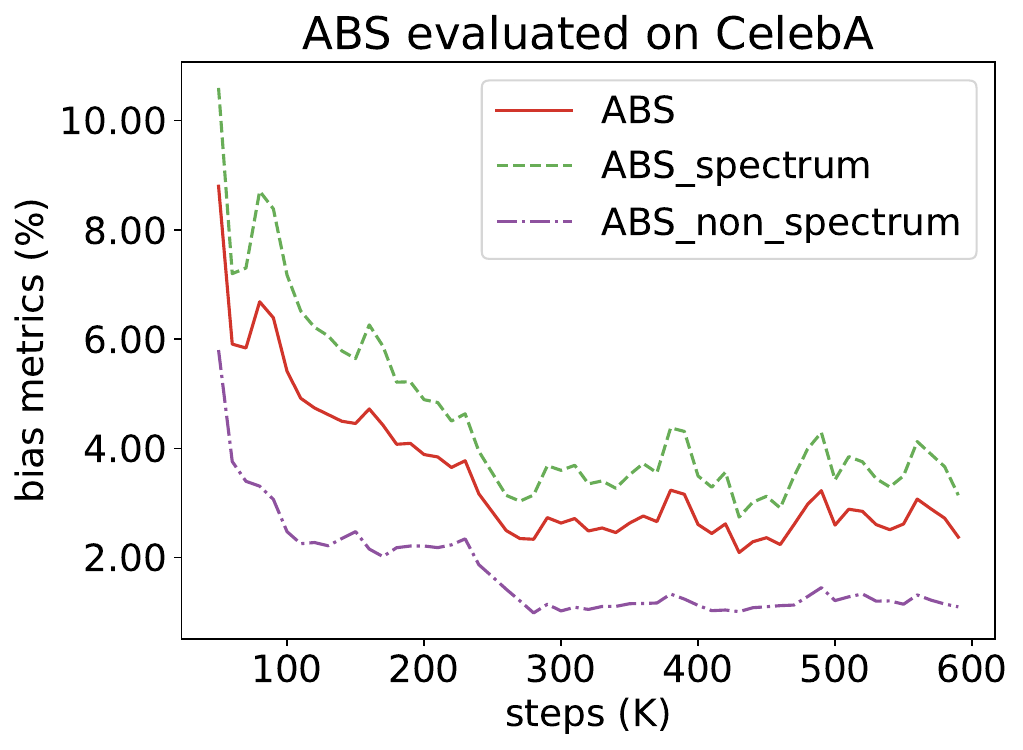}
        \caption{ABS for CelebA}
        \label{fig:bametric_marg_celeba}
    \end{subfigure}
    \begin{subfigure}{0.29\textwidth}
        \centering
        \includegraphics[width=\textwidth]{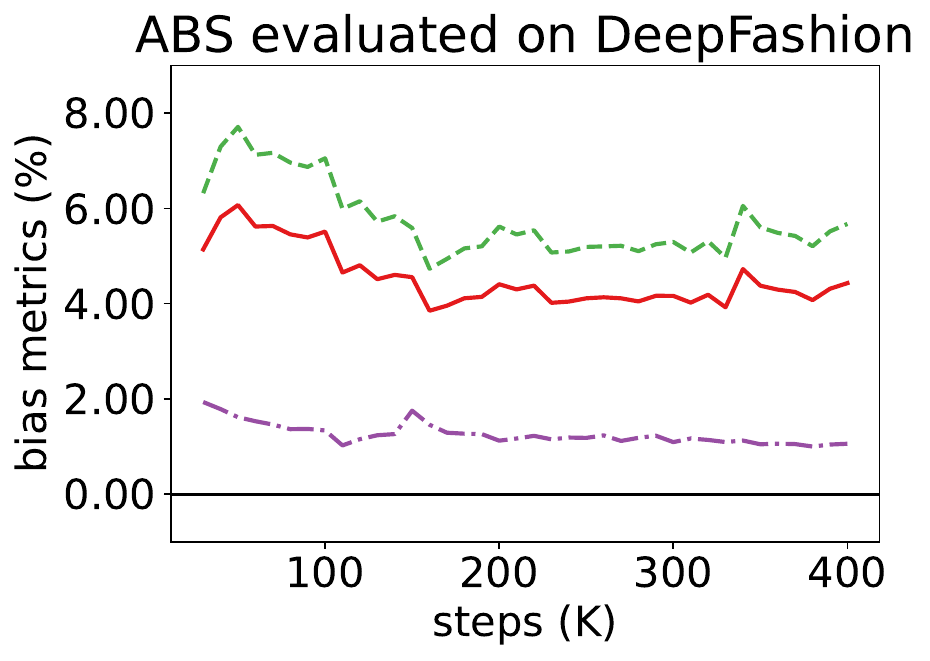}
        \caption{ABS for DeepFashion}
        \label{fig:marg_bametric_dp}
    \end{subfigure}
    \begin{subfigure}{0.28\textwidth}
        \centering
        \includegraphics[width=\linewidth]{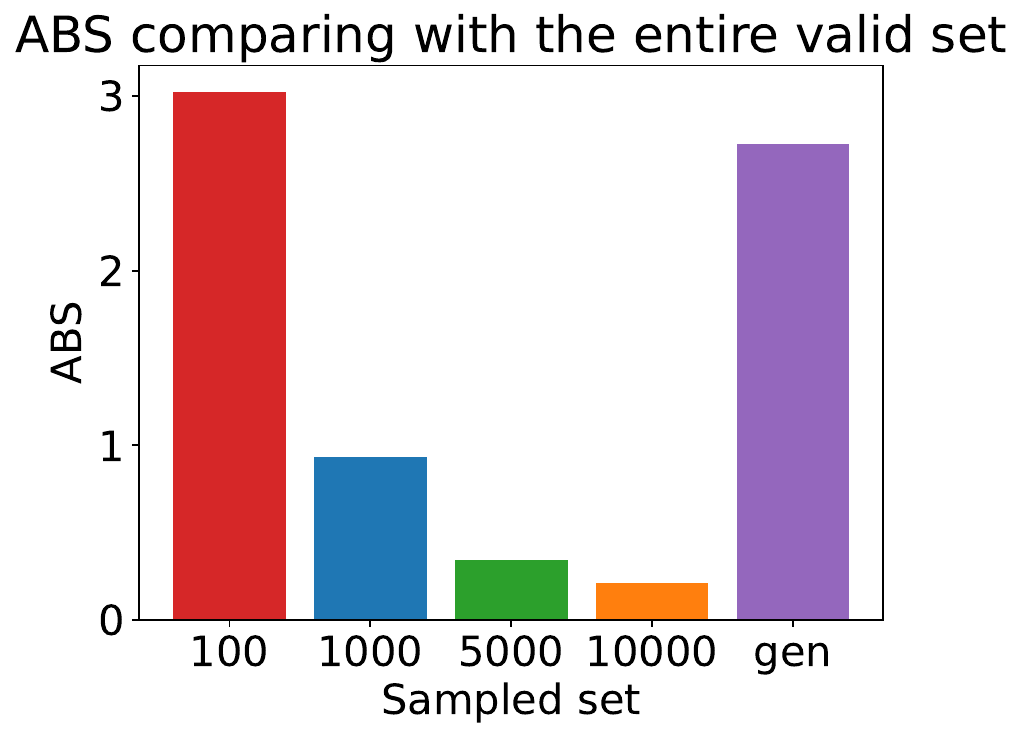}
        \caption{Sampling error}
        \label{fig:sample_error}
    \end{subfigure}
    \caption{\textbf{Average attribute bias shift (ABS) for CelebA and DeepFashion.} For both datasets, shown in Figs.~\ref{fig:bametric_marg_celeba} and \ref{fig:marg_bametric_dp}, ABS over \textit{spectrum-based} attributes show a much larger attribute bias shift than \textit{non-spectrum-based} ones. Fig.~\ref{fig:sample_error} presents that the error coming from sampling of 10K images is small enough, showing that the sampling randomness is not the only cause of attribute bias shifts in generations.}
    \label{fig:bametrics}
\end{figure}

\textbf{Evaluation metrics for image generation. \ \ }
We use common metrics, e.g., FID (Fr\'echet Inception Distance)~\citep{fidmetric} and KID (Kernel Inception Distance)~\citep{kidmetric}, to evaluate the generated images. 
We use FLD (Feature Likelihood Divergence) and generalization gap~\citep{fldmetric} as two additional metrics to gauge the memorization level of the generative models. FLD provides a comprehensive evaluation considering not only quality and diversity, but also novelty (i.e., difference from the training samples) of generated samples. 
Positive generalization gap shows no overfitting to the training set.
We adopt the implementation\footnote{\url{https://github.com/marcojira/FLD}} \cite{fldmetric} and follow their suggestion of using DINOv2~\citep{DINO} as the feature extractor to calculate FID, KID, and FLD. We also use a conventional FID implementation\footnote{\url{https://github.com/mseitzer/pytorch-fid}} to give a comparable value of how well the trained models are.

\subsection{Backbone models performance}
\textbf{Diffusion models. \ \ }
Figure~\ref{fig:overallmetrics} shows the image generation evaluation metrics for CelebA and DeepFashion datasets. In Figs.~\ref{fig:fldfid_celeba} and \ref{fig:fldfid_deepfashion}, FID and KID converge to small values showing the good quality of generated images and good coverage of modes of the data distribution. FLD agrees with conventional metrics, showing no severe memorization issues in the generation. In Fig.~\ref{fig:gengap_celeba}, the positive or slight negative values of generalization gap indicate that no overfitting is detected in the trained models. More discussions are in Appendix~\ref{app:diffusion}.

\textbf{Classifier. \ \ } For CelebA and DeepFashion datasets, the classification accuracy on the validation set for most attributes is over $80\%$. Overall, the average accuracy across attributes is $91.7\%$ for CelebA and $90.5\%$ for DeepFashion. We discuss the performance of different classifiers in Appendix~\ref{app:cls}.


\vspace{-0.5em}
\subsection{Average attribute bias shift evaluation}
\label{sec:abseval}

Fig.~\ref{fig:bametrics} presents ABS throughout training. The overall ABS is still perceivable when image generation metrics are small, indicating non-negligible attribute bias shifts from the training to generation distributions. 

Attribute bias shifts do not consistently follow the image generation metrics, as illustrated by the comparison between Figs.~\ref{fig:overallmetrics} and~\ref{fig:bametrics}. This misalignment highlights that models with superior image generation metrics are not necessarily less biased. Bias should be treated as an independent issue, distinct from quality and diversity. While diversity metrics typically assess the coverage of modes in the generated distribution, bias evaluation should focus on the relative proportions of these modes.
For CelebA dataset, the bias evaluation metrics exhibit more fluctuations after 300K steps, while the image generation metrics plateau.
Similarly, for DeepFashion dataset, the image generation metrics continue improving during the whole training, while ABS for both \textit{spectrum-based} and \textit{non-spectrum-based} attributes are stable with minimal increases after approximately 200K steps.

\begin{figure}[!tbp]
      \centering
      \begin{subfigure}{0.23\textwidth}
		\includegraphics[width=\textwidth]{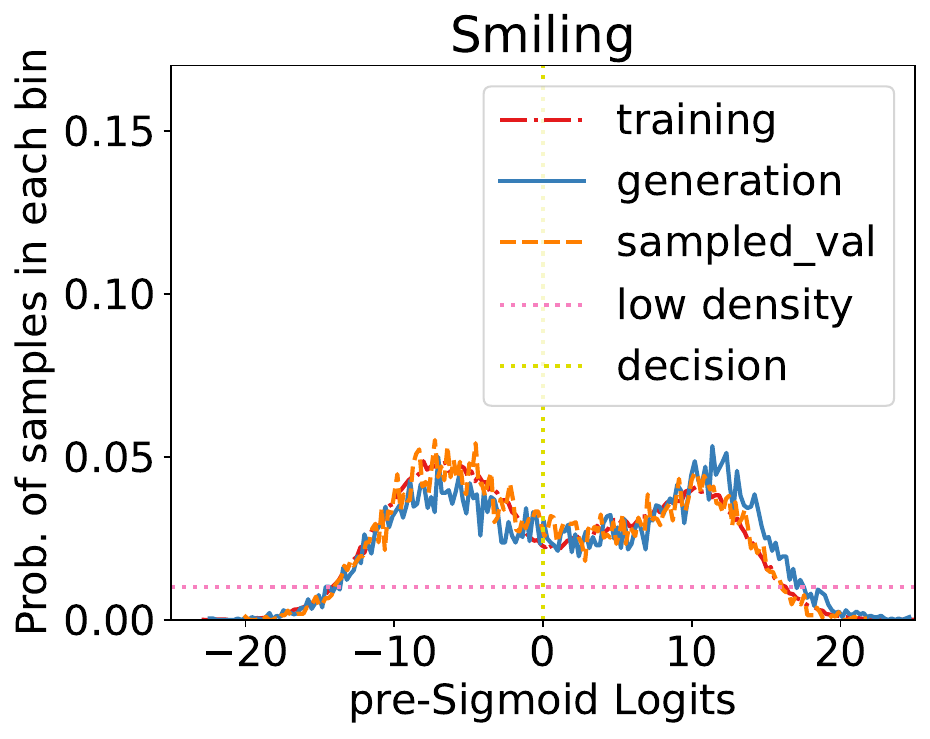}
		\caption{\textit{spectrum}: \\ Smiling}
		\label{fig:presig-smiling}
	    \end{subfigure}
	   \begin{subfigure}{0.23\textwidth}
		\includegraphics[width=\textwidth]{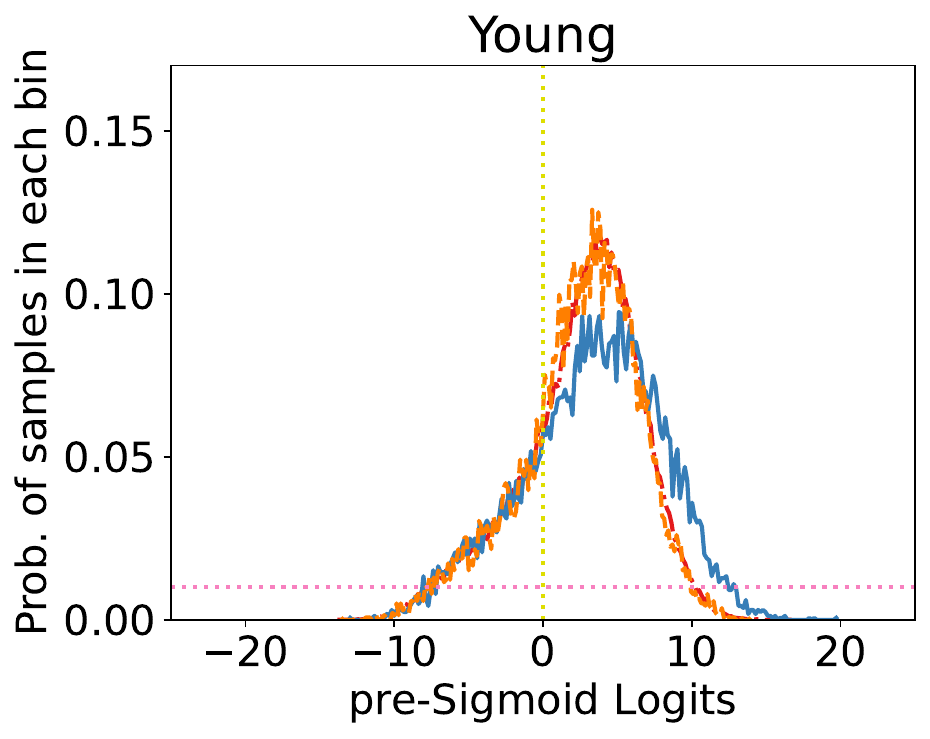}
		\caption{\textit{spectrum-based}: \\ Young}
		\label{fig:presig-young}
	   \end{subfigure}
     \begin{subfigure}{0.23\textwidth}
         \includegraphics[width=\textwidth]{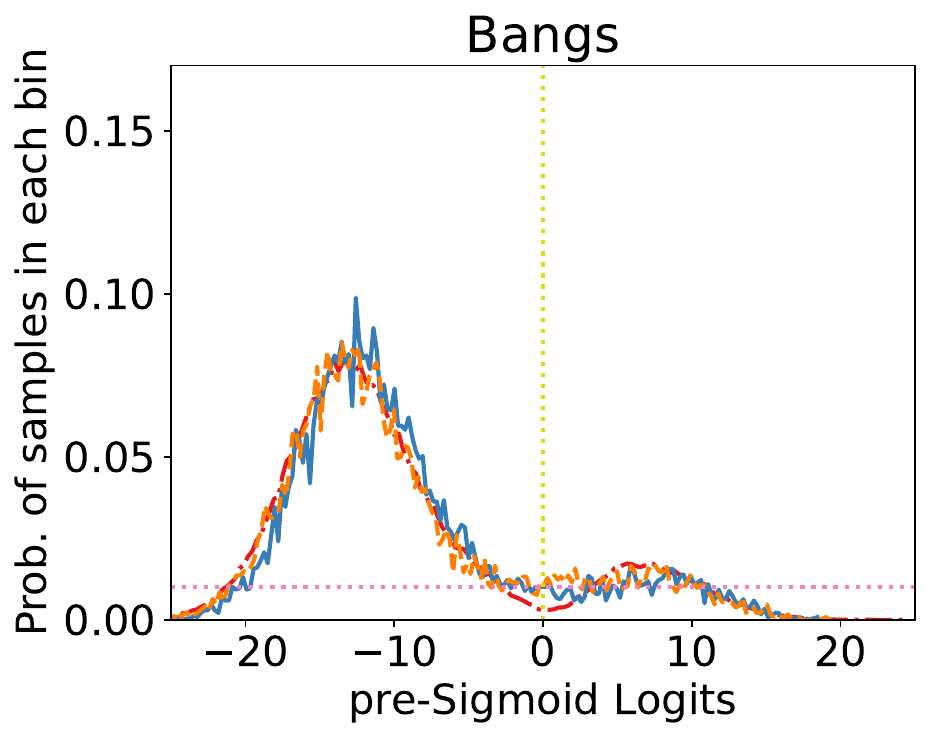}
         \caption{\textit{non-spectrum}: \\ Bangs}
         \label{fig:presig-bangs}
     \end{subfigure}
     \begin{subfigure}{0.23\textwidth}
         \includegraphics[width=\textwidth]{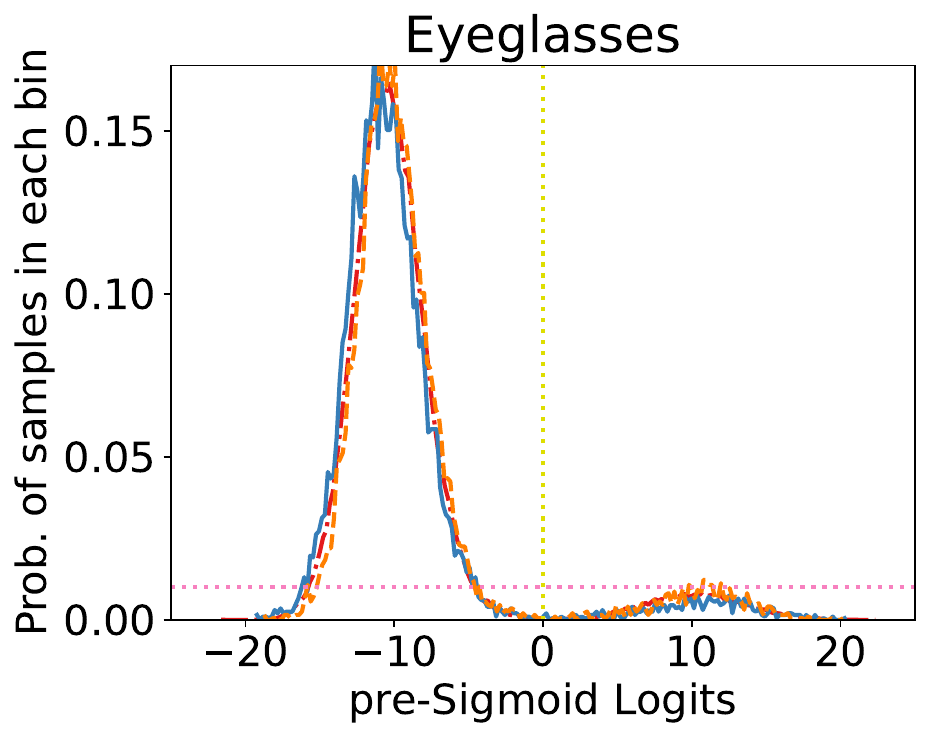}
         \caption{\textit{non-spectrum}: \\ Eyeglasses}
         \label{fig:presig-eyeglasses}
     \end{subfigure}
	\caption{\textbf{CelebA classifier's pre-sigmoid logits distributions of selected \textit{spectrum-based} and \textit{non-spectrum-based} attributes.} The decision boundary for \textit{spectrum-based} attributes (Fig.~\ref{fig:presig-smiling}, \ref{fig:presig-young}) always falls in a high-density region, while that for \textit{non-spectrum-based} attributes (Fig.~\ref{fig:presig-bangs} and \ref{fig:presig-eyeglasses}) falls in a low-density region.}
	\label{fig:attr_cls_dist_celeba}
\end{figure}

\begin{figure}[!tbp]
    \centering
    \begin{subfigure}{0.23\textwidth}
        \centering
		\includegraphics[width=\textwidth]{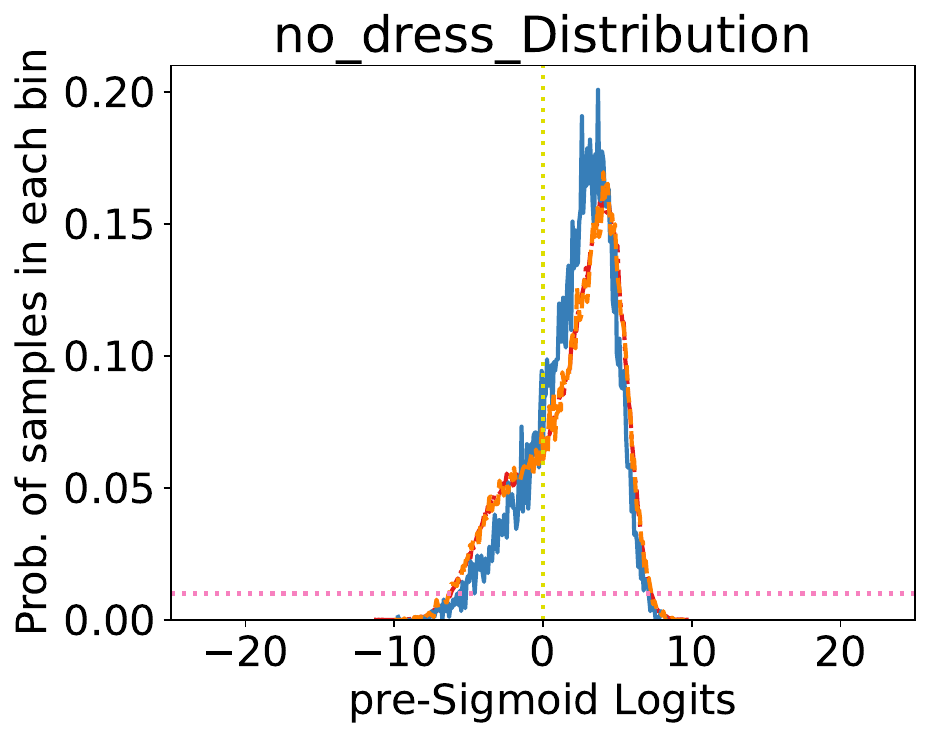}
		\caption{\textit{spectrum}: \\ No Dress}
		\label{fig:presig-no_dress}
	   \end{subfigure}
	   \begin{subfigure}{0.23\textwidth}
            \centering
		\includegraphics[width=\textwidth]{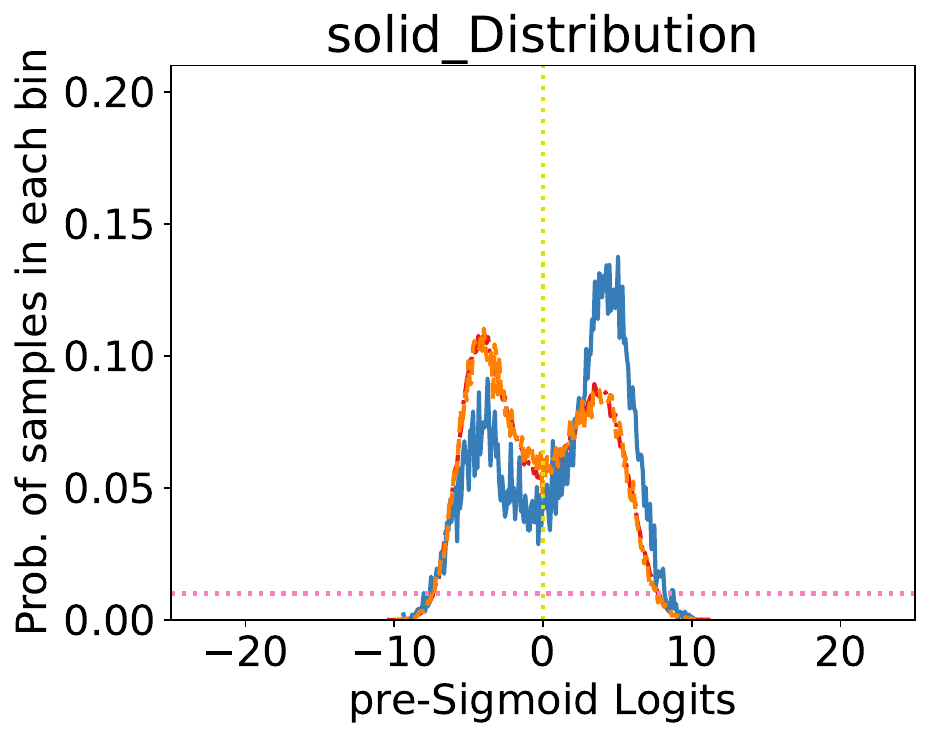}
		\caption{\textit{spectrum}: \\ Solid}
		\label{fig:presig-solid}
	    \end{subfigure}
     \begin{subfigure}{0.23\textwidth}
            \centering
		\includegraphics[width=\textwidth]{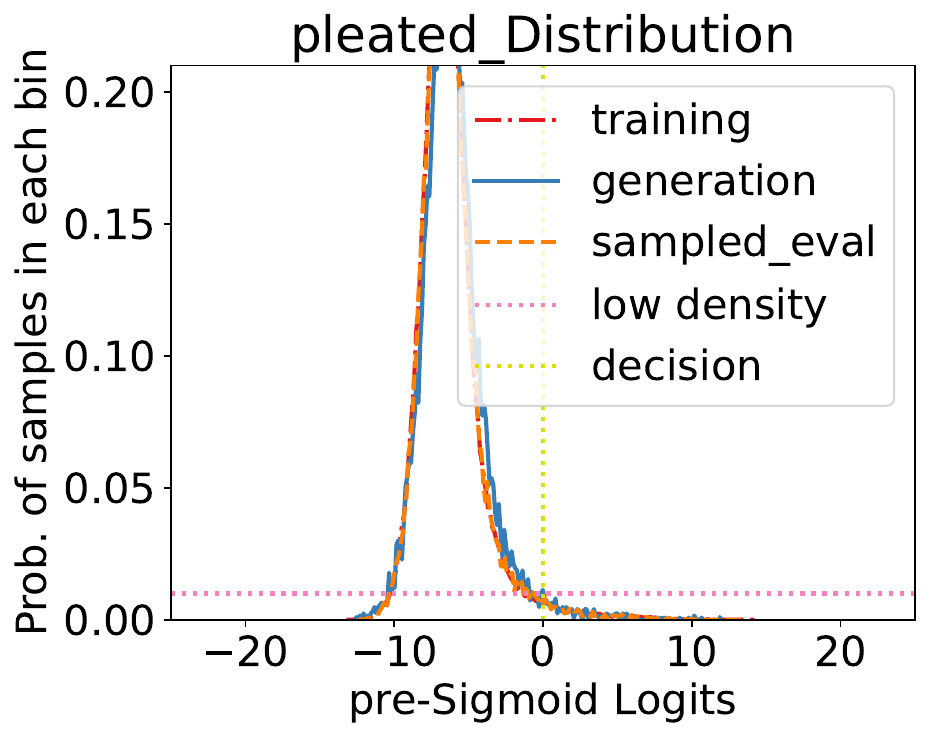}
		\caption{\textit{non-spectrum}: \\ Pleated}
		\label{fig:presig-pleated}
	    \end{subfigure}
    \caption{\textbf{DeepFashion classifier's pre-sigmoid logits distributions of selected \textit{spectrum-based} and \textit{non-spectrum-based} attribute.} The decision boundary for \textit{spectrum-based} attributes (Fig.~\ref{fig:presig-no_dress}, \ref{fig:presig-solid}) falls in a high-density region, while that for \textit{non-spectrum-based} attributes (Fig.~\ref{fig:presig-pleated}) falls in a low-density region.}
    \label{fig:attr_cls_dist_dp}
\end{figure}

To demonstrate that sampling 10,000 image generation is sufficient for a reliable statistical estimation, we present ABS between sampled subsets of the validation set and the full validation set on CelebA dataset in Fig.~\ref{fig:sample_error}. Additionally, we plot ABS between the generation set at the final checkpoint and the full validation set. The generation set has a much larger ABS compared to the sampled validation set with 10,000 images, emphasizing that the attribute bias shifts observed in Figs.~\ref{fig:bametric_marg_celeba} and \ref{fig:marg_bametric_dp} exceed the variance introduced by the sampling process. This also suggests that using 10,000 images is sufficient to estimate attribute bias shifts with minor errors.

Looking closer into attribute bias shift for each attribute (Figs.~\ref{fig:attr_bias_celeba} and \ref{fig:attr_bias_dp} in Section~\ref{sec:perattr}),
we can categorize all attributes into two categories: \textit{spectrum-based} and \textit{non-spectrum-based}.
We present the categorization of attributes in Table~\ref{tab:attr_cat}. In the following section~\ref{sec:cls_revisiting}, we will talk about the criteria for the attributes categorization.
ABS for \textit{non-spectrum-based} attributes (purple dashed lines in Fig.~\ref{fig:bametrics}) converges to small values for both datasets, reaching $0.71\%$ for CelebA and $0.98\%$ for DeepFashion. However, \textit{spectrum-based} attributes exhibit significantly larger ABS, achieving minima of $3.25\%$ for CelebA and $4.73\%$ for DeepFashion.

\subsection{Attribute bias shifts' sensitivity relates to decision boundary} 

\label{sec:cls_revisiting}

In this section, we analyze the classifier to explain why some attributes experience greater attribute bias shifts than others, leading to the attribute categorization presented in Table~\ref{tab:attr_cat}.

Figs.~\ref{fig:attr_cls_dist_celeba} and \ref{fig:attr_cls_dist_dp} show the trained classifier's pre-sigmoid logits distribution for some attributes of CelebA and DeepFashion respectively. The distributions for all attributes are in Appendix~\ref{app:cls}. 
These plots provide visualizations of how the data points are distributed in a projected uni-dimensional space. To estimate the empirical distributions, we use all the training images, 10,000 images sampled from the validation set, and all the 10,000 images in the generation set.

\begin{figure}[!tbp]
    \centering
    \begin{subfigure}{0.24\textwidth}
        \centering
		\includegraphics[width=0.93\textwidth]{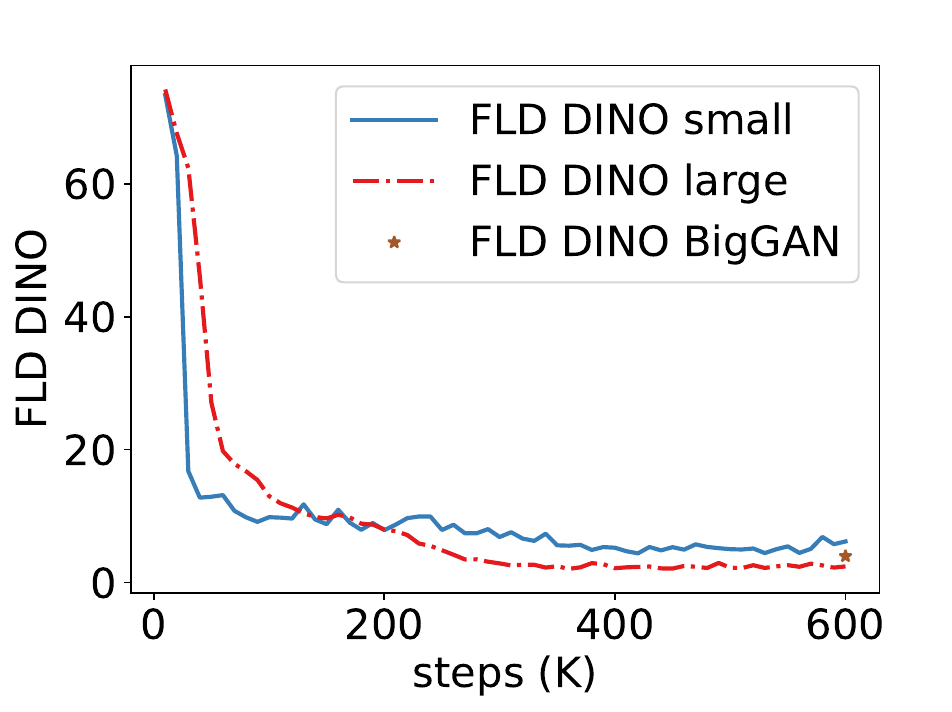}
		\caption{FLD}
		\label{fig:fld_cmp}
	    \end{subfigure}
    \begin{subfigure}{0.24\textwidth}
        \centering
		\includegraphics[width=\textwidth]{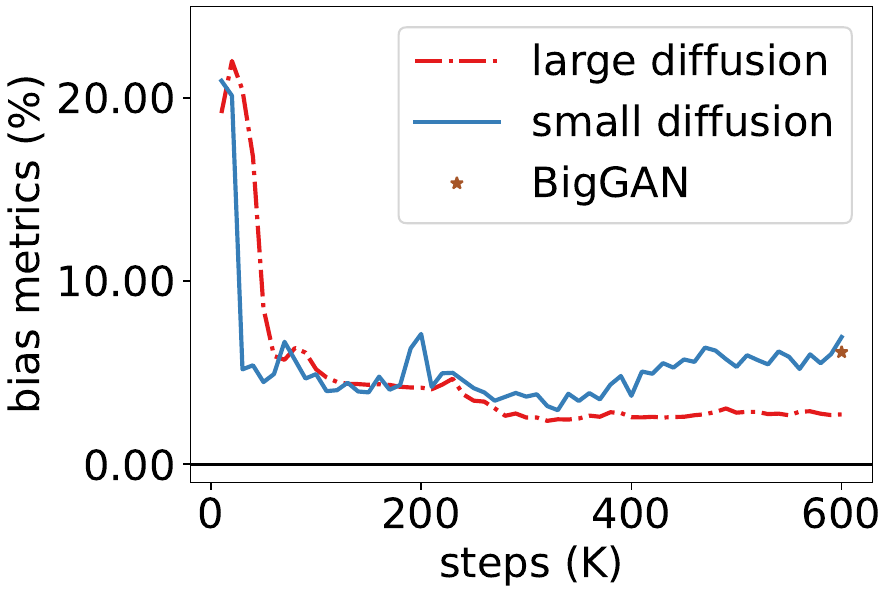}
		\caption{ABS overall}
		\label{fig:bsr_overall}
	   \end{subfigure}
    \begin{subfigure}{0.24\textwidth}
        \centering
		\includegraphics[width=\textwidth]{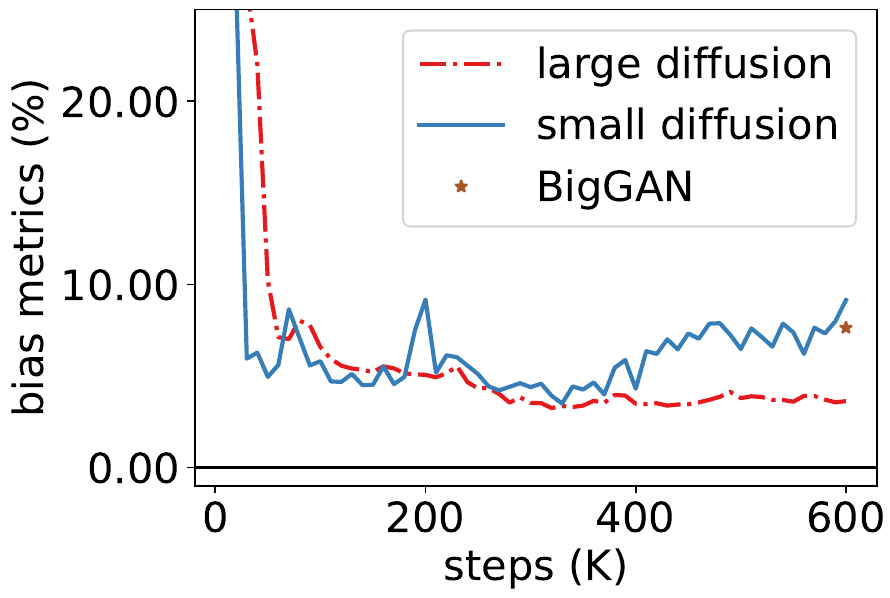}
		\caption{\scriptsize ABS spectrum-based}
		\label{fig:bsr_separate}
	    \end{subfigure}
     \begin{subfigure}{0.24\textwidth}
        \centering
		\includegraphics[width=\textwidth]{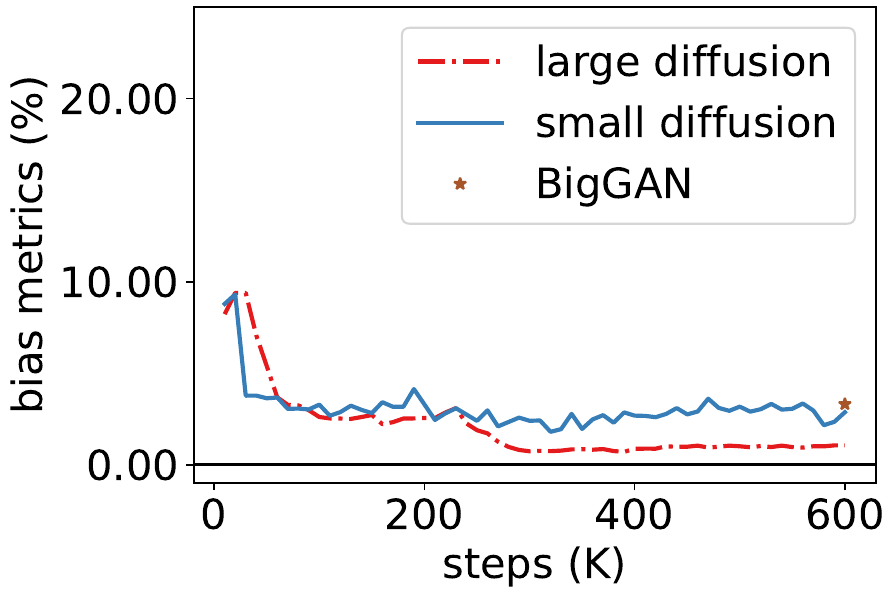}
		\caption{\scriptsize ABS non-spectrum-based}
		\label{fig:bsr_obj}
	    \end{subfigure}
    \caption{\textbf{FLD and ABS of different generative models on CelebA.} The small diffusion model has slightly worse image generation quality but much larger ABS for both \textit{spectrum-based} and \textit{non-spectrum-based} attributes compared to the large diffusion model. BigGAN has a similar FLD as the large diffusion model but has larger attribute bias shifts.}
    \label{fig:bias_compare}
\end{figure}

The main difference between \textit{small attribute bias shift} and \textit{large attribute bias shift} attributes is the density at the decision boundary. The distribution shifts for different attributes can manifest in various ways, but the decision boundaries for \textit{large attribute bias shift} attributes consistently fall in higher density regions compared to those for \textit{small attribute bias shift} ones. 
We thus use the density where the decision boundary falls in the validation distribution to categorize the attributes. Those with density more than 0.01 are categorized as \textit{spectrum-based}, and vice versa. This density selection aligns with previous studies, as discussed in Appendix~\ref{app:category_alignment}.

Attribute bias shifts of \textit{spectrum-based} attributes are more sensitive to distribution shifts compared to \textit{non-spectrum-based} attributes. The distributions for \textit{non-spectrum-based} attributes still change between training and generation sets, but their effects on attribute bias shifts are small. 
Since the decision boundary falls in a low-density region, it is more difficult to transport the density mass from one side of the boundary to the other. We find empirically that the distribution shifts between the training and generation distributions generally have low earth mover's distance (EMD)~\citep{emd}. Significant reweighting of well-separated modes would constitute a significant EMD between training and generated distributions.


\begin{table}[!tbp]
    \centering
    \tiny
    \caption{Attribute categorization of \textit{spectrum-based} and \textit{non-spectrum-based} for each dataset.}
    \begin{tabular}{c|l|l}
    \toprule
       Dataset  & \textit{spectrum-based} attributes & \textit{non-spectrum-based} attributes \\
       \midrule
       \multirow{5}{*}{CelebA} 
       & \texttt{Rosy\_Cheeks}, \texttt{Big\_Nose}, \texttt{No\_Beard}, \texttt{Narrow\_Eyes}, \texttt{Arched\_Eyebrows}, & \texttt{5-o-Clock\_Shadow}, \texttt{Bangs}, \\
        &  \texttt{High\_Cheekbones}, \texttt{Bushy\_Eyebrows}, \texttt{Black\_Hair}, \texttt{Receding\_Hairline},  & \texttt{Eyeglasses}, \texttt{Bald}, \texttt{Double\_Chin}, \\
        & \texttt{Brown\_Hair}, \texttt{Straight\_Hair}, \texttt{Bags\_Under\_Eyes}, \texttt{Pointy\_Nose}, & \texttt{Wearing\_Hat}, \texttt{Male}, \texttt{Blond\_Hair},  \\
        & \texttt{Big\_Lips}, \texttt{Mouth\_Slightly\_Open}, \texttt{Heavy\_Makeup}, \texttt{Attractive}, & \texttt{Gray\_Hair}, \texttt{Mustache}, \texttt{Chubby}, \\
        & \texttt{Smiling}, \texttt{Wearing\_Lipstick}, \texttt{Wavy\_Hair}, \texttt{Young}, \texttt{Oval\_Face}, & \texttt{Pale\_Skin}, \texttt{Sideburns},\texttt{Goatee},   \\
 \midrule
       \multirow{4}{*}{DeepFashion} 
        &\texttt{Floral}, \texttt{Graphic}, \texttt{Embroidered}, \texttt{Solid}, \texttt{Long\_sleeve}, \texttt{Short\_sleeve}, &  \texttt{Striped}, \texttt{Pleated}, \\
        &   \texttt{Sleeveless}, \texttt{Knit}, \texttt{Chiffon}, \texttt{Cotton},  \texttt{Maxi\_length}, \texttt{Mini\_length}, & \texttt{Leather}, \texttt{Faux}, \\
        &   \texttt{No\_dress}, \texttt{Crew\_neckline},\texttt{V\_neckline}, \texttt{No\_neckline},   & \texttt{Square\_neckline}, \\
        & \texttt{Loose}, \texttt{Tight}, \texttt{Conventional} &  \texttt{Lattice}, \texttt{Denim},  \\
       \bottomrule
    \end{tabular}
    \label{tab:attr_cat}
\end{table}

\subsection{Attribute bias shift in different generative models}
\label{sec:differentsizes}

In this section, we compare the attribute bias shifts for different sizes of diffusion models by changing the number of channels in the bottleneck layer of U-Net (32 for tiny, 64 for small, and 256 for large), and BigGAN model. Fig.~\ref{fig:bias_compare} shows image generation metrics and bias evaluation metrics for different generative models. The tiny diffusion model cannot generate realistic images (check Appendix~\ref{app:gen_samples} for sampled images), making it unsuitable for our bias analysis framework. FLD for the small diffusion model is worse than the large diffusion model, while BigGAN achieves a similar FLD as the large diffusion model. However, ABS shows clear differences among generative models (See Figs.~\ref{fig:bsr_overall}, \ref{fig:bsr_separate} and \ref{fig:bsr_obj}).

\begin{figure}[!tbp]
      \centering
      \begin{subfigure}{0.23\textwidth}
            \centering
		\includegraphics[width=\textwidth]{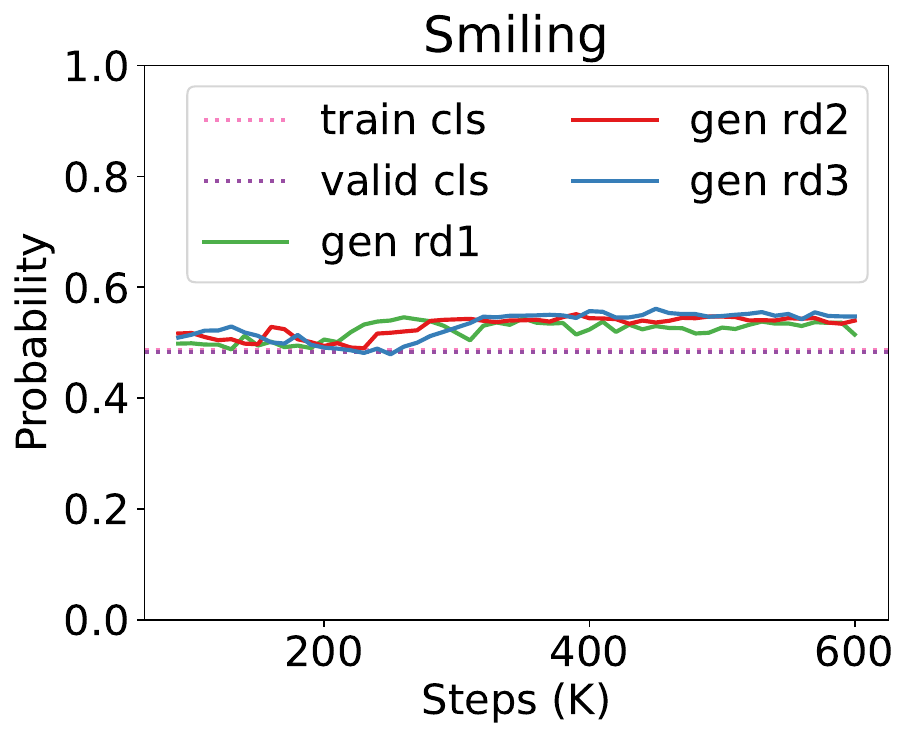}
		\caption{\textit{spectrum}: \\ Smiling}
		\label{fig:smiling}
	    \end{subfigure}
	   \begin{subfigure}{0.23\textwidth}
            \centering
		\includegraphics[width=\textwidth]{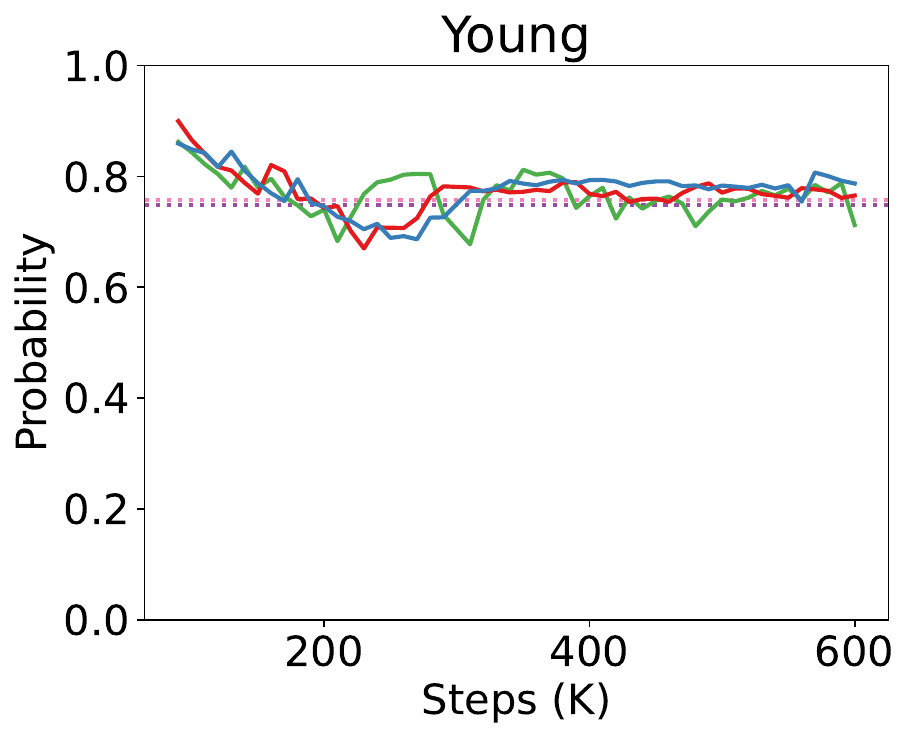}
		\caption{\textit{spectrum}: \\ Young}
		\label{fig:young}
	   \end{subfigure}
     \begin{subfigure}{0.23\textwidth}
            \centering
         \includegraphics[width=\textwidth]{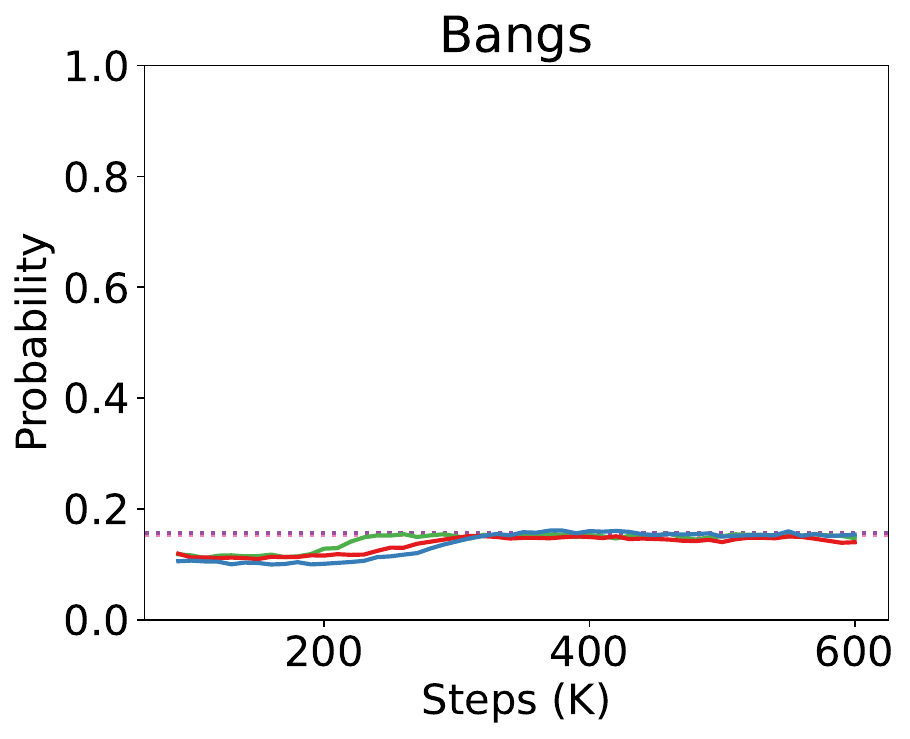}
         \caption{\textit{non-spectrum}: \\ Bangs}
          \label{fig:bangs}
     \end{subfigure}
     \begin{subfigure}{0.23\textwidth}
            \centering
         \includegraphics[width=\textwidth]{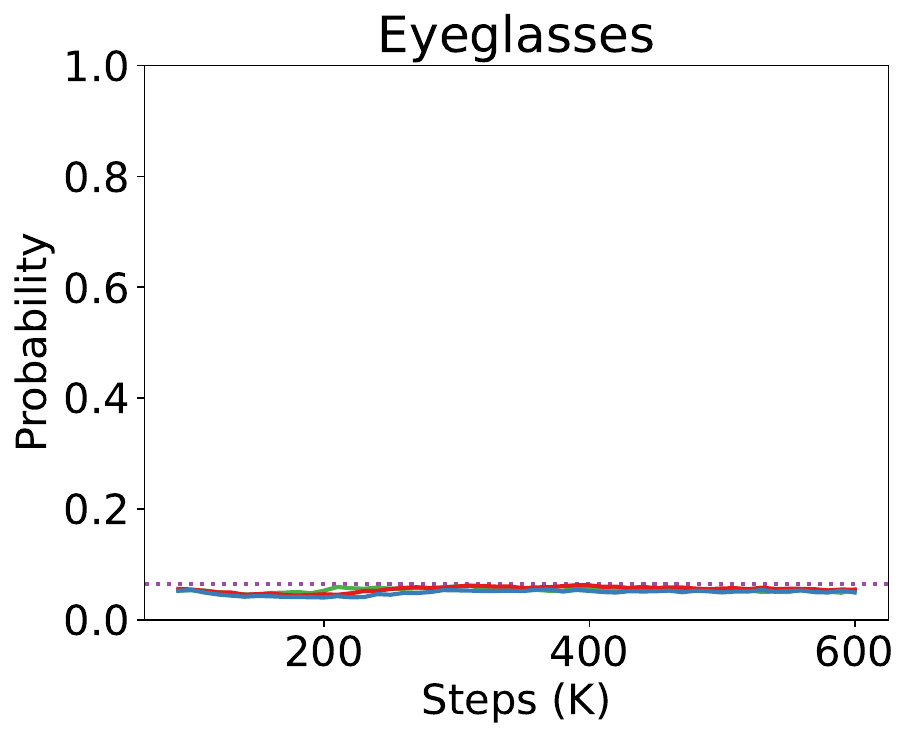}
         \caption{\textit{non-spectrum}: \\ Eyeglasses}
         \label{fig:eyeglasses}
     \end{subfigure}
	\caption{\textbf{Probabilities of selected \textit{spectrum-based} and \textit{non-spectrum-based} attributes during training on the CelebA dataset with three different random seeds.} The probabilities of \textit{spectrum-based} attributes (Fig.~\ref{fig:smiling} and \ref{fig:young}) present a gap between generation and validation data while \textit{non-spectrum-based} ones (Fig.~\ref{fig:bangs} and \ref{fig:eyeglasses}) do not.}
	\label{fig:attr_bias_celeba}
    \vspace{-1.em}
\end{figure}

\begin{figure}[!tbp]
    \centering
    \begin{subfigure}{0.23\textwidth}
            \centering
		\includegraphics[width=\textwidth]{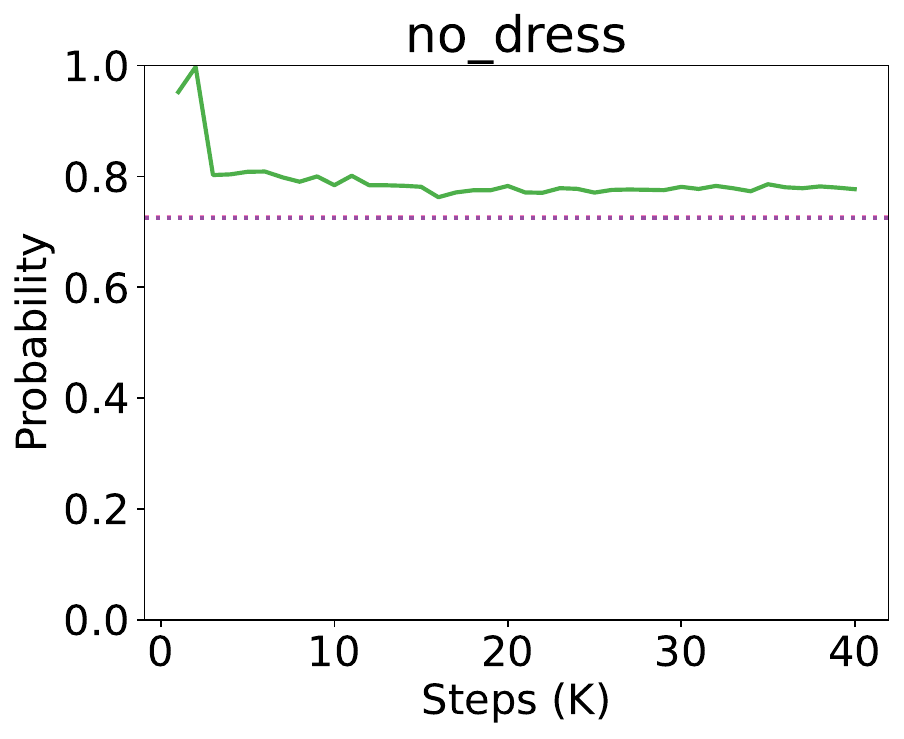}
		\caption{\textit{spectrum}: \\ No Dress}
		\label{fig:no_dress}
	   \end{subfigure}
    \begin{subfigure}{0.23\textwidth}
            \centering
		\includegraphics[width=\textwidth]{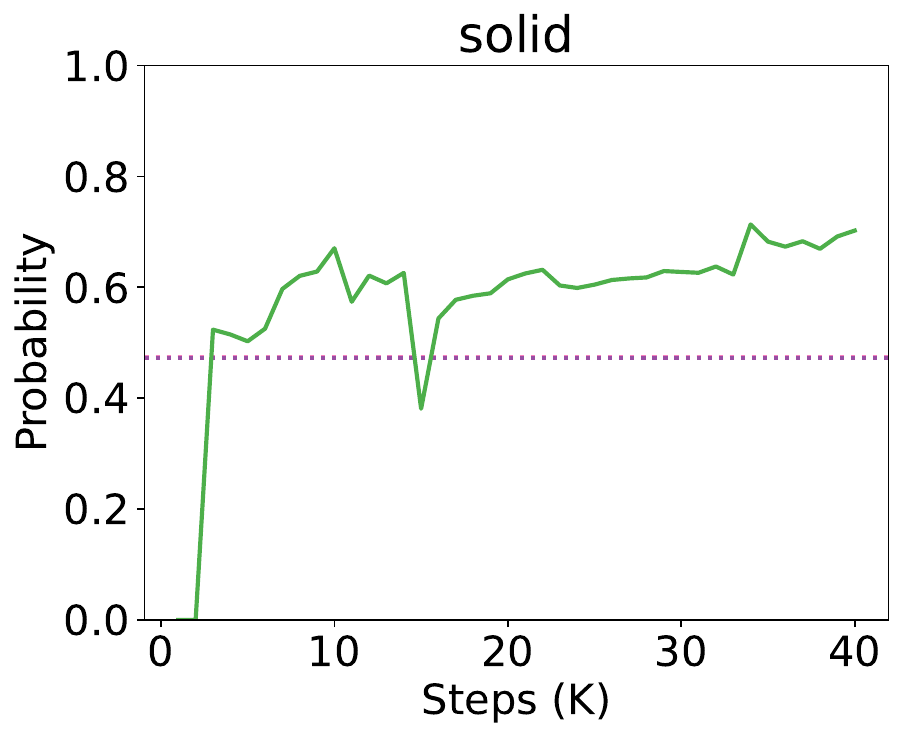}
		\caption{\textit{spectrum}: \\ Solid}
		\label{fig:solid}
	    \end{subfigure}
     \begin{subfigure}{0.23\textwidth}
            \centering
		\includegraphics[width=\textwidth]{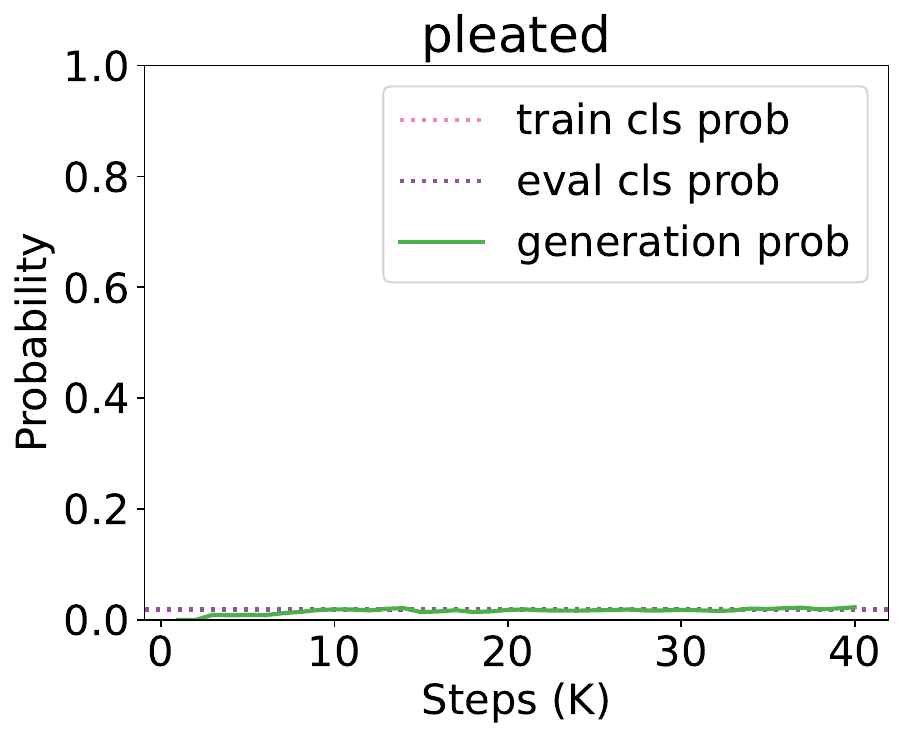}
		\caption{\textit{non-spectrum}: \\ Pleated}
		\label{fig:pleated}
	    \end{subfigure}
    \caption{\textbf{Probabilities of selected \textit{spectrum-based} and \textit{non-spectrum-based} attribute during training on the DeepFashion dataset.} The probabilities of \textit{spectrum-based} attributes (Fig.~\ref{fig:no_dress}, \ref{fig:solid}) present a gap between generation and validation data while \textit{non-spectrum-based} one (Fig.~\ref{fig:pleated}) does not.}
    \label{fig:attr_bias_dp}
\end{figure}

Diffusion models have matched—or even surpassed—GAN models in terms of image synthesis performance.~\citep{adm21}. 
We evaluate whether diffusion models also perform better than BigGAN in terms of attribute bias shifts. We observe that BigGAN exhibits a considerably larger ABS compared to the large diffusion model, despite having only slightly worse image generation performance according to FLD. 
This finding may be attributed to the well-known issue of mode collapse in GAN models~\citep{wgan}.


We train different sizes of diffusion models to study the influence of the model size on attribute bias shifts. The \textit{small} diffusion model exhibits larger bias shifts compared to the \textit{large} diffusion model.
For the \textit{small} diffusion model, we notice that the average attribute bias shift increases after 300K steps although the FLD value does not change significantly. While BigGAN has better FLD, the attribute bias shifts are similar to those of the \textit{small} diffusion model at the end of the training.
We also observe more fluctuations in bias evaluation metrics for the small diffusion model.

We present image samples generated from different models in Appendix~\ref{app:gen_samples}. The images generated by BigGAN and the small diffusion model are more ``washed out'' than those produced by the large diffusion model, showing fewer variations and less details. 

\subsection{Per-attribute bias shift}
\label{sec:perattr}
Figs.~\ref{fig:attr_bias_celeba} and \ref{fig:attr_bias_dp} show probabilities for selected attributes in the generated data during training. Plots for other attributes are in Appendix~\ref{app:perattribute}. Probabilities of \textit{spectrum-based} attributes generally exhibit values distinct from the classifier-predicted validation probabilities, resulting in attribute bias shifts in Fig.~\ref{fig:bametrics}. 

\textit{Spectrum-based} attributes exhibit more fluctuations throughout training compared to \textit{non-spectrum-based} ones. While the probabilities for many attributes converge before 300K steps, \texttt{young} (Fig.~\ref{fig:young}) still has fluctuations. A similar pattern is also witnessed in DeepFashion, where \texttt{solid} (Fig.~\ref{fig:solid}), as a \textit{spectrum-based} attribute, also exhibits perceivable fluctuations. This suggests that extra caution is needed when handling certain \textit{spectrum-based} attributes using generative models. 

We conduct several runs of training using three different random seeds on CelebA dataset. 
There is randomness across different random seeds as the curves for each random seed vary. However, the probabilities of each attribute from distinct random seeds generally converge to the same value. 
Therefore, we report results for only one random seed in other experiments.

\vspace{-0.5em}
\section{Conclusion}
\vspace{-0.5em}
\label{sec:conclusion}
This study examines bias shifts caused by the inductive biases of unconditional image generative models. Our experimental results show that different attributes have varying bias shifts in response to distribution changes. Attributes for which the classifier's decision boundary falls in a low-density area tend to have small bias shifts. We thus categorize attributes into \textit{spectrum-based} and \textit{non-spectrum-based} sets. Our analysis results in the following observations: 1) Biases shift between training and generation distributions for unconditional image generative models. 2) Selecting a model checkpoint with the best image generation metrics does not guarantee the smallest bias shifts. 3) BigGAN models and small diffusion models have larger bias shifts compared to large diffusion models, despite having similar image generation metric values.


We hope that our analysis of unconditional generative models will enable researchers to systematically study the effects of additional sources of bias—such as conditioning (e.g., with ground-truth labels or text), guidance, and pretrained modules—on bias shift.

\textbf{Limitations. } This analysis is conducted in a controlled setting using classifier-predicted labels, which may not reflect real-world ambiguity in attribute interpretation. The reliance on binary annotations introduces limitation for spectrum-based attributes. Our findings, especially the small bias shift from training to generation, may be limited to unconditional generative models trained on these two datasets, CelebA and DeepFashion, and may not generalize beyond these settings.

\label{bimpact}
\textbf{Broader Impact. \ \ }Our findings should not be interpreted as evidence that algorithmic bias is not a pressing issue. Rather, to underscore the need for critical reflection on how such tools are applied and interpreted. We explicitly recognize the substantial body of research documenting algorithmic bias and its harmful societal impacts, especially in areas such as facial recognition, hiring, credit scoring, and content moderation (e.g., \cite{barocas2016big, buolamwini2018gender, raji2019actionable, sandvig2014auditing}).

While CelebA offers a controlled environment for analysis, it abstracts away from the sociocultural complexities of real-world distributions. For instance, the dataset presents a highly curated and stylized representation of human faces, collected under relatively uniform conditions and adhering to a binary gender schema. As such, it encodes a simplified `worldview' of human attributes—one that lacks the nuance, ambiguity, and diversity present in more realistic social settings. The attribute labels in CelebA reflect specific cultural and aesthetic norms from the time and context of its curation, and the decisions embedded in its labeling (e.g., what counts as `smiling' or `attractive') may not generalize meaningfully across contexts or demographics. Relying on such datasets, without critically engaging with these assumptions, risks drawing misleading conclusions about generative model behavior. This concern aligns with prior critiques on dataset bias and oversimplification in ML (e.g., \cite{Scheuerman, datasheet}). We caution against over-interpreting the findings of CelebA as reflective of broader societal dynamics.




\bibliography{main.bib}
\bibliographystyle{unsrt}

\appendix
\title{Appendix}

\section{Proof of bias shift in translation distribution changes}
\label{app:translationproof}
Given a continuous random variable $C$, its density function $f$, and the decision boundary $C=t$, the probability of this random variable being positive is
\begin{equation}
    P(C \geq t) = \int_t^{+\infty} f_C(x) dx .
\end{equation}
If the distribution of this random variable shifts $+\delta$, using the same decision boundary, the probability of this shifted random variable being positive is
\begin{equation}
    \Tilde{P}(C \geq t) = \int_t^{+\infty} f_C(x-\delta) dx = \int_{t-\delta}^{+\infty} f_C(x) dx .
\end{equation}
Using the definition of bias shift in Section~\ref{sec:framework},
\begin{equation}
    B_{\text{shift}}(C) = \lvert \Tilde{P}(C \geq t) - P(C \geq t) \rvert = \left\lvert \int_{t-\delta}^{t} f_C(x) dx \right\rvert .
\end{equation}
We can see that the bias shift depends on the density function within the $\delta$ neighborhood of the decision boundary. If the decision boundary falls in a low-density region, then the bias shift will be small towards the $\delta$ translation distribution shift. This explains the cancel-out effect of red and blue areas with low-density decision boundaries in Fig.~\ref{fig:cartoon}.

\section{Datasets}
\label{app:datasets}
CelebA~\citep{celebadataset} is a large-scale face attributes dataset with 200,000 celebrity images, each with 40 attribute annotations. The dataset includes 10,000 celebrities with 20 images for each. These attribute annotations cover a wide variety of facial characteristics, ranging from details (e.g., earrings, pointy noise, etc.) to outlines (e.g., hair color, gender, age, etc.). We list all 40 attributes in Table~\ref{tab:dataset_attr}. Before feeding the training images to the model, we centre crop the images and resize them to 128x128 pixels. Because of the crop, some attributes, e.g., \texttt{Wearing\_Necklace}, \texttt{Wearing\_Necktie}, are not visually grounded in the post-process images. \texttt{Blurry} is also an attribute that we do not include since we want the image generation quality to be good. We excluded these attributes in Table~\ref{tab:attr_cat}. We follow the Training/Validation/Test set split in the official release. Training set includes the images of the first eight thousand identities (with 160 thousand images). Validation set contains the images of another one thousand identities (with twenty thousand images). The remaining one thousand identities (with twenty thousand images) go for Test set. In our bias analysis framework, we only use the Training set and the Validadtion set.

DeepFashion~\citep{deepfashiondataset} is a clothes dataset with over 800,000 diverse fashion images, including tops and bottoms. No footwears is in this dataset. Each image is associated with 1000 coarse attribute annotations about texture, fabric, shape, part, and style of the clothes. These attribute annotations are scrapped directly from meta-data of the images. They are thus very noisy and not reliable. Most of the attributes have less than 1\% positive samples, making the classification problem very imbalanced. This dataset also provides a fine-grained annotation subset, where each image is associated with 26 find-grained attribute annotations. These attributes are presented in Table~\ref{tab:dataset_attr}. We train a classifier on this subset and apply this trained classifier to the whole dataset and get classifier-predicted labels for each image. We follow the Training/Validation/Test set split in the official release. Unlike CelebA dataset, the split of DeepFashion dataset is random.

\begin{table}[t]
\small
    \centering
    \caption{Labeled attributes in CelebA and DeepFashion datasets. CelebA has 40 attributes and DeepFashion has 26 attributes.}
    \begin{tabular}{l|l} \toprule
        Dataset & Attributes \\ \midrule
        \multirow{8}{*}{CelebA} & \texttt{5\_o\_Clock\_Shadow}, \texttt{Arched\_Eyebrows}, \texttt{Attractive}, \texttt{Bags\_Under\_Eyes},\\
         &  \texttt{Bald}, \texttt{Bangs}, \texttt{Big\_Lips}, \texttt{Big\_Nose}, \texttt{Black\_Hair}, \texttt{Blond\_Hair}, \texttt{Blurry}, \\
         & \texttt{Brown\_Hair},\texttt{Bushy\_Eyebrows}, \texttt{Chubby}, \texttt{Double\_Chin}, \texttt{Eyeglasses}, \texttt{Goatee},  \\
         & \texttt{Gray\_Hair}, \texttt{Heavy\_Makeup} \texttt{High\_Cheekbones}, \texttt{Male}, \texttt{Mouth\_Slightly\_Open},  \\
         & \texttt{Mustache}, \texttt{Narrow\_Eyes}, \texttt{No\_Beard}, \texttt{Oval\_Face}, \texttt{Pale\_Skin}, \texttt{Pointy\_Nose}, \\
         & \texttt{Receding\_Hairline}, \texttt{Rosy\_Cheeks}, \texttt{Sideburns}, \texttt{Smiling}, \texttt{Straight\_Hair},  \\
         & \texttt{Wavy\_Hair}, \texttt{Wearing\_Earrings}, \texttt{Wearing\_Hat}, \texttt{Wearing\_Lipstick},  \\
         &  \texttt{Wearing\_Necklace}, \texttt{Wearing\_Necktie}, \texttt{Young} \\ \midrule
        \multirow{6}{*}{DeepFashion} & \texttt{floral}, \texttt{graphic}, \texttt{striped}, \texttt{embroidered}, \texttt{pleated}, \texttt{solid}, \texttt{lattice}, \\
        & \texttt{long\_sleeve}, \texttt{short\_sleeve}, \texttt{sleeveless} \\
        & \texttt{maxi\_length}, \texttt{mini\_length}, \texttt{no\_dress}, \\
        & \texttt{crew\_neckline}, \texttt{v\_neckline}, \texttt{square\_neckline}, \texttt{no\_neckline}, \\
        & \texttt{denim}, \texttt{chiffon}, \texttt{cotton}, \texttt{leather}, \texttt{faux}, \texttt{knit}, \\
        & \texttt{tight}, \texttt{loose}, \texttt{conventional}   \\ 
         \bottomrule
    \end{tabular}
    \label{tab:dataset_attr}
\end{table}

\newpage

\section{Licenses for existing assets}
We list licenses for all the models, datasets and libraries used in our paper in Table~\ref{tab:licences}.
\begin{table}[t]
    \centering
    \caption{Licenses for all the models, datasets and libraries used in our paper.}
    \begin{tabular}{c|c|c} \toprule
        \textbf{Asset} & \textbf{Type} & \textbf{License} \\ \midrule
        CelebA~\cite{celebadataset} & Dataset & non-commercial research purposes only \\ 
        DeepFashion~\cite{deepfashiondataset} & Dataset & non-commercial research purposes only \\ 
        ADM~\cite{adm21} & Codebase & MIT License \\
        BigGAN~\cite{biggan}~\tablefootnote{\url{https://github.com/ajbrock/BigGAN-PyTorch}} & Codebase & MIT License \\ \bottomrule
    \end{tabular}
    \label{tab:licences}
\end{table}

\section{Training Details}
\label{app:hyperparams}
\subsection{Diffusion Models}
\label{app:diffusion}
We follow the training setting of \cite{adm21} to train the ablated diffusion models (ADMs). Hyperparameters and architecture selections are in Table~\ref{tab:hyper_diffusion}. We train the diffusion using NVIDIA A100 40GB. The batch size per GPU is set to 16, and we use 8 GPUs to train. During training, we save checkpoint for EMA models every 10K steps. We use half precision (FP16) for training and inference.

\begin{table}[!htbp]
\small
    \centering
    \caption{Hyperparameters and architecture selection for diffusion models}
    \begin{tabular}{ccccccc}
    \toprule
        lr & bsz & channel & res\_block & dropout & diffusion\_step & inference\_step \\ \midrule
        1e-4 & 128 & 256 & 2 & 0.3 & 1000 & ddim100\\
        \bottomrule
    \end{tabular}
    \label{tab:hyper_diffusion}
\end{table}

\begin{figure}[!t]
    \centering
    \includegraphics[width=0.6\linewidth]{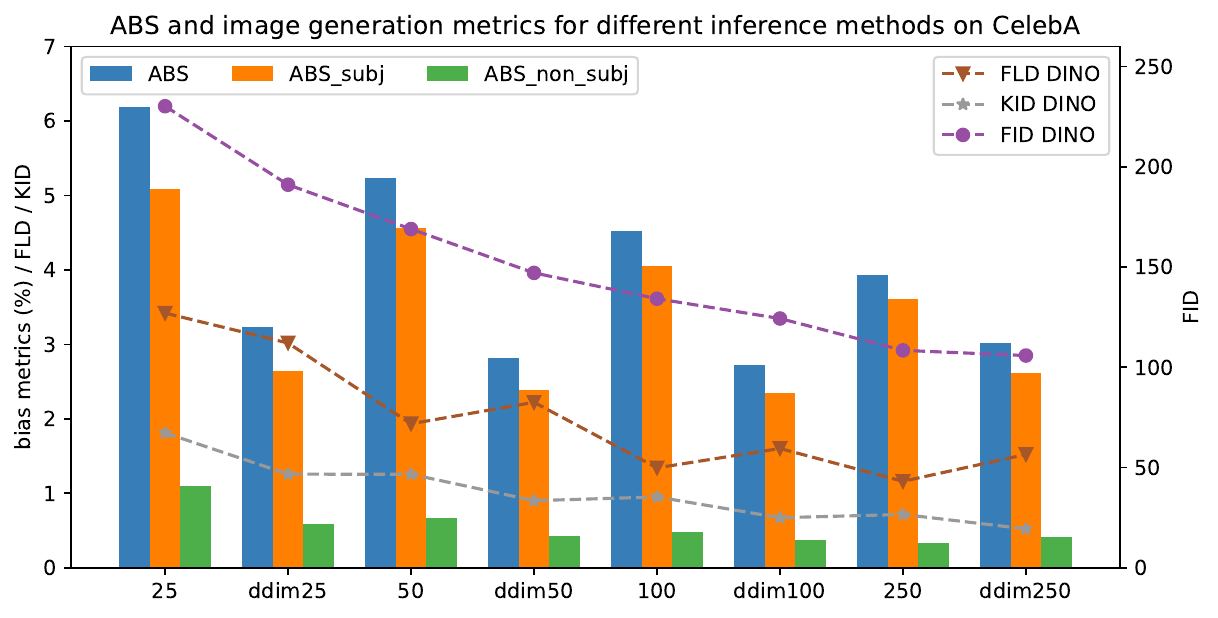}
    \caption{ABS and image generation metrics using different inference methods and inference steps on CelebA dataset. Images generated by DDIM have less bias shifts compared to those by Improved Diffusion Sampler. FID and KID also show the superiority of DDIM sampler.}
    \label{fig:inference_cmp}
\end{figure}

For each saved checkpoint, we employ 100 steps in inference to generate 10K images from the Gaussian noise. We compare the two inference methods used in ADM~\citep{adm21}, one proposed by improved diffusion model~\citep{improvediffusion}, and DDIM~\citep{ddim21}. The results on CelebA dataset are in Fig.~\ref{fig:inference_cmp}. 
Images generated by the improved diffusion sampler exhibit more bias shifts than those from DDIM.
Although FLD shows a slight improvement on improved diffusion sampler, DDIM works better in terms of FID and KID using the same steps of inference. Since we want to test less biased generations, we use DDIM with 100 steps during inference in our experiments.

In Fig.~\ref{fig:gengap_celeba}, generalization gaps for CelebA and DeepFashion datasets are different. This is because the split of the dataset is in different ways. In CelebA dataset, the training and validation sets contain the faces of distinct sets of celebrities. In DeepFashion dataset, the training and validation samples are split randomly. The distribution difference between training and validation sets of CelebA is larger than that of DeepFashion.

\subsection{Classifiers}
\label{app:cls}
We employ pre-trained ResNeXt model~\cite{resnext17} and SwinTransformer model~\cite{swint} as backbone models. For ResNeXt model, we fine-tune the last 6 layers of the pre-trained model as well as the classification layer. For SwinTransformer large model, we fine-tune last 14 layers in stage-3 and all layers in stage-4. We use AdamW optimizer and learning rate at 0.001. 
We follow a standard training procedure for the classifier training. We train the classifier on the train set (with ground truth labels) and choose the best classifier according to the average performance across all the considered attributes on the valid set (with ground truth labels). We use data augmentations to make the classifier more robust. The data augmentations include random horizontal flip, scaling, resizing, etc. This can help the classifier become more reliable when applied to the generation set.

\begin{figure}[tbp]
    \centering
    \begin{subfigure}{0.19\textwidth}
		\includegraphics[width=\textwidth]{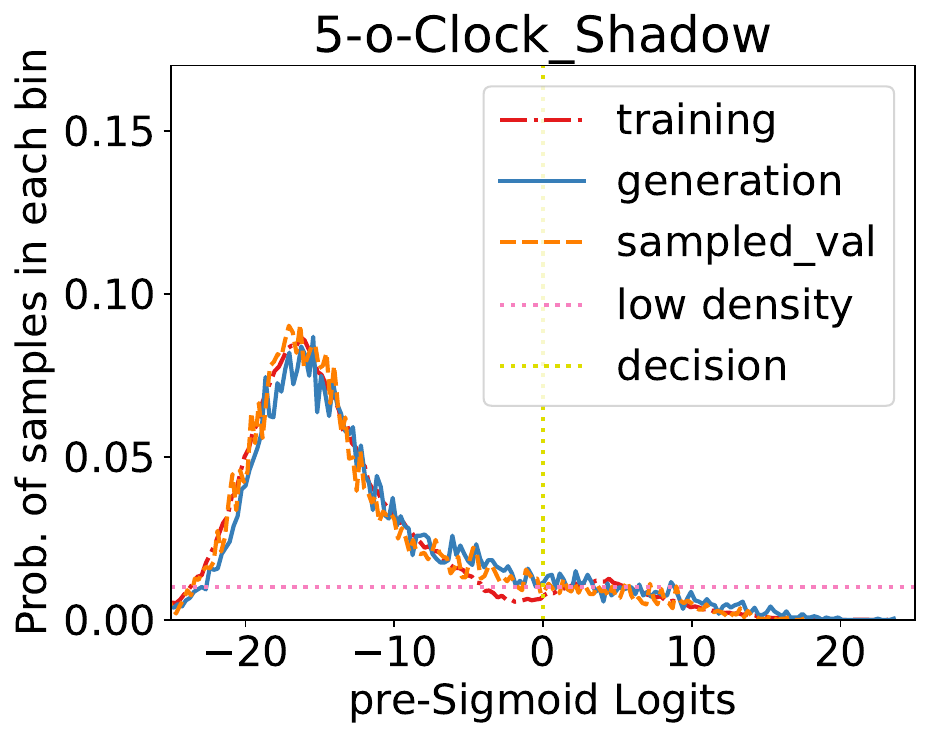}
		\caption{5o Clock Shadow}
    \end{subfigure}
    \begin{subfigure}{0.19\textwidth}
		\includegraphics[width=\textwidth]{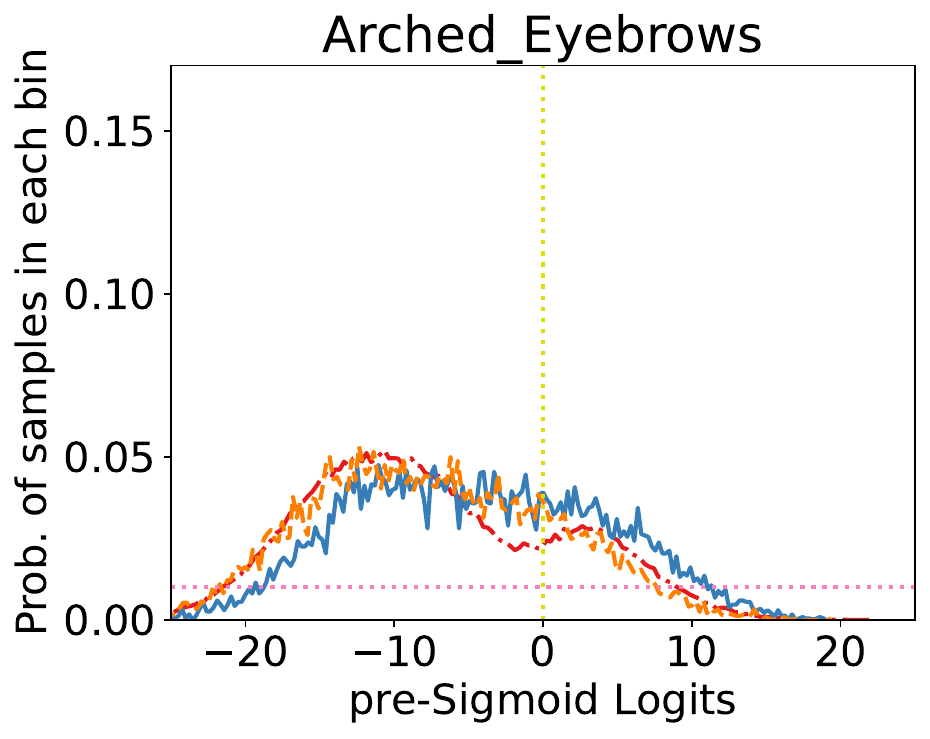}
		\caption{\begin{tiny}
		    Arched Eyebrows
		\end{tiny}}
    \end{subfigure}
     \begin{subfigure}{0.19\textwidth}
		\includegraphics[width=\textwidth]{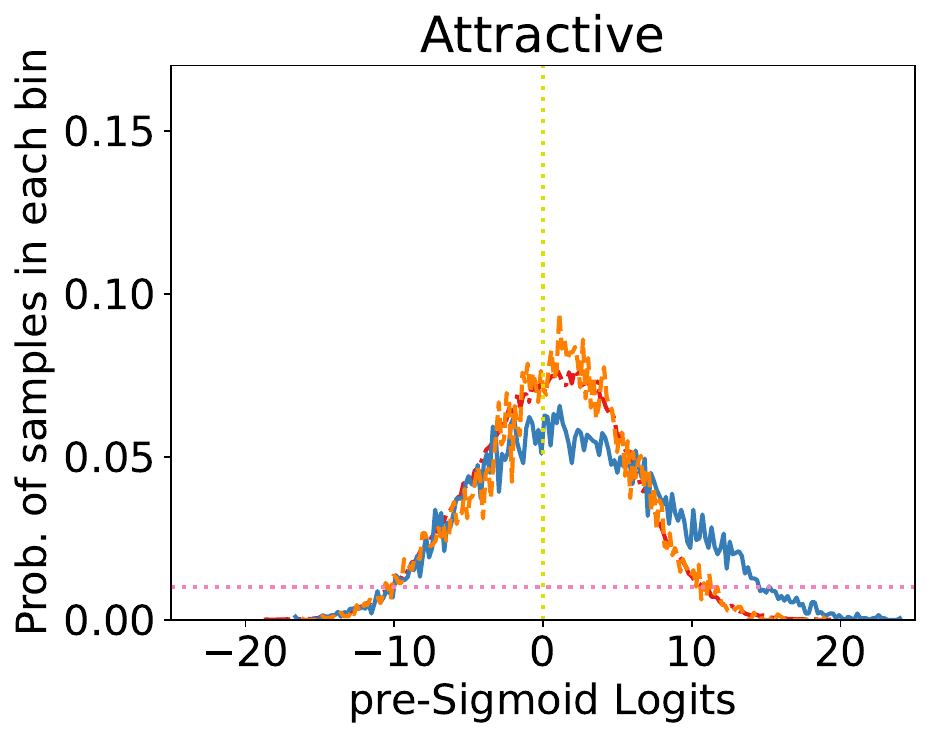}
		\caption{Attractive}
	    \end{subfigure}
    \begin{subfigure}{0.19\textwidth}
		\includegraphics[width=\textwidth]{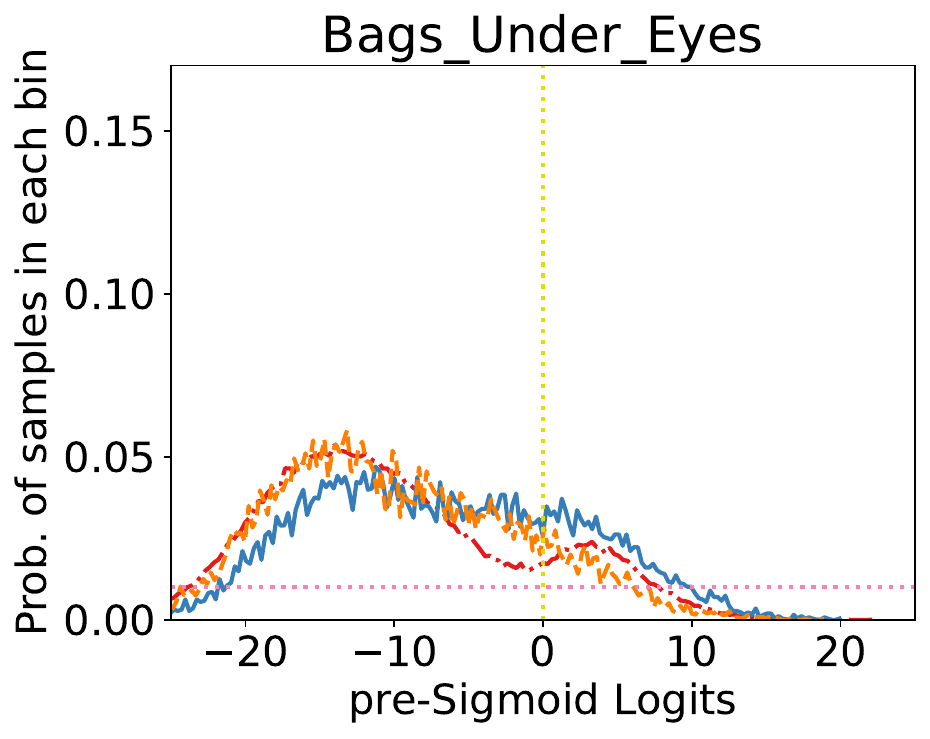}
		\caption{Bags Under Eyes}
	    \end{subfigure}
    \begin{subfigure}{0.19\textwidth}
		\includegraphics[width=\textwidth]{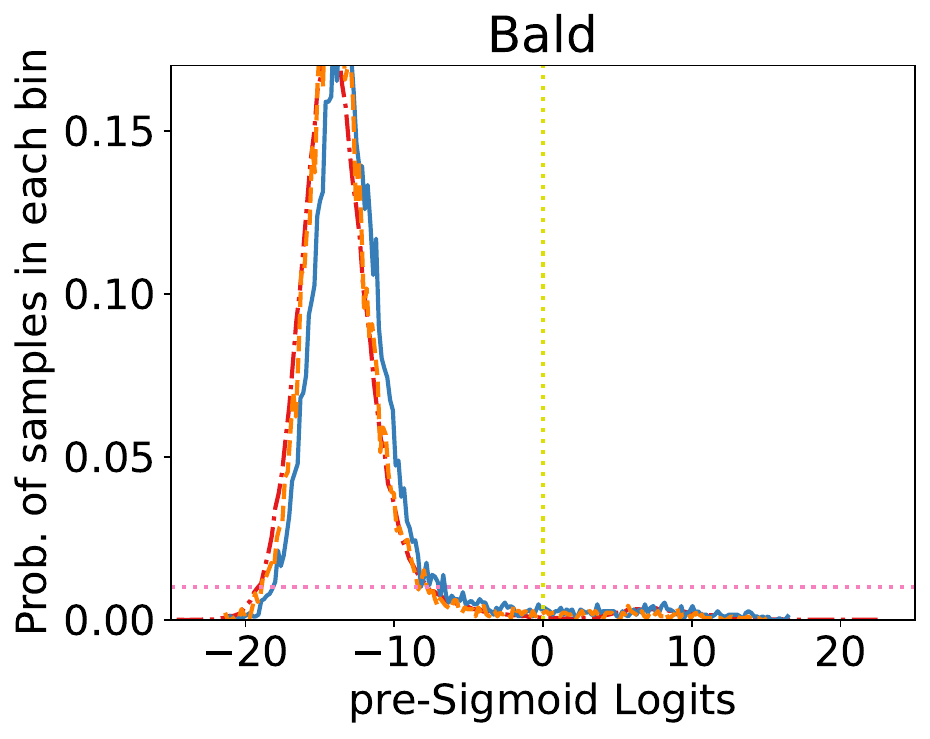}
		\caption{Bald}
    \end{subfigure}
     \vfill
    \begin{subfigure}{0.19\textwidth}
		\includegraphics[width=\textwidth]{Figs/cls_presigmoid_wolegend/Bangs.pdf}
		\caption{Bangs}
    \end{subfigure}
    \begin{subfigure}{0.19\textwidth}
		\includegraphics[width=\textwidth]{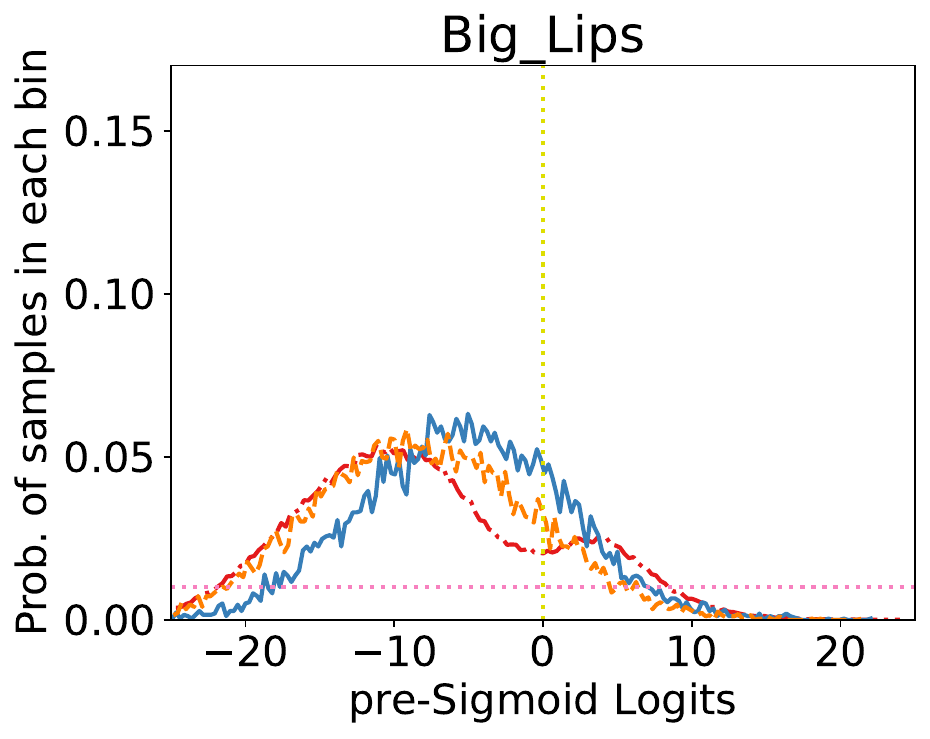}
		\caption{Big Lips}
    \end{subfigure}
     \begin{subfigure}{0.19\textwidth}
		\includegraphics[width=\textwidth]{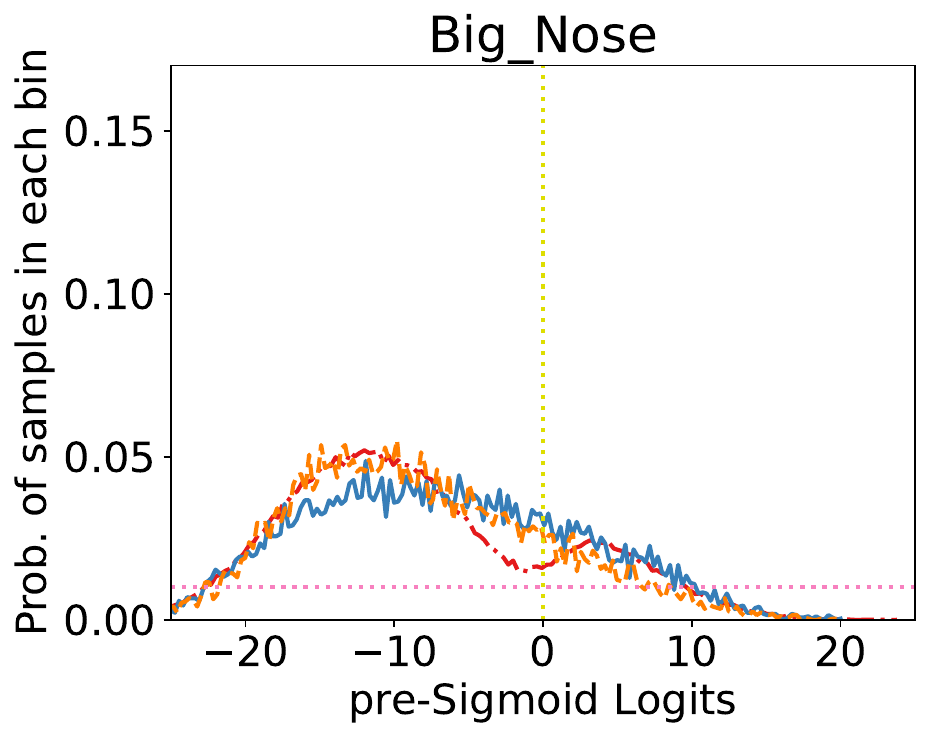}
		\caption{Big Nose}
	    \end{subfigure}
    \begin{subfigure}{0.19\textwidth}
		\includegraphics[width=\textwidth]{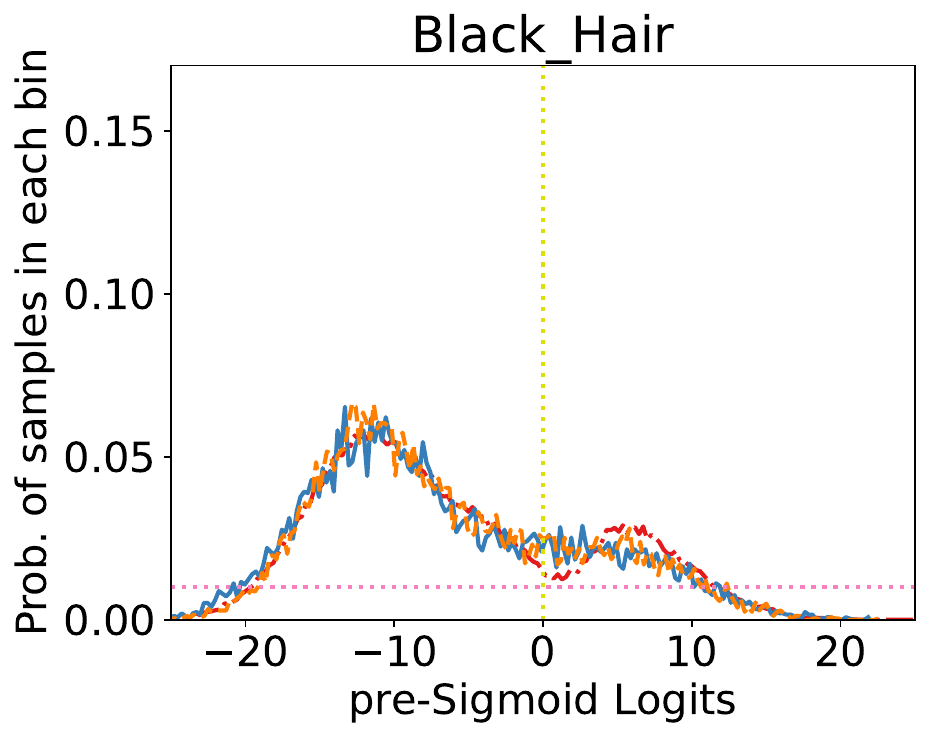}
		\caption{Black Hair}
	    \end{subfigure}
    \begin{subfigure}{0.19\textwidth}
		\includegraphics[width=\textwidth]{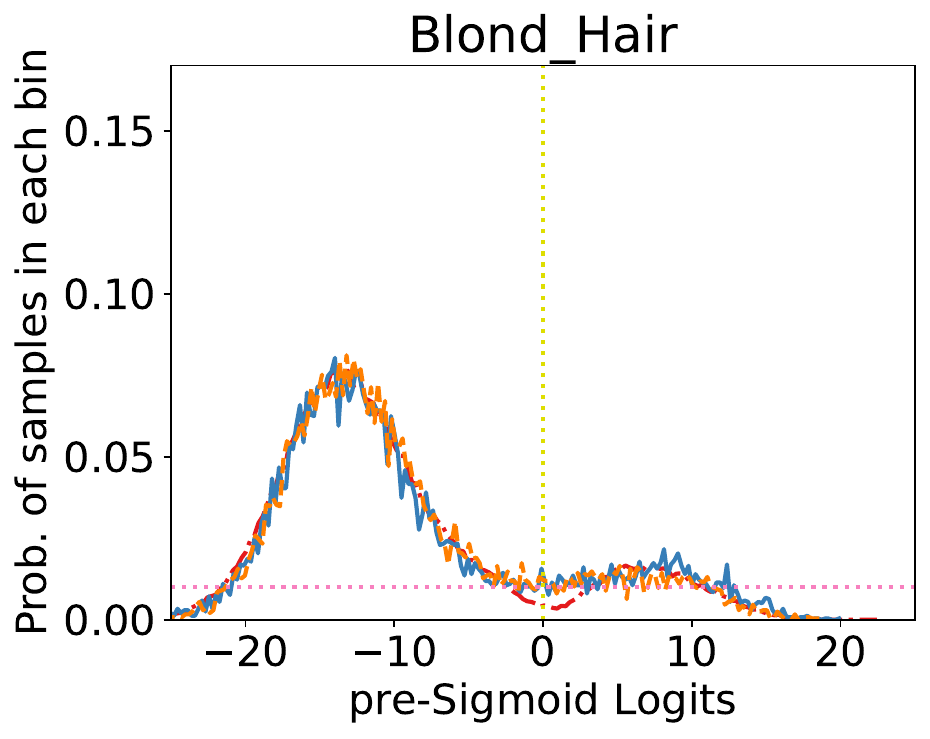}
		\caption{Blond Hair}
    \end{subfigure}
    \vfill
    \begin{subfigure}{0.19\textwidth}
		\includegraphics[width=\textwidth]{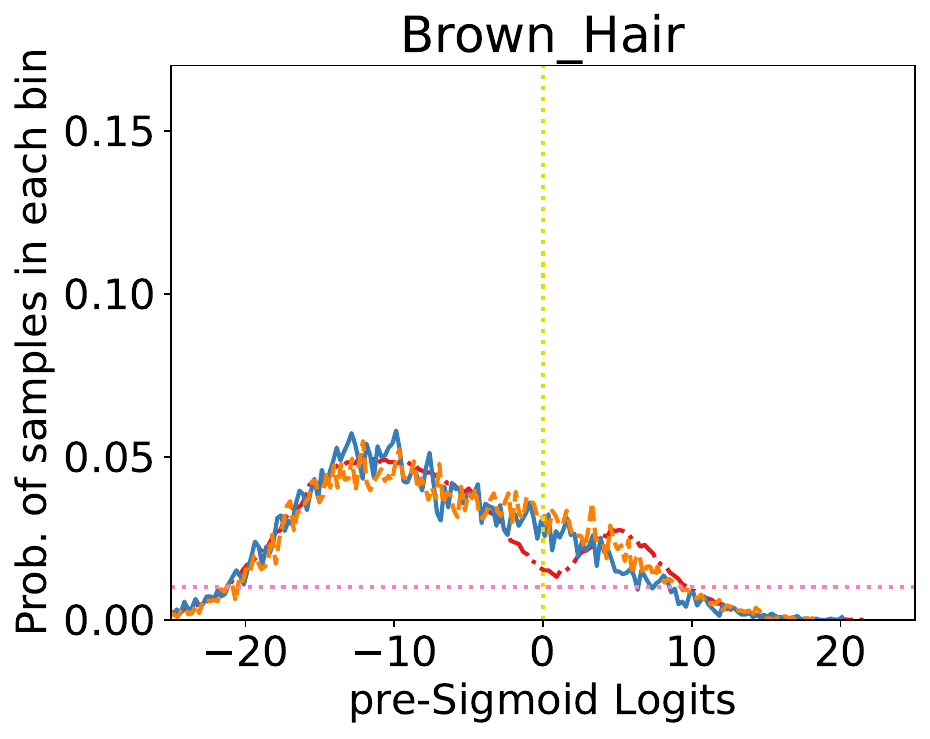}
		\caption{Brown Hair}
    \end{subfigure}
    \begin{subfigure}{0.19\textwidth}
		\includegraphics[width=\textwidth]{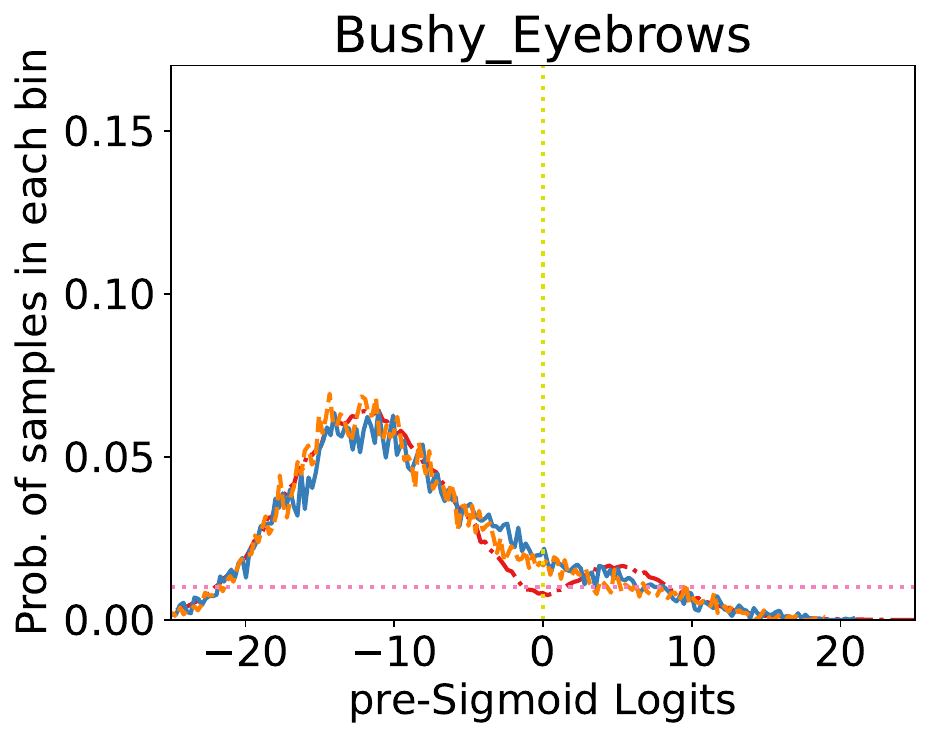}
		\caption{Bushy Eyebrows}
    \end{subfigure}
     \begin{subfigure}{0.19\textwidth}
		\includegraphics[width=\textwidth]{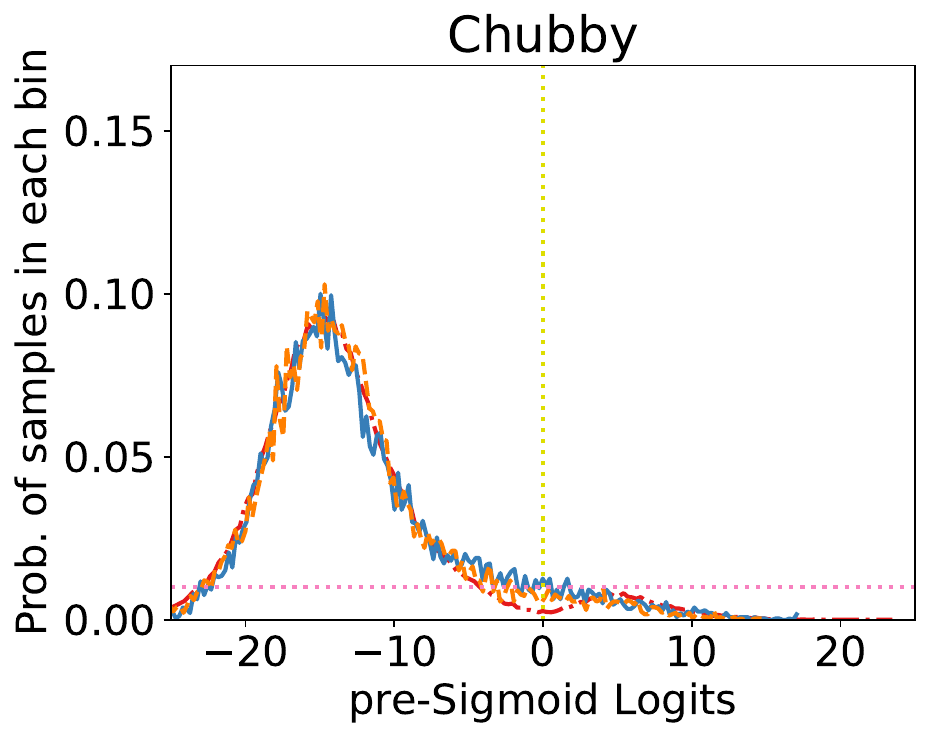}
		\caption{Chubby}
	    \end{subfigure}
    \begin{subfigure}{0.19\textwidth}
		\includegraphics[width=\textwidth]{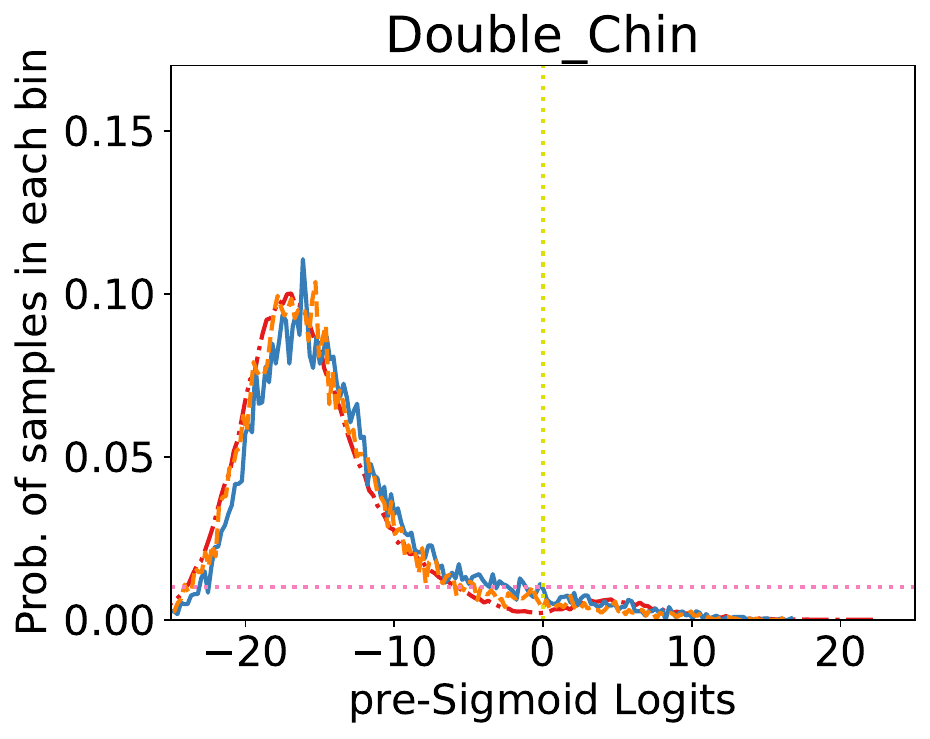}
		\caption{Double Chin}
	    \end{subfigure}
    \begin{subfigure}{0.19\textwidth}
		\includegraphics[width=\textwidth]{Figs/cls_presigmoid_wolegend/Eyeglasses.pdf}
		\caption{Eyeglasses}
    \end{subfigure}
    \vfill
    \begin{subfigure}{0.19\textwidth}
		\includegraphics[width=\textwidth]{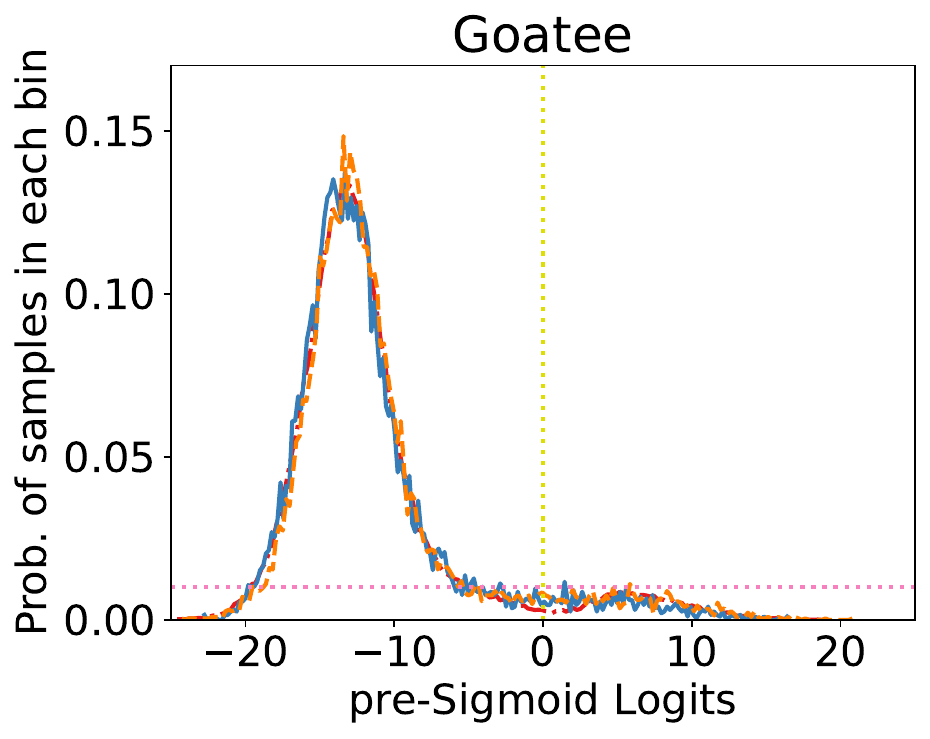}
		\caption{Goatee}
    \end{subfigure}
    \begin{subfigure}{0.19\textwidth}
		\includegraphics[width=\textwidth]{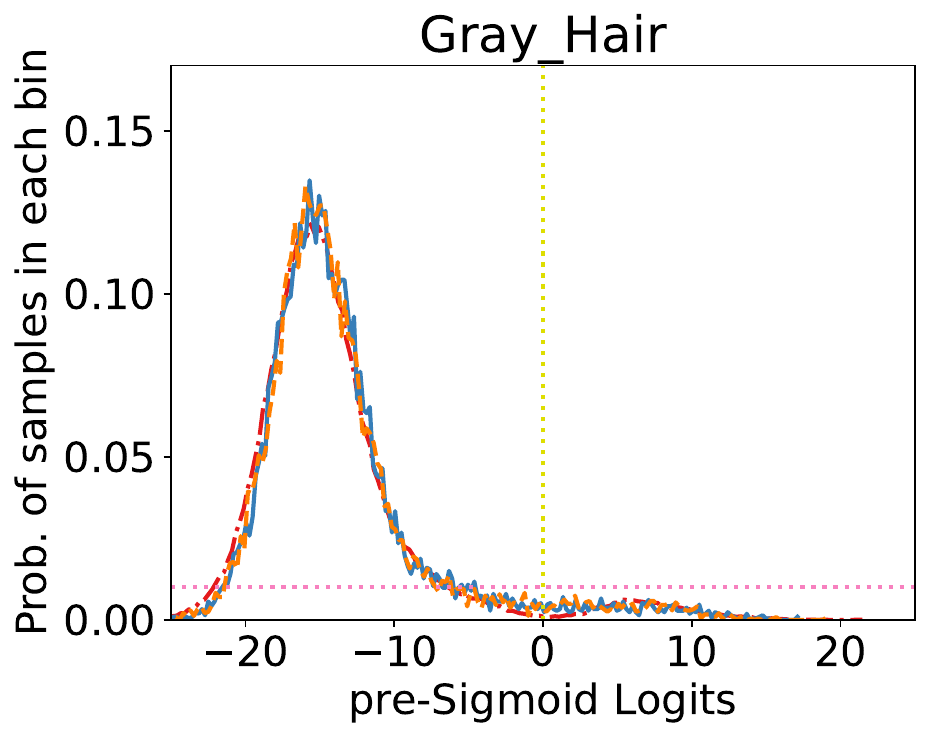}
		\caption{Gray Hair}
    \end{subfigure}
     \begin{subfigure}{0.19\textwidth}
		\includegraphics[width=\textwidth]{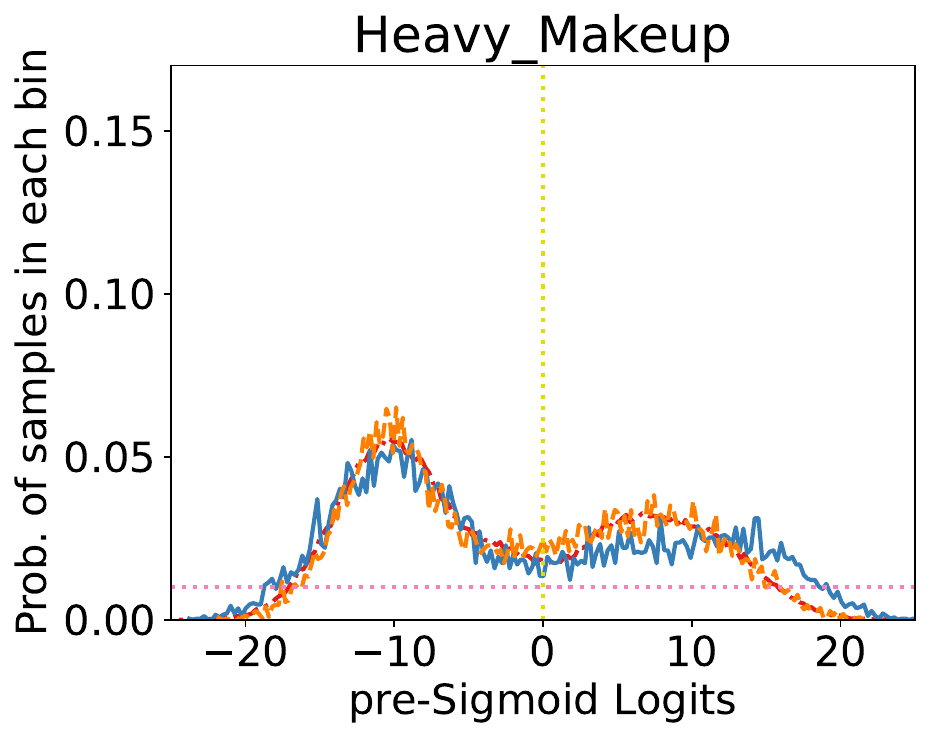}
		\caption{Heavy Makeup}
	    \end{subfigure}
    \begin{subfigure}{0.19\textwidth}
		\includegraphics[width=\textwidth]{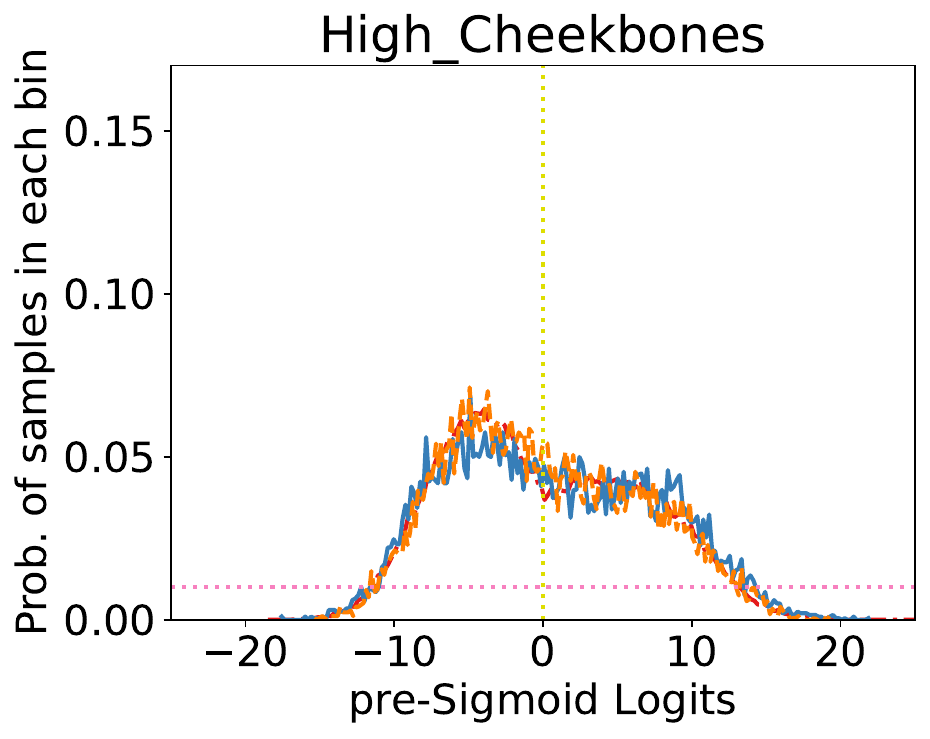}
		\caption{High Cheekbones}
	    \end{subfigure}
    \begin{subfigure}{0.19\textwidth}
		\includegraphics[width=\textwidth]{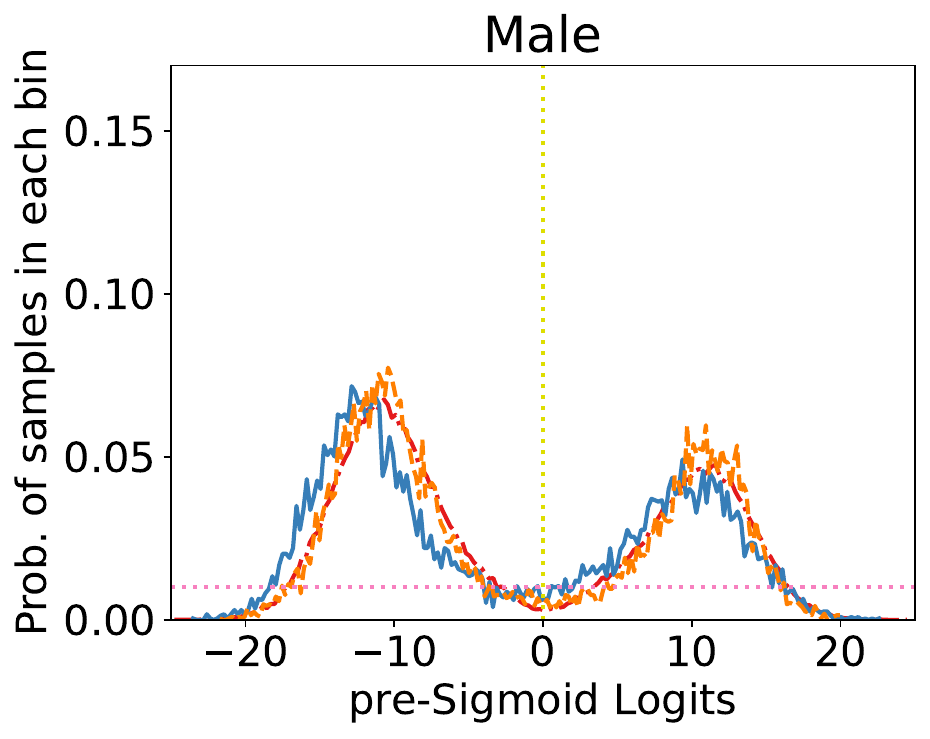}
		\caption{Male}
    \end{subfigure}
    \vfill
    \begin{subfigure}{0.19\textwidth}
		\includegraphics[width=\textwidth]{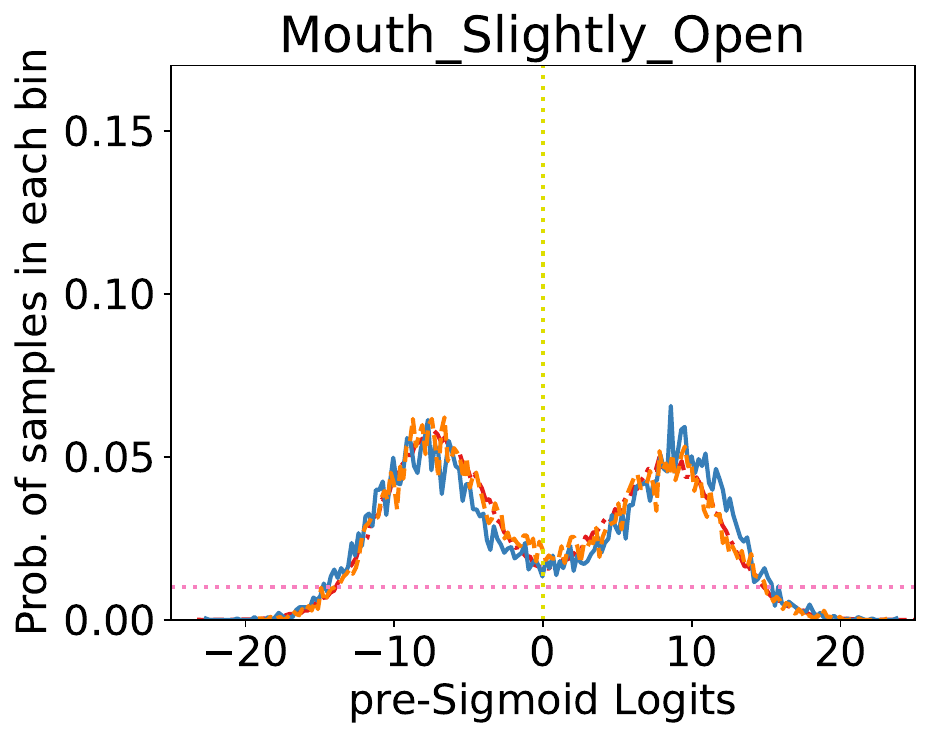}
		\caption{\begin{tiny}
		    Mouth Slightly Open
		\end{tiny}}
    \end{subfigure}
    \begin{subfigure}{0.19\textwidth}
		\includegraphics[width=\textwidth]{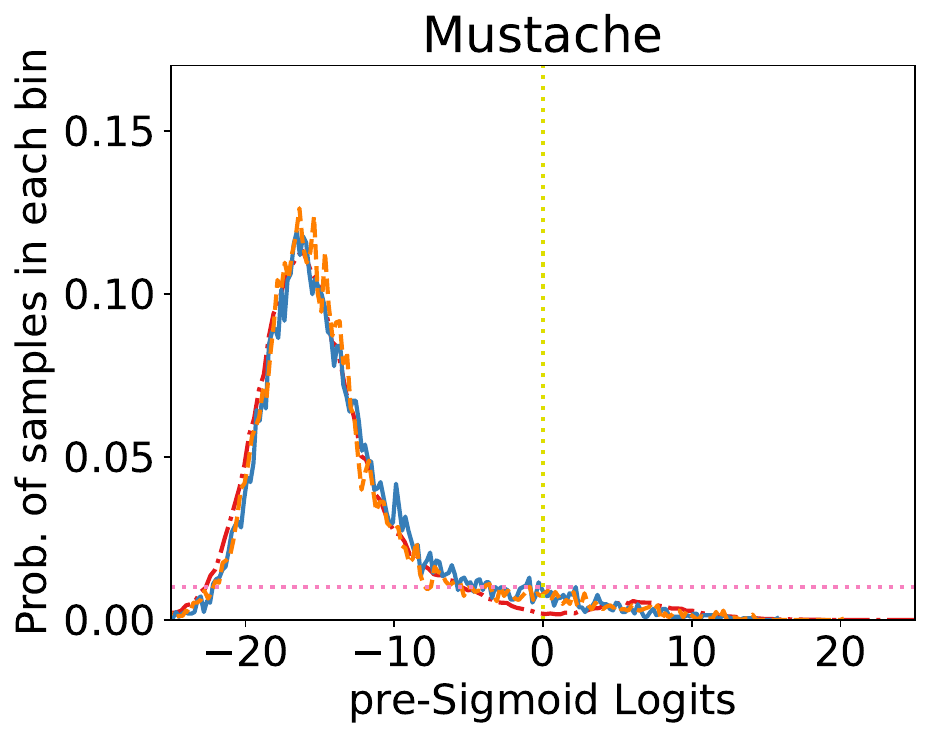}
		\caption{Mustache}
    \end{subfigure}
     \begin{subfigure}{0.19\textwidth}
		\includegraphics[width=\textwidth]{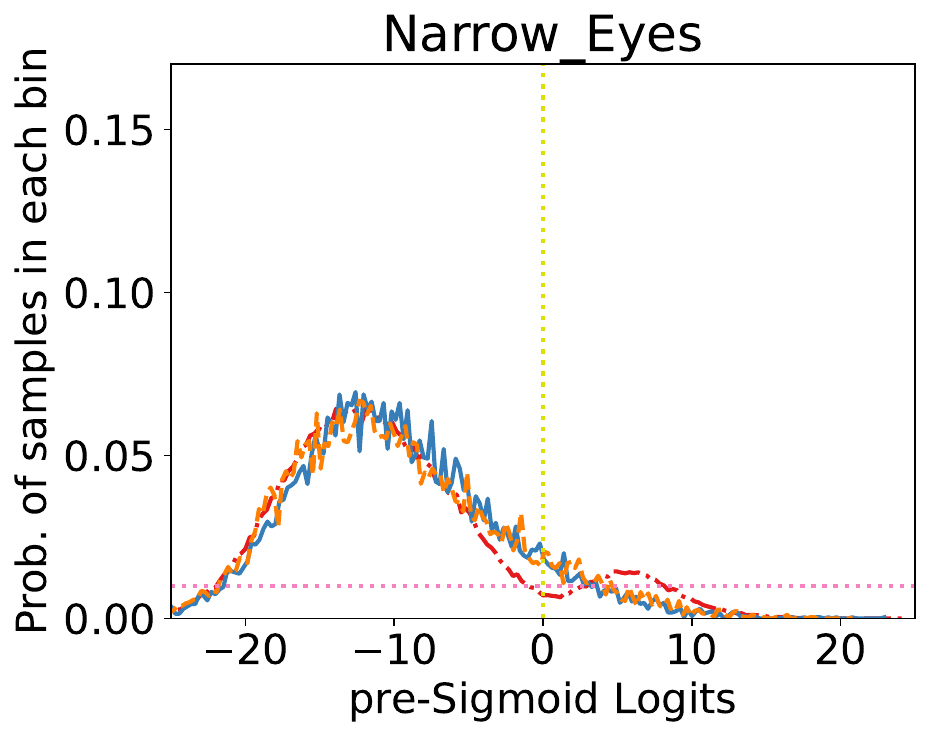}
		\caption{Narrow Eyes}
	    \end{subfigure}
    \begin{subfigure}{0.19\textwidth}
		\includegraphics[width=\textwidth]{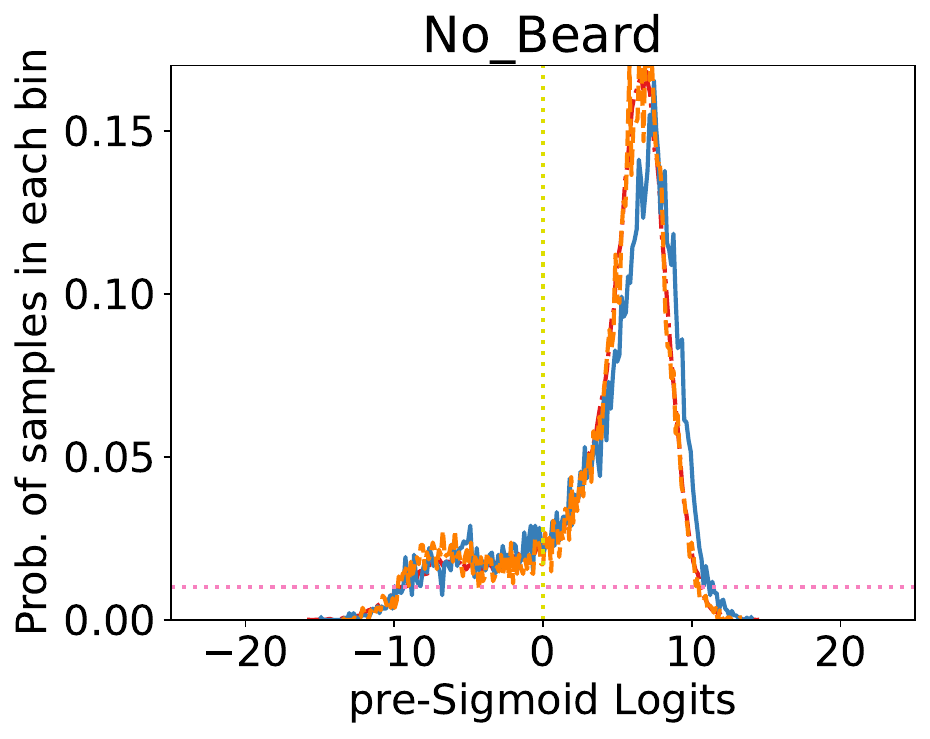}
		\caption{No Beard}
	    \end{subfigure}
    \begin{subfigure}{0.19\textwidth}
		\includegraphics[width=\textwidth]{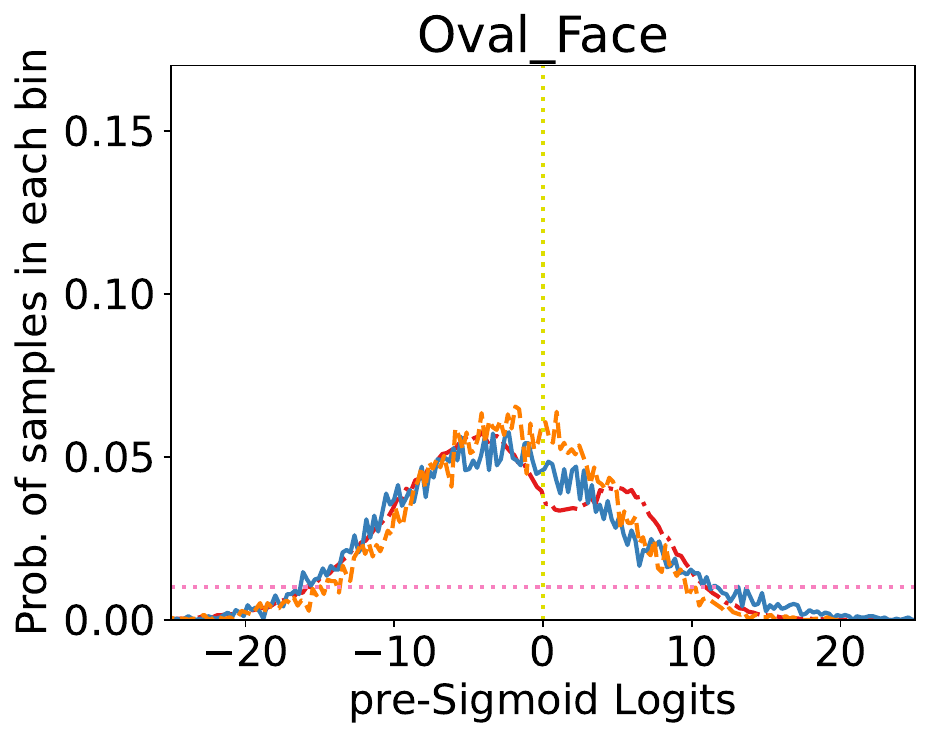}
		\caption{Oval Face}
    \end{subfigure}
    \vfill
    \begin{subfigure}{0.19\textwidth}
		\includegraphics[width=\textwidth]{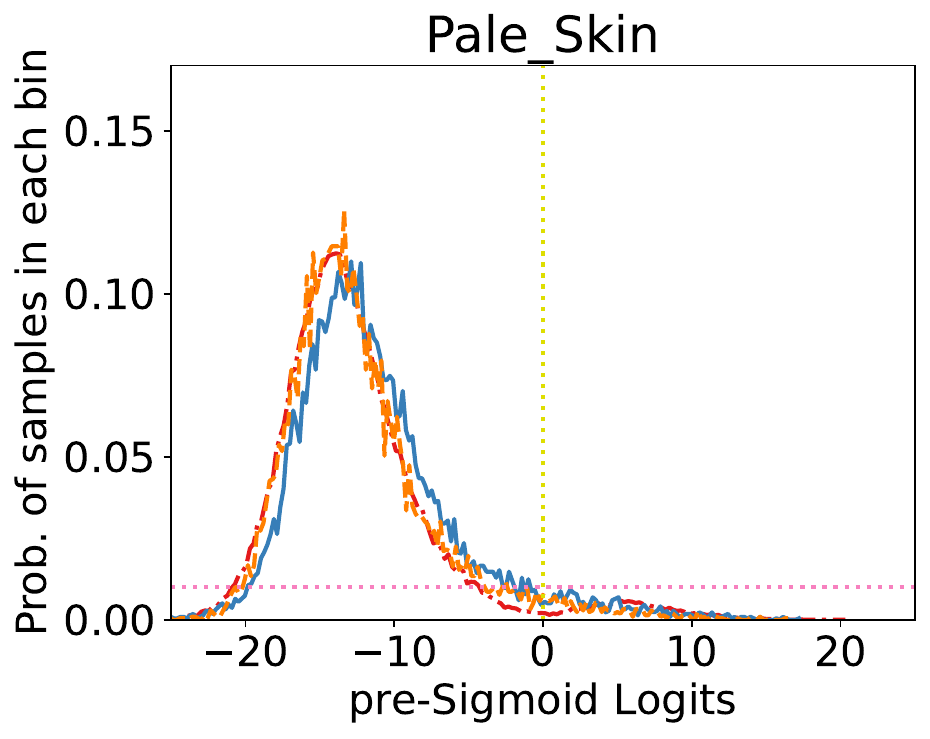}
		\caption{Pale Skin}
    \end{subfigure}
    \begin{subfigure}{0.19\textwidth}
		\includegraphics[width=\textwidth]{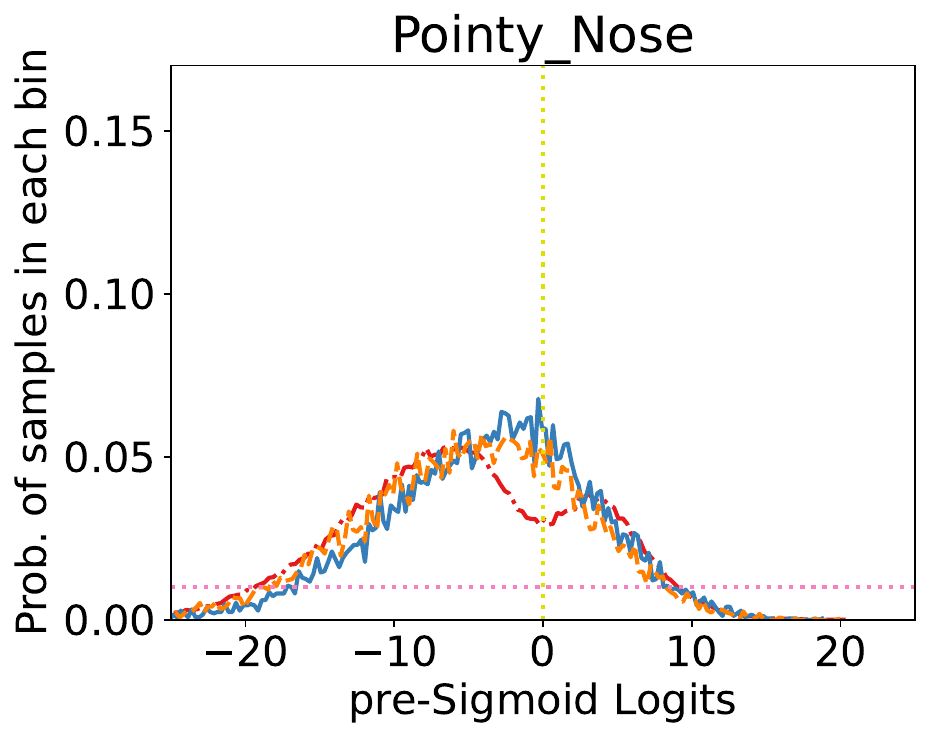}
		\caption{Pointy Nose}
    \end{subfigure}
     \begin{subfigure}{0.19\textwidth}
		\includegraphics[width=\textwidth]{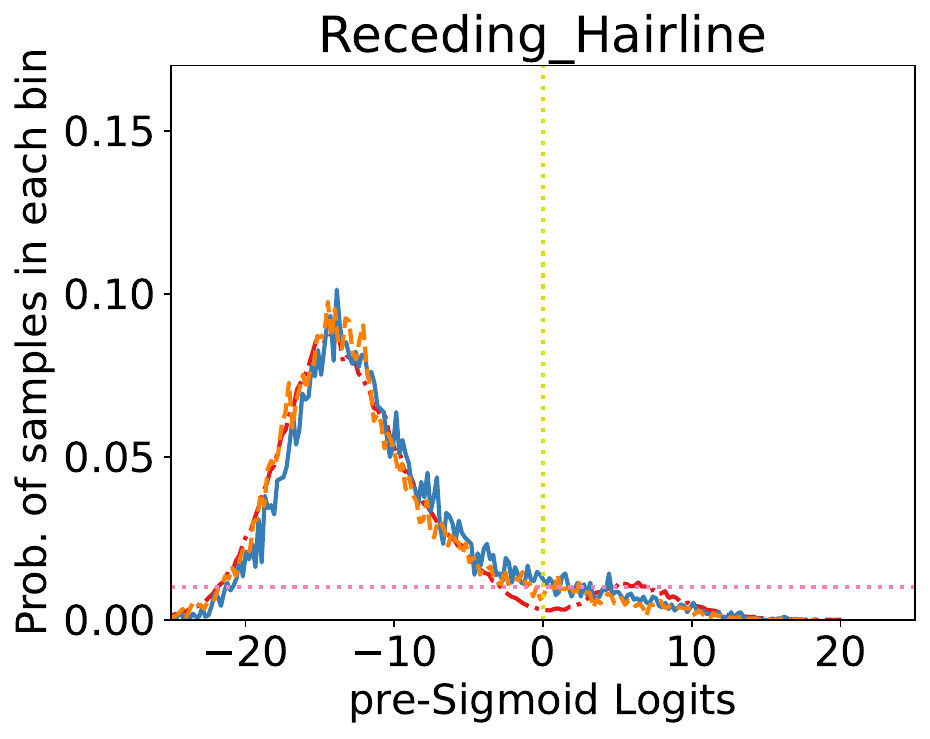}
		\caption{\begin{tiny}
		    Receding Hairline
		\end{tiny}}
	    \end{subfigure}
    \begin{subfigure}{0.19\textwidth}
		\includegraphics[width=\textwidth]{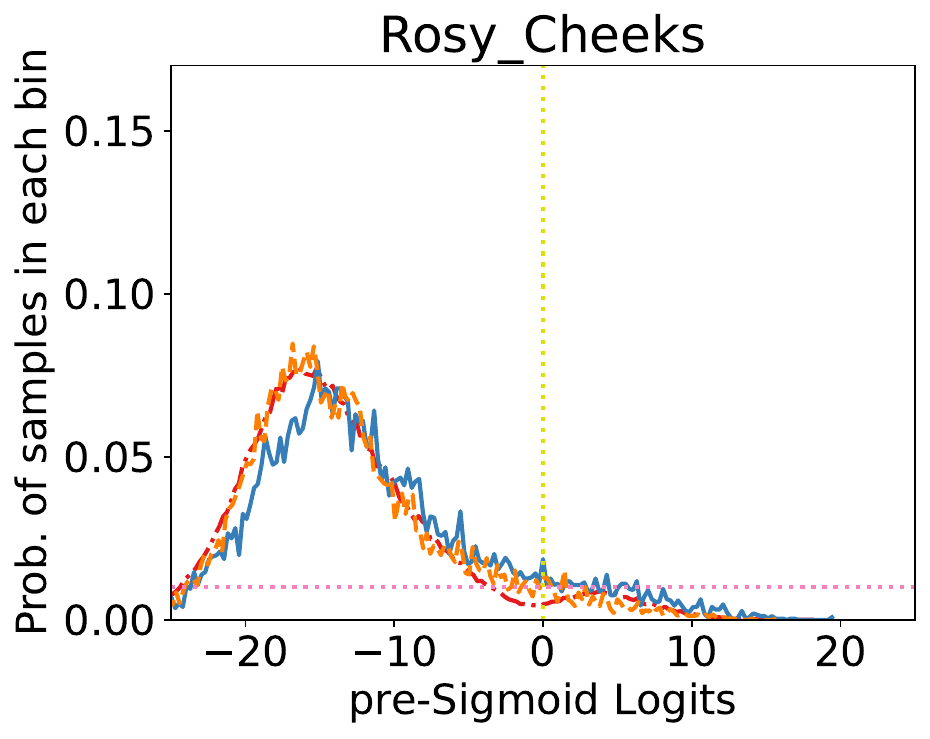}
		\caption{Rosy Cheeks}
	    \end{subfigure}
    \begin{subfigure}{0.19\textwidth}
		\includegraphics[width=\textwidth]{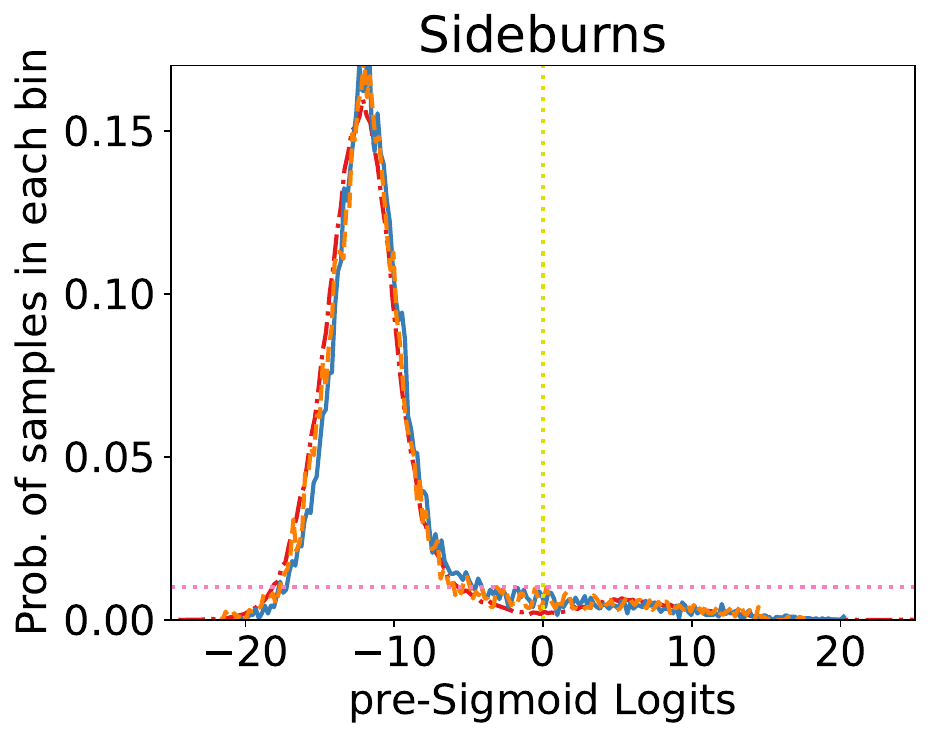}
		\caption{Sideburns}
    \end{subfigure}
    \vfill
    \begin{subfigure}{0.19\textwidth}
		\includegraphics[width=\textwidth]{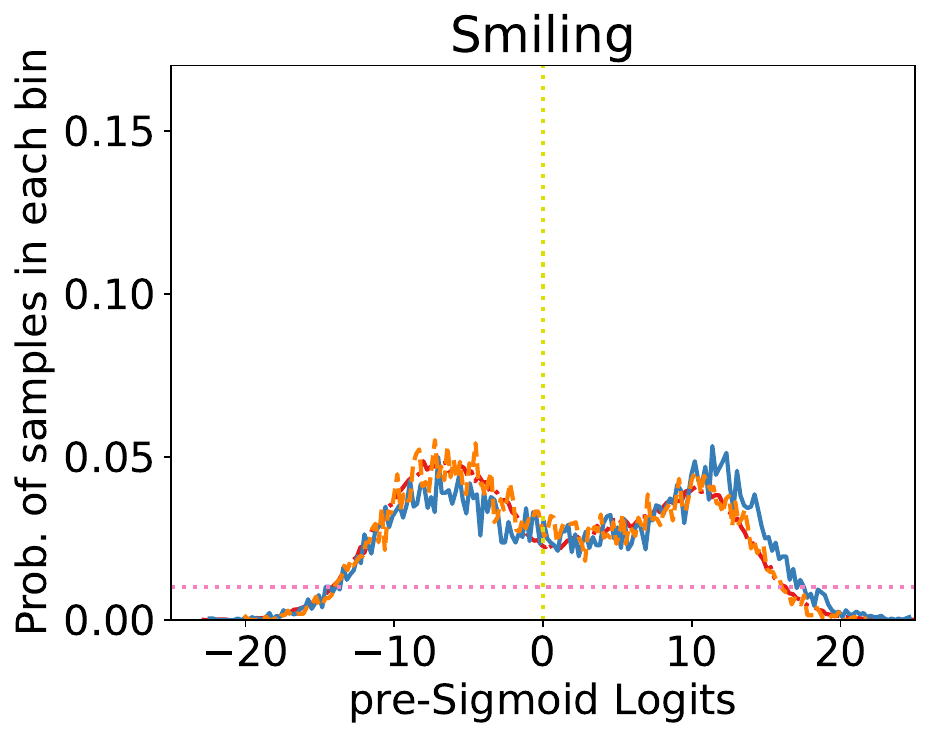}
		\caption{Smiling}
    \end{subfigure}
    \begin{subfigure}{0.19\textwidth}
		\includegraphics[width=\textwidth]{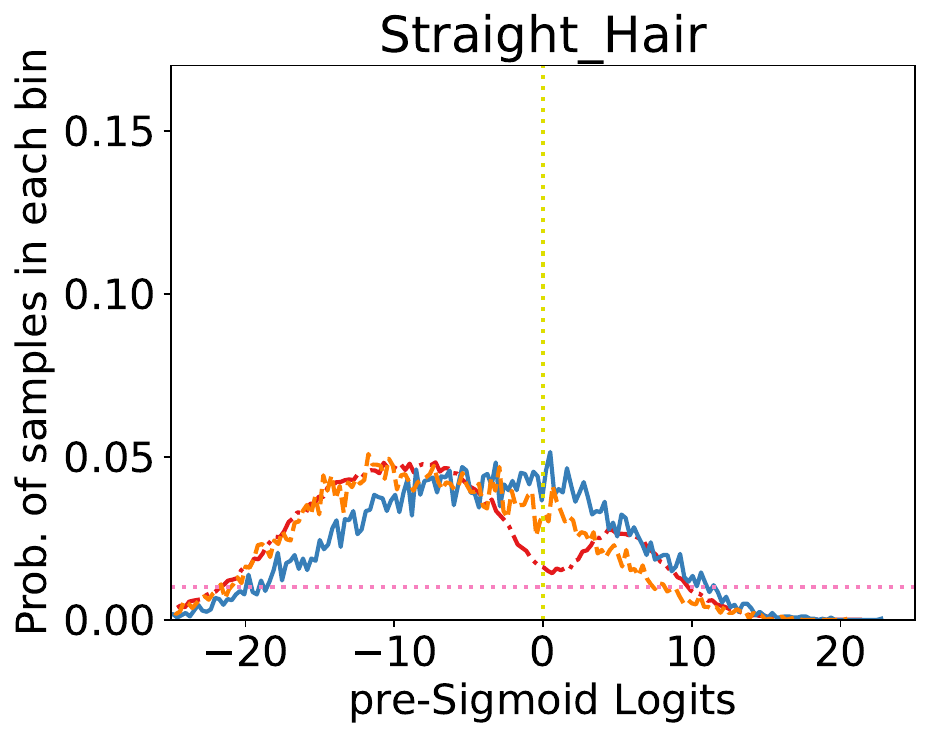}
		\caption{Straight hair}
    \end{subfigure}
     \begin{subfigure}{0.19\textwidth}
		\includegraphics[width=\textwidth]{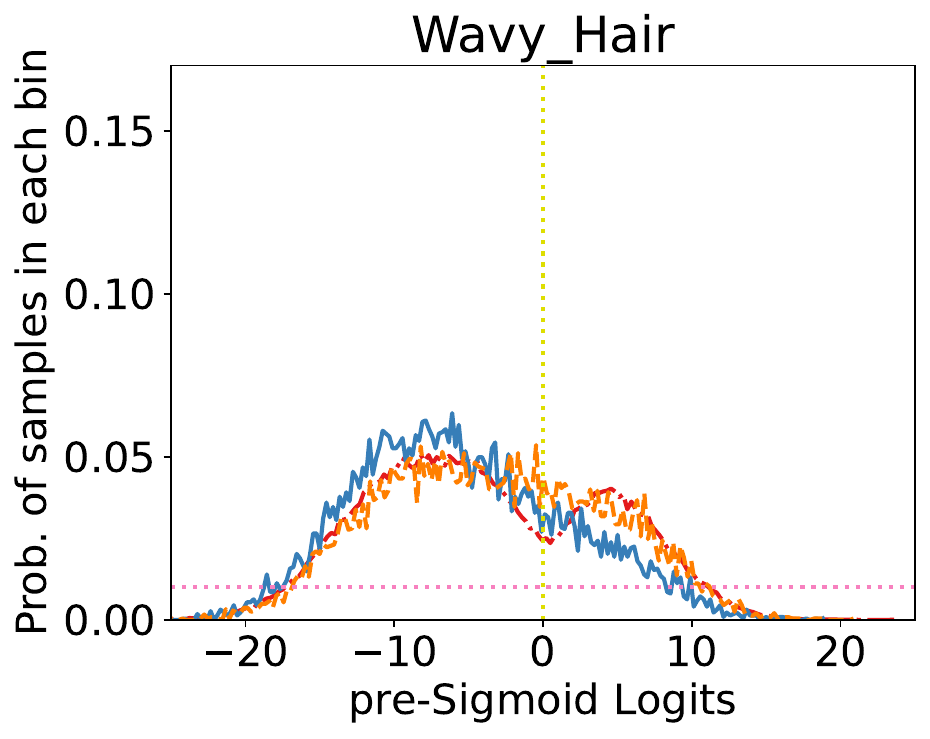}
		\caption{Wavy Hair}
	    \end{subfigure}
    \begin{subfigure}{0.19\textwidth}
		\includegraphics[width=\textwidth]{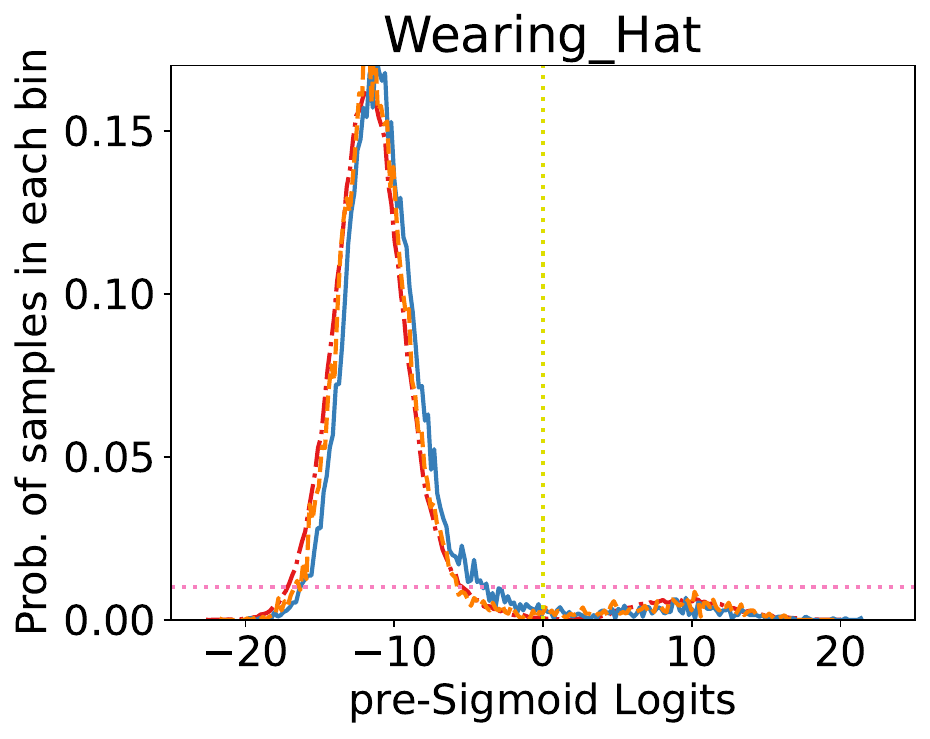}
		\caption{Wearing Hat}
    \end{subfigure}
    \begin{subfigure}{0.19\textwidth}
		\includegraphics[width=\textwidth]{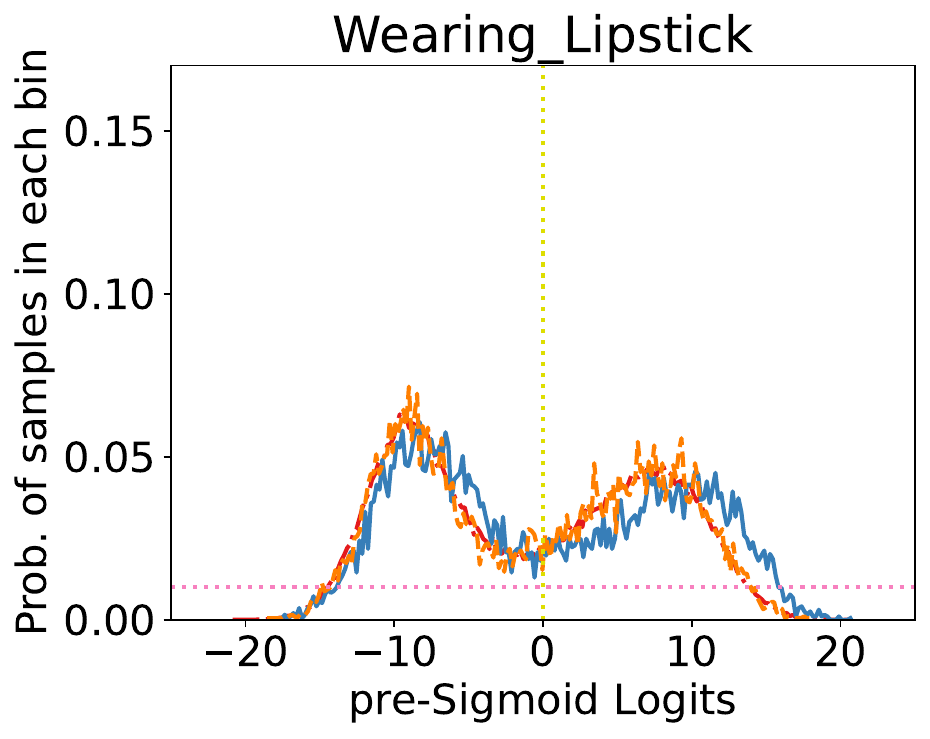}
		\caption{\begin{tiny}
		    Wearing Lipstick
		\end{tiny}}
    \end{subfigure}
    \vfill
    \begin{subfigure}{0.19\textwidth}
		\includegraphics[width=\textwidth]{Figs/cls_presigmoid_wolegend/Young.pdf}
		\caption{Young}
    \end{subfigure}
	\caption{The pre-sigmoid logits distribution of each attribute in CelebA (ResNeXt-based classifier).}
	\label{fig:attr_dist_celeba_all}
\end{figure}

\begin{figure}[tbp]
    \centering
    \begin{subfigure}{0.19\textwidth}
		\includegraphics[width=\textwidth]{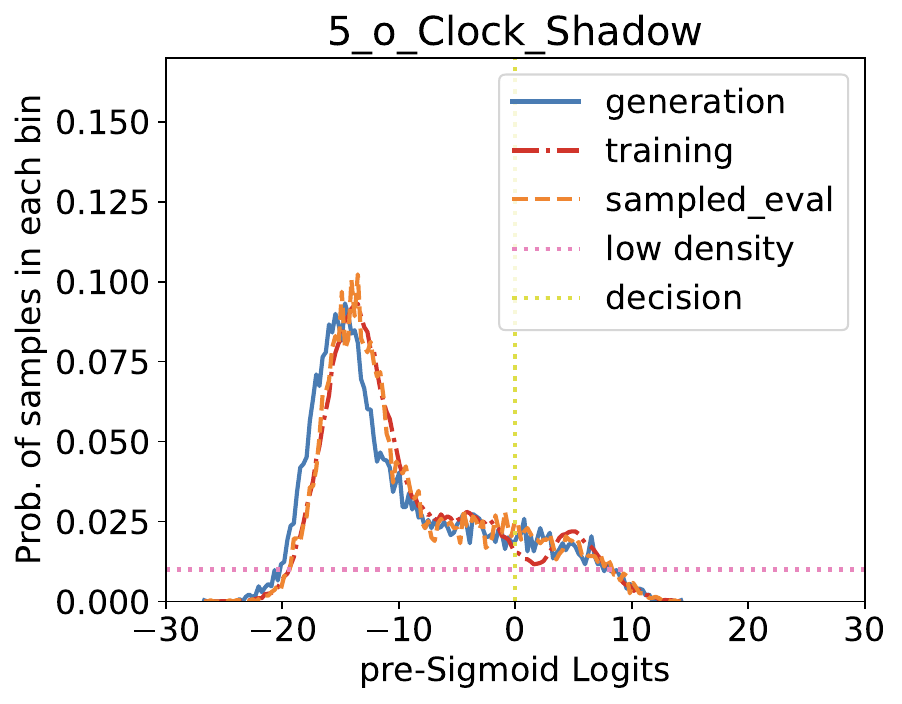}
		\caption{5o Clock Shadow}
    \end{subfigure}
    \begin{subfigure}{0.19\textwidth}
		\includegraphics[width=\textwidth]{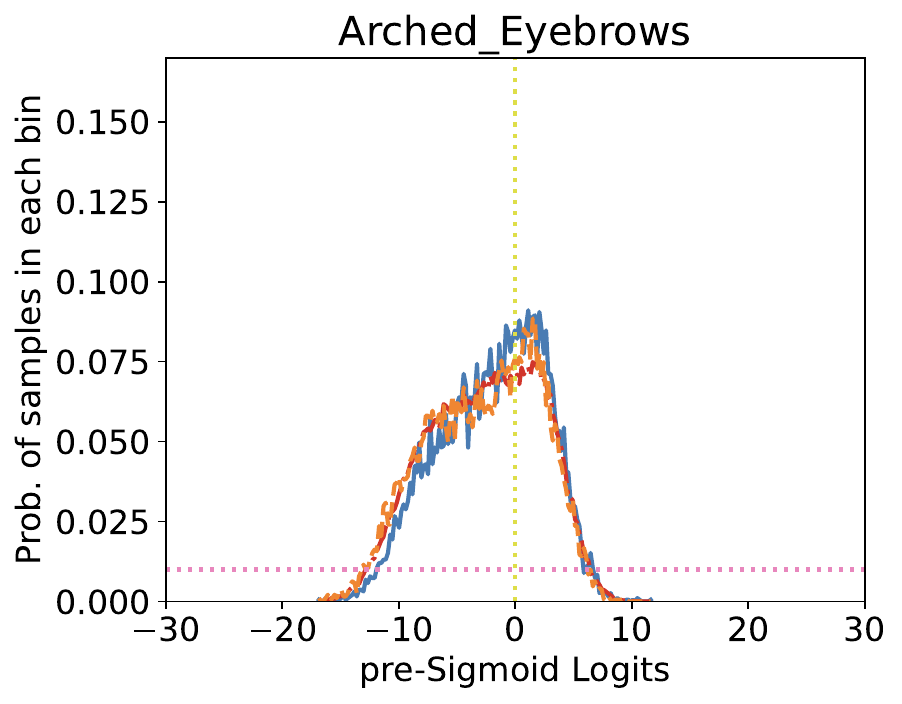}
		\caption{\begin{tiny}
		    Arched Eyebrows
		\end{tiny}}
    \end{subfigure}
     \begin{subfigure}{0.19\textwidth}
		\includegraphics[width=\textwidth]{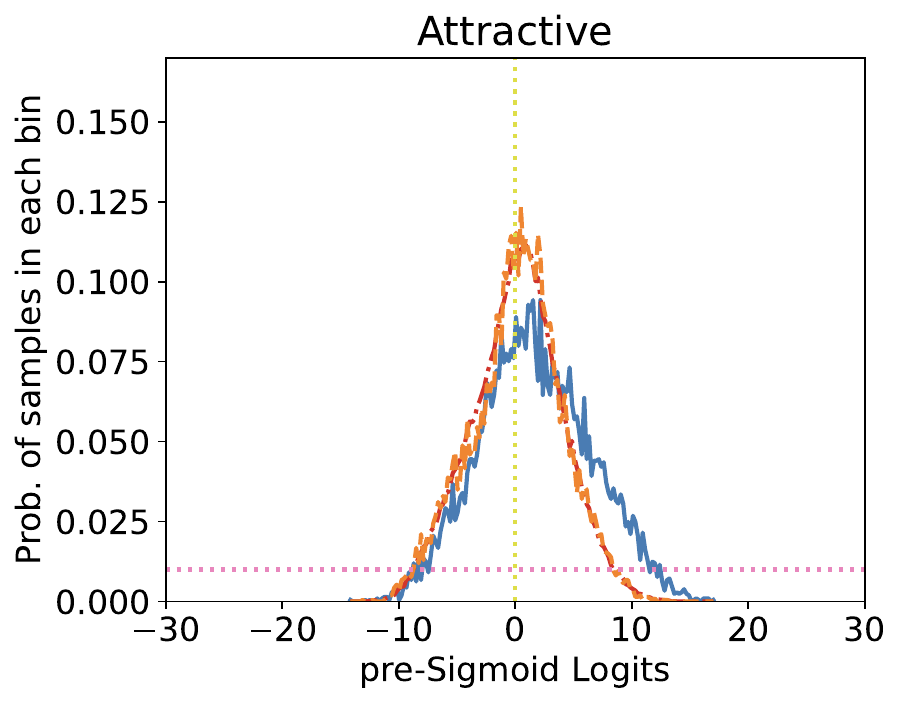}
		\caption{Attractive}
	    \end{subfigure}
    \begin{subfigure}{0.19\textwidth}
		\includegraphics[width=\textwidth]{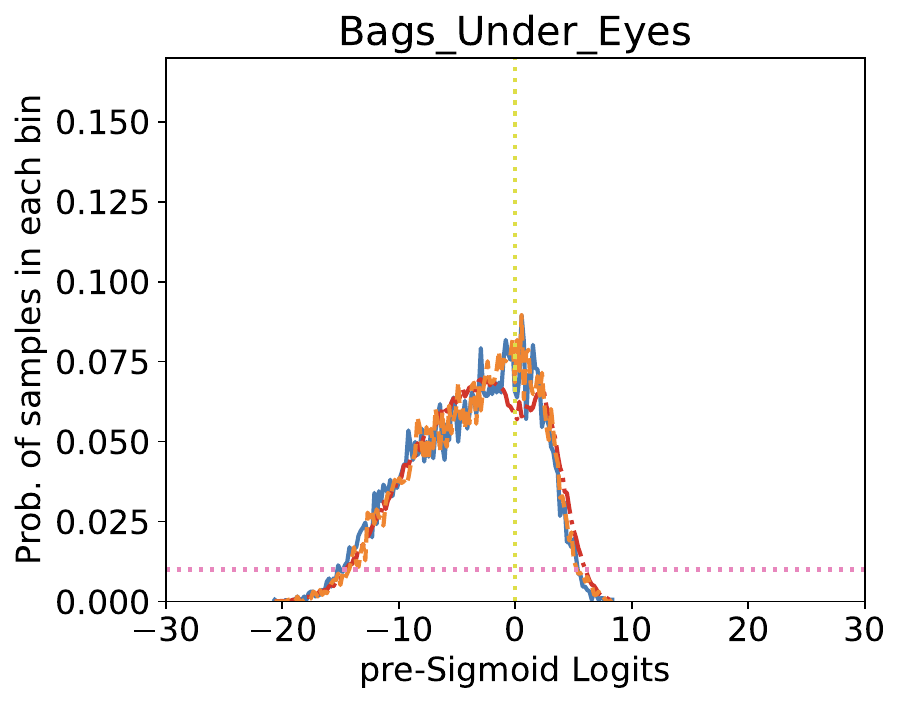}
		\caption{Bags Under Eyes}
	    \end{subfigure}
    \begin{subfigure}{0.19\textwidth}
		\includegraphics[width=\textwidth]{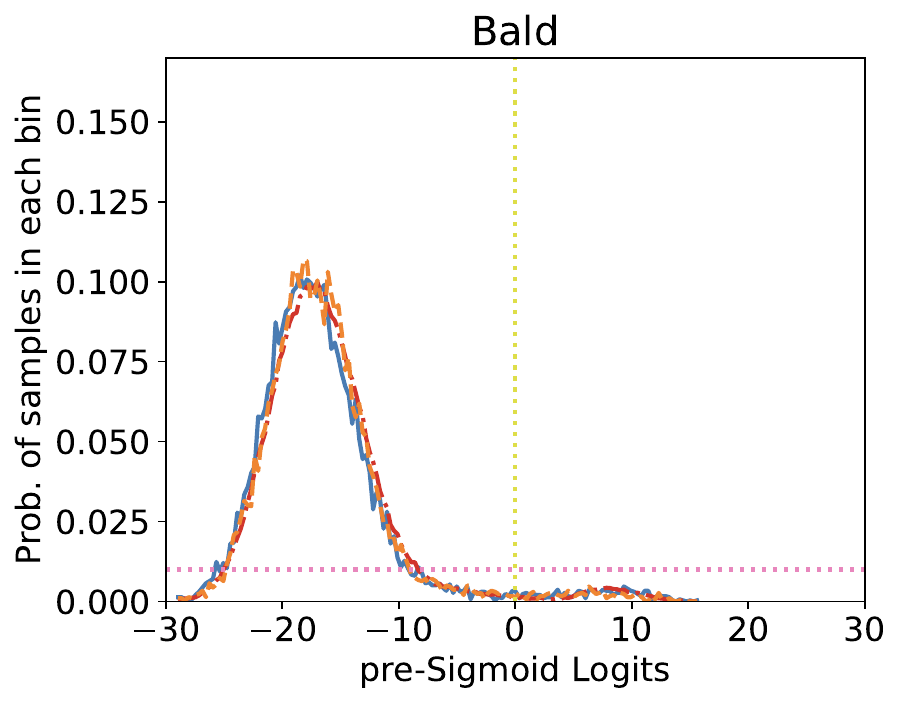}
		\caption{Bald}
    \end{subfigure}
     \vfill
    \begin{subfigure}{0.19\textwidth}
		\includegraphics[width=\textwidth]{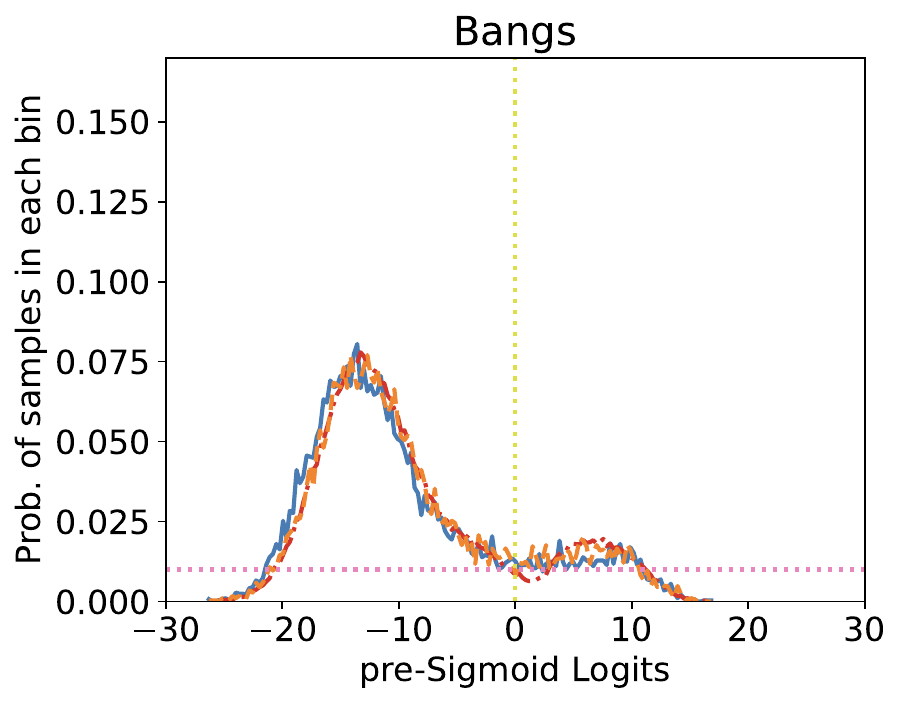}
		\caption{Bangs}
    \end{subfigure}
    \begin{subfigure}{0.19\textwidth}
		\includegraphics[width=\textwidth]{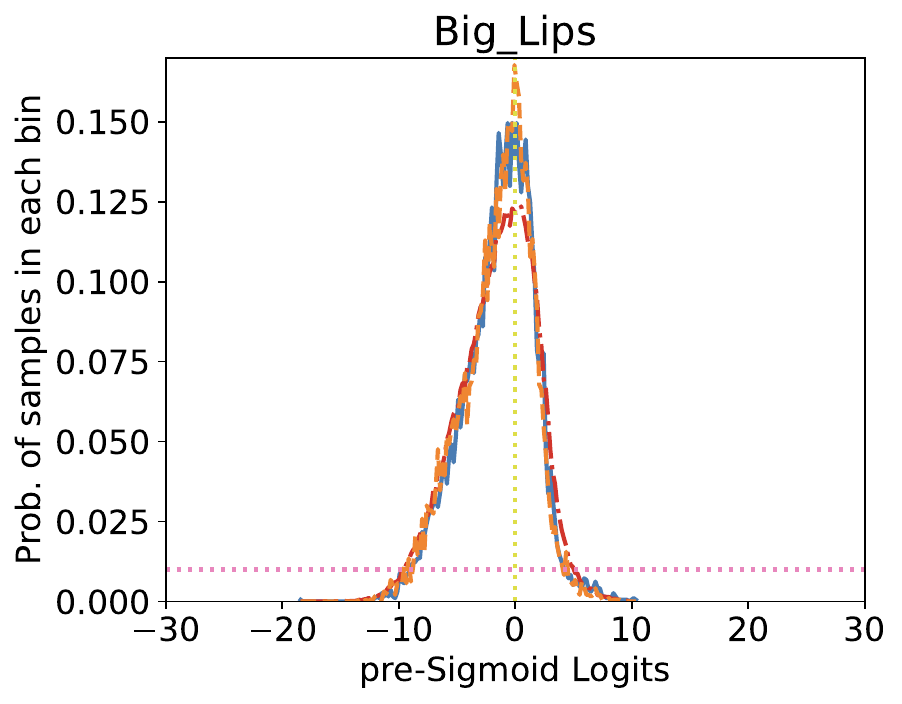}
		\caption{Big Lips}
    \end{subfigure}
     \begin{subfigure}{0.19\textwidth}
		\includegraphics[width=\textwidth]{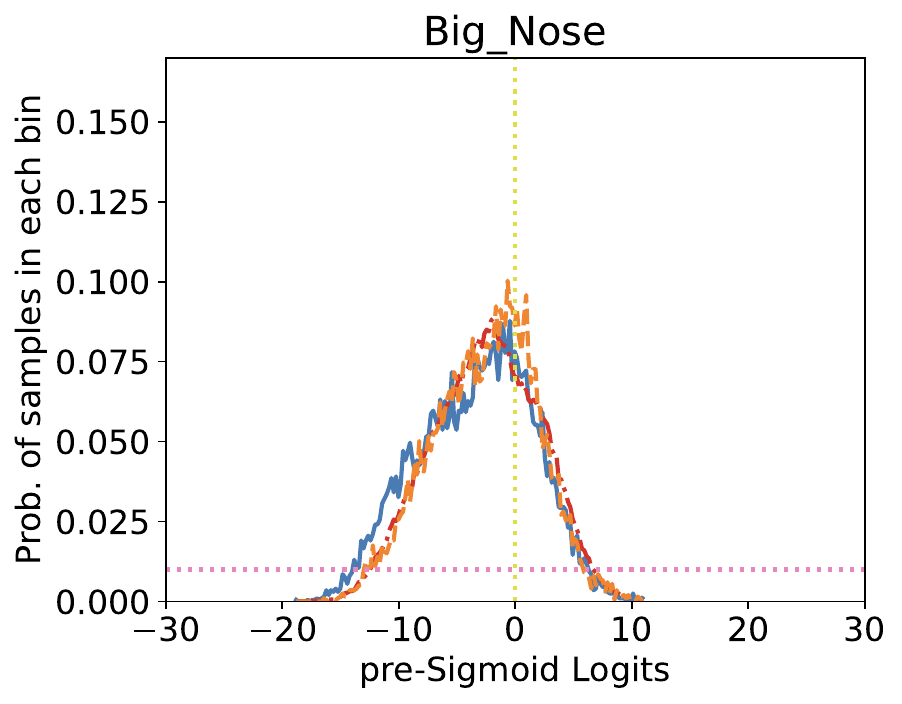}
		\caption{Big Nose}
	    \end{subfigure}
    \begin{subfigure}{0.19\textwidth}
		\includegraphics[width=\textwidth]{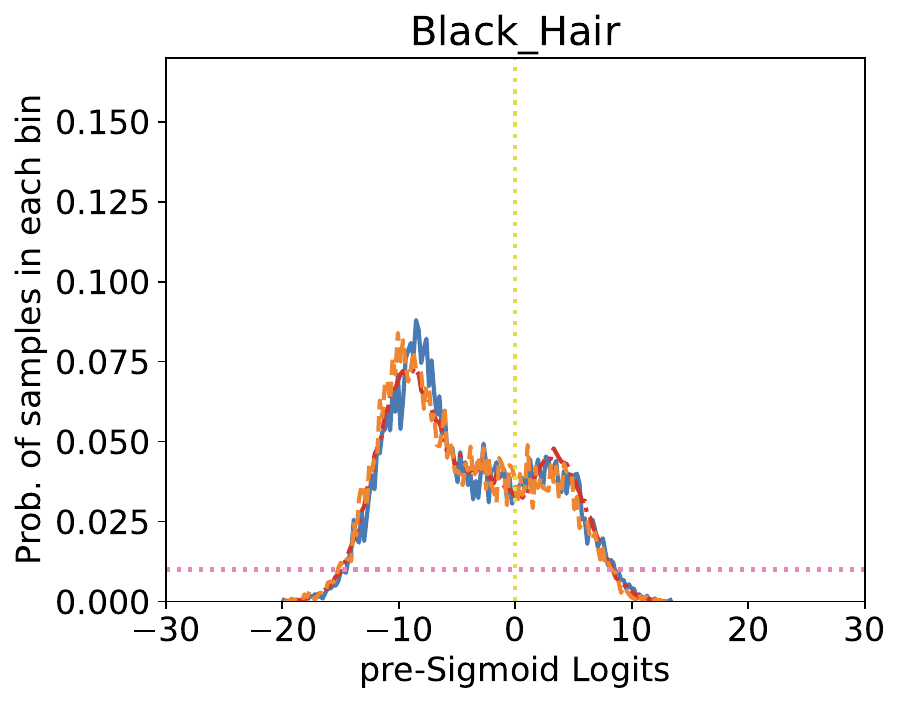}
		\caption{Black Hair}
	    \end{subfigure}
    \begin{subfigure}{0.19\textwidth}
		\includegraphics[width=\textwidth]{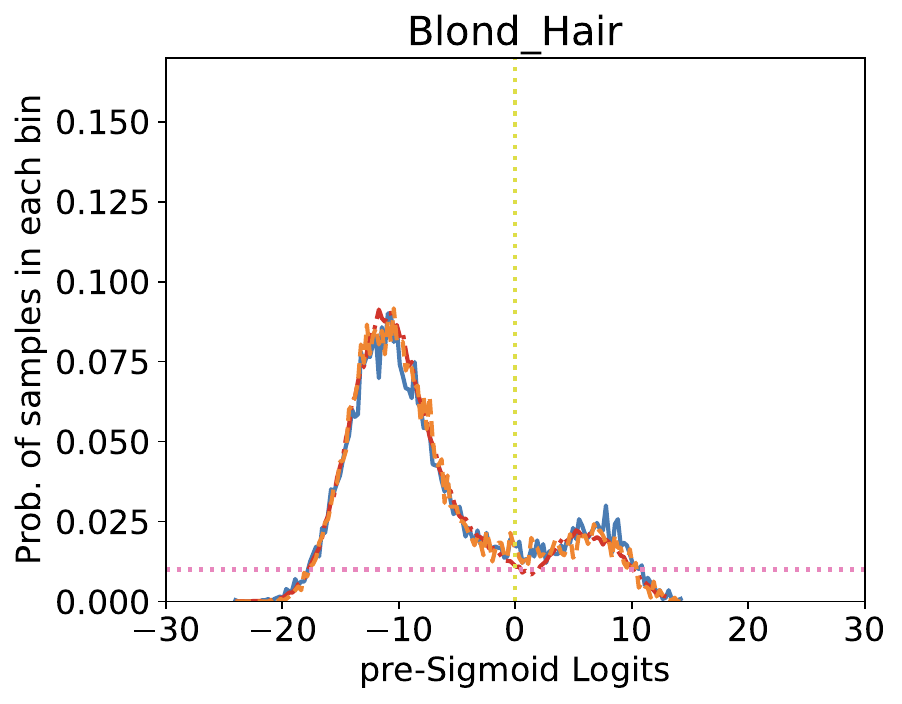}
		\caption{Blond Hair}
    \end{subfigure}
    \vfill
    \begin{subfigure}{0.19\textwidth}
		\includegraphics[width=\textwidth]{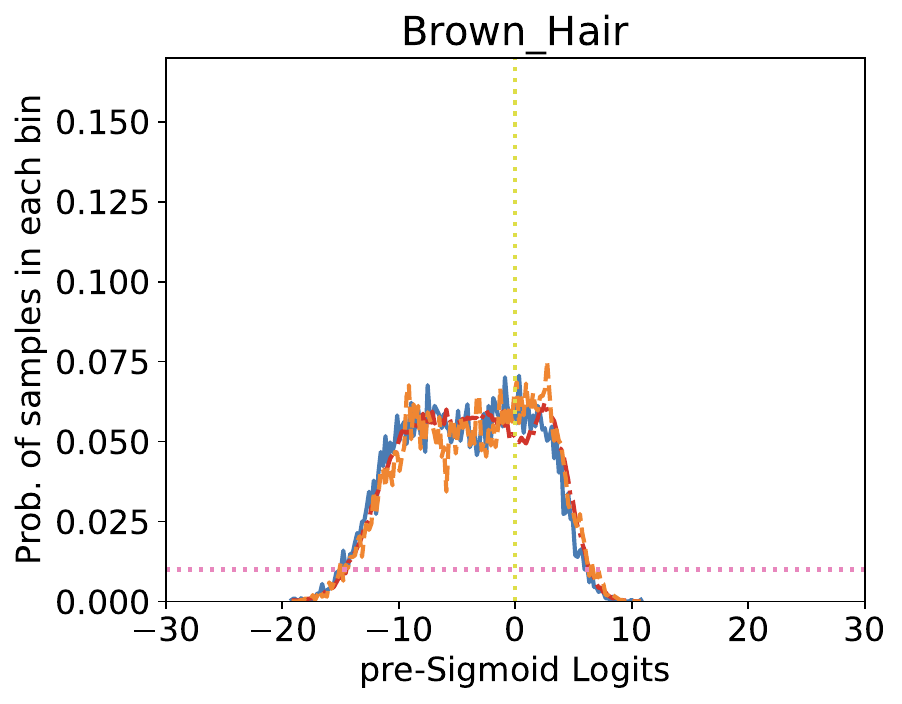}
		\caption{Brown Hair}
    \end{subfigure}
    \begin{subfigure}{0.19\textwidth}
		\includegraphics[width=\textwidth]{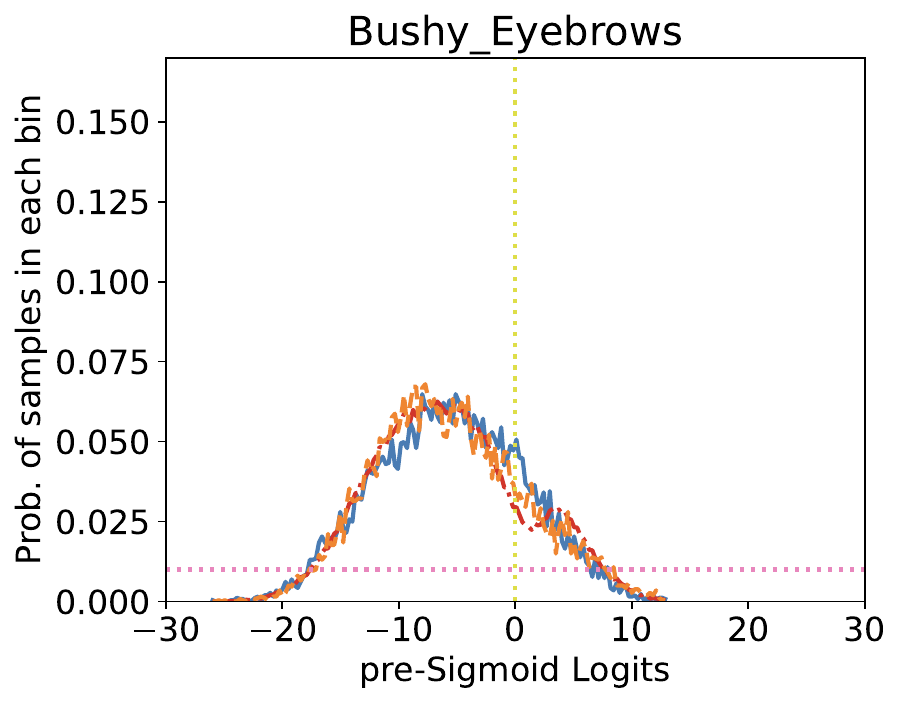}
		\caption{Bushy Eyebrows}
    \end{subfigure}
     \begin{subfigure}{0.19\textwidth}
		\includegraphics[width=\textwidth]{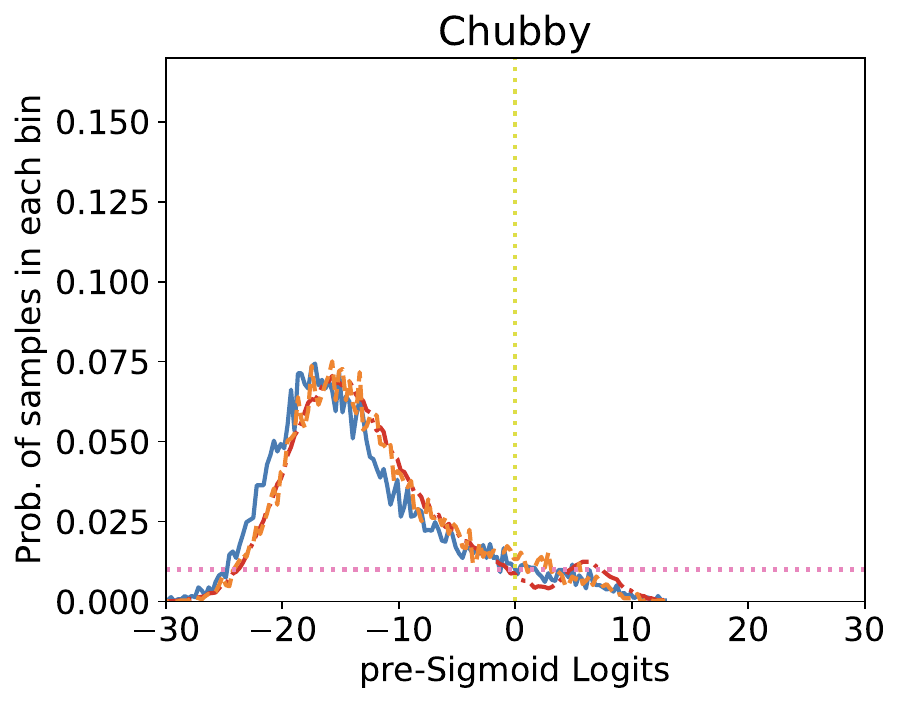}
		\caption{Chubby}
	    \end{subfigure}
    \begin{subfigure}{0.19\textwidth}
		\includegraphics[width=\textwidth]{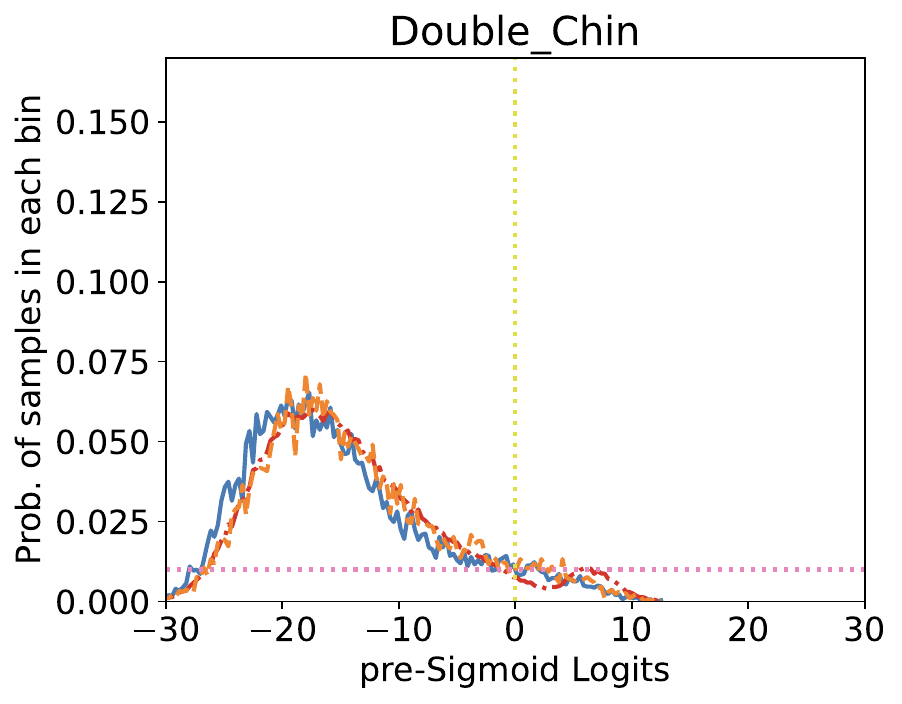}
		\caption{Double Chin}
	    \end{subfigure}
    \begin{subfigure}{0.19\textwidth}
		\includegraphics[width=\textwidth]{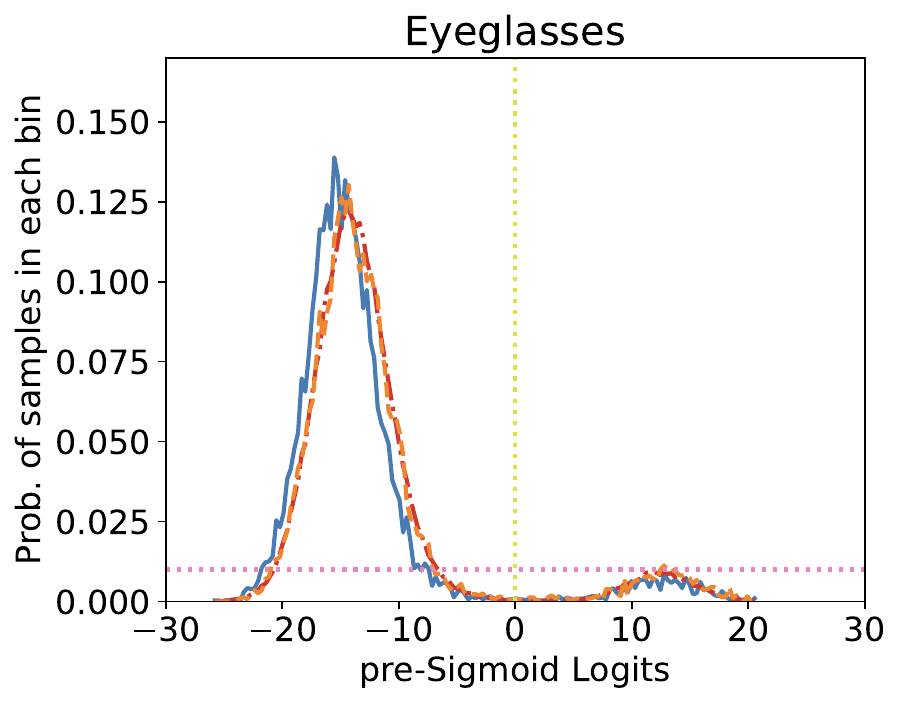}
		\caption{Eyeglasses}
    \end{subfigure}
    \vfill
    \begin{subfigure}{0.19\textwidth}
		\includegraphics[width=\textwidth]{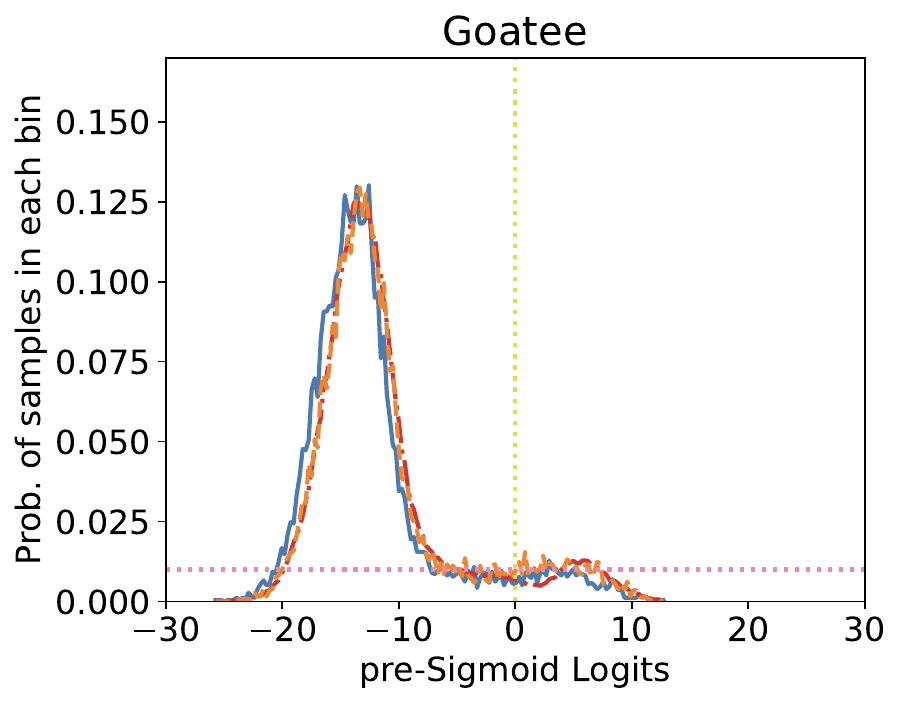}
		\caption{Goatee}
    \end{subfigure}
    \begin{subfigure}{0.19\textwidth}
		\includegraphics[width=\textwidth]{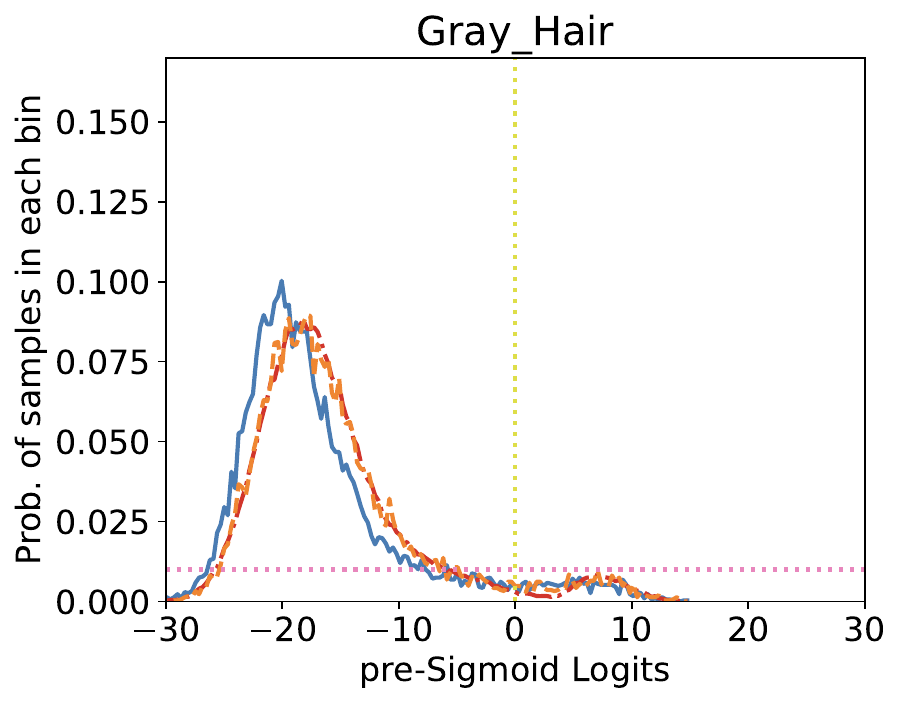}
		\caption{Gray Hair}
    \end{subfigure}
     \begin{subfigure}{0.19\textwidth}
		\includegraphics[width=\textwidth]{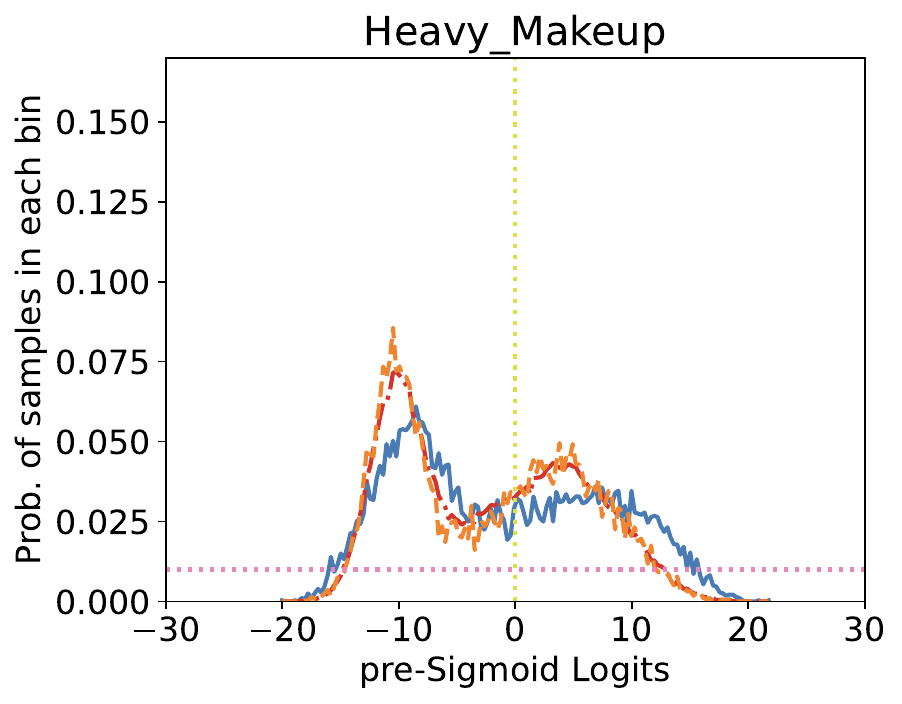}
		\caption{Heavy Makeup}
	    \end{subfigure}
    \begin{subfigure}{0.19\textwidth}
		\includegraphics[width=\textwidth]{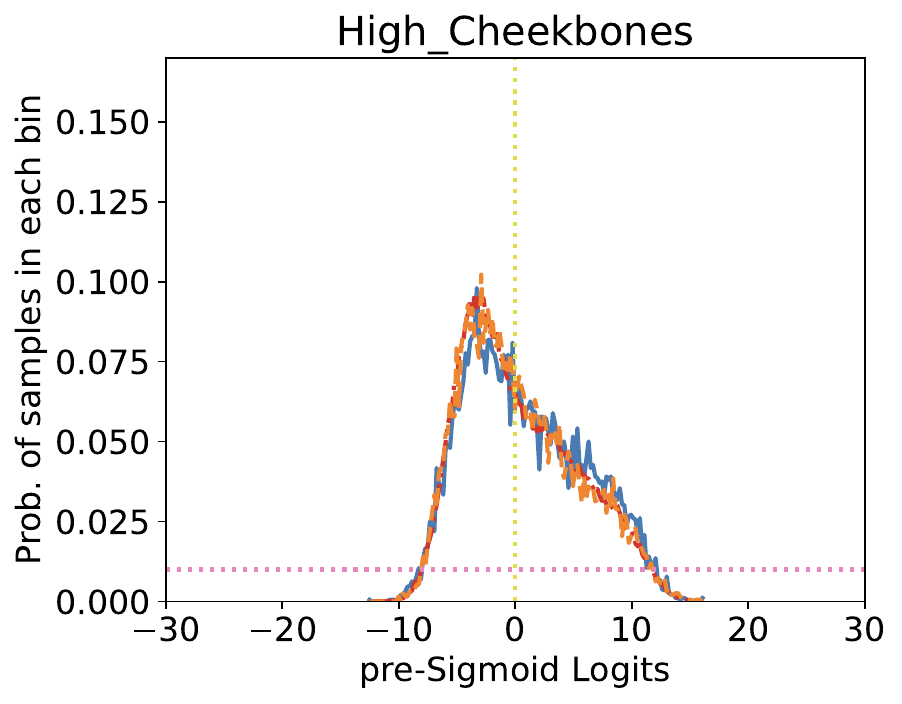}
		\caption{High Cheekbones}
	    \end{subfigure}
    \begin{subfigure}{0.19\textwidth}
		\includegraphics[width=\textwidth]{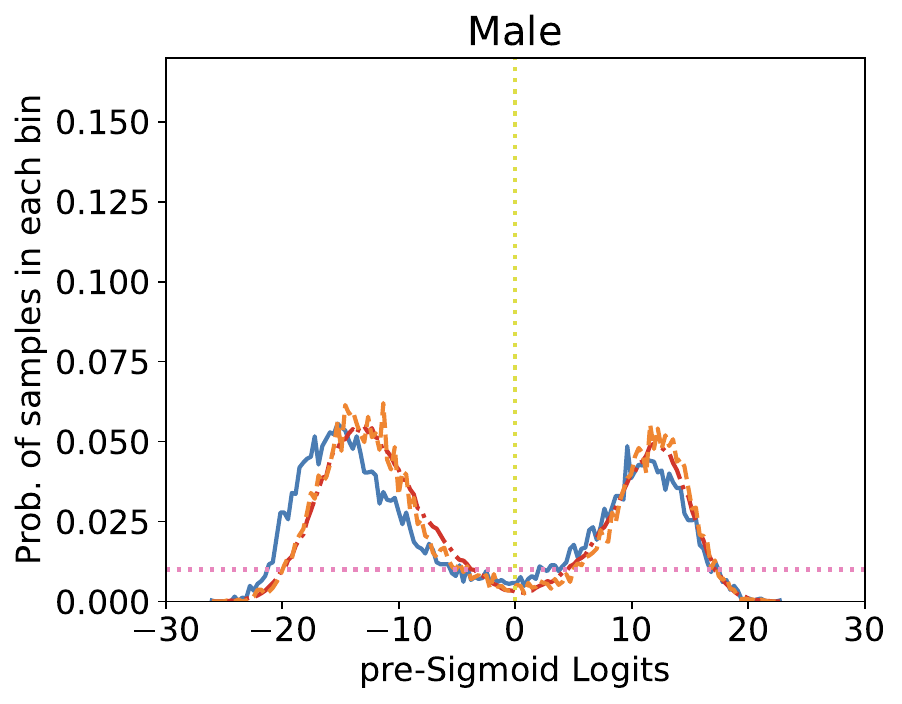}
		\caption{Male}
    \end{subfigure}
    \vfill
    \begin{subfigure}{0.19\textwidth}
		\includegraphics[width=\textwidth]{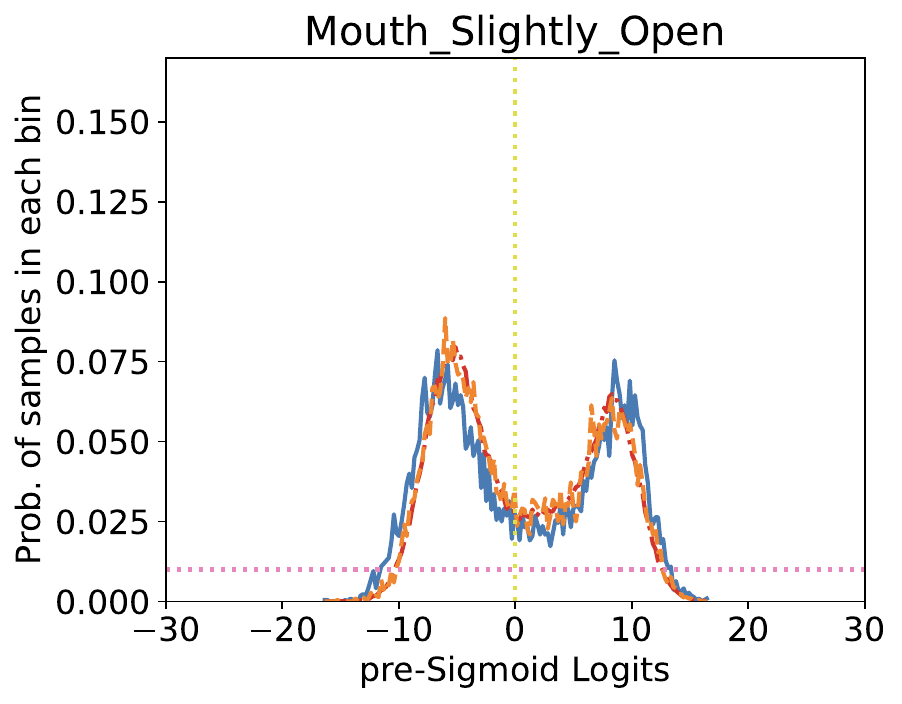}
		\caption{\begin{tiny}
		    Mouth Slightly Open
		\end{tiny}}
    \end{subfigure}
    \begin{subfigure}{0.19\textwidth}
		\includegraphics[width=\textwidth]{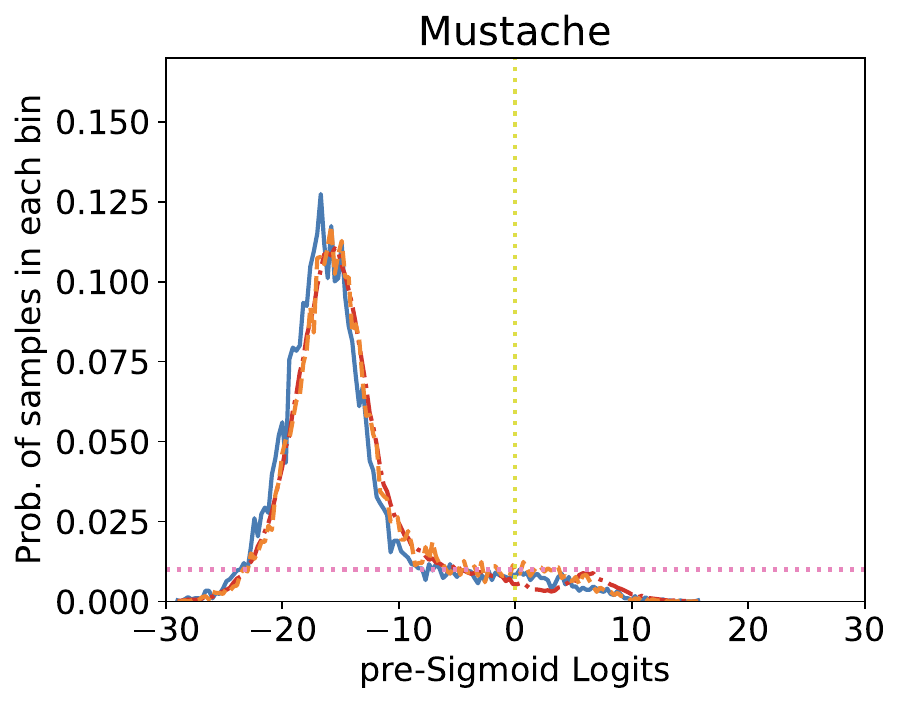}
		\caption{Mustache}
    \end{subfigure}
     \begin{subfigure}{0.19\textwidth}
		\includegraphics[width=\textwidth]{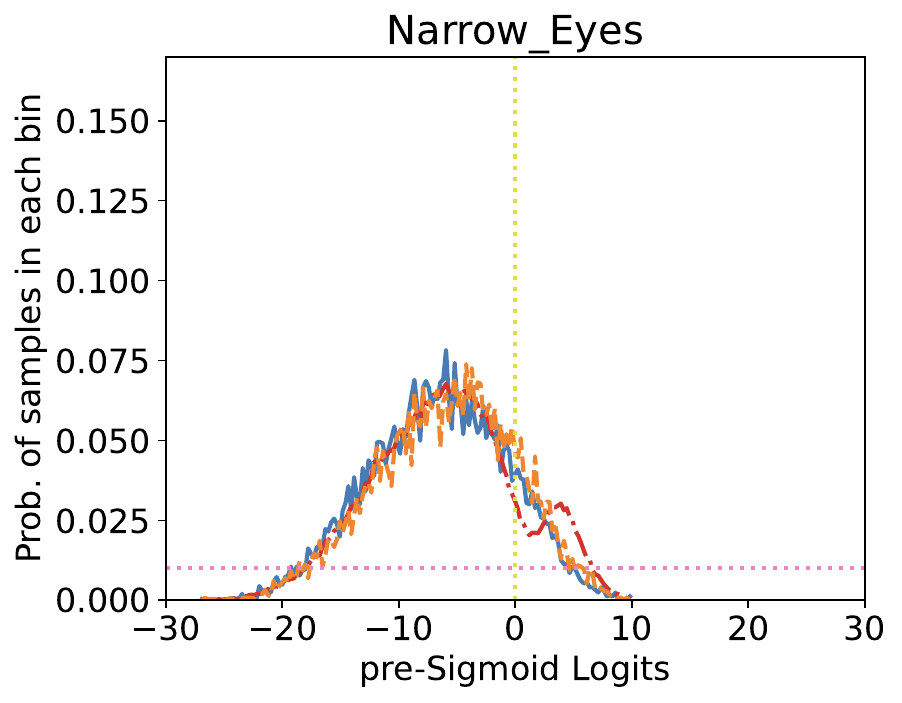}
		\caption{Narrow Eyes}
	    \end{subfigure}
    \begin{subfigure}{0.19\textwidth}
		\includegraphics[width=\textwidth]{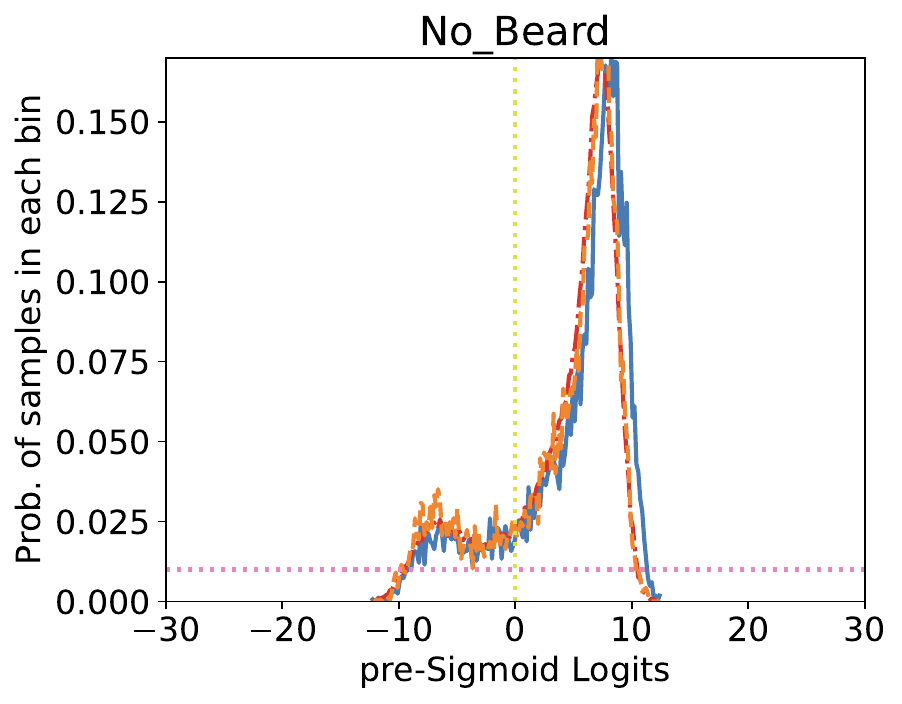}
		\caption{No Beard}
	    \end{subfigure}
    \begin{subfigure}{0.19\textwidth}
		\includegraphics[width=\textwidth]{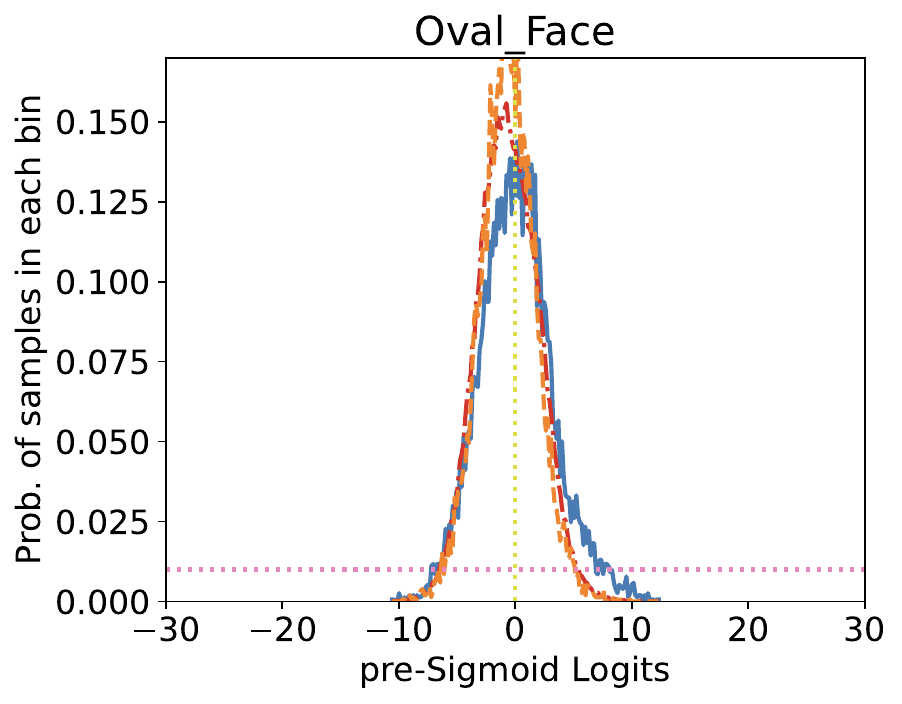}
		\caption{Oval Face}
    \end{subfigure}
    \vfill
    \begin{subfigure}{0.19\textwidth}
		\includegraphics[width=\textwidth]{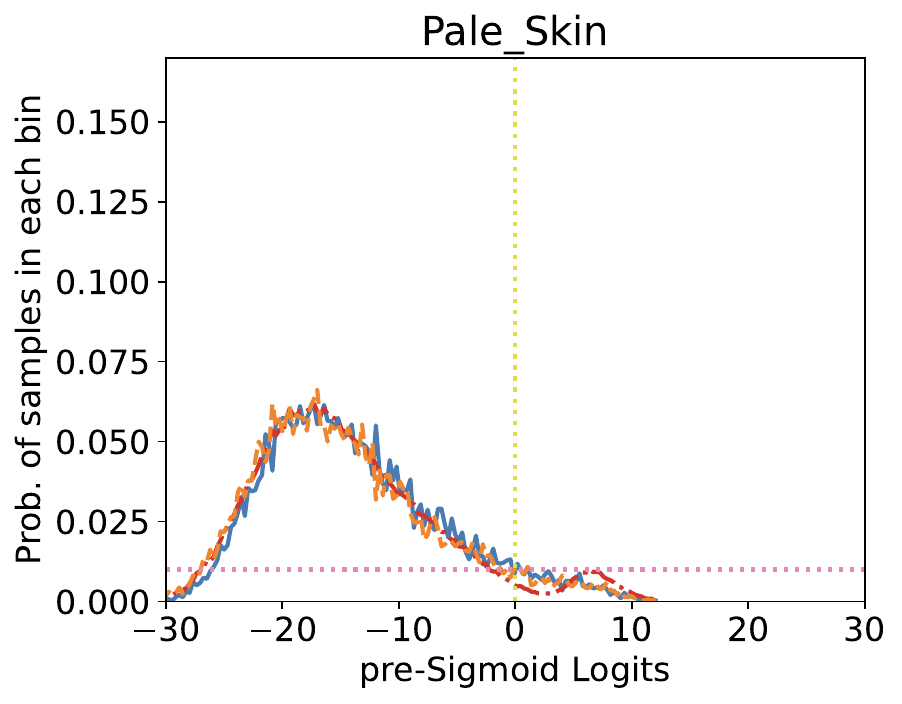}
		\caption{Pale Skin}
    \end{subfigure}
    \begin{subfigure}{0.19\textwidth}
		\includegraphics[width=\textwidth]{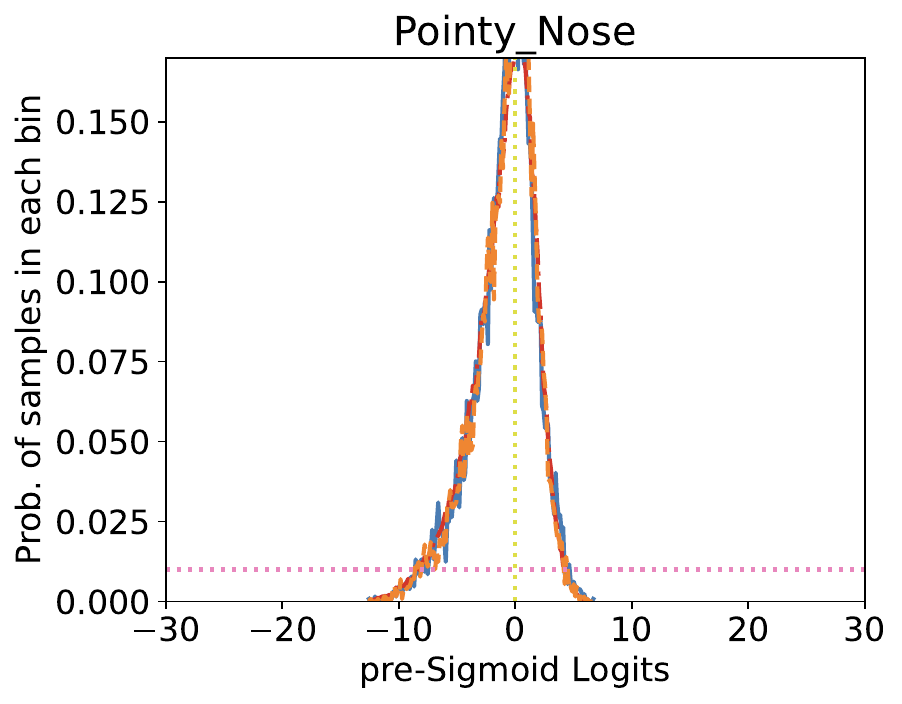}
		\caption{Pointy Nose}
    \end{subfigure}
     \begin{subfigure}{0.19\textwidth}
		\includegraphics[width=\textwidth]{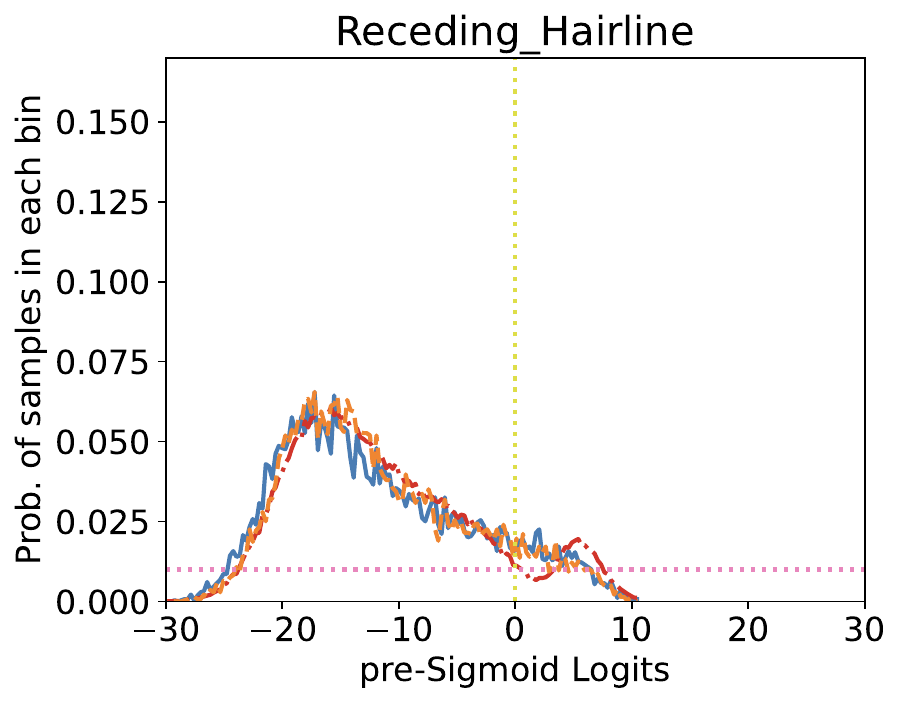}
		\caption{\begin{tiny}
		    Receding Hairline
		\end{tiny}}
	    \end{subfigure}
    \begin{subfigure}{0.19\textwidth}
		\includegraphics[width=\textwidth]{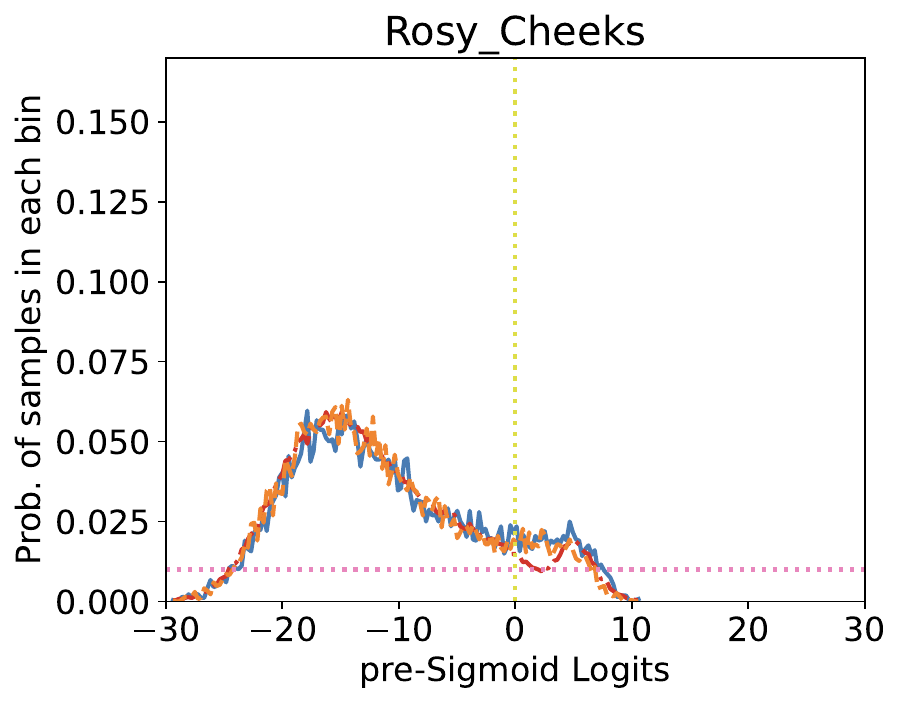}
		\caption{Rosy Cheeks}
	    \end{subfigure}
    \begin{subfigure}{0.19\textwidth}
		\includegraphics[width=\textwidth]{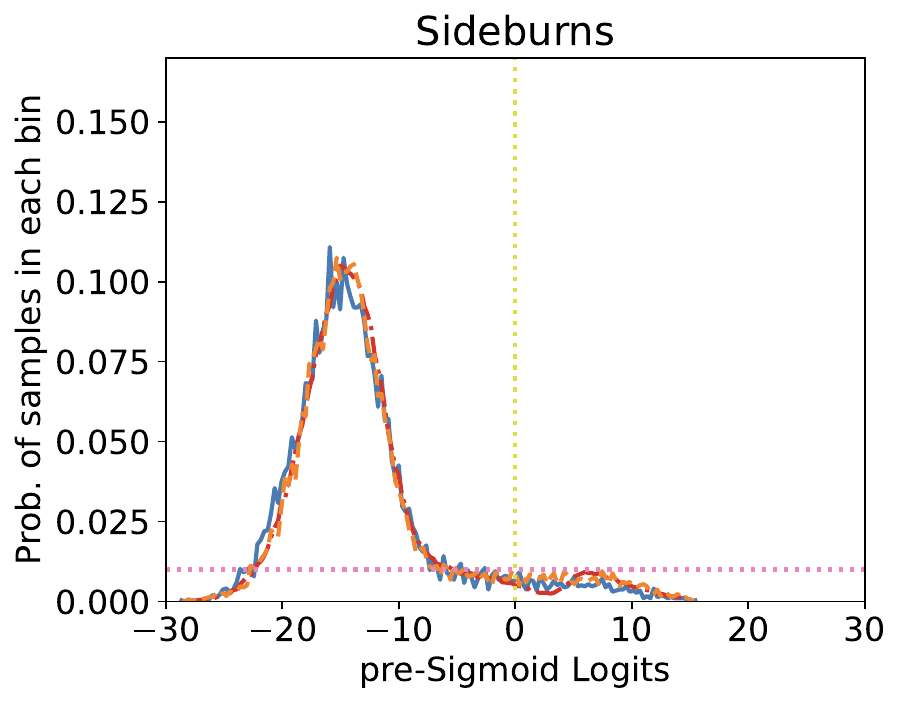}
		\caption{Sideburns}
    \end{subfigure}
    \vfill
    \begin{subfigure}{0.19\textwidth}
		\includegraphics[width=\textwidth]{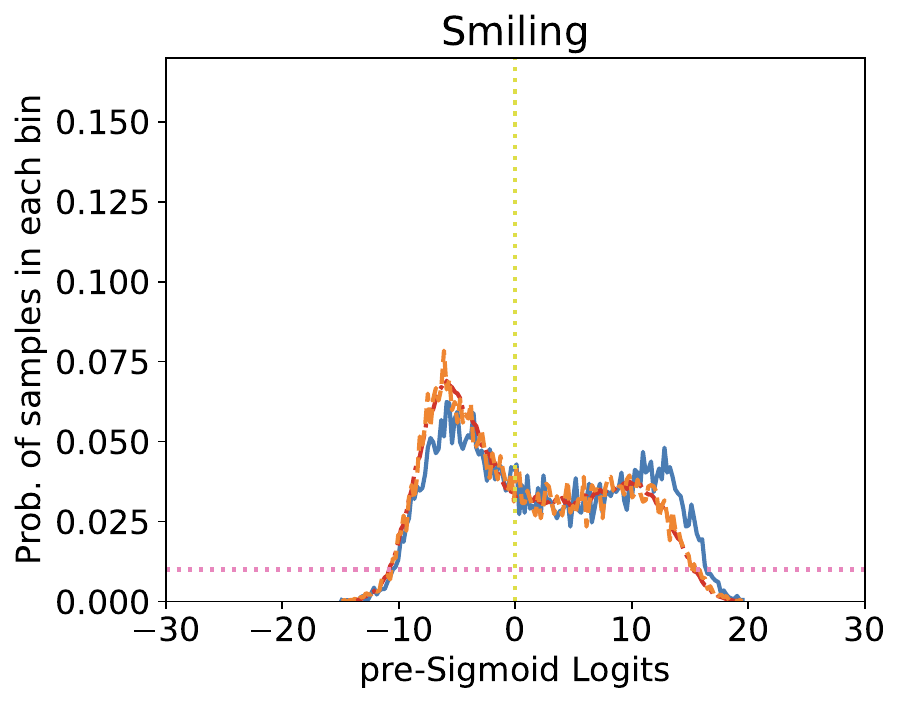}
		\caption{Smiling}
    \end{subfigure}
    \begin{subfigure}{0.19\textwidth}
		\includegraphics[width=\textwidth]{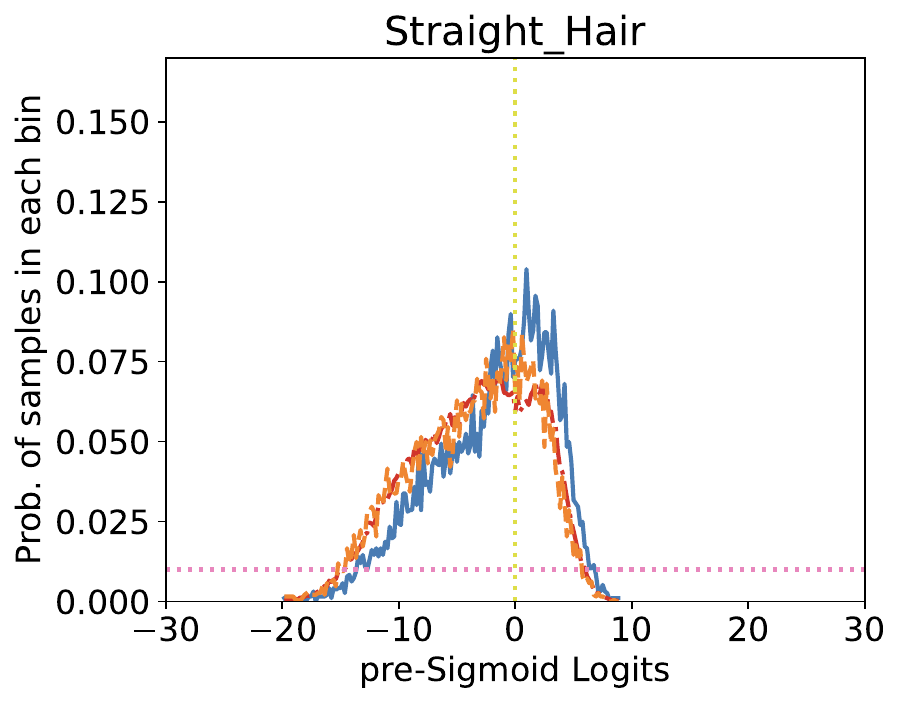}
		\caption{Straight hair}
    \end{subfigure}
     \begin{subfigure}{0.19\textwidth}
		\includegraphics[width=\textwidth]{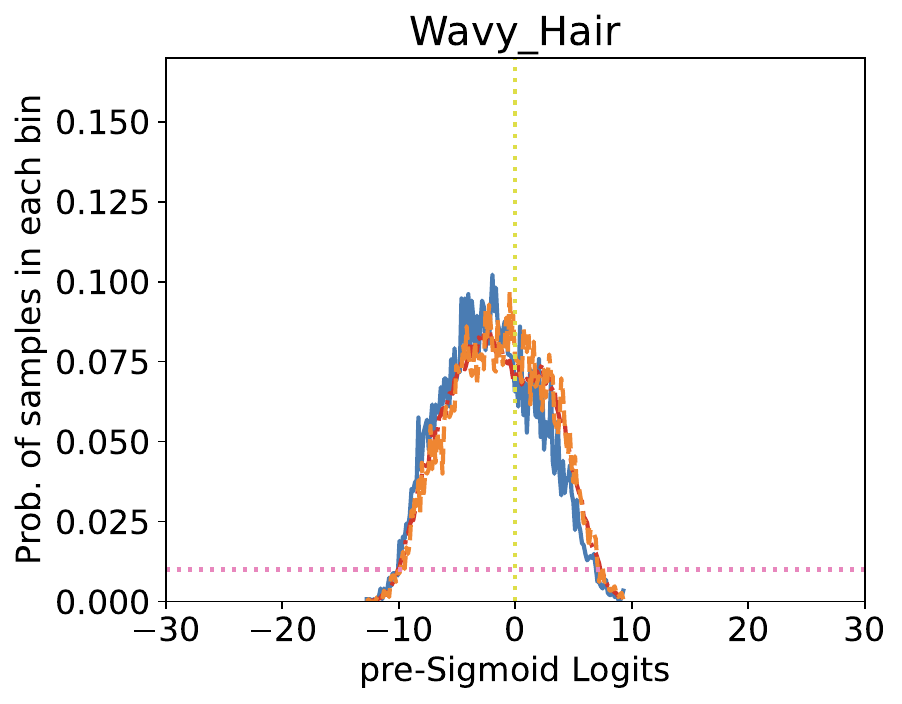}
		\caption{Wavy Hair}
	    \end{subfigure}
    \begin{subfigure}{0.19\textwidth}
		\includegraphics[width=\textwidth]{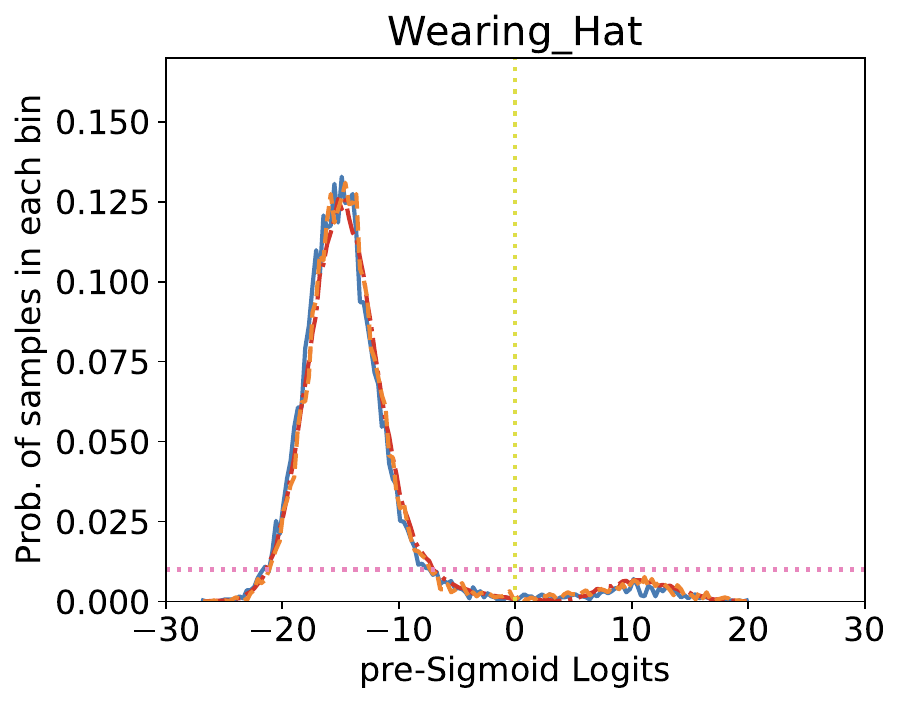}
		\caption{Wearing Hat}
    \end{subfigure}
    \begin{subfigure}{0.19\textwidth}
		\includegraphics[width=\textwidth]{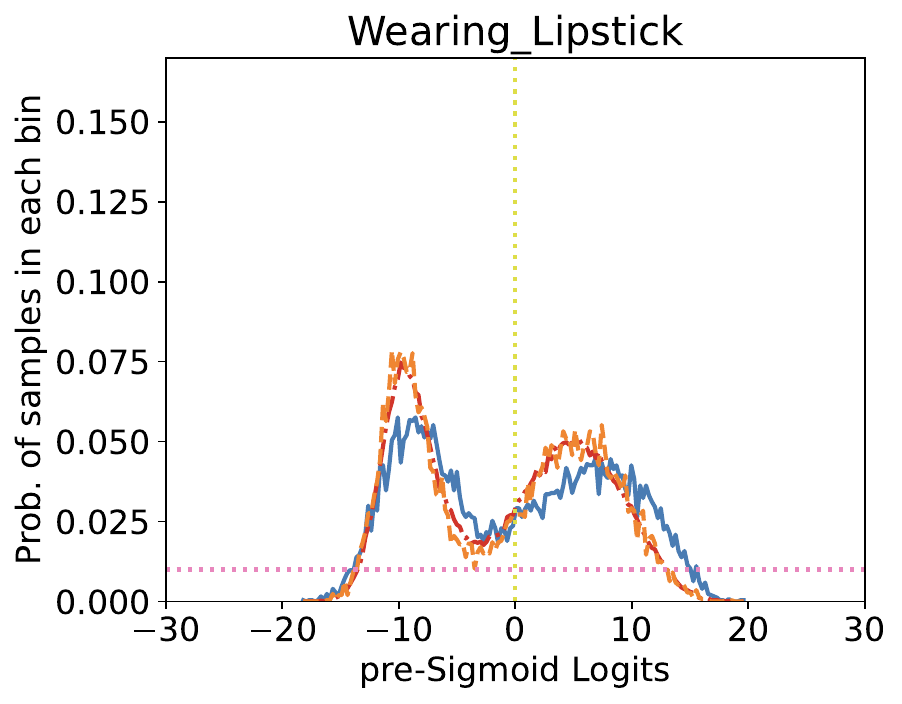}
		\caption{\begin{tiny}
		    Wearing Lipstick
		\end{tiny}}
    \end{subfigure}
    \vfill
    \begin{subfigure}{0.19\textwidth}
		\includegraphics[width=\textwidth]{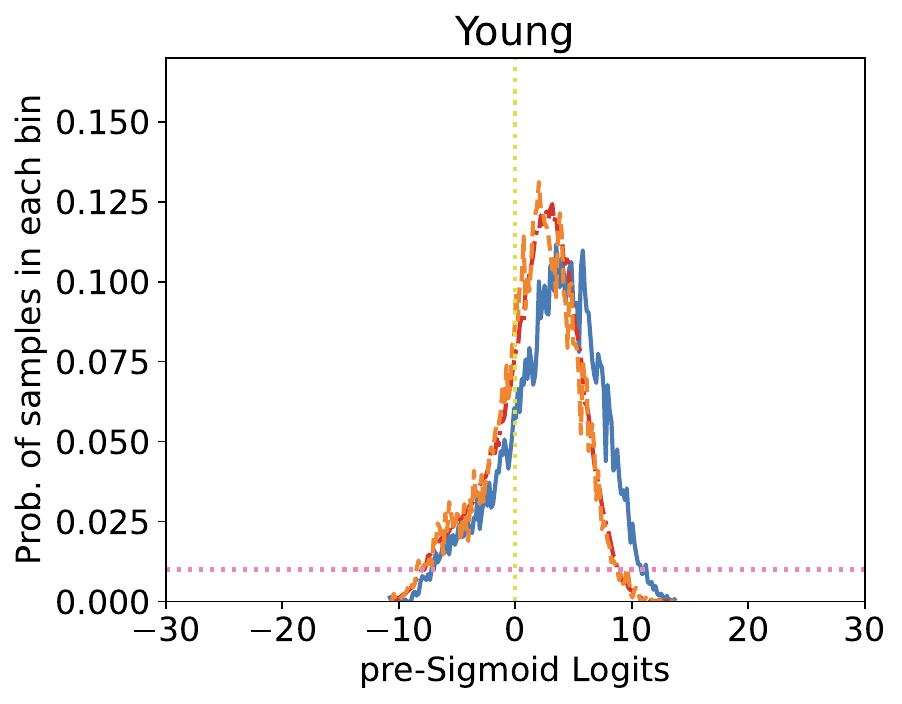}
		\caption{Young}
    \end{subfigure}
	\caption{The pre-sigmoid logits distribution of each attribute in CelebA (SwinTransformer-based classifier).}
	\label{fig:attr_dist_celeba_all_swint}
\end{figure}

\begin{figure}[tbp]
    \centering
    \begin{subfigure}{0.19\textwidth}
		\includegraphics[width=\textwidth]{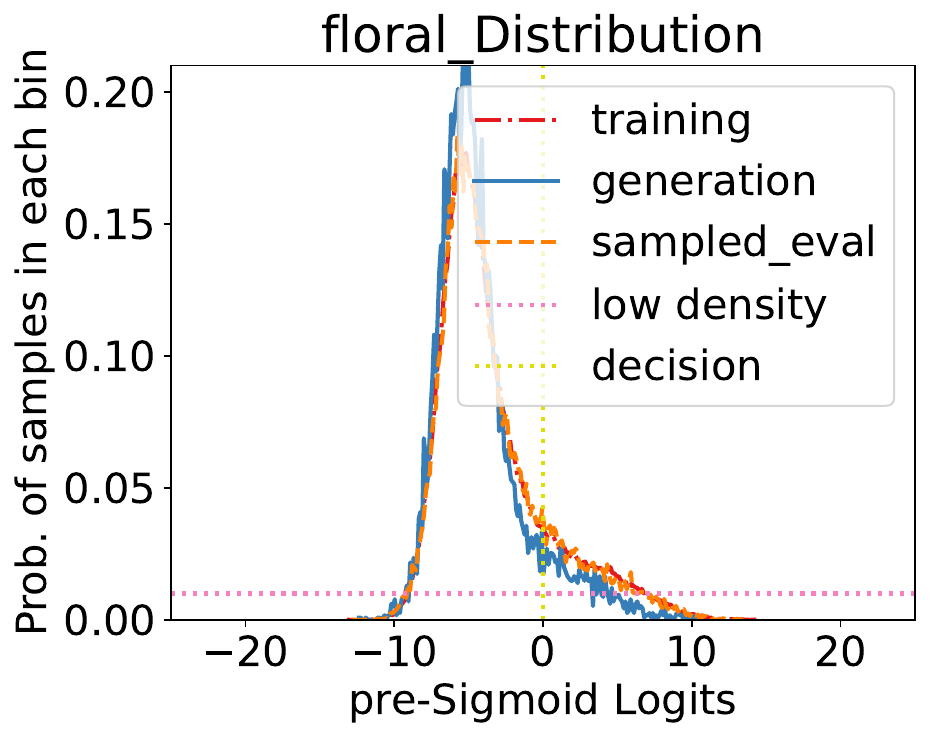}
		\caption{Floral}
    \end{subfigure}
    \begin{subfigure}{0.19\textwidth}
		\includegraphics[width=\textwidth]{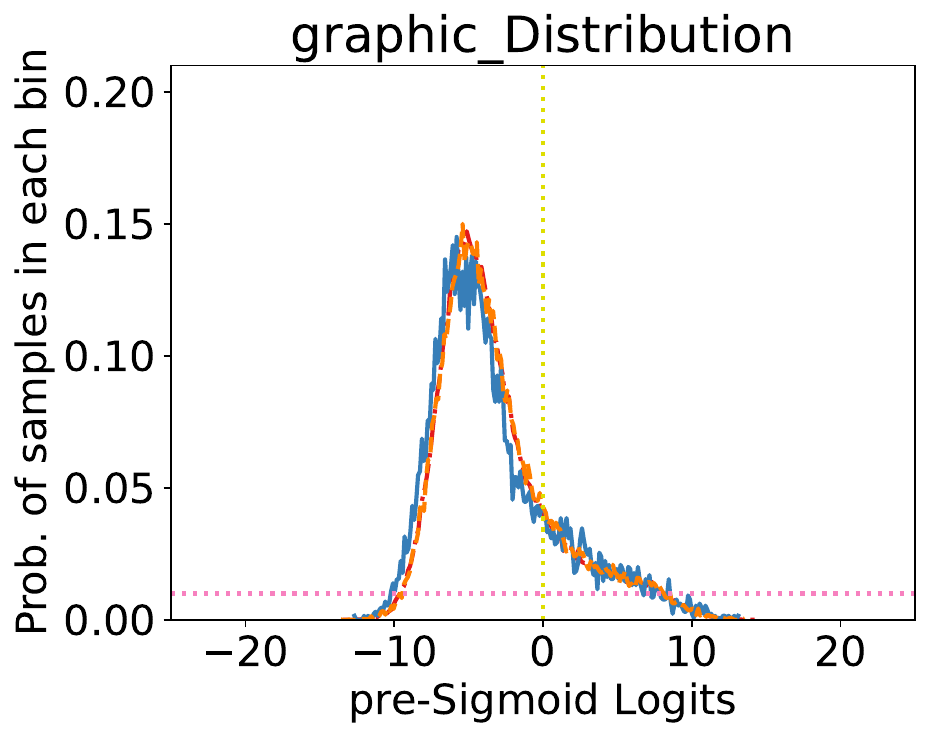}
		\caption{Graphic}
    \end{subfigure}
     \begin{subfigure}{0.19\textwidth}
		\includegraphics[width=\textwidth]{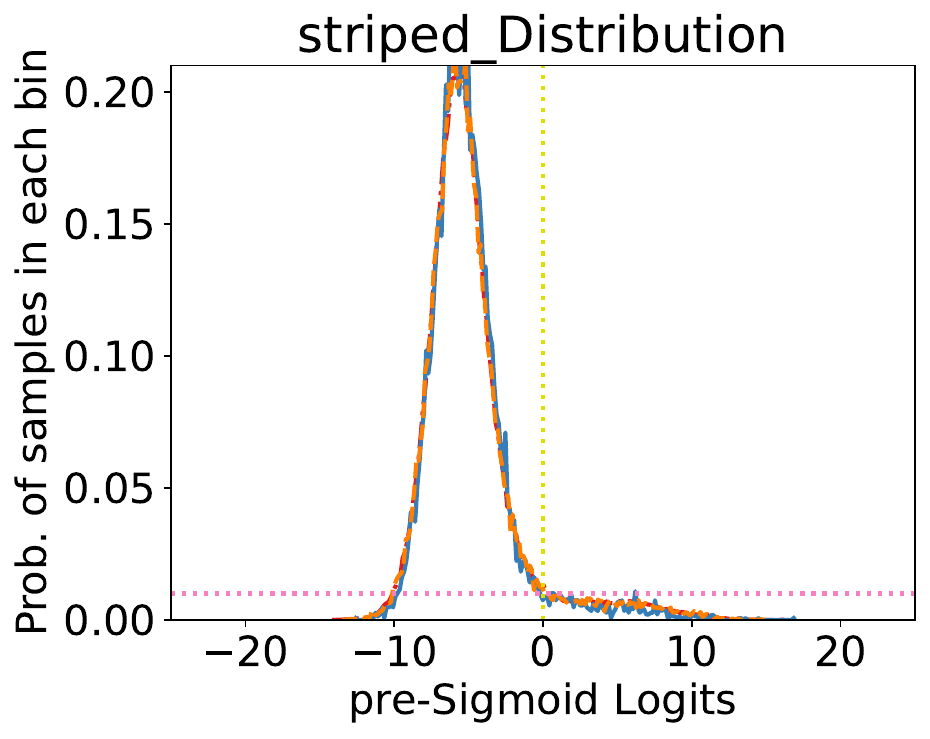}
		\caption{Striped}
	    \end{subfigure}
    \begin{subfigure}{0.19\textwidth}
		\includegraphics[width=\textwidth]{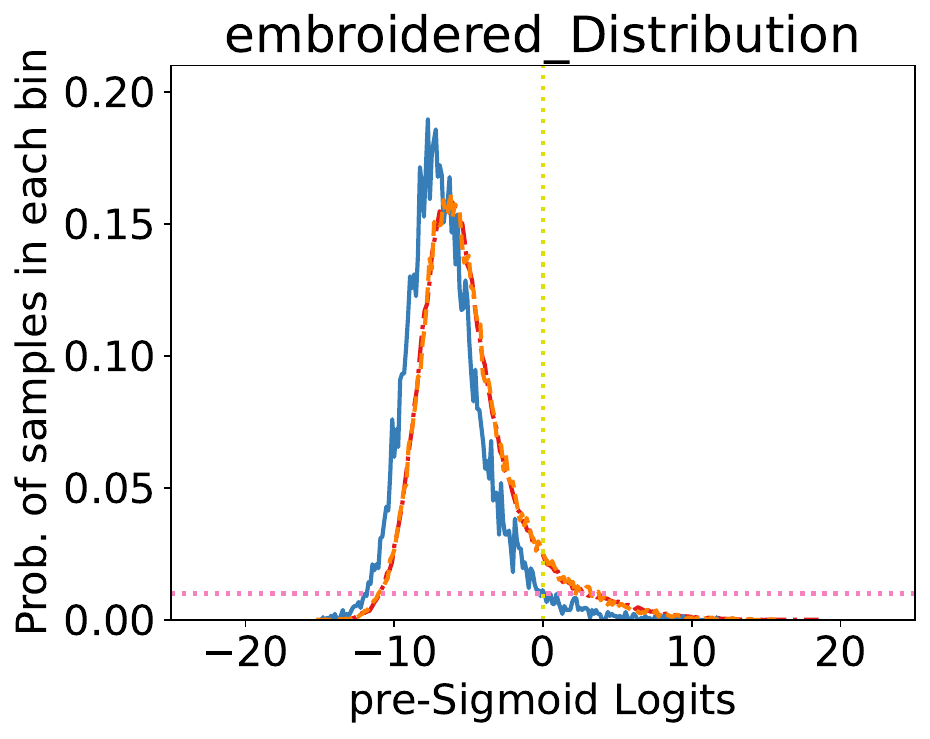}
		\caption{Embroidered}
	    \end{subfigure}
    \begin{subfigure}{0.19\textwidth}
		\includegraphics[width=\textwidth]{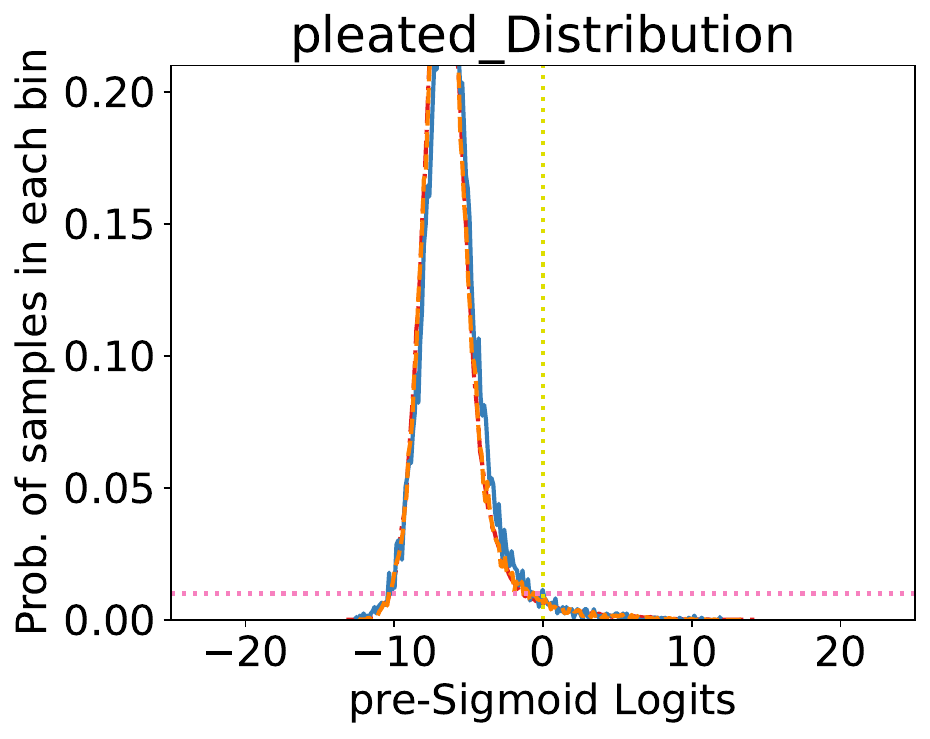}
		\caption{Pleated}
    \end{subfigure}
     \vfill
    \begin{subfigure}{0.19\textwidth}
		\includegraphics[width=\textwidth]{Figs/cls_presigmoid_dp_wolegend/solid.pdf}
		\caption{Solid}
    \end{subfigure}
    \begin{subfigure}{0.19\textwidth}
		\includegraphics[width=\textwidth]{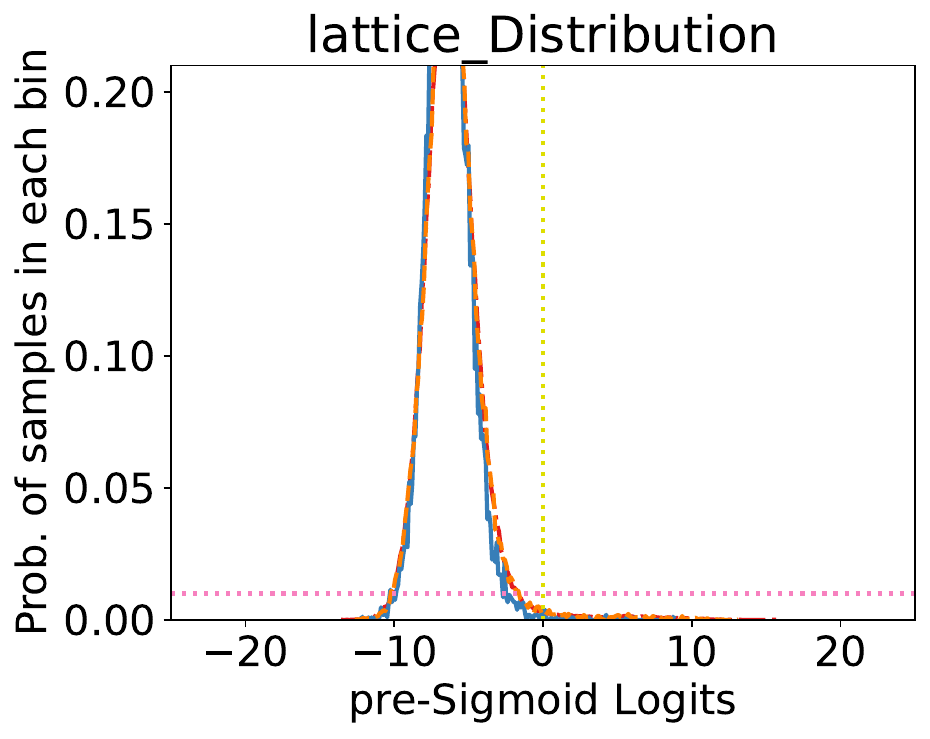}
		\caption{Lattice}
    \end{subfigure}
     \begin{subfigure}{0.19\textwidth}
		\includegraphics[width=\textwidth]{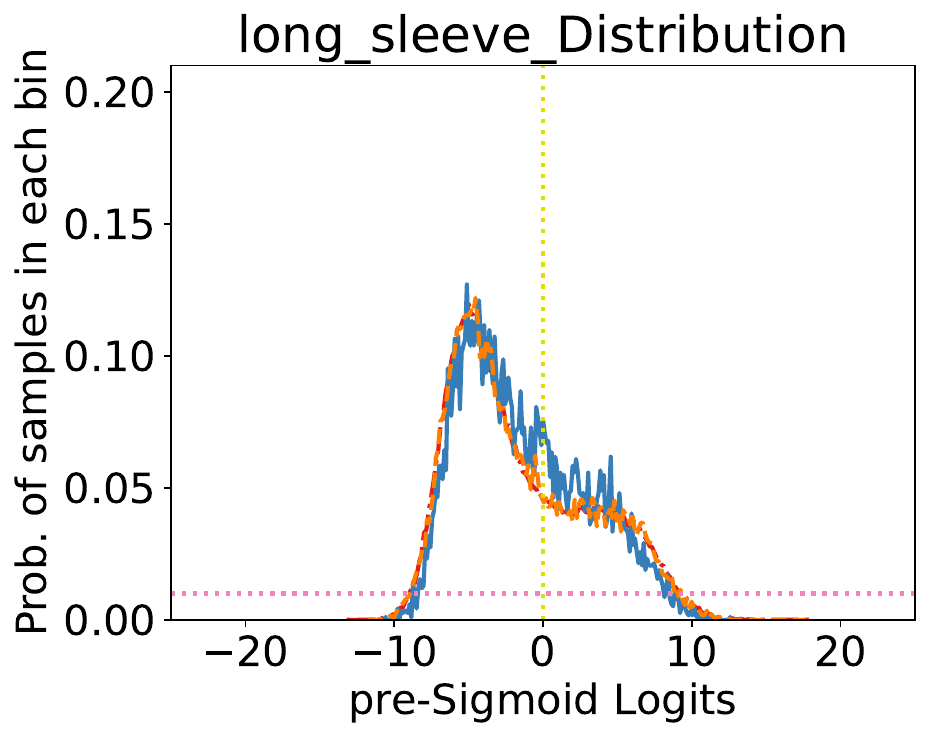}
		\caption{Long Sleeve}
	    \end{subfigure}
    \begin{subfigure}{0.19\textwidth}
		\includegraphics[width=\textwidth]{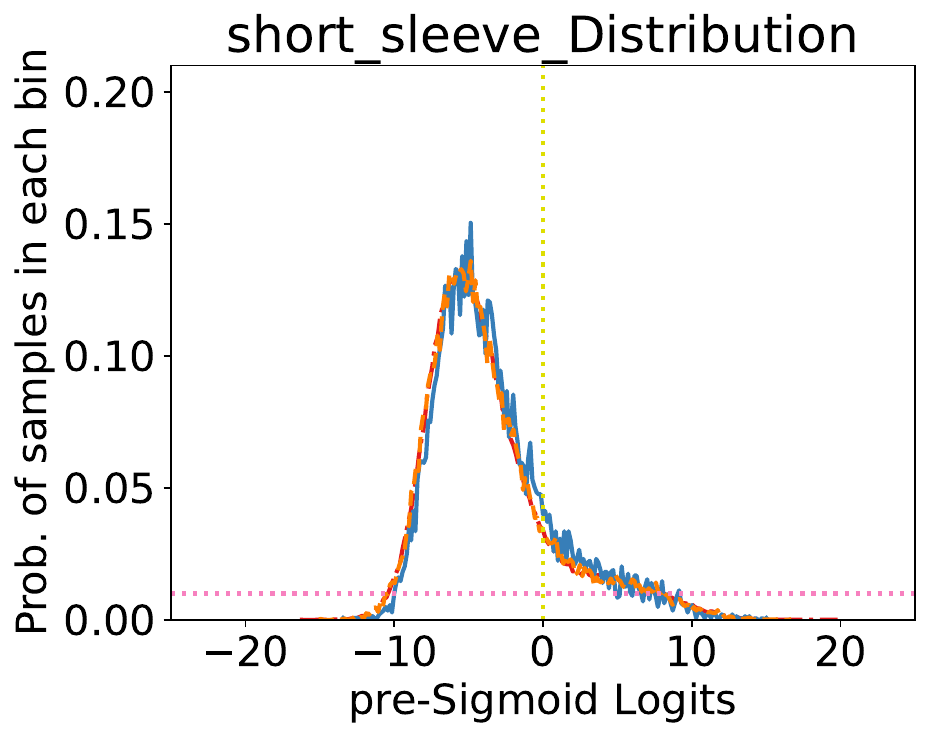}
		\caption{Short Sleeve}
	    \end{subfigure}
    \begin{subfigure}{0.19\textwidth}
		\includegraphics[width=\textwidth]{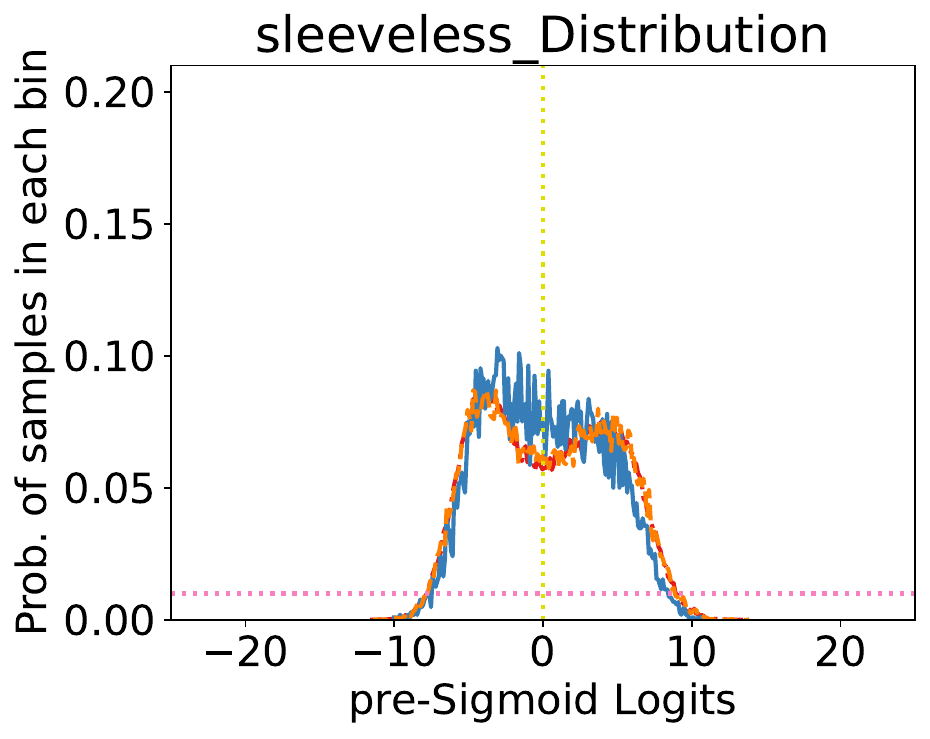}
		\caption{Sleeveless}
    \end{subfigure}
    \vfill
    \begin{subfigure}{0.19\textwidth}
		\includegraphics[width=\textwidth]{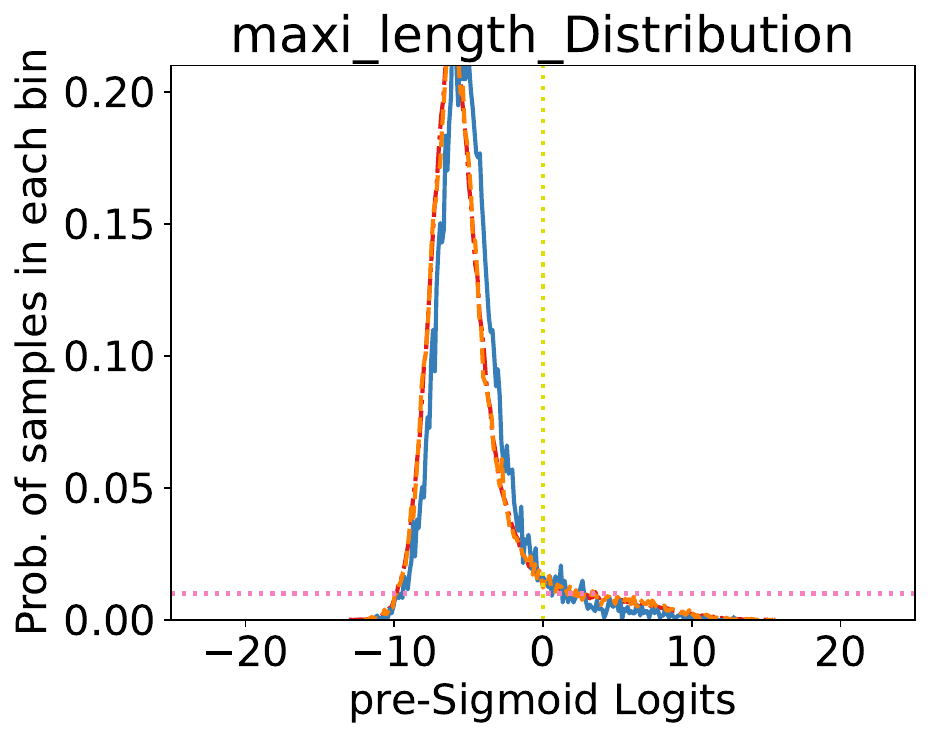}
		\caption{Maxi Length}
    \end{subfigure}
    \begin{subfigure}{0.19\textwidth}
		\includegraphics[width=\textwidth]{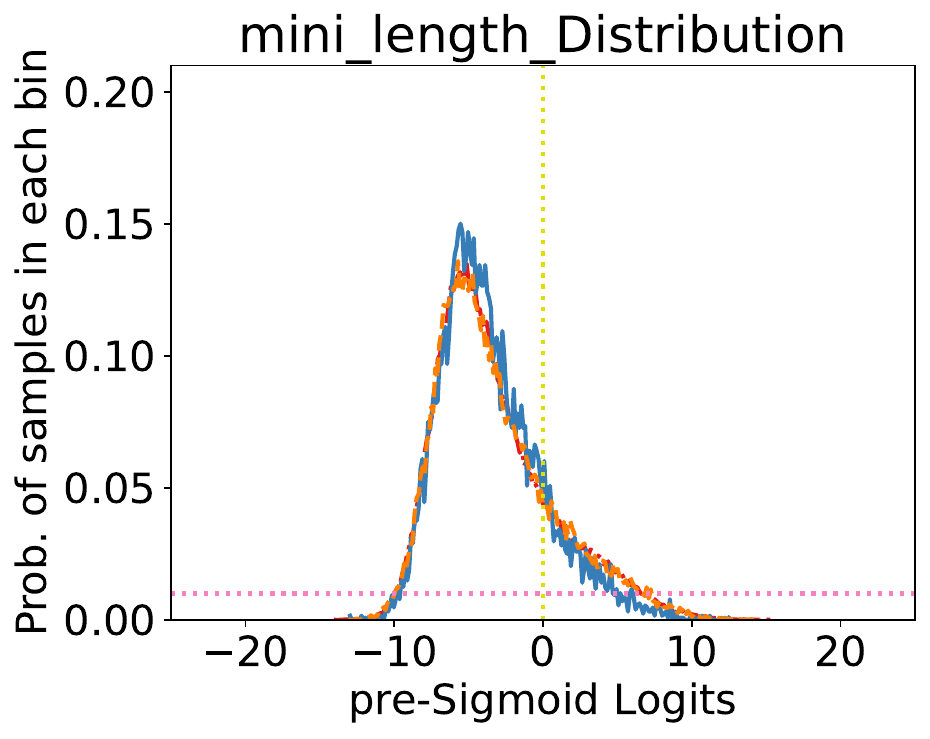}
		\caption{Mini Length}
    \end{subfigure}
     \begin{subfigure}{0.19\textwidth}
		\includegraphics[width=\textwidth]{Figs/cls_presigmoid_dp_wolegend/no_dress.pdf}
		\caption{No Dress}
	    \end{subfigure}
    \begin{subfigure}{0.19\textwidth}
		\includegraphics[width=\textwidth]{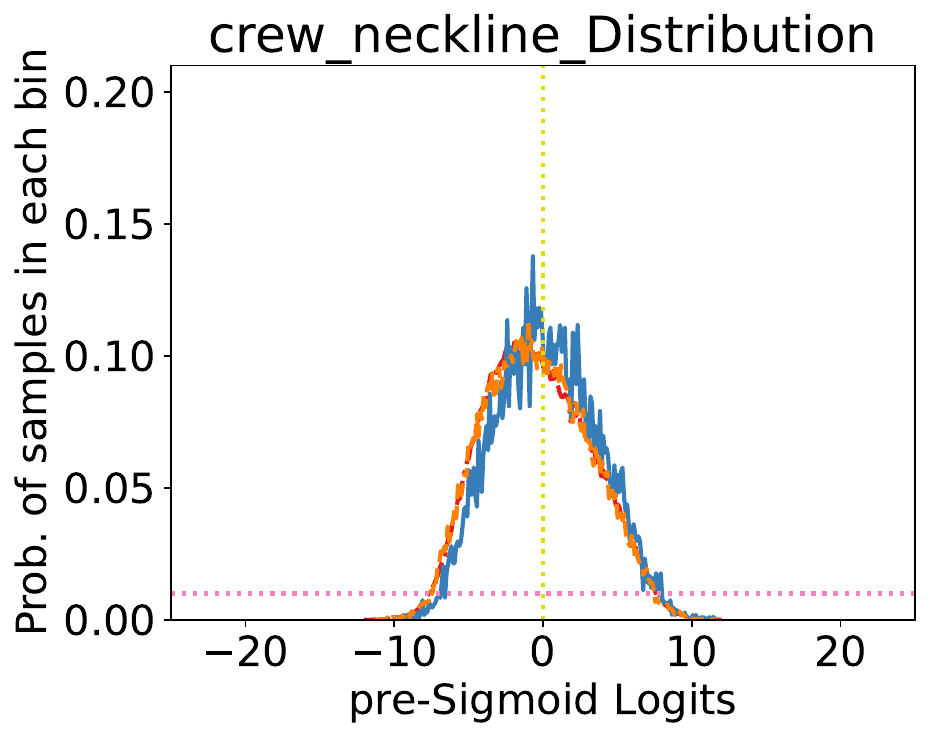}
		\caption{Crew Neckline}
	    \end{subfigure}
    \begin{subfigure}{0.19\textwidth}
		\includegraphics[width=\textwidth]{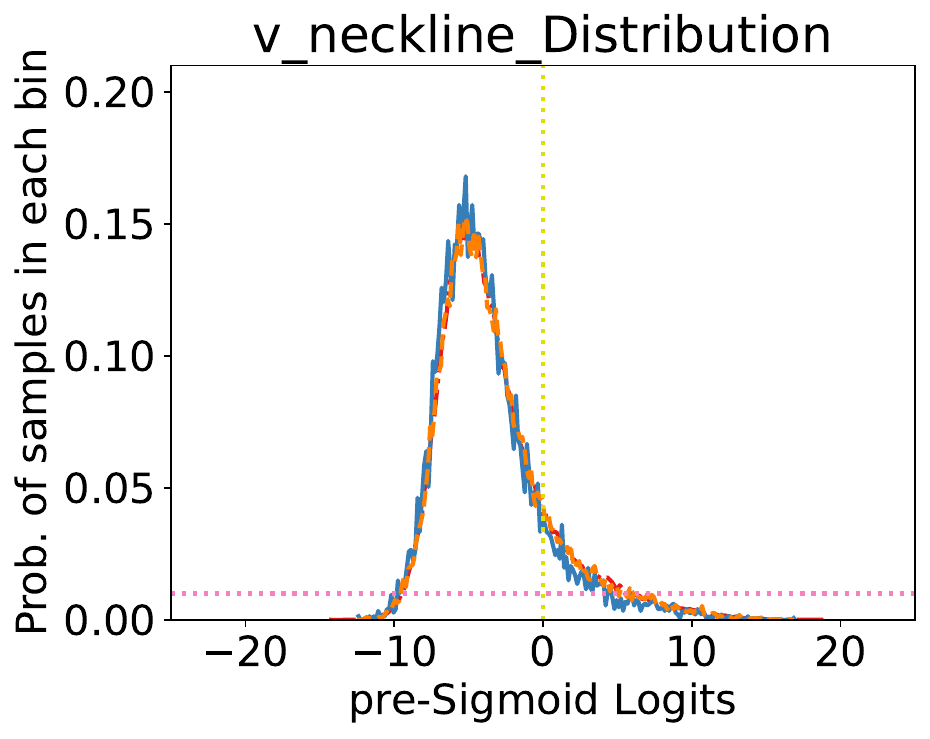}
		\caption{V Neckline}
    \end{subfigure}
    \vfill
    \begin{subfigure}{0.19\textwidth}
		\includegraphics[width=\textwidth]{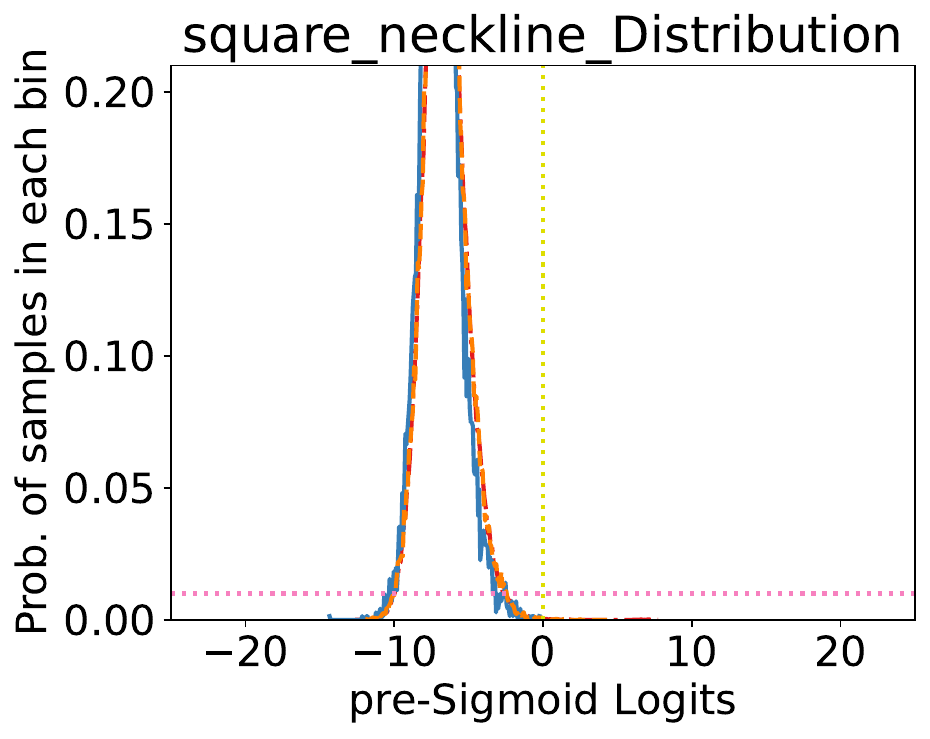}
		\caption{Square Neckline}
    \end{subfigure}
    \begin{subfigure}{0.19\textwidth}
		\includegraphics[width=\textwidth]{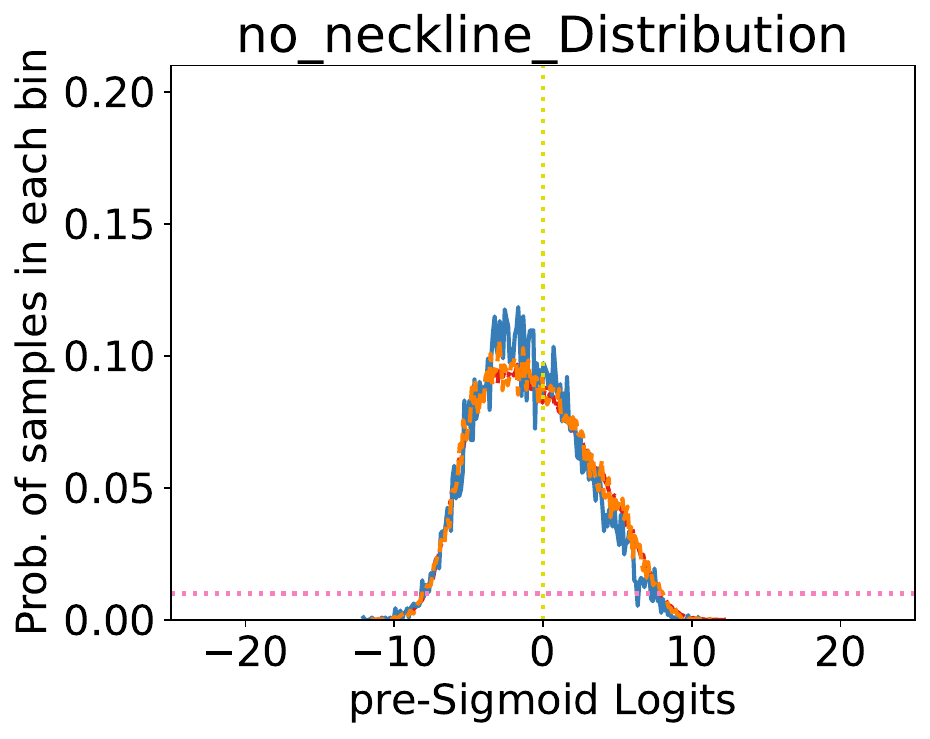}
		\caption{No Neckline}
    \end{subfigure}
     \begin{subfigure}{0.19\textwidth}
		\includegraphics[width=\textwidth]{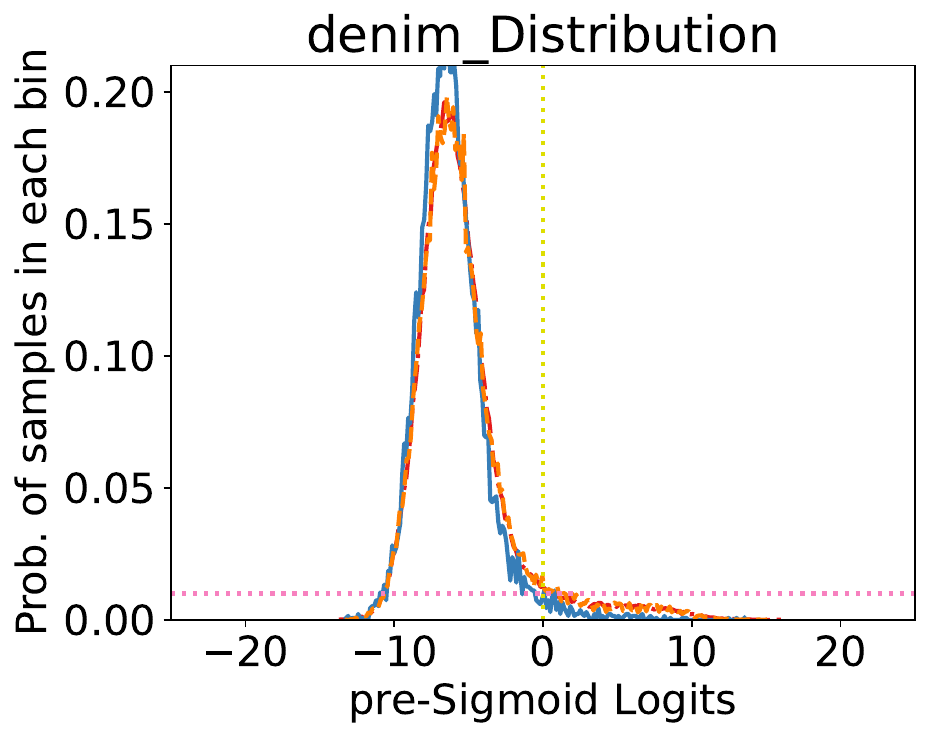}
		\caption{Denim}
	    \end{subfigure}
    \begin{subfigure}{0.19\textwidth}
		\includegraphics[width=\textwidth]{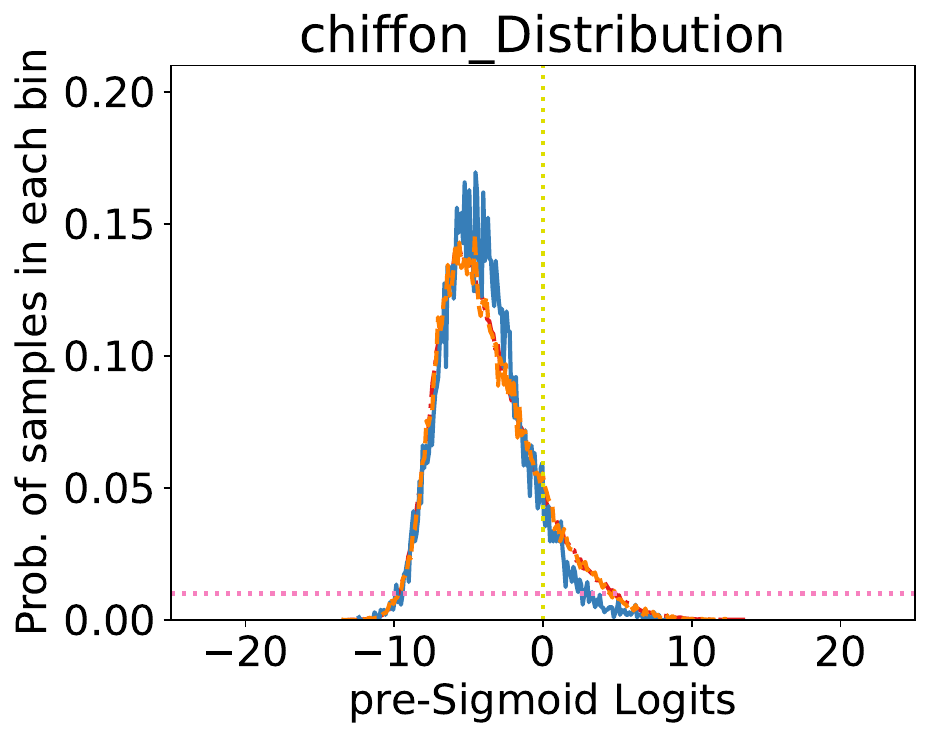}
		\caption{Chiffon}
	    \end{subfigure}
    \begin{subfigure}{0.19\textwidth}
		\includegraphics[width=\textwidth]{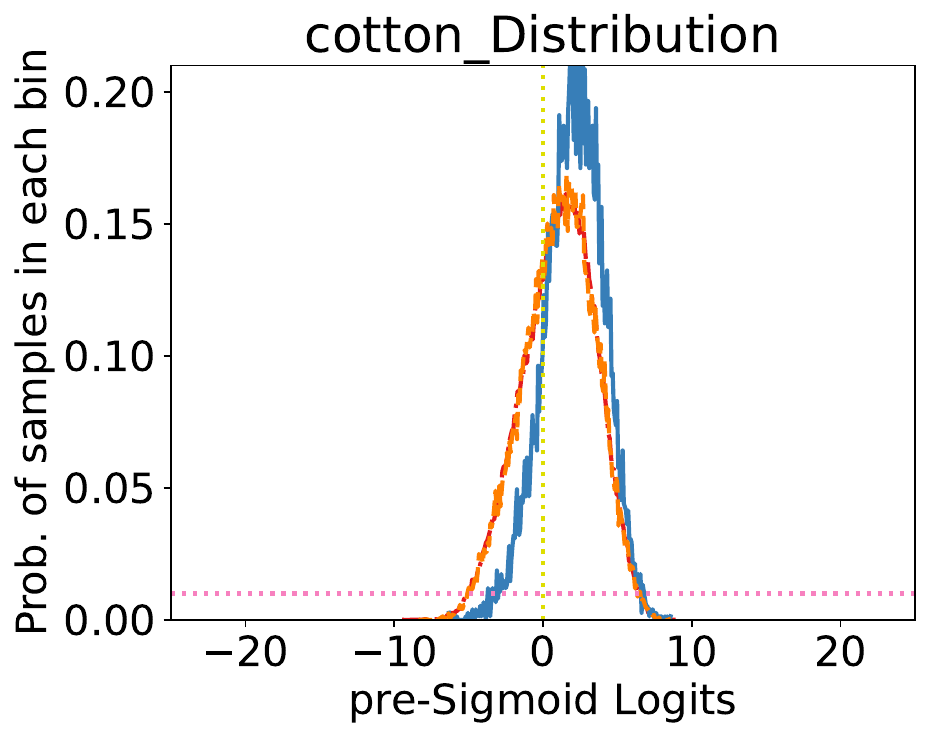}
		\caption{Cotton}
    \end{subfigure}
    \vfill
    \begin{subfigure}{0.19\textwidth}
		\includegraphics[width=\textwidth]{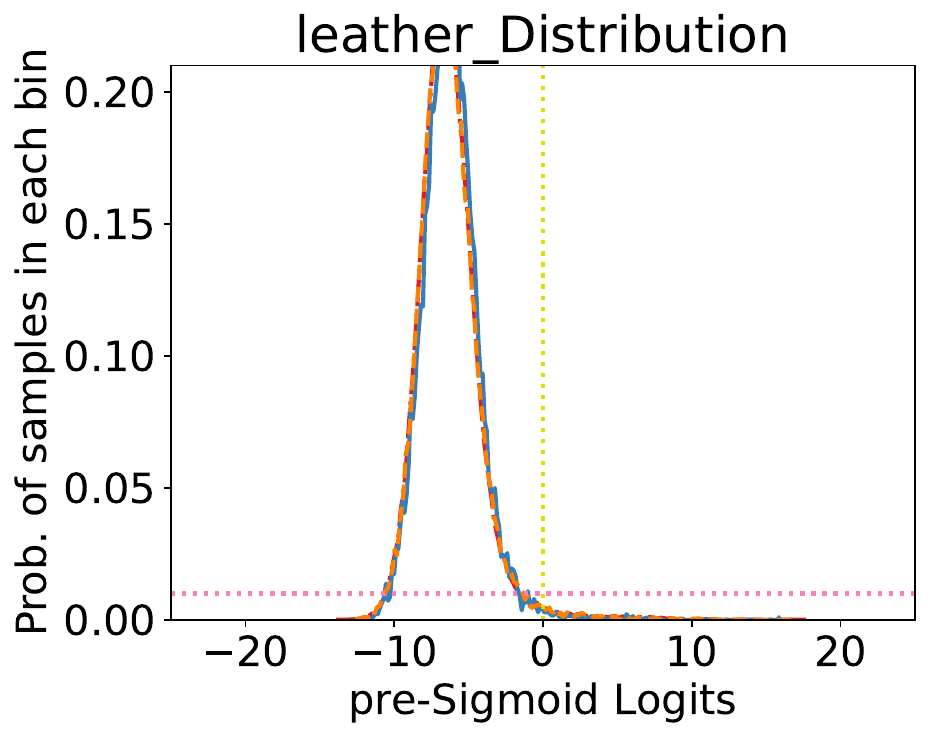}
		\caption{Leather}
    \end{subfigure}
    \begin{subfigure}{0.19\textwidth}
		\includegraphics[width=\textwidth]{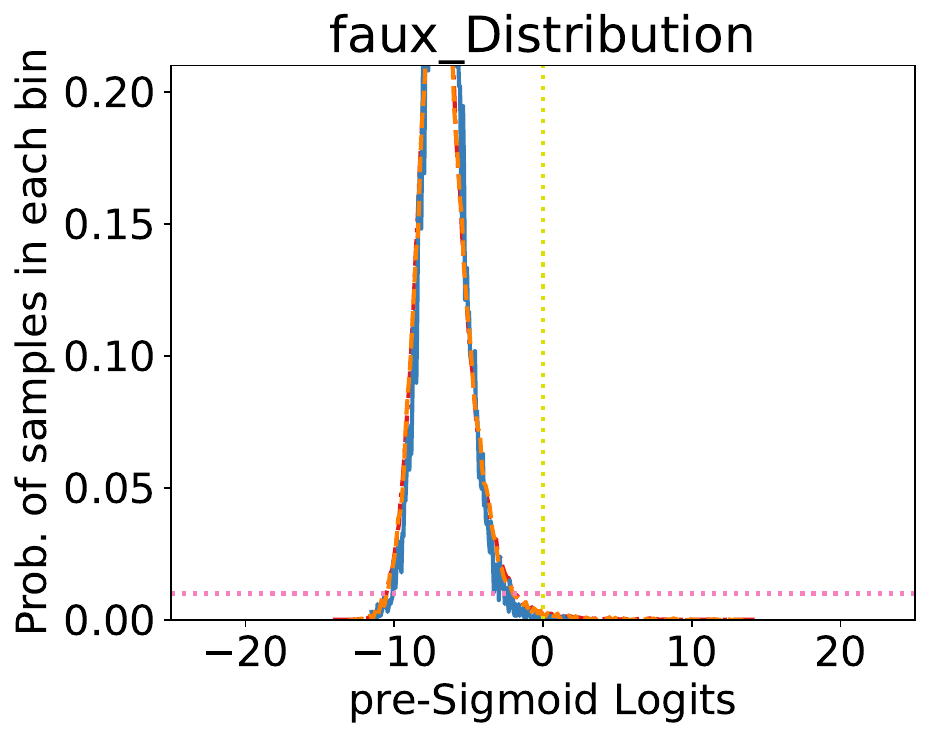}
		\caption{Faux}
    \end{subfigure}
     \begin{subfigure}{0.19\textwidth}
		\includegraphics[width=\textwidth]{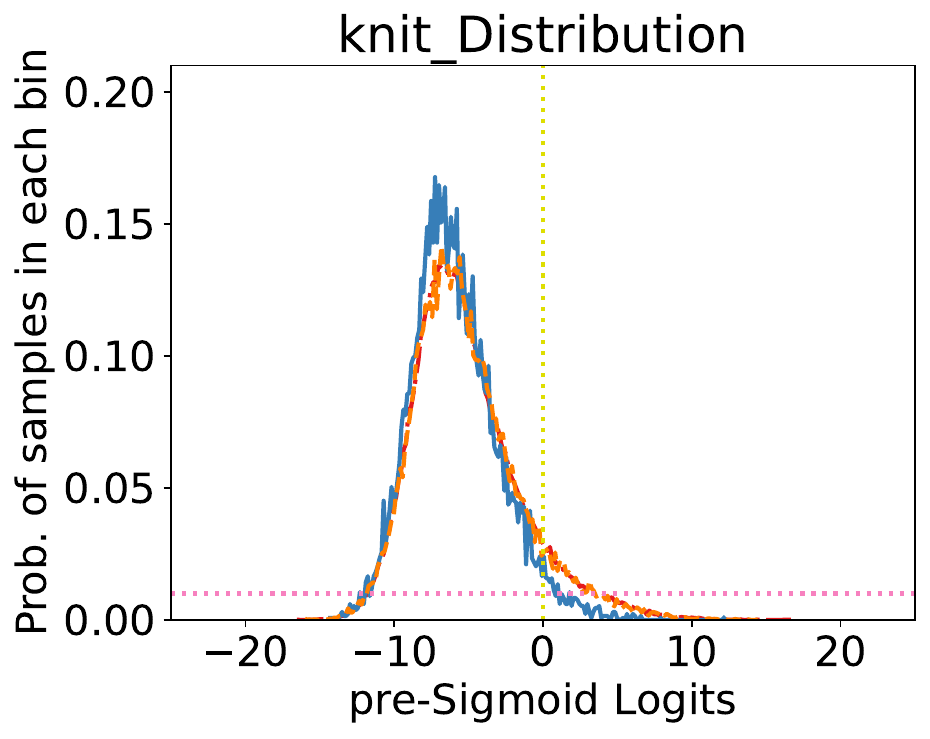}
		\caption{Knit}
	    \end{subfigure}
    \begin{subfigure}{0.19\textwidth}
		\includegraphics[width=\textwidth]{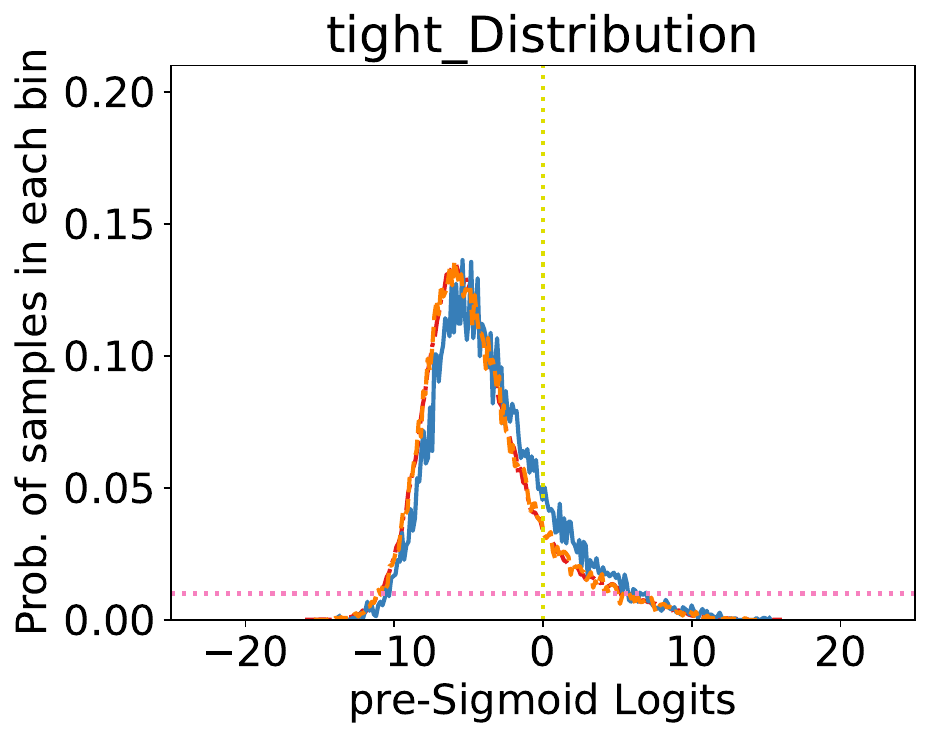}
		\caption{Tight}
	    \end{subfigure}
    \begin{subfigure}{0.19\textwidth}
		\includegraphics[width=\textwidth]{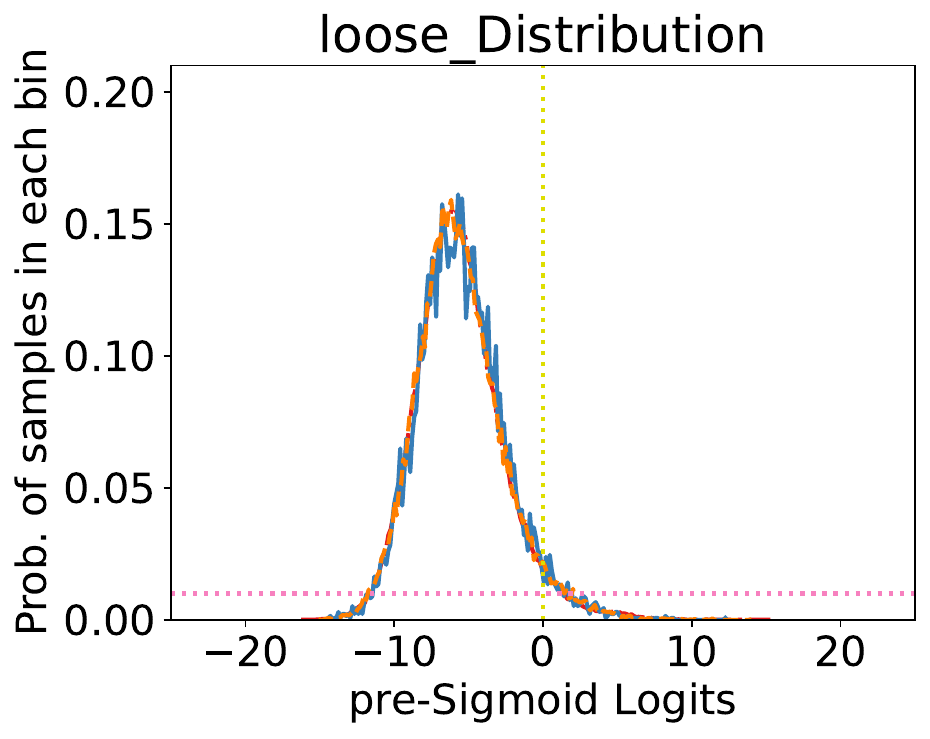}
		\caption{Loose}
    \end{subfigure}
    \vfill
    \begin{subfigure}{0.19\textwidth}
		\includegraphics[width=\textwidth]{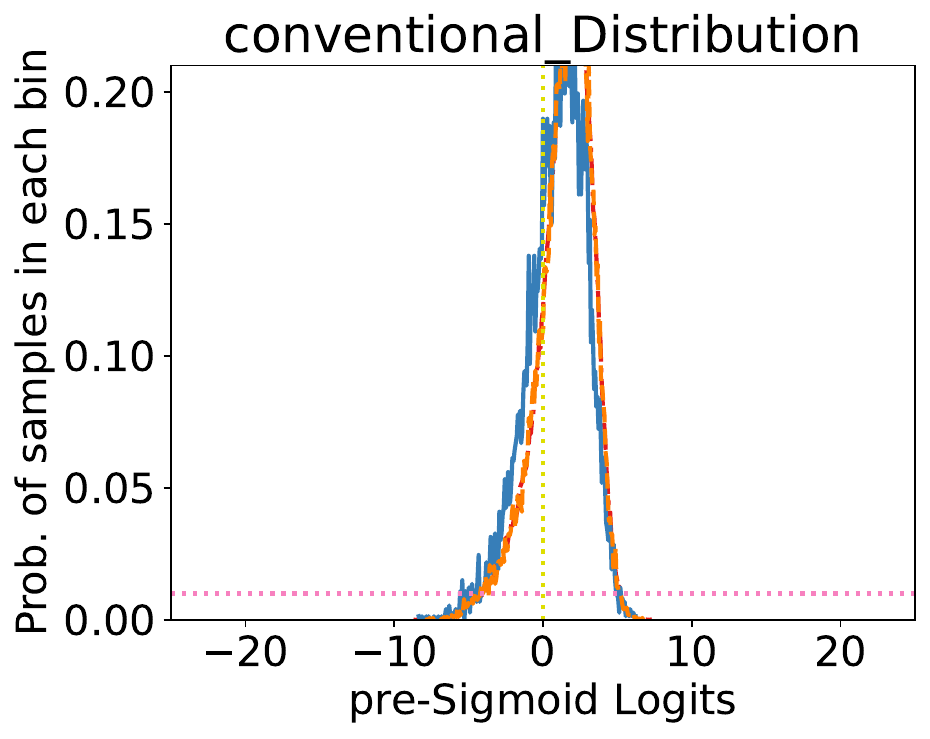}
		\caption{Conventional}
    \end{subfigure}
	\caption{The pre-sigmoid logits distribution of each attribute in DeepFashion.}
	\label{fig:attr_dist_dp_all}
\end{figure}

Previous work indicates that classifiers can amplify the discriminative biases in the training set~\citep{menshopping2017,biasampmeta}.
We use the positive and negative sample ratio to reweigh the cross entropy loss terms. This acts as an upsampling of the minority samples and alleviates the label imbalance issue. We don't see the discriminative biases being amplified for most attributes according to Figs.~\ref{fig:attr_bias_celeba_all} and \ref{fig:attr_bias_dp_all} comparing the training ground truth probability and the validation classifier-predicted probability.
The classifiers' performances for each attribute are listed in Tables~\ref{tab:cls_perform_celeba}, \ref{tab:cls_perform_celeba_swint} and \ref{tab:cls_perform_dp}. For both dataset, the accuracy for most attributes is over 80\%. In addition, the SwinTransformer-based classifier and the ResNeXt-based classifier perform similarly. The accuracy differences for most of the attributes are within 2\% difference. Figs.~\ref{fig:attr_dist_celeba_all} and \ref{fig:attr_dist_celeba_all_swint} show the pre-sigmoid logits distributions for each attribute in CelebA dataset using ResNeXt and SwinTransformer backbones respectively. Fig.~\ref{fig:attr_dist_dp_all} depict the pre-sigmoid logits distributions for each attribute in DeepFashion dataset using ResNeXt backbone.

\begin{figure}[tbp]
	\centering
	\begin{subfigure}{0.3\textwidth}
		\includegraphics[width=\textwidth]{Figs/subobj_bias_change_plot_3187_resnext.pdf}
		\caption{ResNeXt-based}
    \end{subfigure}
    \begin{subfigure}{0.3\textwidth}
		\includegraphics[width=\textwidth]{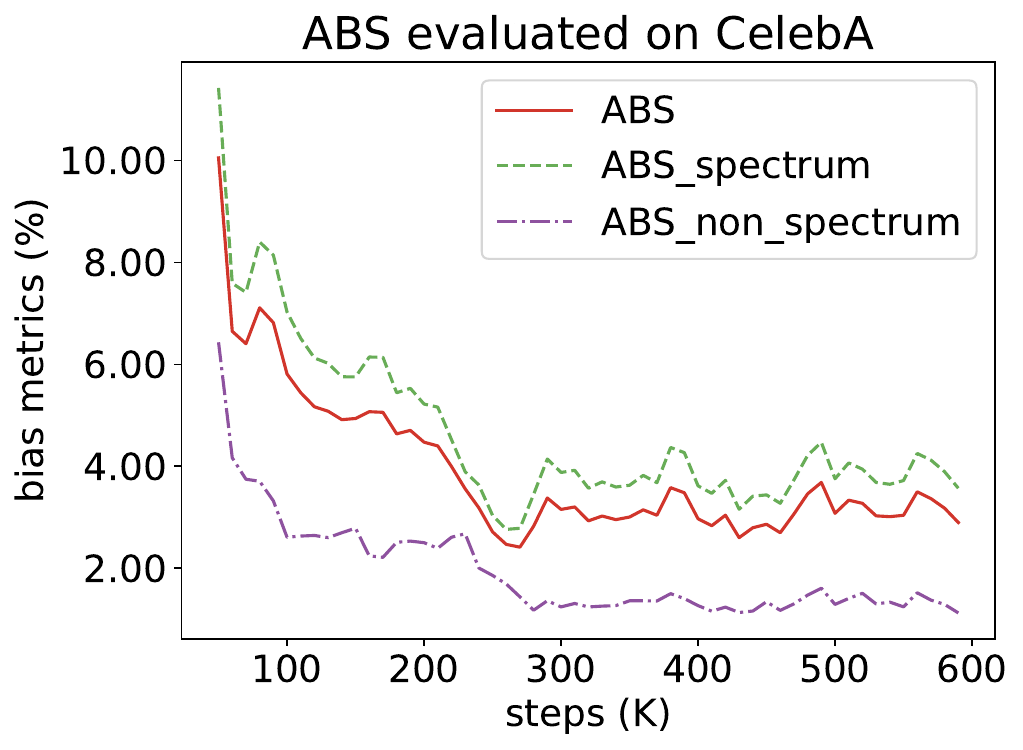}
		\caption{SwinTransformer-based}
    \end{subfigure}
	\caption{ABS for CelebA dataset using different classifiers.}
	\label{fig:abs_classifiers}
\end{figure}

To verify the reliability of our findings with respect to the choice of classifier, we employ two different types of classifier backbones: a Convolutional Neural Network (CNN)-based model (ResNeXt) and a Transformer-based model (Swin Transformer).
As shown in Fig.~\ref{fig:abs_classifiers}, we observe similar bias shifts in both cases.
This comparison indicates that our findings are not overly dependent on the specific classifier architecture used.

To compute spectrum-based and non-spectrum-based bias shift for SwinTransformer-based classifier, we use the same criteria as mentioned in Section~\ref{sec:exp}. The categorization is shown in Table~\ref{tab:attr_cat_swint}. Compared to Table~\ref{tab:attr_cat}, \texttt{5-o-clock\_shadow}, \texttt{Bangs}, and \texttt{Blond\_Hair} are categorized as spectrum-based instead. This is perhaps not surprising since the density of the pre-sigmoid logits distribution from the ResNeXt-based classifier of these attributes is very close to the criteria boundary (0.01), as shown in Fig~\ref{fig:attr_dist_celeba_all}.



We additionally evaluate the Llama 3.2-Vision 11B model\footnote{\url{https://ai.meta.com/blog/llama-3-2-connect-2024-vision-edge-mobile-devices/}} for attribute classification. Using Ollama\footnote{\url{https://ollama.com/}}, we perform zero-shot visual question answering (VQA) tasks to predict attribute labels for each facial image in the validation set. Following this, we compute the classification accuracy for each attribute. All experiments are conducted on an NVIDIA V100 32GB GPU. 
Our findings indicate that the zero-short VQA approach using multi-modal foundation models for attribute classification performs poorly overall, yielding an average accuracy of 77.39\%. Notably, it underperforms on several attributes—including attractive, heavy makeup, mouth slightly open, high cheekbones, oval face, pale skin, pointy nose, and young—each with accuracy below 70\%.

\section{Threshold Selection Informed by Human Labeling Consistency}
\label{app:category_alignment}
We chose our 0.01 decision boundary threshold for our categorization of attributes in alignment with observed human inconsistencies in subjective labeling found in a previous work, Wu et al.~\cite{wu2023consistency}, that studies inter-human agreement in the attribute annotations of the CelebA dataset. 
Specifically, \cite{wu2023consistency} analyzed the CelebA dataset by having two annotators classify 1,000 images and measuring their agreement levels. Attributes are categorized into three groups based on agreement levels: $\geq$95\%, 85–95\%, and $\leq$85\%. We find strong alignment between their agreement-based attribute categorization and our attribute categorization listed in Table~\ref{tab:attr_cat}. 
Most of the non-subjective attributes in our study have $>$85\% agreement and most of the subjective attributes have $\leq$85\% agreement.
Only two non-subjective attributes - pale skin and bangs - have $\leq$85\% agreement, with bangs' agreement being exactly 85\%, leaving pale skin as the sole notable outlier in our comparison. Other exceptions include smiling (level 85-95\%), no beard (level $\geq$95\%) and mouth slightly open (MSO) (level $\geq$95\%), which are categorized as subjective in our study. Interestingly, in Table 3 of Wu et al.~\cite{wu2023consistency} where they conducted a deeper study for attributes in level $\geq$95\%, no beard and MSO - despite their high agreement levels - exhibit large errors relative to the CelebA ground truth, supporting their classification as subjective attributes under our criteria. 

\begin{table}[tbp]
    \centering
    \tiny
    \caption{Attribute categorization of \textit{spectrum-based} and \textit{non-spectrum-based} for CelebA based on SwinTransformer-based classifier.}
    \begin{tabular}{c|l|l}
    \toprule
       Dataset  & \textit{spectrum-based} attributes & \textit{non-spectrum-based} attributes \\
       \midrule
       \multirow{5}{*}{CelebA} 
       & \texttt{Rosy\_Cheeks}, \texttt{Big\_Nose}, \texttt{No\_Beard}, \texttt{Narrow\_Eyes}, \texttt{Arched\_Eyebrows}, &  \\
        &  \texttt{High\_Cheekbones}, \texttt{Bushy\_Eyebrows}, \texttt{Black\_Hair}, \texttt{Receding\_Hairline},  & \texttt{Eyeglasses}, \texttt{Bald}, \texttt{Double\_Chin}, \\
        & \texttt{Brown\_Hair}, \texttt{Straight\_Hair}, \texttt{Bags\_Under\_Eyes}, \texttt{Pointy\_Nose}, & \texttt{Wearing\_Hat}, \texttt{Male},  \\
        & \texttt{Big\_Lips}, \texttt{Mouth\_Slightly\_Open}, \texttt{Heavy\_Makeup}, \texttt{Attractive}, & \texttt{Gray\_Hair}, \texttt{Mustache}, \texttt{Chubby}, \\
        & \texttt{Smiling}, \texttt{Wearing\_Lipstick}, \texttt{Wavy\_Hair}, \texttt{Young}, \texttt{Oval\_Face}, & \texttt{Pale\_Skin}, \texttt{Sideburns},\texttt{Goatee},   \\
        & \texttt{5-o-Clock\_Shadow}, \texttt{Bangs}, \texttt{Blond\_Hair} & \\
       \bottomrule
    \end{tabular}
    \label{tab:attr_cat_swint}
\end{table}

\begin{table}[tbp]
\small
\centering
\caption{Classifier (ResNeXt-based) performance on validation set of CelebA.}
\begin{tabular}{llllll} \toprule
	Attr & Accuracy & Precision & Recall & F1 & AUPR \\ \midrule
	Eyeglasses & 99.58 & 96.90 & 97.11 & 97.00 & 98.74 \\
	Wearing\_Hat & 99.10 & 87.95 & 93.94 & 90.84 & 94.94 \\
	Bald & 98.86 & 70.98 & 76.16 & 73.47 & 78.84 \\
	Male & 98.40 & 97.88 & 98.37 & 98.12 & 99.82 \\
	Gray\_Hair & 97.73 & 75.42 & 79.32 & 77.32 & 83.19 \\
	Sideburns & 97.27 & 79.65 & 81.05 & 80.35 & 88.54 \\
	Goatee & 96.52 & 73.61 & 82.31 & 77.72 & 83.77 \\
	Pale\_Skin & 96.43 & 58.31 & 59.81 & 59.05 & 61.02 \\
	Blurry & 96.18 & 58.69 & 65.00 & 61.69 & 62.35 \\
	Double\_Chin & 96.10 & 60.08 & 61.44 & 60.75 & 62.38 \\
	Mustache & 95.98 & 60.53 & 58.72 & 59.62 & 62.67 \\
	Bangs & 95.57 & 83.24 & 87.41 & 85.27 & 91.16 \\
	Chubby & 95.29 & 61.80 & 60.28 & 61.03 & 62.52 \\
	Blond\_Hair & 95.15 & 82.48 & 86.91 & 84.64 & 91.59 \\
	No\_Beard & 95.14 & 98.54 & 95.51 & 97.00 & 99.69 \\
	Wearing\_Necktie & 95.02 & 65.17 & 67.43 & 66.28 & 69.49 \\
	Rosy\_Cheeks & 94.54 & 59.06 & 65.54 & 62.13 & 63.69 \\
	Receding\_Hairline & 93.79 & 56.03 & 63.40 & 59.49 & 61.72 \\
	5\_o\_Clock\_Shadow & 93.49 & 71.44 & 74.67 & 73.02 & 80.00 \\
	Mouth\_Slightly\_Open & 93.11 & 93.11 & 92.56 & 92.83 & 97.99 \\
	Wearing\_Lipstick & 92.37 & 88.20 & 95.69 & 91.79 & 97.72 \\
	Smiling & 92.17 & 92.42 & 91.28 & 91.85 & 97.90 \\
	Heavy\_Makeup & 91.70 & 88.30 & 90.75 & 89.51 & 96.50 \\
	Bushy\_Eyebrows & 91.58 & 70.71 & 69.83 & 70.27 & 76.57 \\
	Narrow\_Eyes & 90.90 & 42.38 & 58.91 & 49.30 & 47.99 \\
	Wearing\_Earings & 90.56 & 75.01 & 75.75 & 75.38 & 81.54 \\
	Black\_Hair & 89.94 & 73.51 & 80.96 & 77.06 & 83.49 \\
	High\_Cheekbones & 86.60 & 84.89 & 85.37 & 85.13 & 93.84 \\
	Young & 85.58 & 92.38 & 87.94 & 90.11 & 96.98 \\
	Wearing\_Necklace & 85.25 & 38.43 & 37.10 & 37.76 & 35.80 \\
	Brown\_Hair & 84.29 & 68.13 & 65.53 & 66.81 & 71.82 \\
	Arched\_Eyebrows & 83.88 & 67.57 & 72.32 & 69.87 & 76.08 \\
	Wavy\_Hair & 82.91 & 65.67 & 80.04 & 72.15 & 79.15 \\
	Bags\_Under\_Eyes & 82.27 & 57.18 & 57.75 & 57.47 & 58.43 \\
	Straight\_Hair & 81.95 & 55.87 & 58.16 & 56.99 & 59.65 \\
	Big\_Nose & 81.57 & 64.12 & 58.89 & 61.39 & 68.16 \\
	Attractive & 80.32 & 80.99 & 81.23 & 81.11 & 90.79 \\
	Big\_Lips & 76.61 & 31.35 & 44.25 & 36.70 & 34.41 \\
	Pointy\_Nose & 72.71 & 51.99 & 54.73 & 53.33 & 55.88 \\
	Oval\_Face & 70.91 & 48.00 & 46.18 & 47.07 & 50.47 \\ \bottomrule
\end{tabular}
\label{tab:cls_perform_celeba}
\end{table}

\begin{table}[tbp]
\small
\centering
\caption{Classifier (SwinTransformer-based) performance on validation set of CelebA.}
\begin{tabular}{llllll} \toprule
Attr & Accuracy & Precision & Recall & F1 & AUPR \\ \midrule
Eyeglasses & 99.52 & 95.87 & 97.32 & 96.59 & 98.84 \\
Wearing\_Hat & 99.16 & 87.71 & 95.64 & 91.50 & 96.27 \\
Bald & 98.83 & 66.98 & 85.40 & 75.08 & 82.05 \\
Male & 98.68 & 98.21 & 98.69 & 98.45 & 99.89 \\
Gray\_Hair & 97.65 & 71.69 & 85.63 & 78.04 & 84.49 \\
Sideburns & 96.83 & 71.40 & 90.05 & 79.65 & 89.08 \\
Goatee & 96.17 & 68.34 & 89.48 & 77.49 & 85.30 \\
Pale\_Skin & 95.77 & 50.68 & 69.63 & 58.66 & 62.08 \\
Mustache & 95.58 & 54.52 & 75.17 & 63.20 & 65.21 \\
Blurry & 95.57 & 52.26 & 73.94 & 61.23 & 64.76 \\
Double\_Chin & 95.17 & 50.55 & 74.77 & 60.32 & 64.65 \\
Bangs & 95.14 & 78.53 & 92.08 & 84.76 & 92.90 \\
No\_Beard & 94.95 & 98.88 & 94.93 & 96.86 & 99.72 \\
Blond\_Hair & 94.67 & 77.02 & 93.13 & 84.31 & 93.24 \\
Chubby & 94.44 & 53.42 & 71.30 & 61.08 & 63.34 \\
Wearing\_Necktie & 93.96 & 55.90 & 79.76 & 65.73 & 72.23 \\
Mouth\_Slightly\_Open & 93.39 & 92.92 & 93.40 & 93.16 & 98.37 \\
Receding\_Hairline & 93.21 & 51.94 & 75.86 & 61.66 & 64.94 \\
Rosy\_Cheeks & 92.87 & 48.71 & 81.96 & 61.10 & 66.06 \\
Smiling & 92.75 & 92.35 & 92.67 & 92.51 & 98.30 \\
Wearing\_Lipstick & 92.50 & 87.51 & 97.02 & 92.02 & 98.08 \\
5\_o\_Clock\_Shadow & 92.32 & 62.53 & 87.16 & 72.82 & 82.13 \\
Heavy\_Makeup & 91.74 & 86.18 & 93.87 & 89.86 & 96.96 \\
Bushy\_Eyebrows & 89.87 & 60.81 & 81.38 & 69.61 & 79.60 \\
Black\_Hair & 89.47 & 69.10 & 89.58 & 78.02 & 86.79 \\
Wearing\_Earings & 89.46 & 67.65 & 85.72 & 75.62 & 84.35 \\
High\_Cheekbones & 87.77 & 85.01 & 88.36 & 86.65 & 94.96 \\
Narrow\_Eyes & 87.20 & 34.05 & 75.13 & 46.86 & 50.46 \\
Young & 86.28 & 92.99 & 88.28 & 90.58 & 97.27 \\
Brown\_Hair & 83.42 & 61.89 & 81.47 & 70.34 & 75.91 \\
Arched\_Eyebrows & 83.25 & 63.11 & 84.73 & 72.34 & 78.19 \\
Wavy\_Hair & 81.88 & 61.98 & 89.23 & 73.15 & 83.71 \\
Attractive & 81.34 & 80.60 & 84.44 & 82.47 & 91.93 \\
Straight\_Hair & 81.17 & 52.96 & 75.45 & 62.24 & 65.96 \\
Bags\_Under\_Eyes & 81.02 & 53.01 & 74.79 & 62.04 & 62.81 \\
Big\_Nose & 80.36 & 58.64 & 71.45 & 64.42 & 70.26 \\
Wearing\_Necklace & 79.96 & 32.65 & 62.27 & 42.84 & 39.86 \\
Big\_Lips & 70.71 & 29.46 & 65.41 & 40.62 & 38.58 \\
Pointy\_Nose & 70.17 & 48.40 & 71.24 & 57.64 & 60.14 \\
Oval\_Face & 69.12 & 46.29 & 63.85 & 53.67 & 56.70 \\ \bottomrule
\end{tabular}
\label{tab:cls_perform_celeba_swint}
\end{table}

\begin{table}[tbp]
\small
\centering
\caption{Classifier performance on validation set of DeepFashion.}
\begin{tabular}{l|ccccc}
\toprule
Attr             & Acc   & Precision & Recall & F1    & AUPR  \\ \midrule
lattice          & 99.48 & 100.00    & 50.00  & 66.67 & 50.52 \\
square\_neckline & 98.97 & 0.00      & 0.00   & 0.00  & 1.03  \\
faux             & 98.45 & 50.00     & 33.33  & 40.00 & 17.70 \\
leather          & 97.94 & 0.00      & 0.00   & 0.00  & 1.03  \\
pleated          & 97.42 & 40.00     & 50.00  & 44.45 & 21.03 \\
maxi\_length     & 96.91 & 96.00     & 82.76  & 88.89 & 82.03 \\
denim            & 96.91 & 87.50     & 58.33  & 70.00 & 53.62 \\
striped          & 96.39 & 55.56     & 62.50  & 58.82 & 36.27 \\
loose            & 94.33 & 60.00     & 25.00  & 35.29 & 19.64 \\
knit             & 92.27 & 52.63     & 62.50  & 57.14 & 35.99 \\
mini\_length     & 91.24 & 75.61     & 81.58  & 78.48 & 65.29 \\
graphic          & 90.72 & 69.70     & 74.19  & 71.88 & 55.83 \\
embroidered      & 90.72 & 36.36     & 26.67  & 30.77 & 15.37 \\
long\_sleeve     & 90.72 & 82.54     & 88.14  & 85.25 & 76.36 \\
short\_sleeve    & 90.21 & 66.67     & 73.33  & 69.84 & 53.01 \\
no\_dress        & 90.21 & 90.91     & 94.49  & 92.66 & 89.51 \\
solid            & 88.14 & 88.89     & 88.00  & 88.44 & 88.41 \\
floral           & 87.63 & 61.90     & 76.47  & 68.42 & 51.46 \\
tight            & 87.63 & 61.29     & 61.29  & 61.29 & 43.75 \\
chiffon          & 87.11 & 57.69     & 51.72  & 54.55 & 37.06 \\
v\_neckline      & 86.60 & 70.83     & 47.22  & 56.67 & 43.24 \\
sleeveless       & 86.08 & 86.79     & 87.62  & 87.20 & 82.75 \\
conventional     & 80.93 & 86.54     & 89.40  & 87.95 & 85.62 \\
no\_neckline     & 75.26 & 71.26     & 72.94  & 72.09 & 63.84 \\
cotton           & 75.26 & 81.34     & 82.58  & 81.95 & 79.03 \\
crew\_neckline   & 71.65 & 59.30     & 71.83  & 64.97 & 52.91 \\ \bottomrule
\end{tabular}
\label{tab:cls_perform_dp}
\end{table}


\clearpage

\section{Bias Shift Analysis Per Attribute}
\label{app:perattribute}
\begin{figure}[!htbp]
    \centering
    \begin{subfigure}{0.19\textwidth}
		\includegraphics[width=\textwidth]{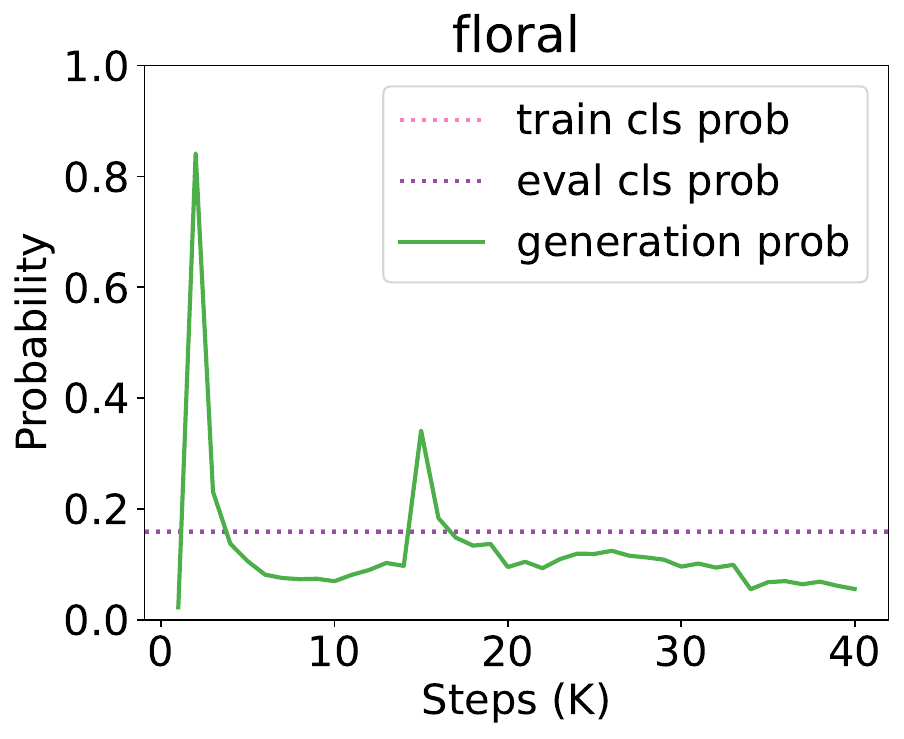}
		\caption{Floral}
    \end{subfigure}
    \begin{subfigure}{0.19\textwidth}
		\includegraphics[width=\textwidth]{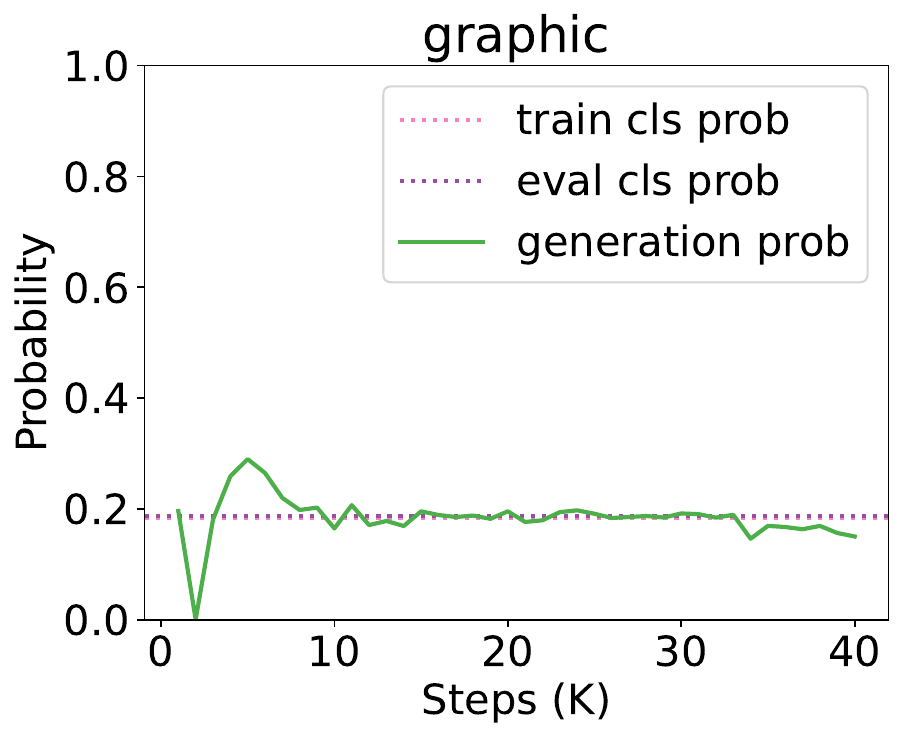}
		\caption{Graphic}
    \end{subfigure}
     \begin{subfigure}{0.19\textwidth}
		\includegraphics[width=\textwidth]{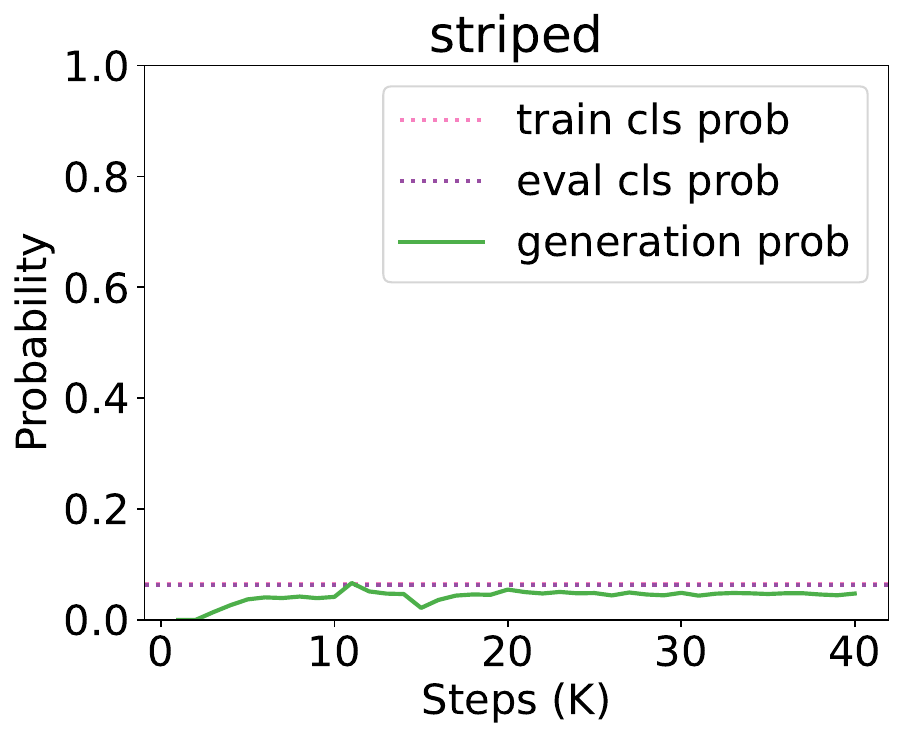}
		\caption{Striped}
	    \end{subfigure}
    \begin{subfigure}{0.19\textwidth}
		\includegraphics[width=\textwidth]{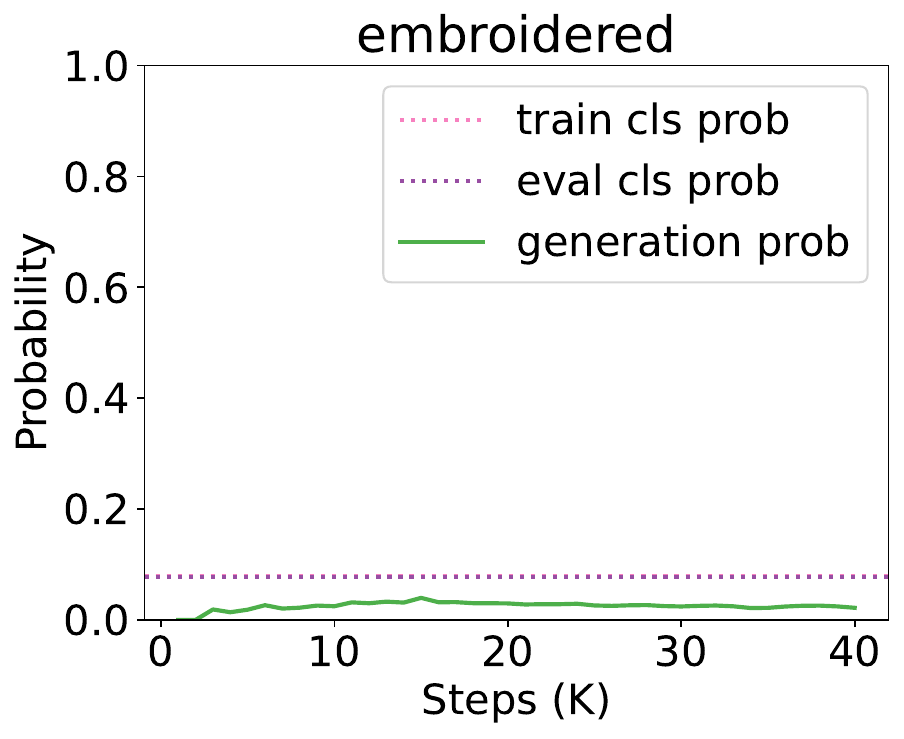}
		\caption{Embroidered}
	    \end{subfigure}
    \begin{subfigure}{0.19\textwidth}
		\includegraphics[width=\textwidth]{Figs/marginal_bias_curves_dp/pleated.pdf}
		\caption{Pleated}
    \end{subfigure}
     \vfill
    \begin{subfigure}{0.19\textwidth}
		\includegraphics[width=\textwidth]{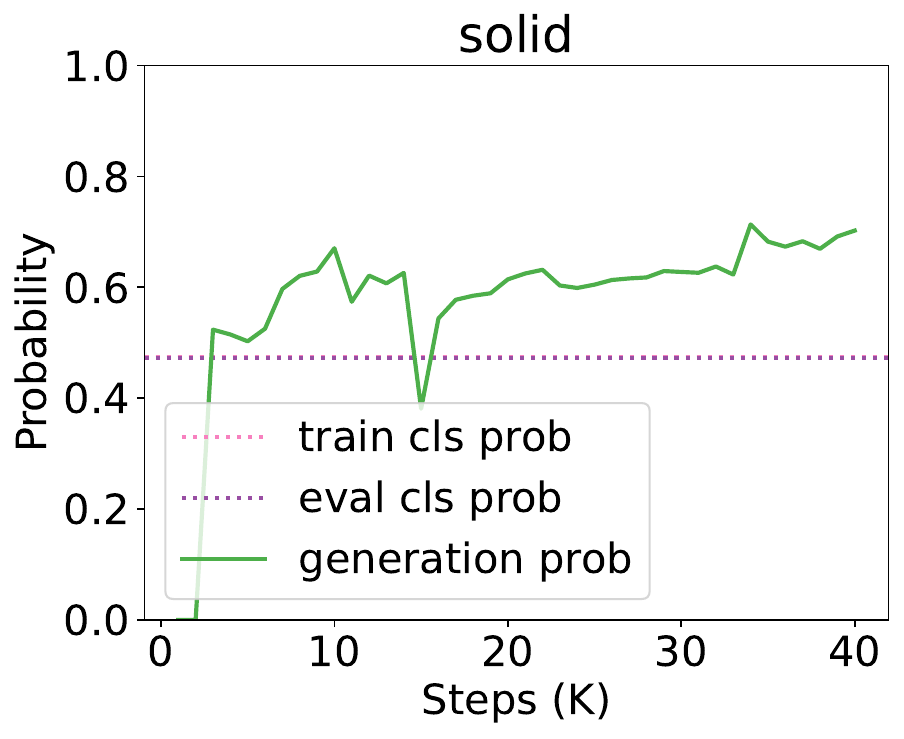}
		\caption{Solid}
    \end{subfigure}
    \begin{subfigure}{0.19\textwidth}
		\includegraphics[width=\textwidth]{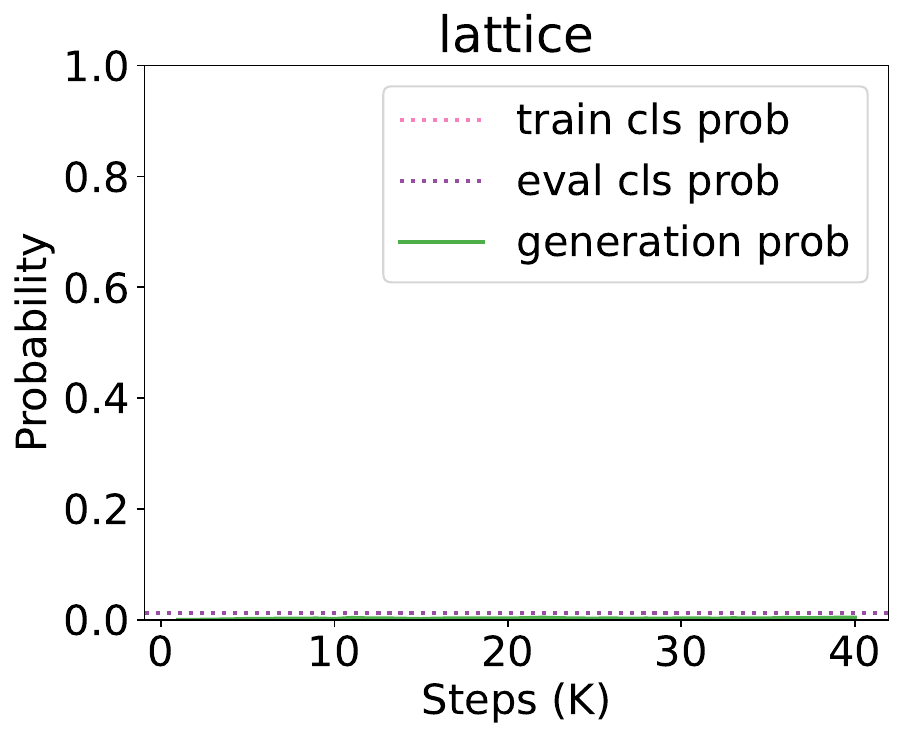}
		\caption{Lattice}
    \end{subfigure}
     \begin{subfigure}{0.19\textwidth}
		\includegraphics[width=\textwidth]{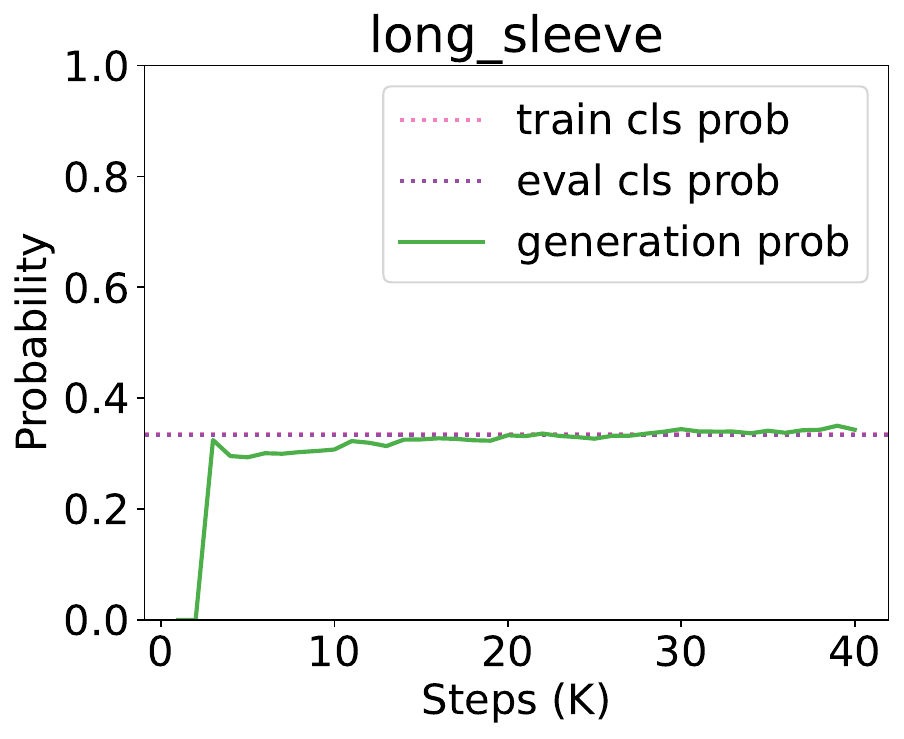}
		\caption{Long Sleeve}
	    \end{subfigure}
    \begin{subfigure}{0.19\textwidth}
		\includegraphics[width=\textwidth]{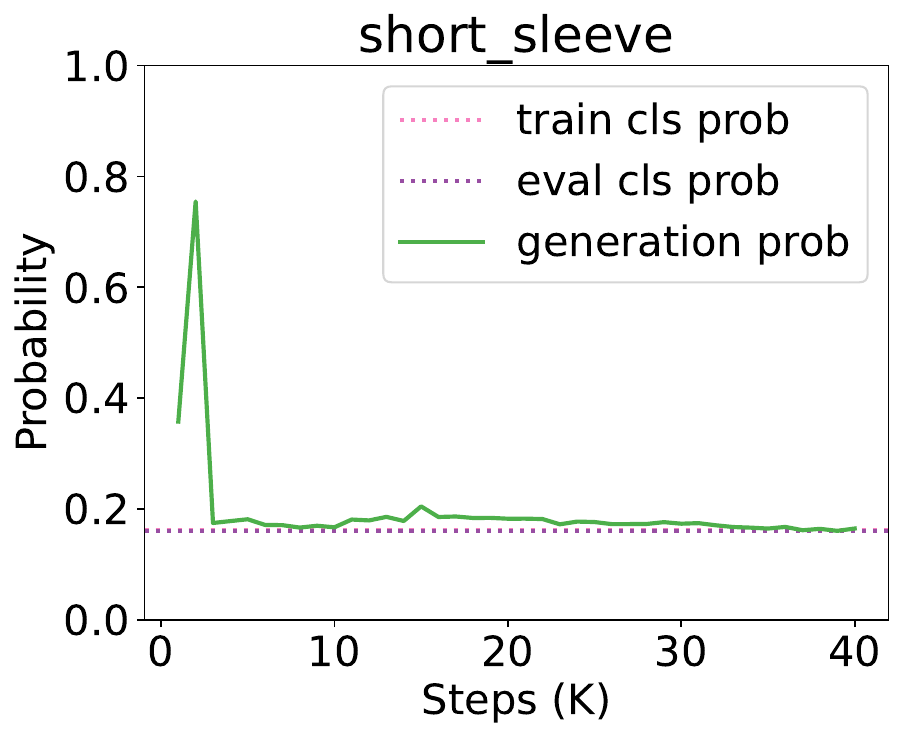}
		\caption{Short Sleeve}
	    \end{subfigure}
    \begin{subfigure}{0.19\textwidth}
		\includegraphics[width=\textwidth]{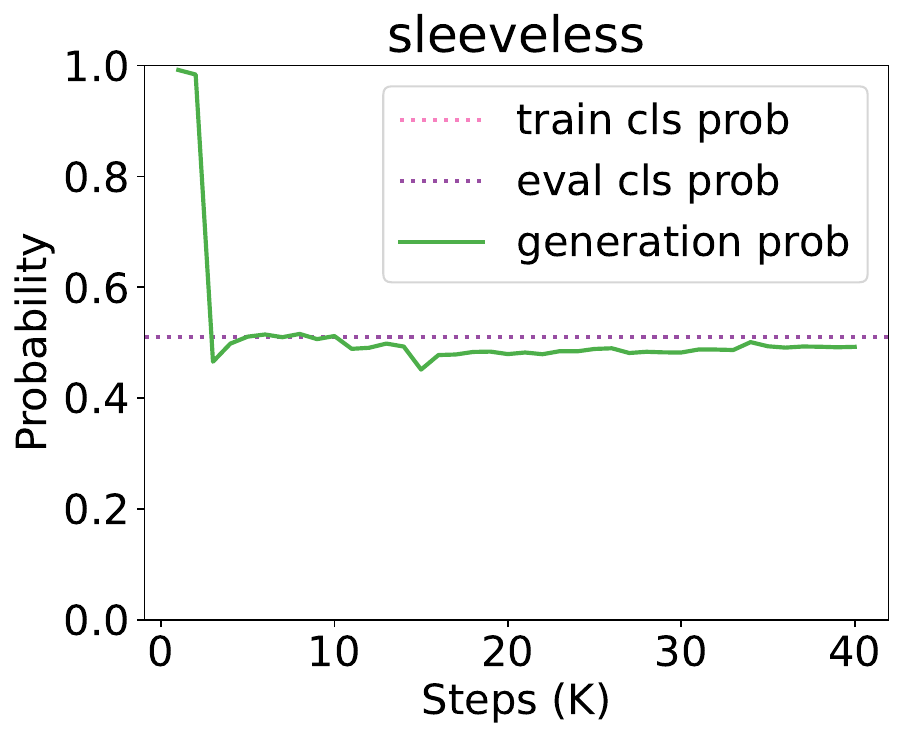}
		\caption{Sleeveless}
    \end{subfigure}
    \vfill
    \begin{subfigure}{0.19\textwidth}
		\includegraphics[width=\textwidth]{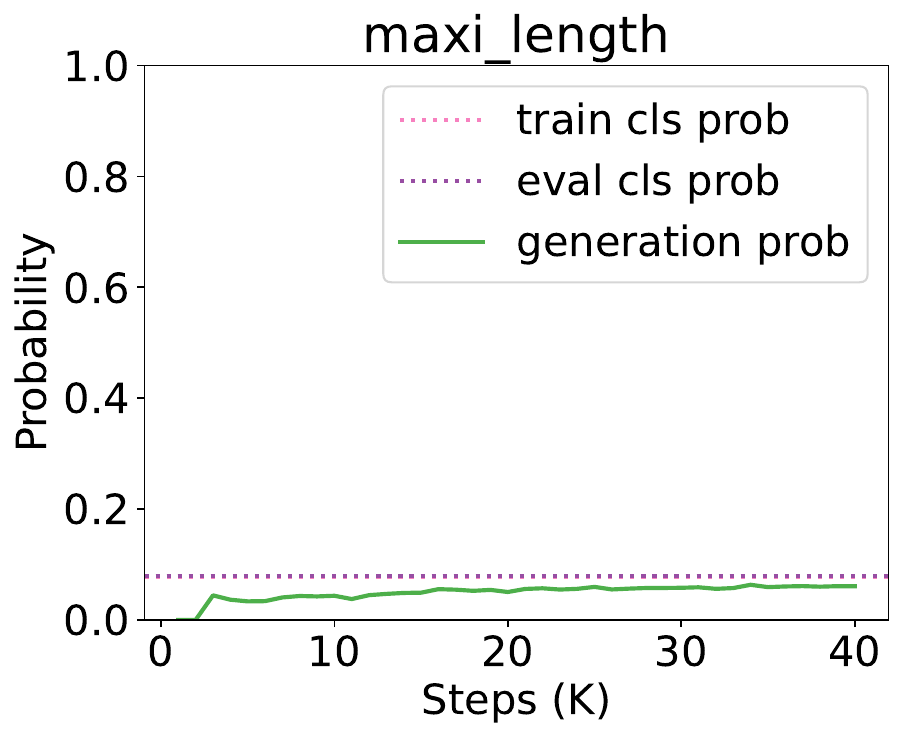}
		\caption{Maxi Length}
    \end{subfigure}
    \begin{subfigure}{0.19\textwidth}
		\includegraphics[width=\textwidth]{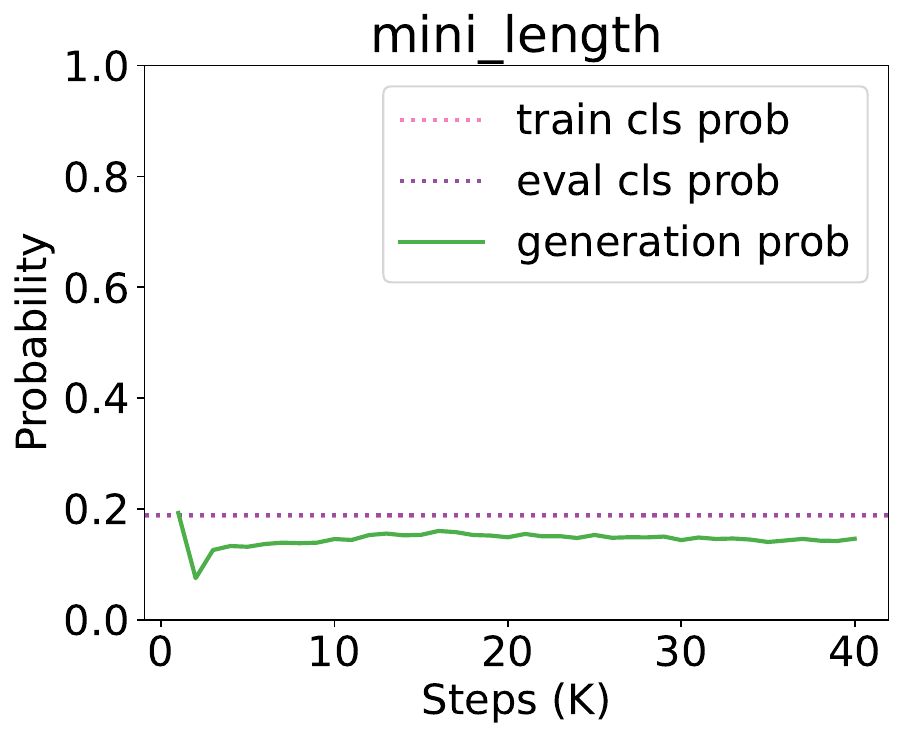}
		\caption{Mini Length}
    \end{subfigure}
     \begin{subfigure}{0.19\textwidth}
		\includegraphics[width=\textwidth]{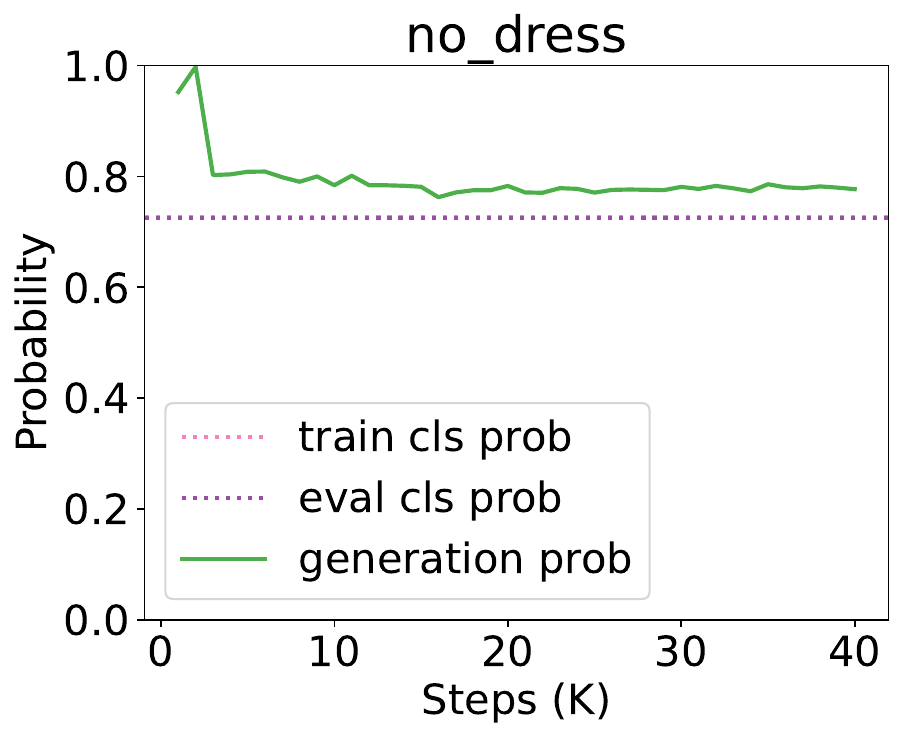}
		\caption{No Dress}
	    \end{subfigure}
    \begin{subfigure}{0.19\textwidth}
		\includegraphics[width=\textwidth]{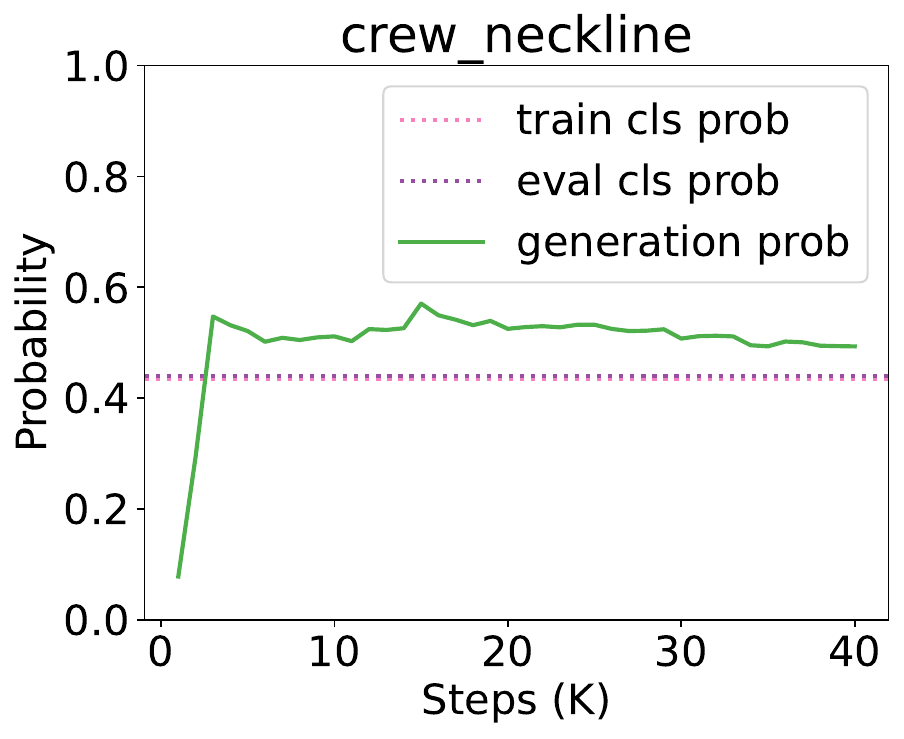}
		\caption{Crew Neckline}
	    \end{subfigure}
    \begin{subfigure}{0.19\textwidth}
		\includegraphics[width=\textwidth]{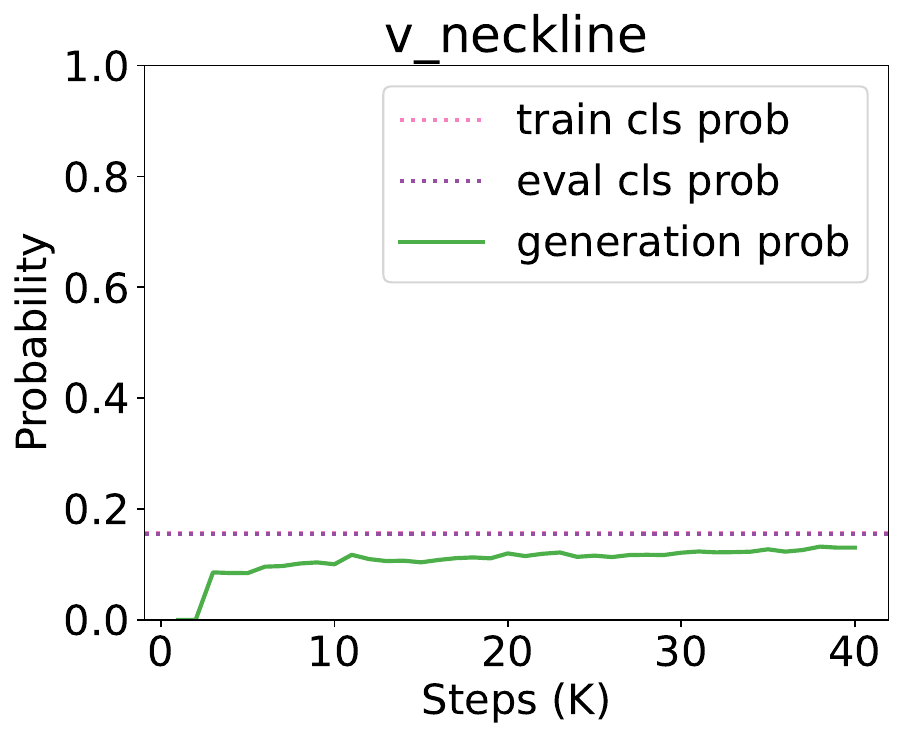}
		\caption{V Neckline}
    \end{subfigure}
    \vfill
    \begin{subfigure}{0.19\textwidth}
		\includegraphics[width=\textwidth]{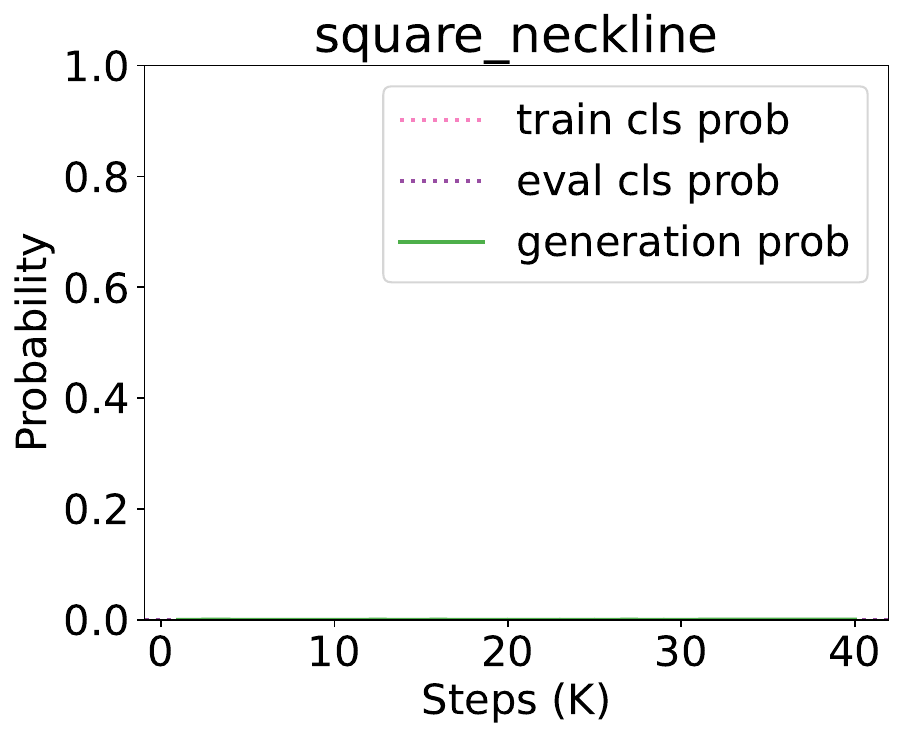}
		\caption{Square Neckline}
    \end{subfigure}
    \begin{subfigure}{0.19\textwidth}
		\includegraphics[width=\textwidth]{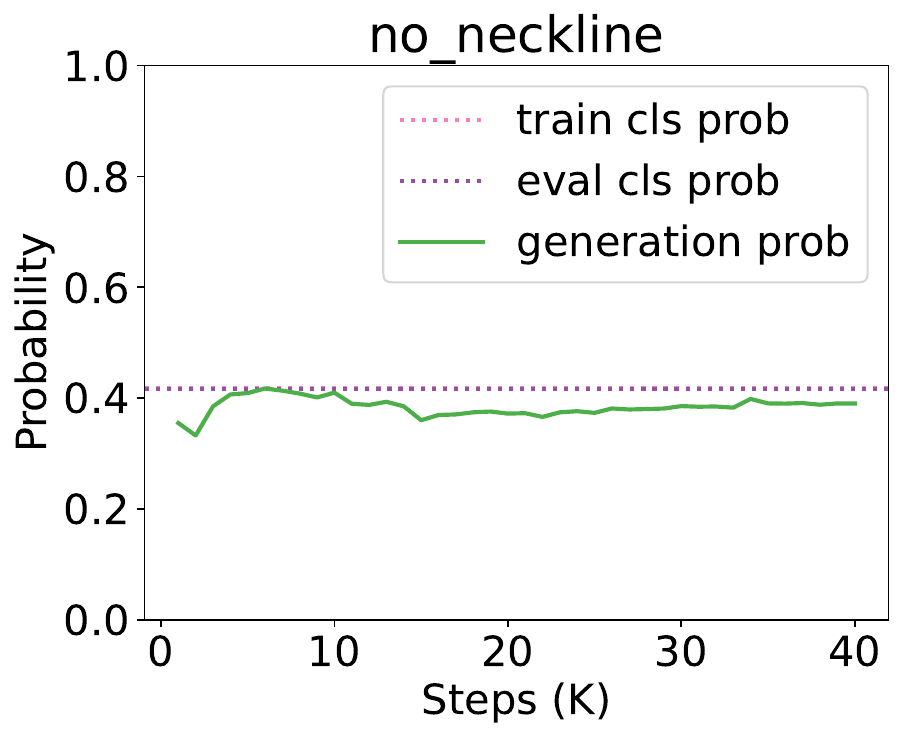}
		\caption{No Neckline}
    \end{subfigure}
     \begin{subfigure}{0.19\textwidth}
		\includegraphics[width=\textwidth]{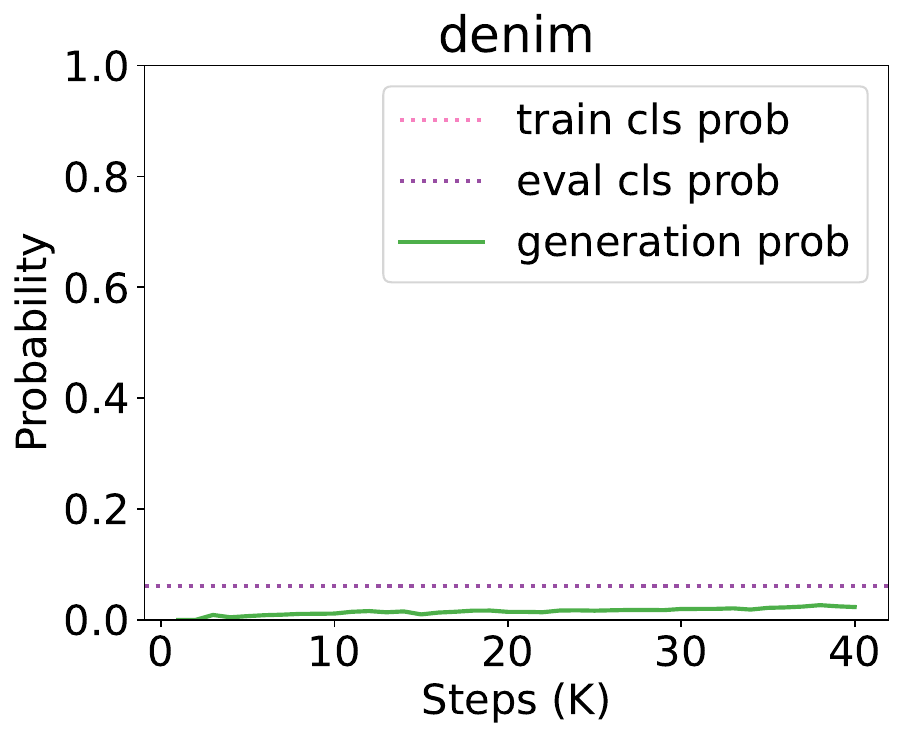}
		\caption{Denim}
	    \end{subfigure}
    \begin{subfigure}{0.19\textwidth}
		\includegraphics[width=\textwidth]{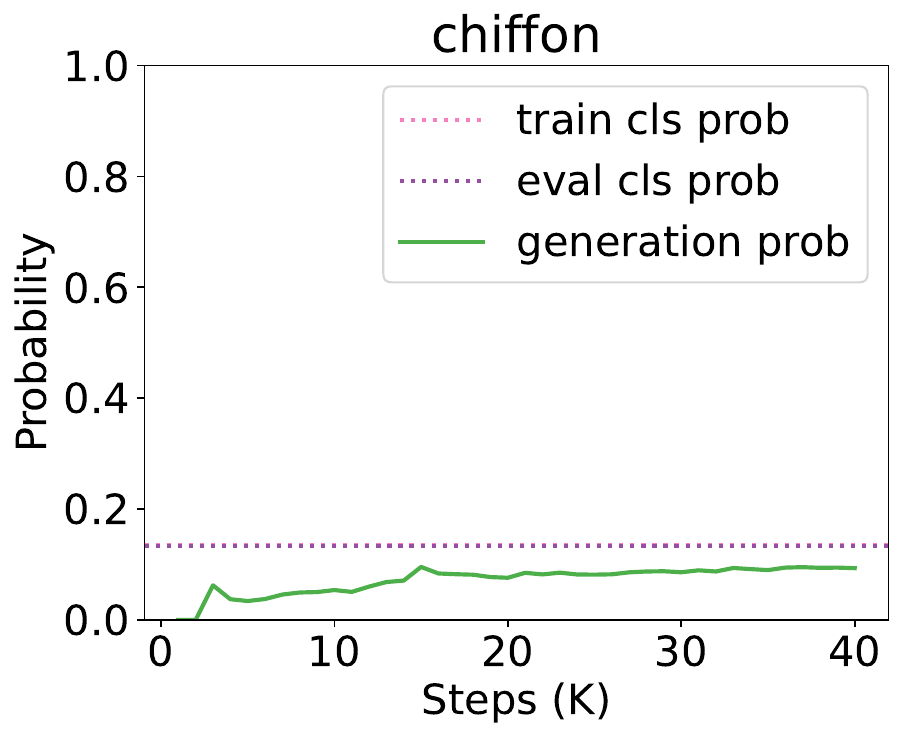}
		\caption{Chiffon}
	    \end{subfigure}
    \begin{subfigure}{0.19\textwidth}
		\includegraphics[width=\textwidth]{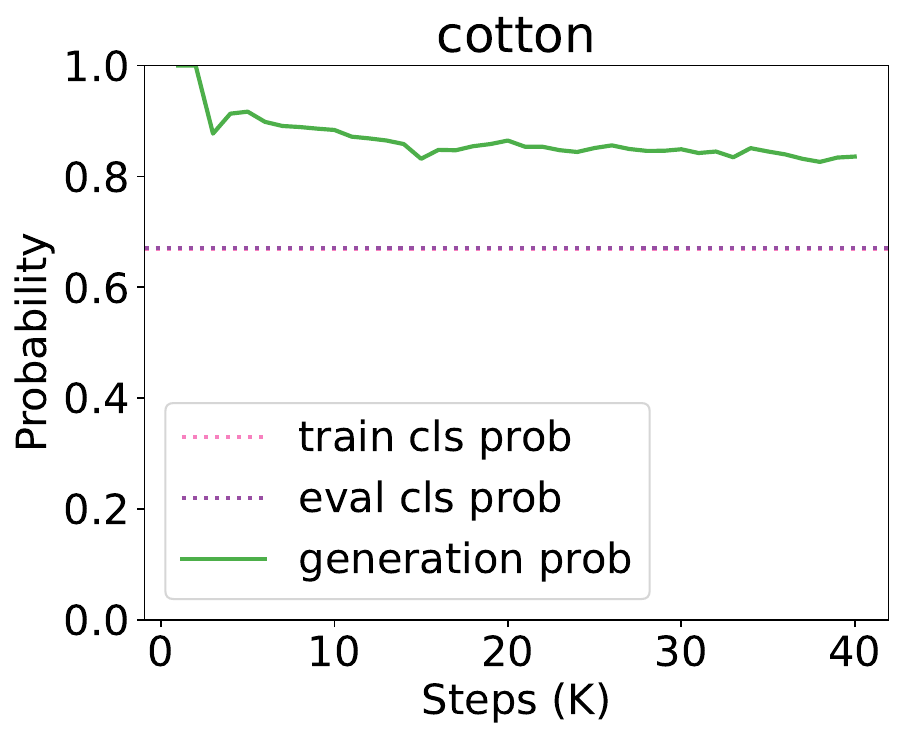}
		\caption{Cotton}
    \end{subfigure}
    \vfill
    \begin{subfigure}{0.19\textwidth}
		\includegraphics[width=\textwidth]{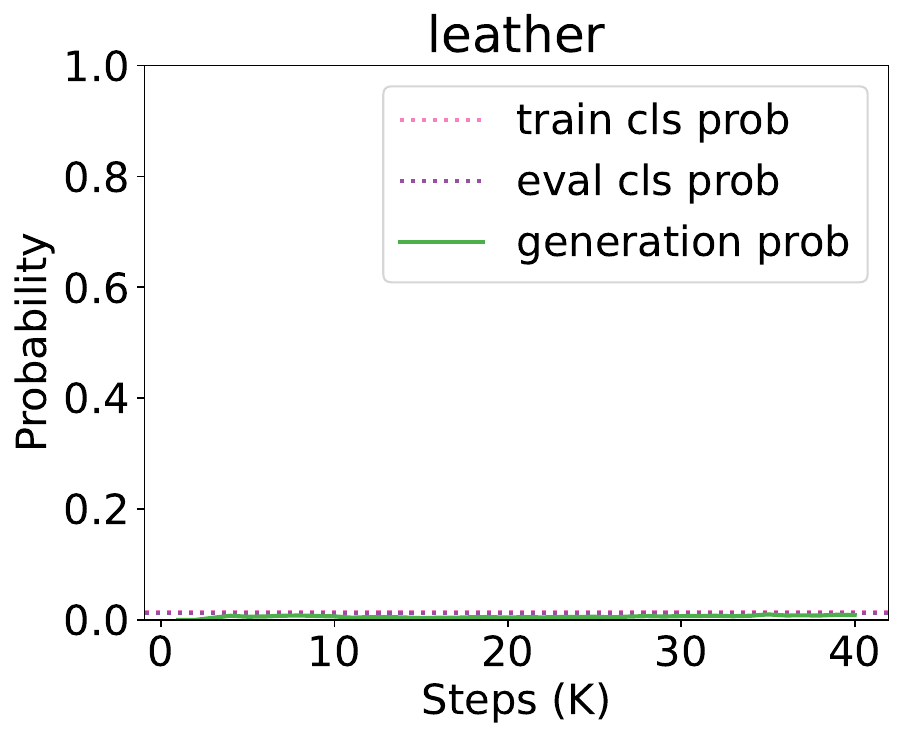}
		\caption{Leather}
    \end{subfigure}
    \begin{subfigure}{0.19\textwidth}
		\includegraphics[width=\textwidth]{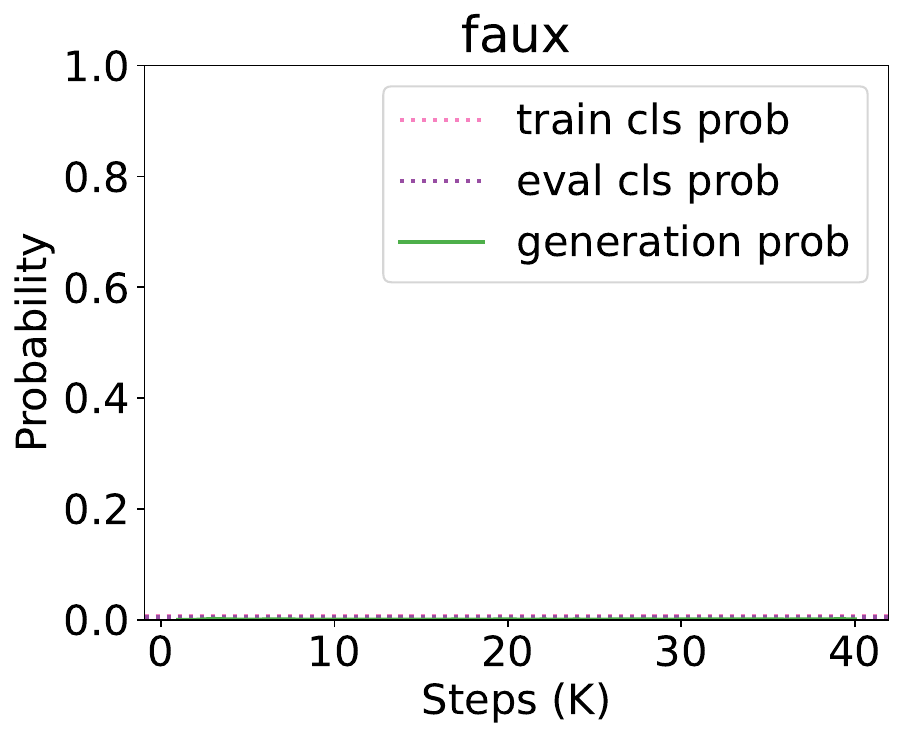}
		\caption{Faux}
    \end{subfigure}
     \begin{subfigure}{0.19\textwidth}
		\includegraphics[width=\textwidth]{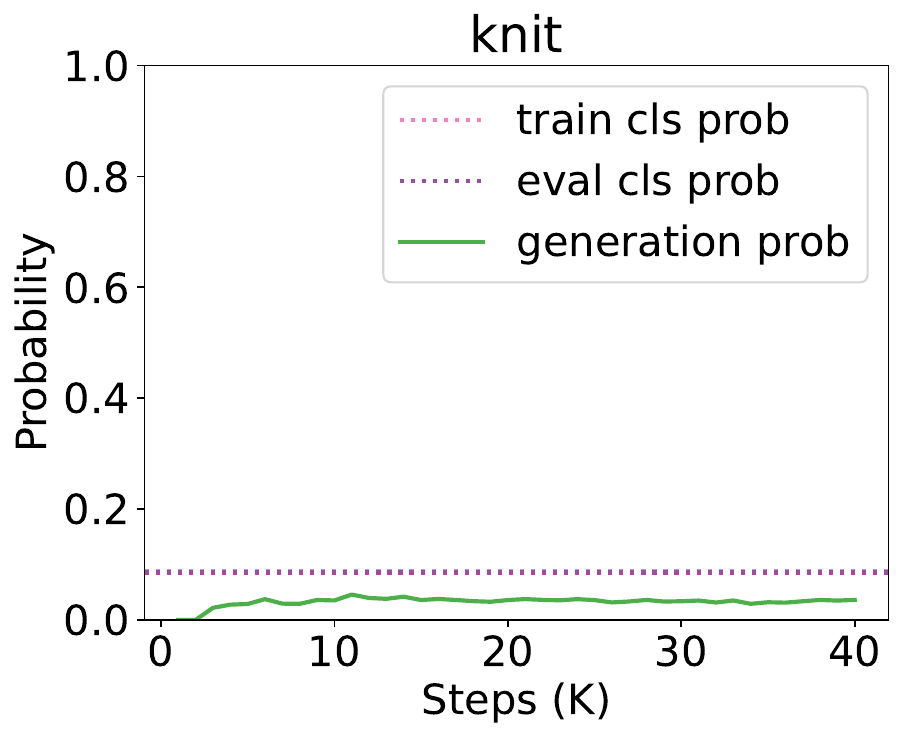}
		\caption{Knit}
	    \end{subfigure}
    \begin{subfigure}{0.19\textwidth}
		\includegraphics[width=\textwidth]{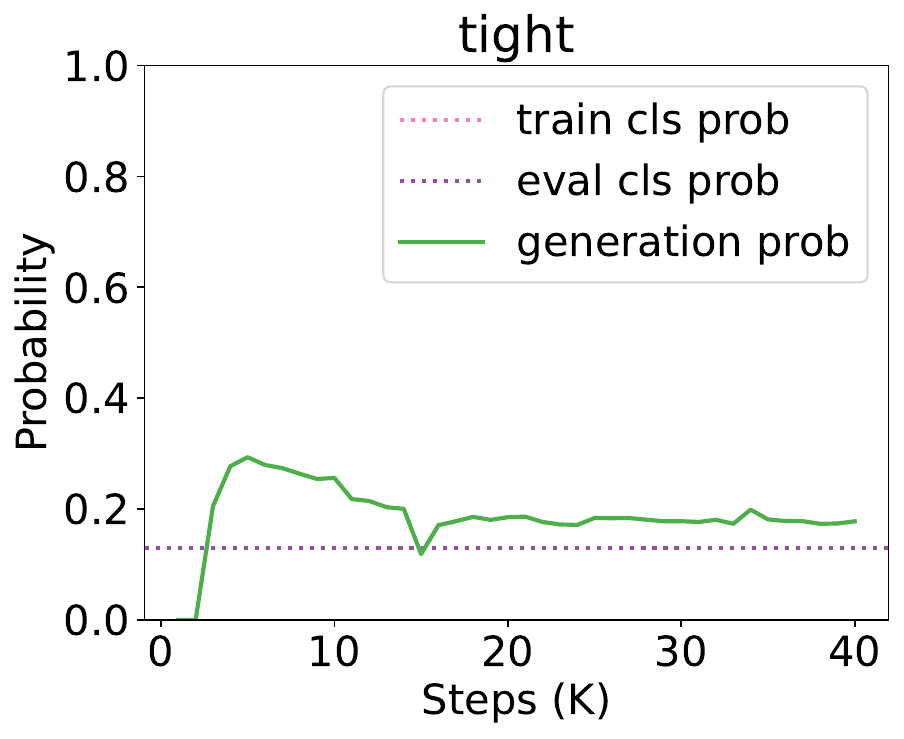}
		\caption{Tight}
	    \end{subfigure}
    \begin{subfigure}{0.19\textwidth}
		\includegraphics[width=\textwidth]{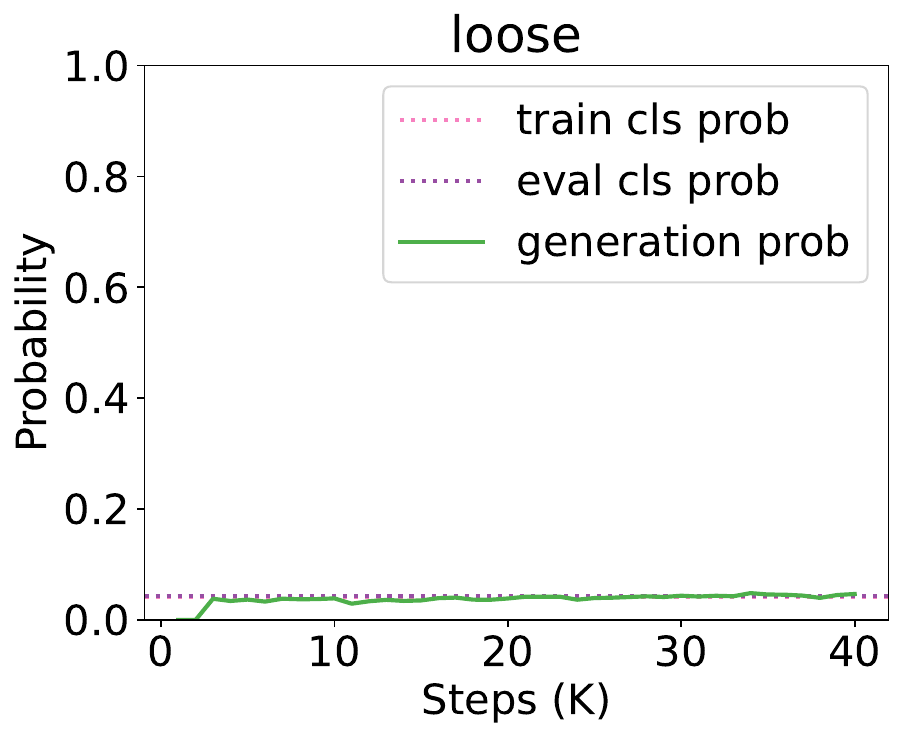}
		\caption{Loose}
    \end{subfigure}
    \vfill
    \begin{subfigure}{0.19\textwidth}
		\includegraphics[width=\textwidth]{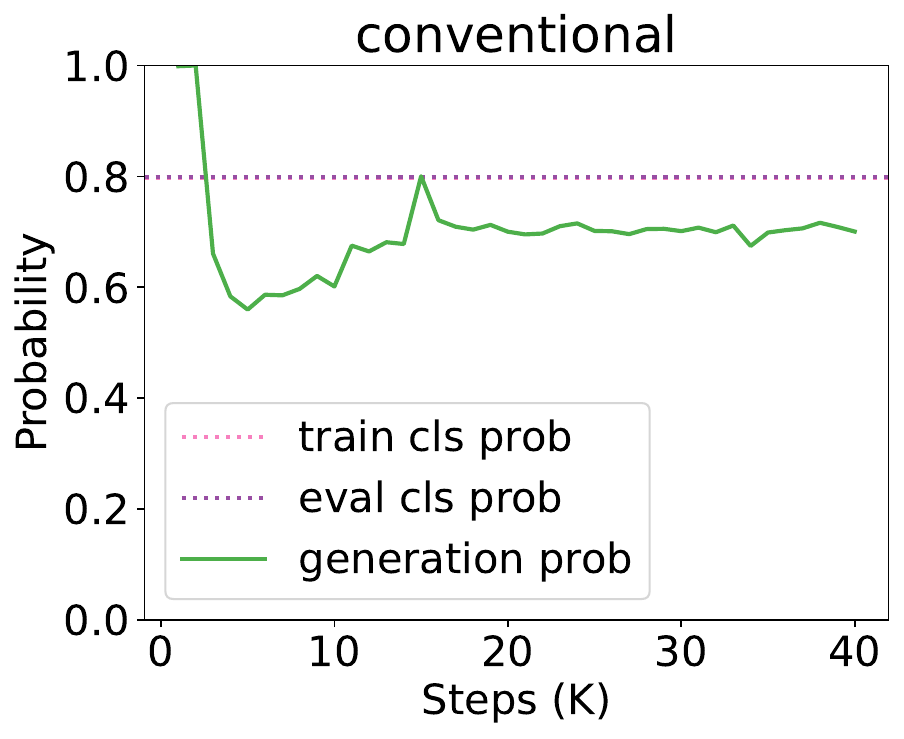}
		\caption{Conventional}
        \label{fig:conventional}
    \end{subfigure}
	\caption{Probabilities of attributes for DeepFashion dataset during training. Please note that it might seem like some of the subplots are missing the probability lines; they are actually very close to the x-axis, especially for \textit{Square Neckline} and \textit{Faux}.}
	\label{fig:attr_bias_dp_all}
\end{figure}

\begin{figure}[!tbp]
    \centering
    \begin{subfigure}{0.19\textwidth}
		\includegraphics[width=\textwidth]{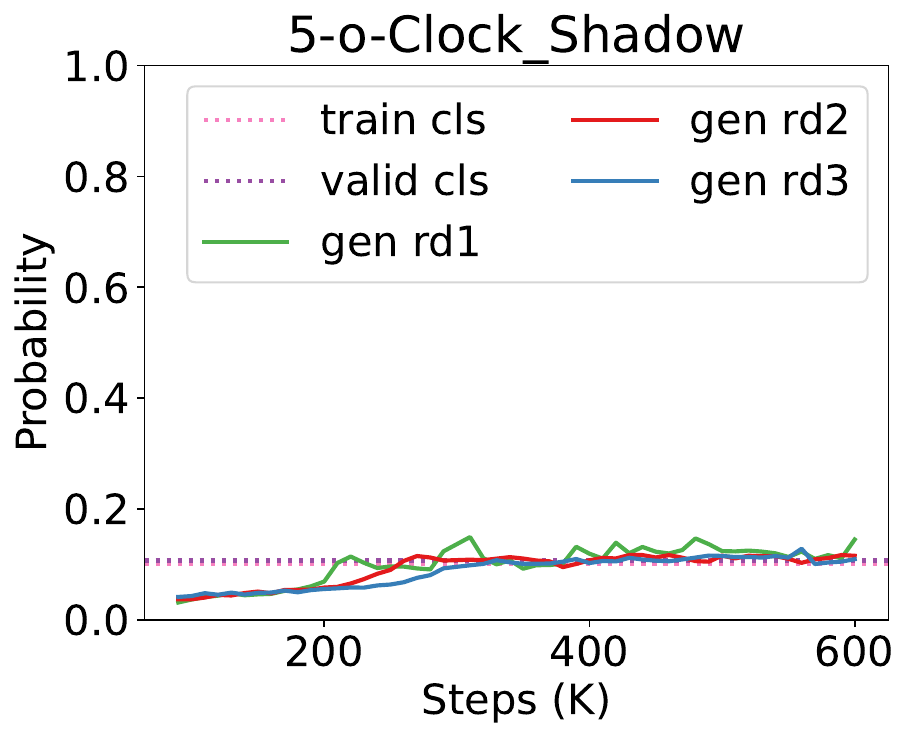}
		\caption{5o Clock Shadow}
    \end{subfigure}
    \begin{subfigure}{0.19\textwidth}
		\includegraphics[width=\textwidth]{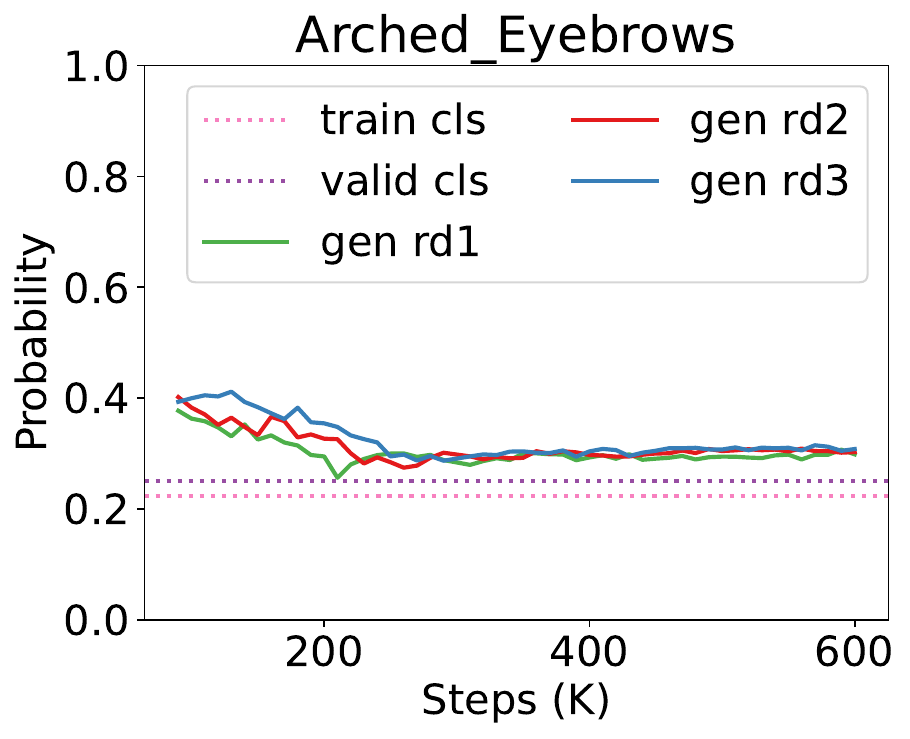}
		\caption{\begin{tiny}
		    Arched Eyebrows
		\end{tiny}}
        \label{fig:arched_eyebrows}
    \end{subfigure}
     \begin{subfigure}{0.19\textwidth}
		\includegraphics[width=\textwidth]{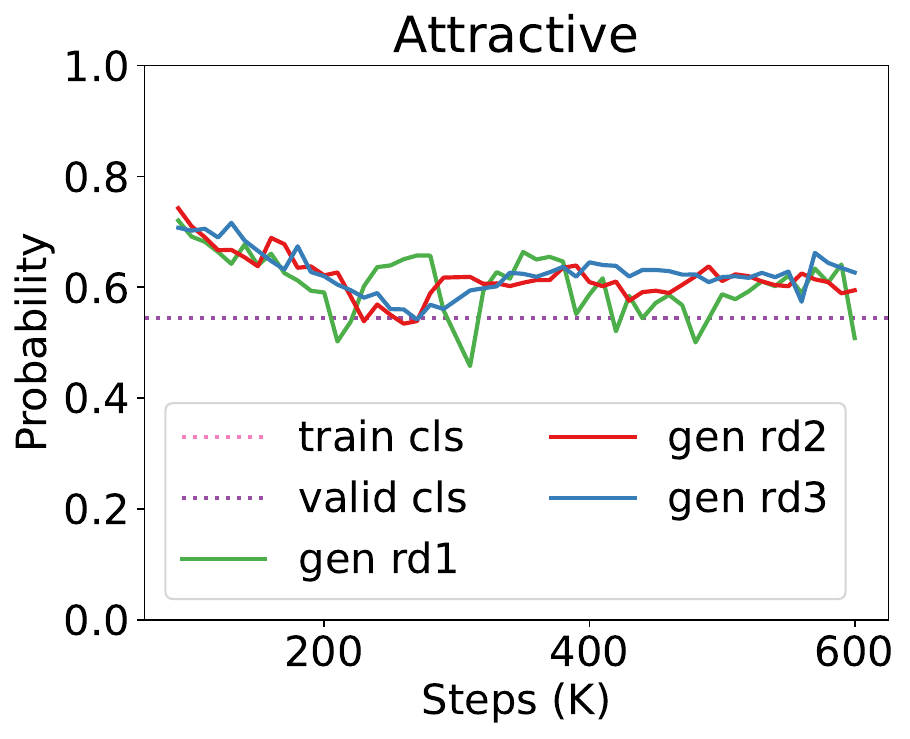}
		\caption{Attractive}
	    \end{subfigure}
    \begin{subfigure}{0.19\textwidth}
		\includegraphics[width=\textwidth]{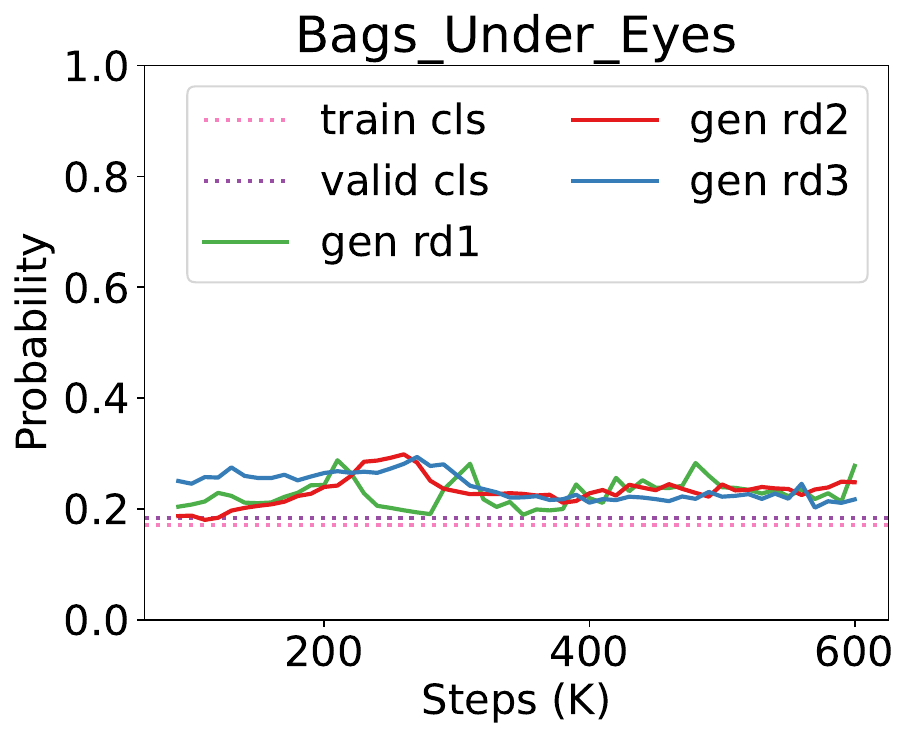}
		\caption{Bags Under Eyes}
	    \end{subfigure}
    \begin{subfigure}{0.19\textwidth}
		\includegraphics[width=\textwidth]{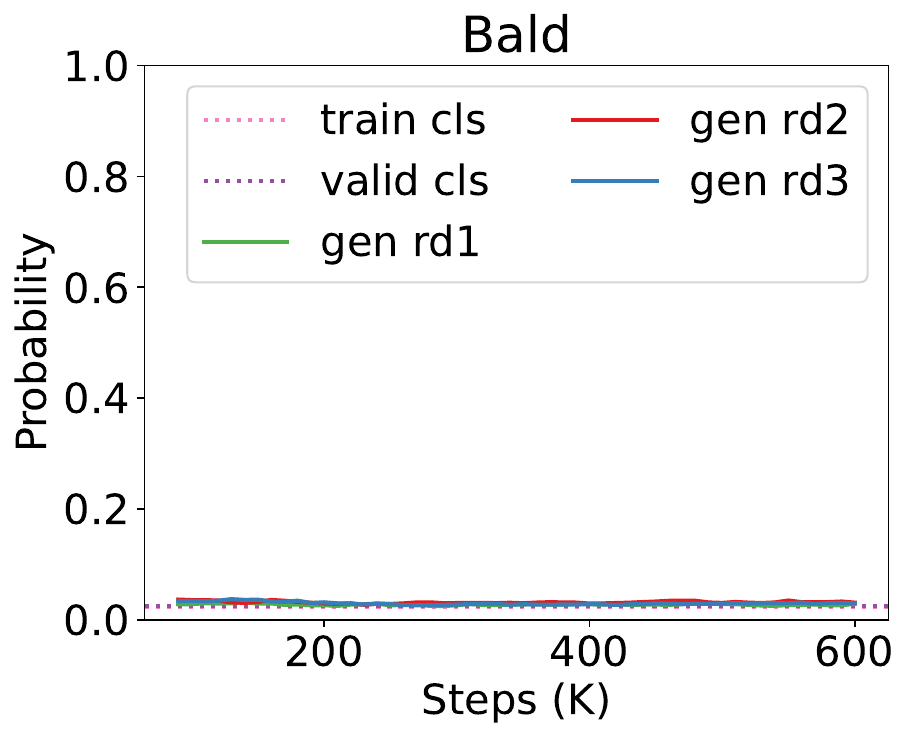}
		\caption{Bald}
    \end{subfigure}
     \vfill
    \begin{subfigure}{0.19\textwidth}
		\includegraphics[width=\textwidth]{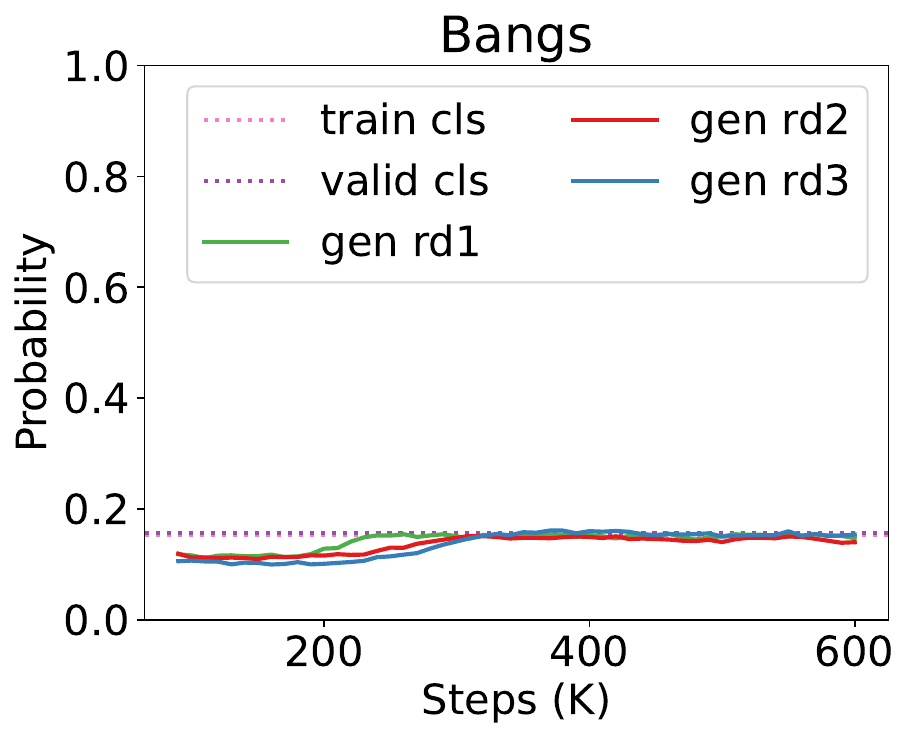}
		\caption{Bangs}
    \end{subfigure}
    \begin{subfigure}{0.19\textwidth}
		\includegraphics[width=\textwidth]{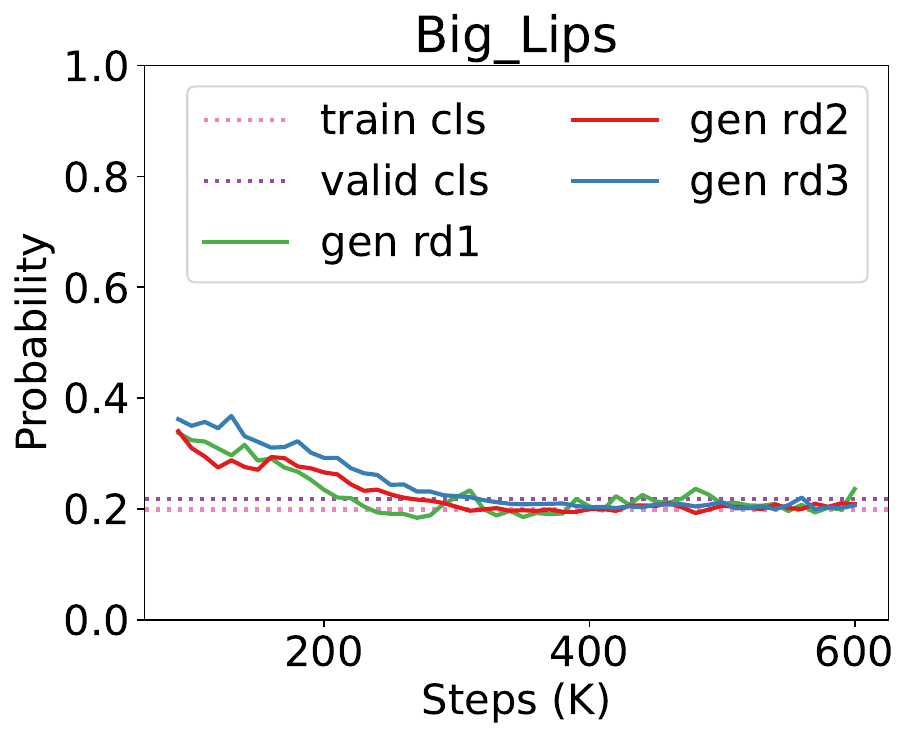}
		\caption{Big Lips}
    \end{subfigure}
     \begin{subfigure}{0.19\textwidth}
		\includegraphics[width=\textwidth]{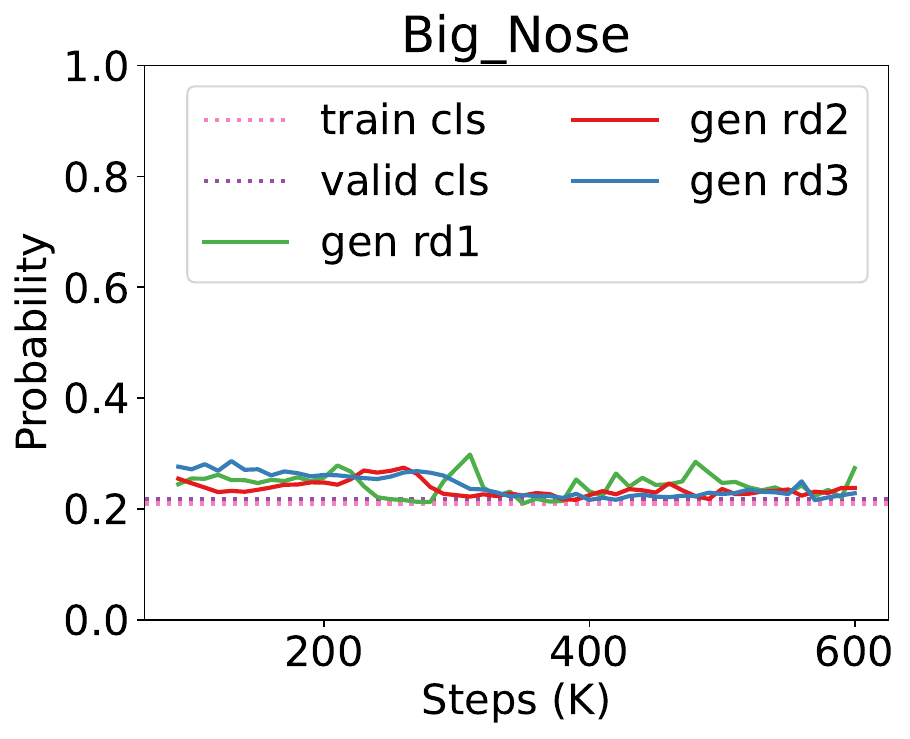}
		\caption{Big Nose}
	    \end{subfigure}
    \begin{subfigure}{0.19\textwidth}
		\includegraphics[width=\textwidth]{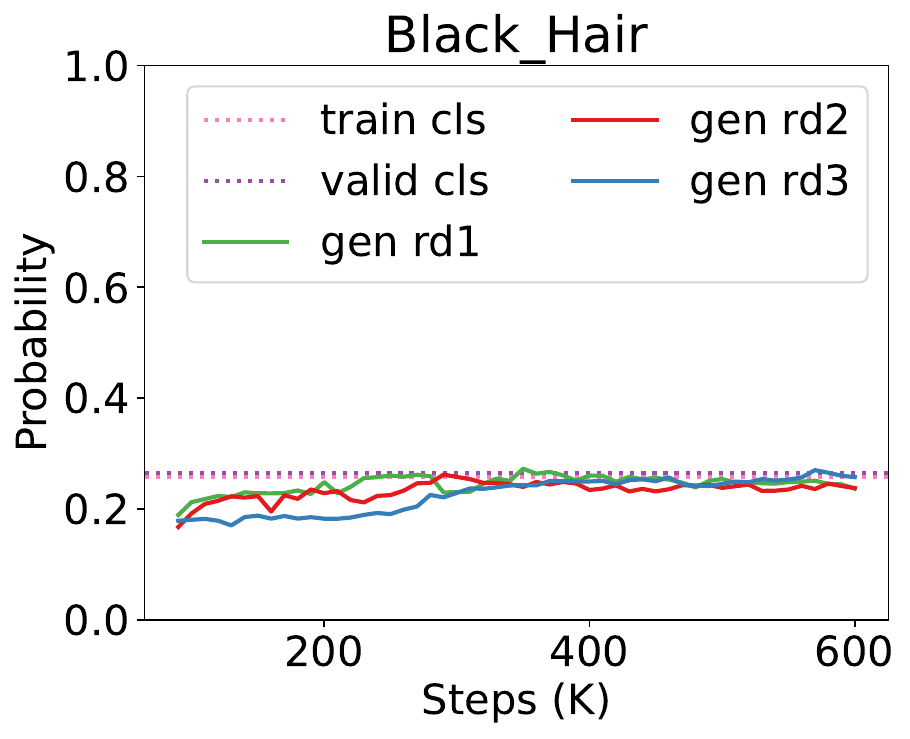}
		\caption{Black Hair}
	    \end{subfigure}
    \begin{subfigure}{0.19\textwidth}
		\includegraphics[width=\textwidth]{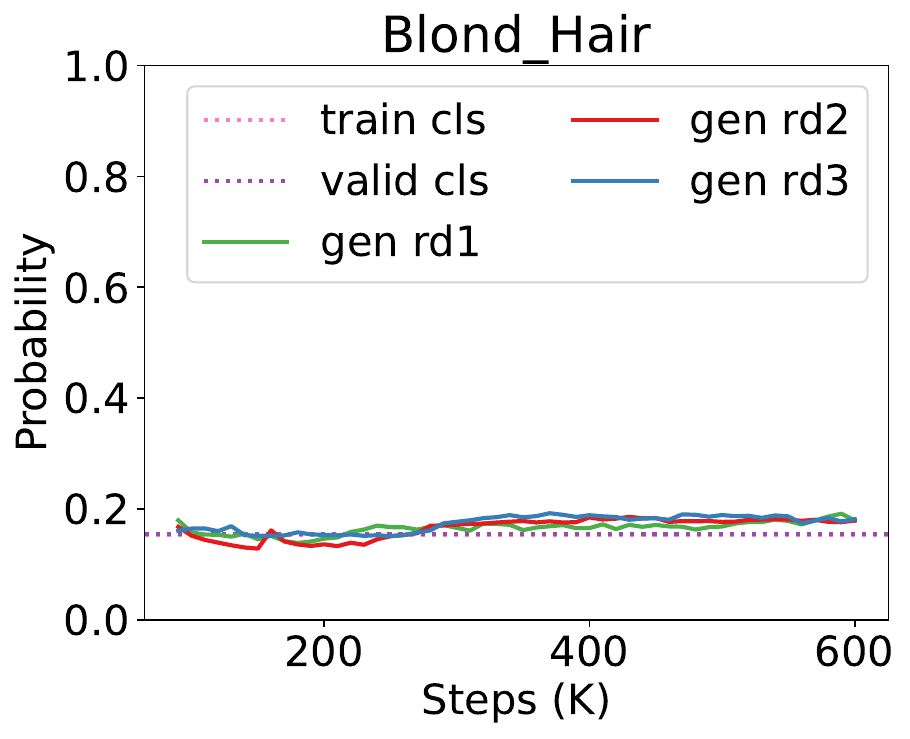}
		\caption{Blond Hair}
    \end{subfigure}
    \vfill
    \begin{subfigure}{0.19\textwidth}
		\includegraphics[width=\textwidth]{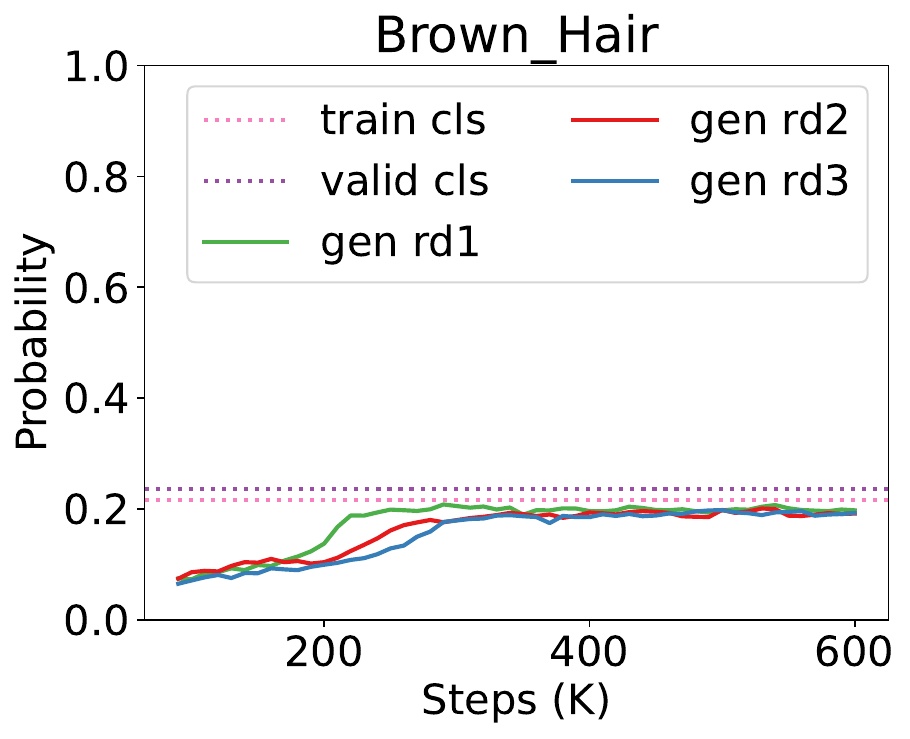}
		\caption{Brown Hair}
    \end{subfigure}
    \begin{subfigure}{0.19\textwidth}
		\includegraphics[width=\textwidth]{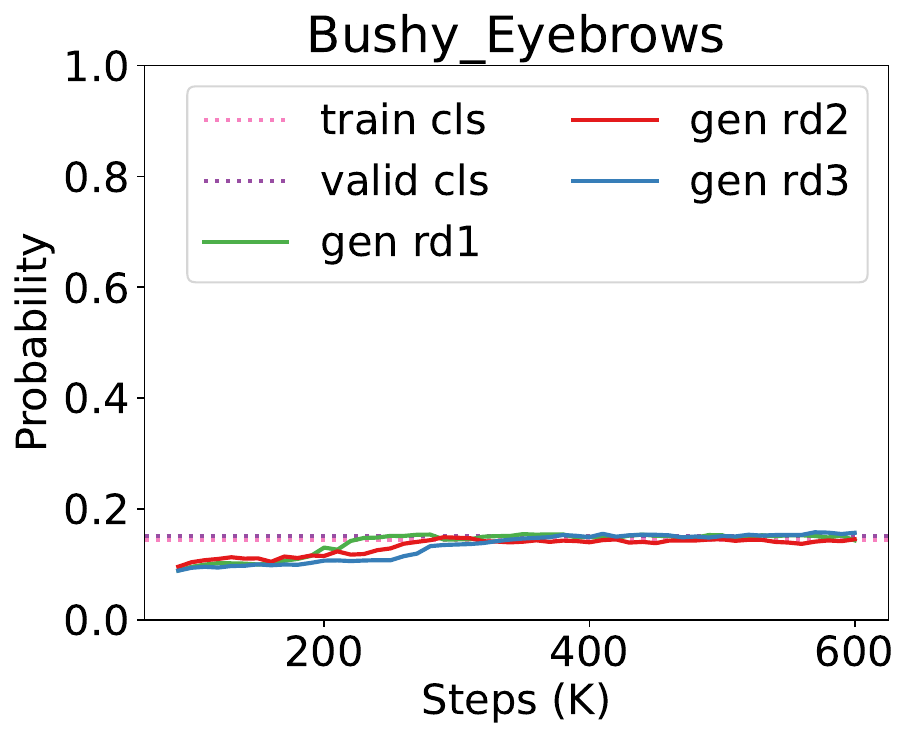}
		\caption{Bushy Eyebrows}
    \end{subfigure}
     \begin{subfigure}{0.19\textwidth}
		\includegraphics[width=\textwidth]{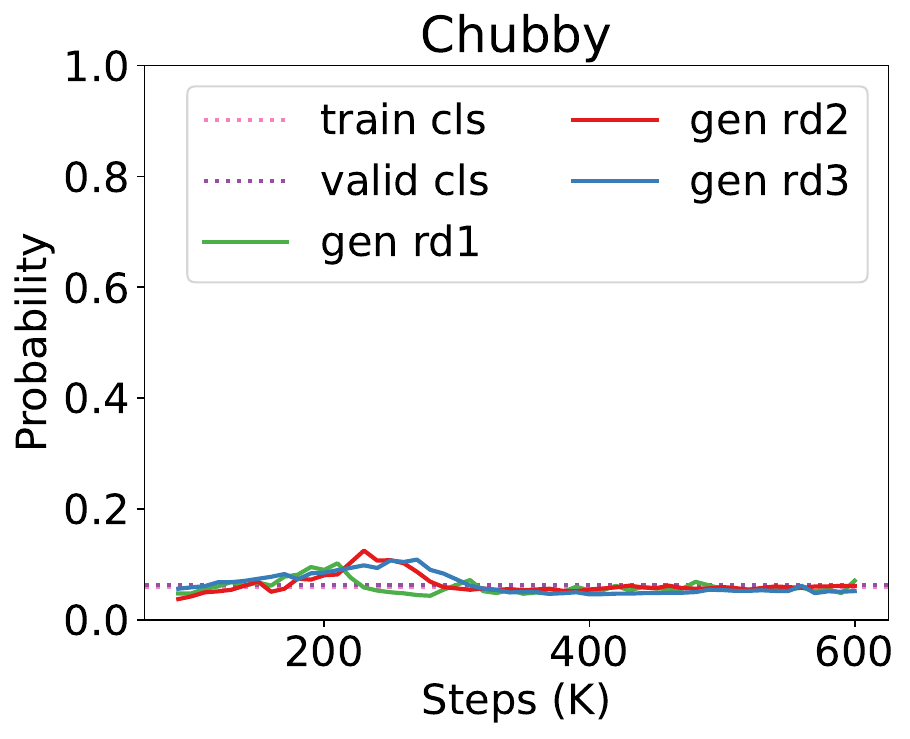}
		\caption{Chubby}
	    \end{subfigure}
    \begin{subfigure}{0.19\textwidth}
		\includegraphics[width=\textwidth]{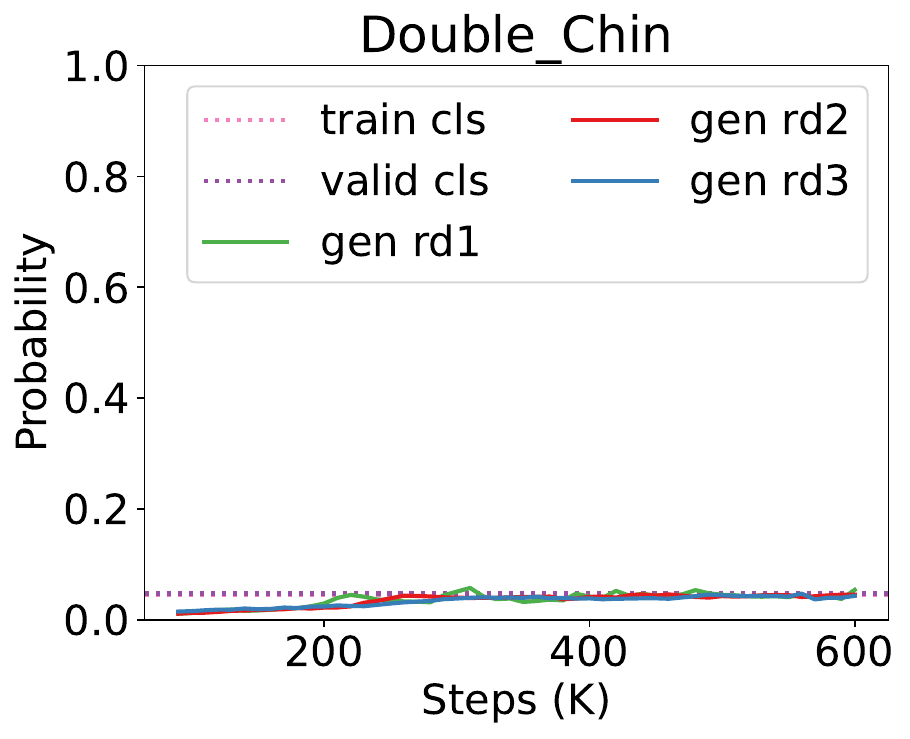}
		\caption{Double Chin}
	    \end{subfigure}
    \begin{subfigure}{0.19\textwidth}
		\includegraphics[width=\textwidth]{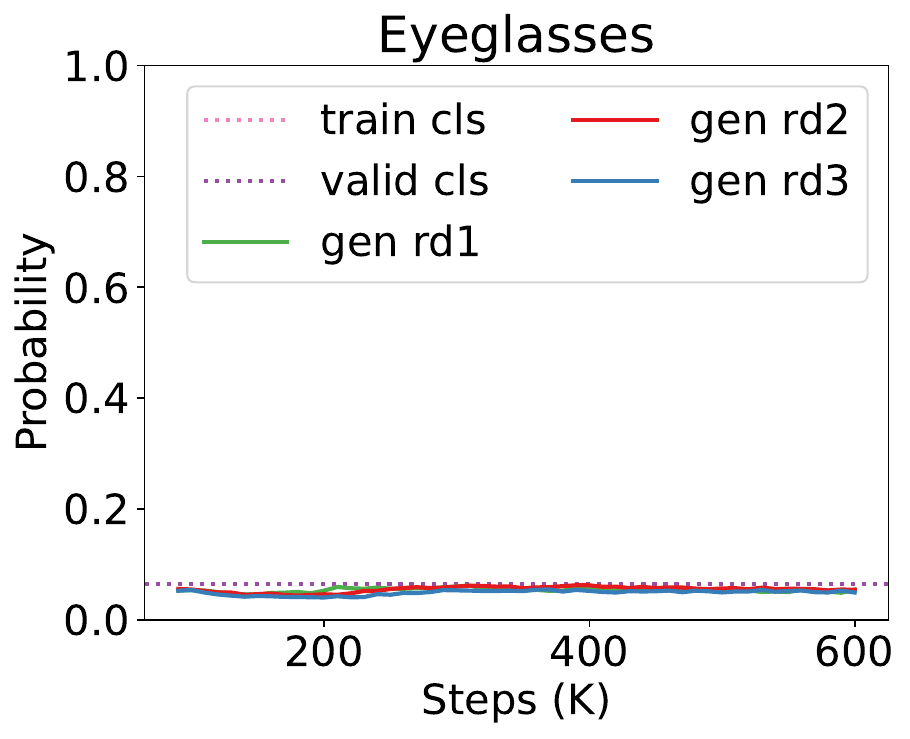}
		\caption{Eyeglasses}
    \end{subfigure}
    \vfill
    \begin{subfigure}{0.19\textwidth}
		\includegraphics[width=\textwidth]{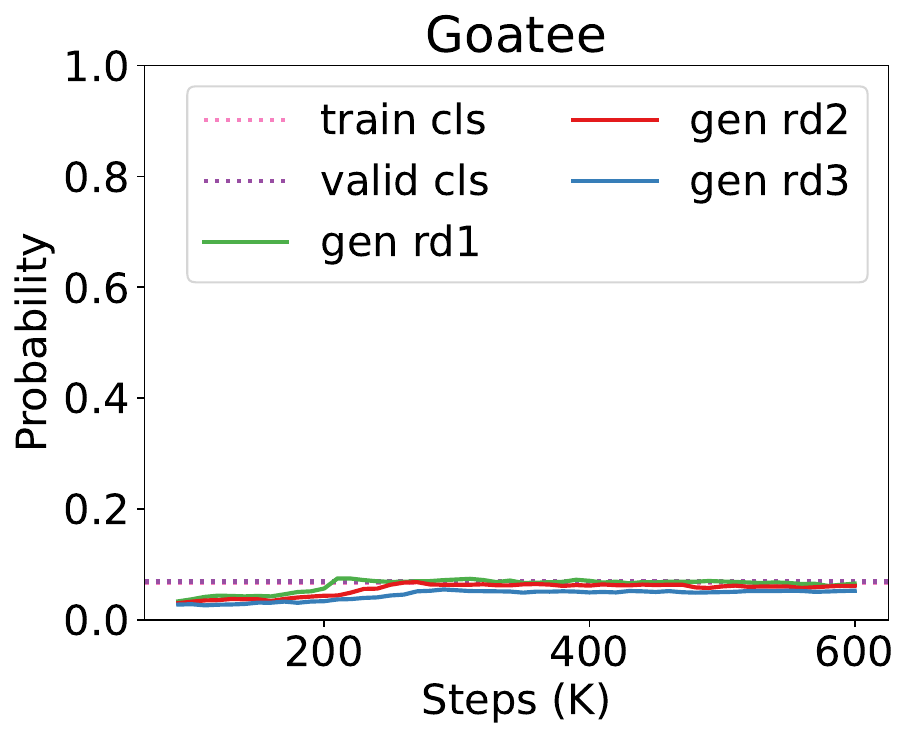}
		\caption{Goatee}
    \end{subfigure}
    \begin{subfigure}{0.19\textwidth}
		\includegraphics[width=\textwidth]{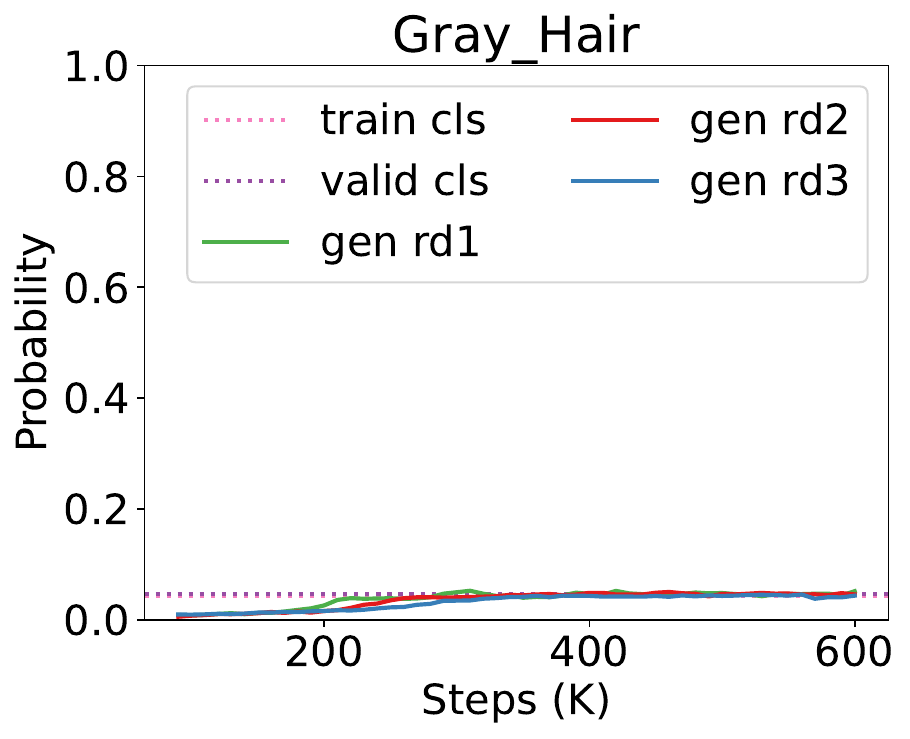}
		\caption{Gray Hair}
    \end{subfigure}
     \begin{subfigure}{0.19\textwidth}
		\includegraphics[width=\textwidth]{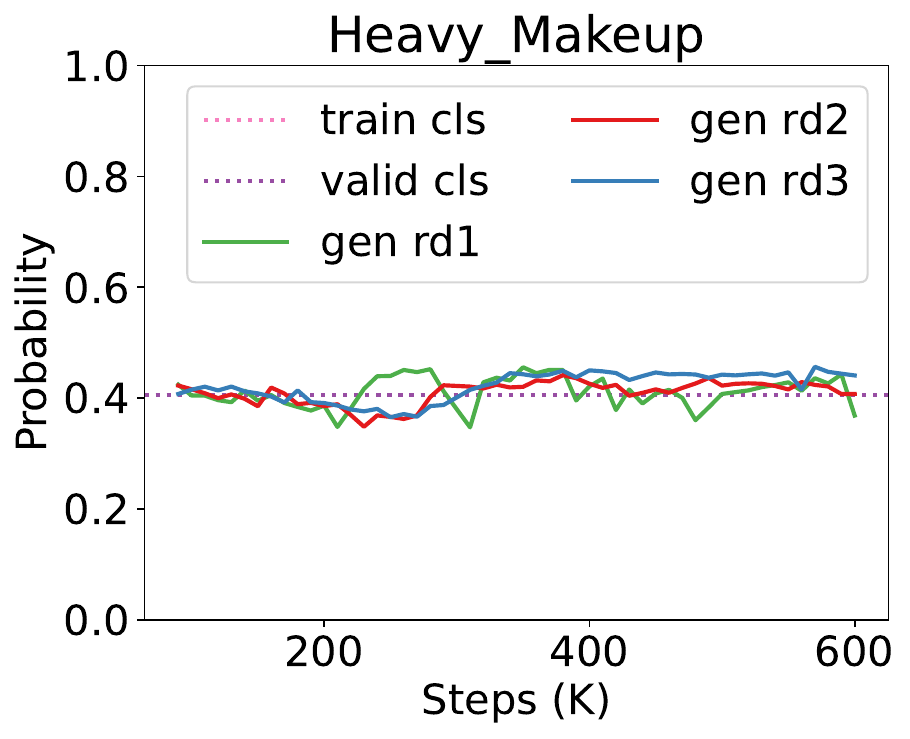}
		\caption{Heavy Makeup}
	    \end{subfigure}
    \begin{subfigure}{0.19\textwidth}
		\includegraphics[width=\textwidth]{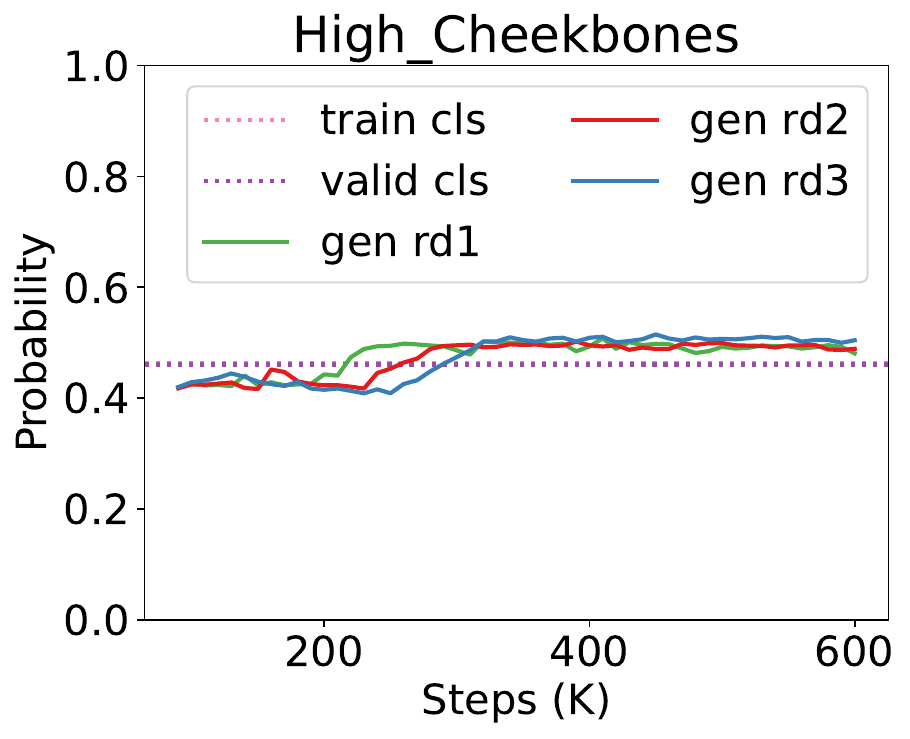}
		\caption{High Cheekbones}
	    \end{subfigure}
    \begin{subfigure}{0.19\textwidth}
		\includegraphics[width=\textwidth]{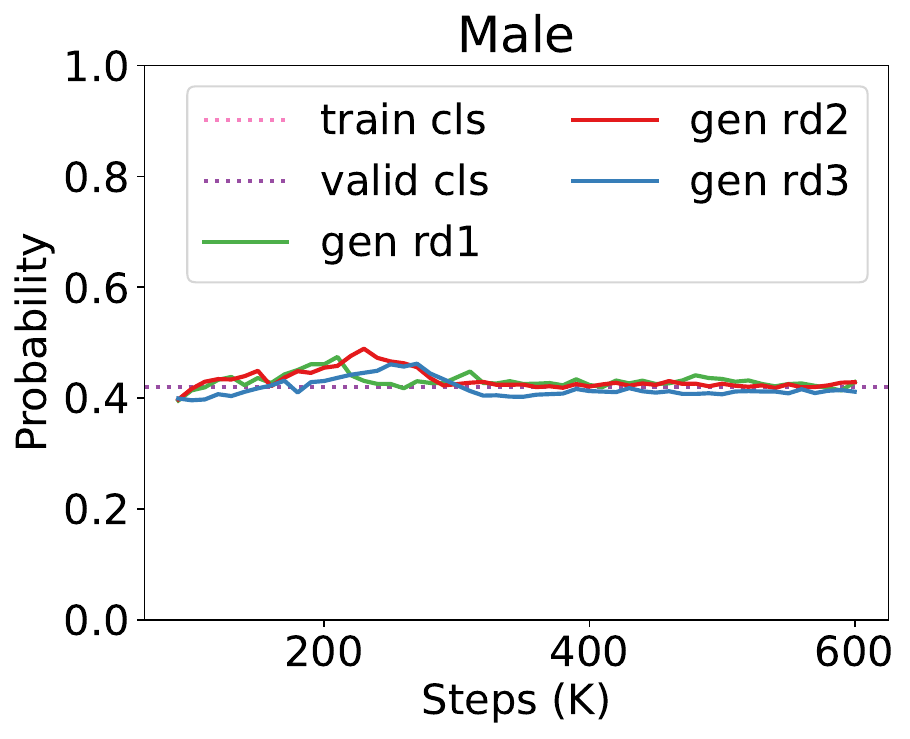}
		\caption{Male}
    \end{subfigure}
    \vfill
    \begin{subfigure}{0.19\textwidth}
		\includegraphics[width=\textwidth]{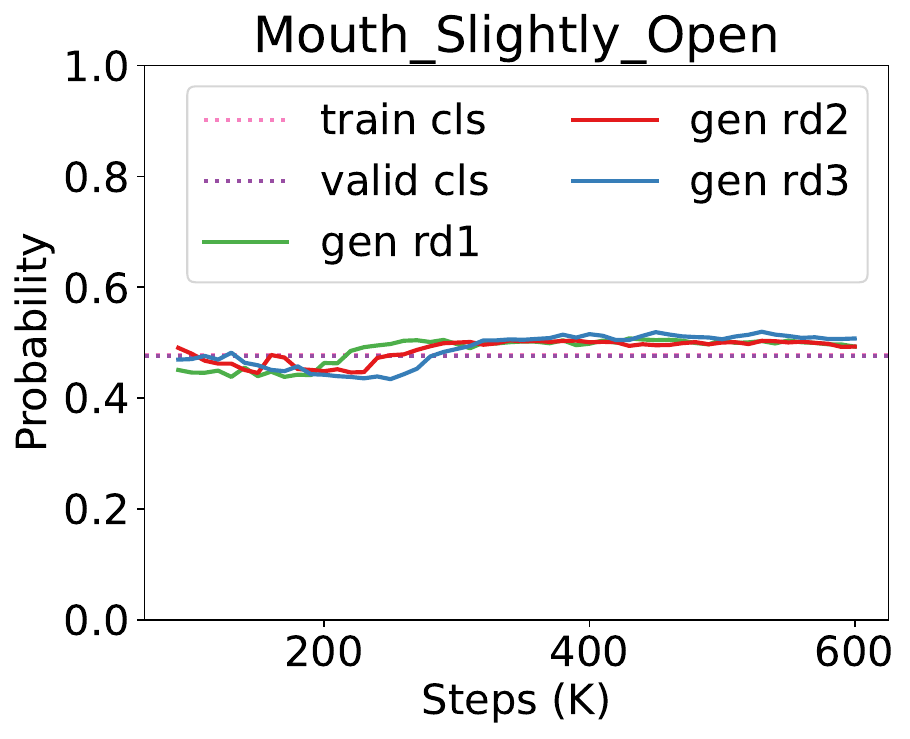}
		\caption{\begin{tiny}
		    Mouth Slightly Open
		\end{tiny}}
    \end{subfigure}
    \begin{subfigure}{0.19\textwidth}
		\includegraphics[width=\textwidth]{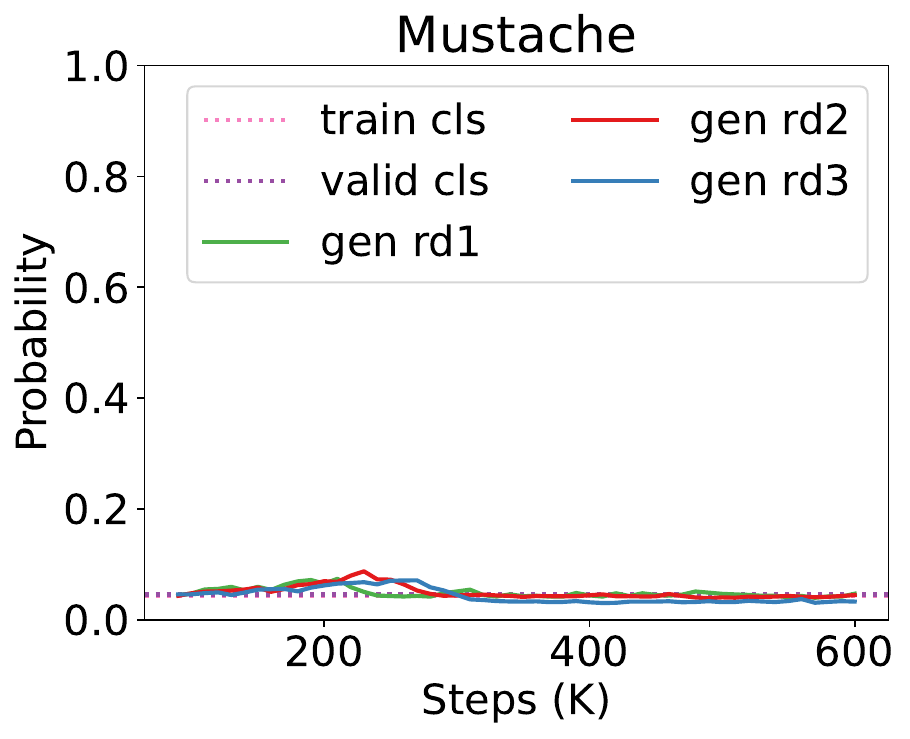}
		\caption{Mustache}
    \end{subfigure}
     \begin{subfigure}{0.19\textwidth}
		\includegraphics[width=\textwidth]{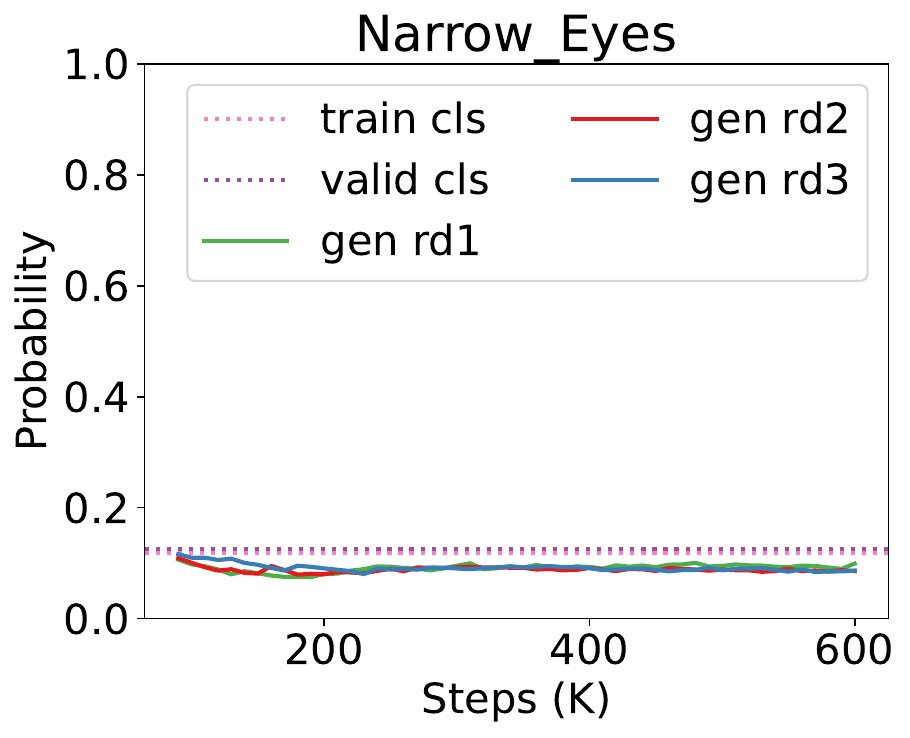}
		\caption{Narrow Eyes}
	    \end{subfigure}
    \begin{subfigure}{0.19\textwidth}
		\includegraphics[width=\textwidth]{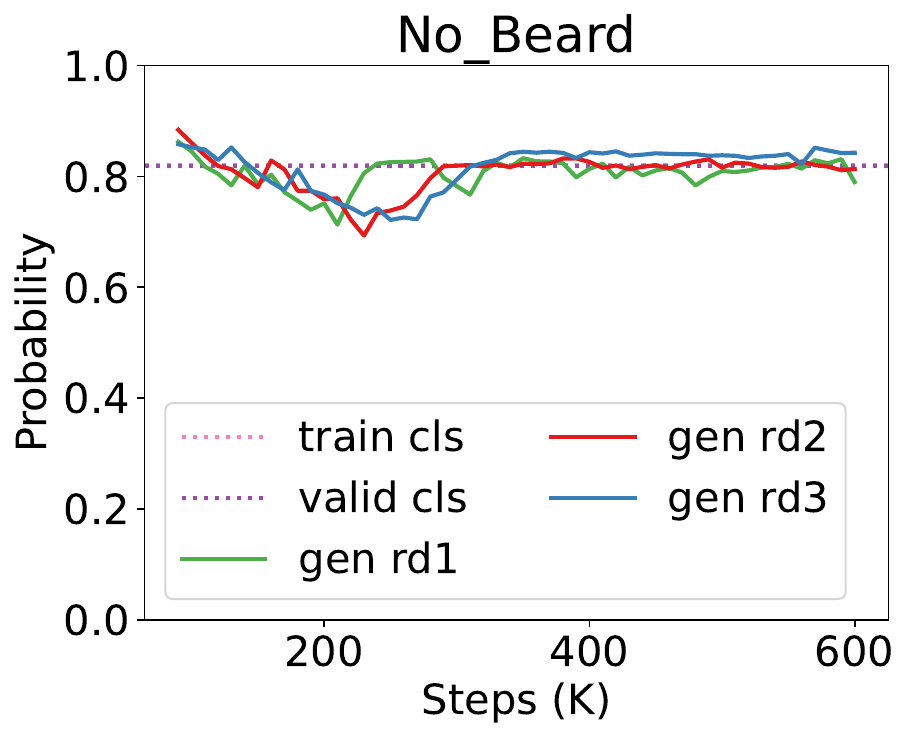}
		\caption{No Beard}
	    \end{subfigure}
    \begin{subfigure}{0.19\textwidth}
		\includegraphics[width=\textwidth]{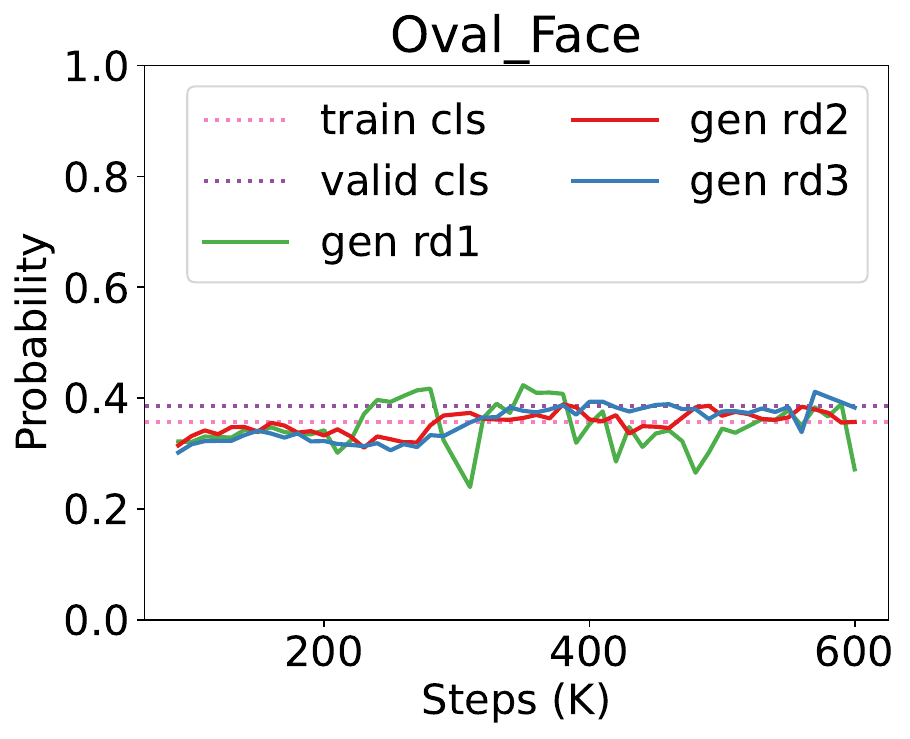}
		\caption{Oval Face}
    \end{subfigure}
    \vfill
    \begin{subfigure}{0.19\textwidth}
		\includegraphics[width=\textwidth]{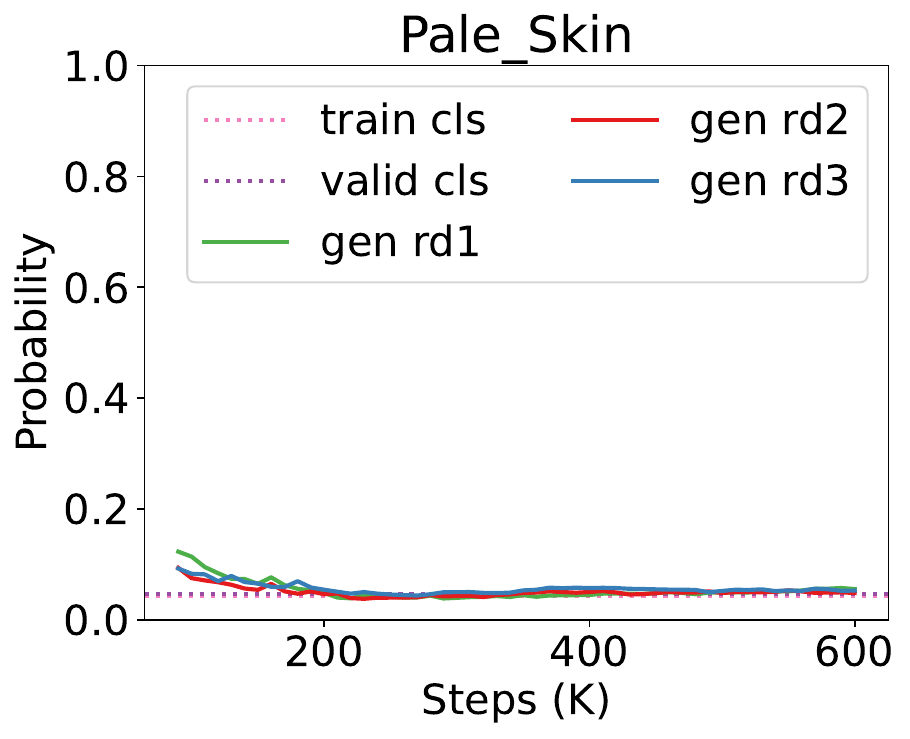}
		\caption{Pale Skin}
    \end{subfigure}
    \begin{subfigure}{0.19\textwidth}
		\includegraphics[width=\textwidth]{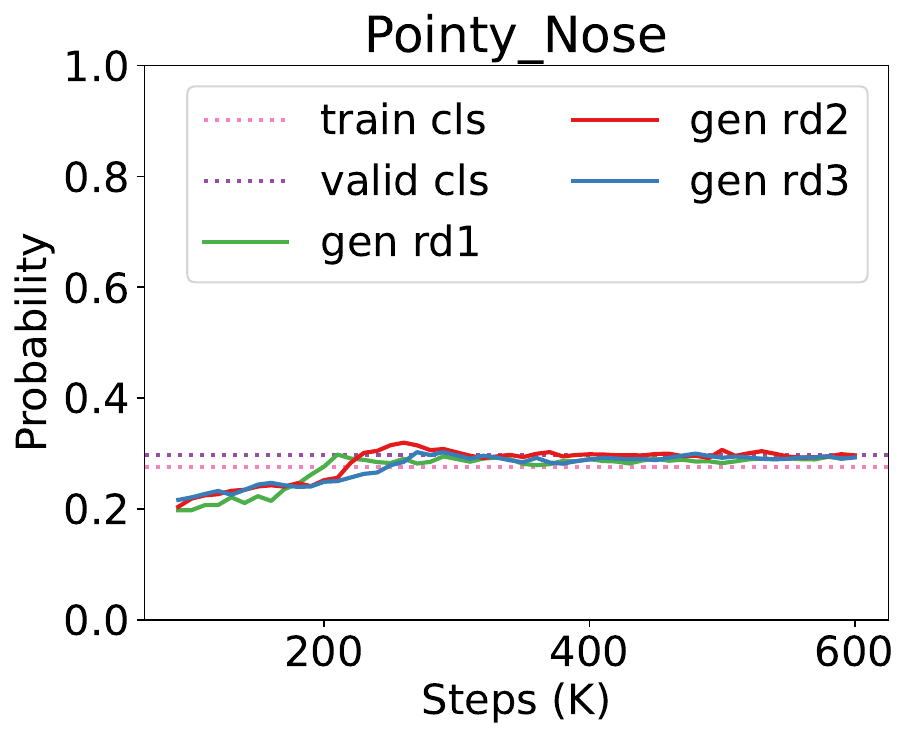}
		\caption{Pointy Nose}
    \end{subfigure}
     \begin{subfigure}{0.19\textwidth}
		\includegraphics[width=\textwidth]{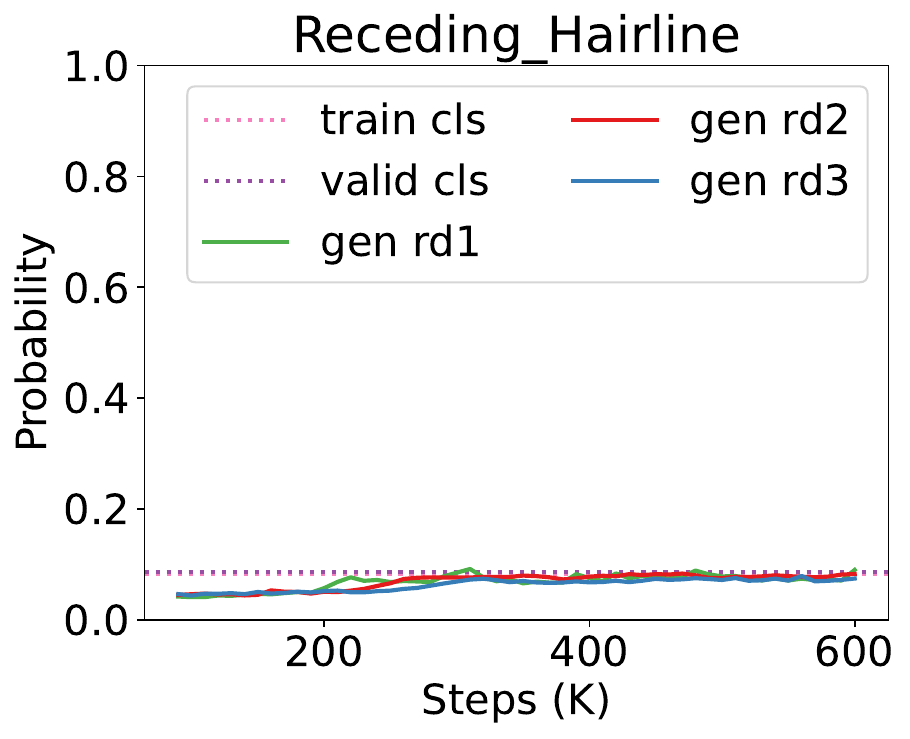}
		\caption{\begin{tiny}
		    Receding Hairline
		\end{tiny}}
	    \end{subfigure}
    \begin{subfigure}{0.19\textwidth}
		\includegraphics[width=\textwidth]{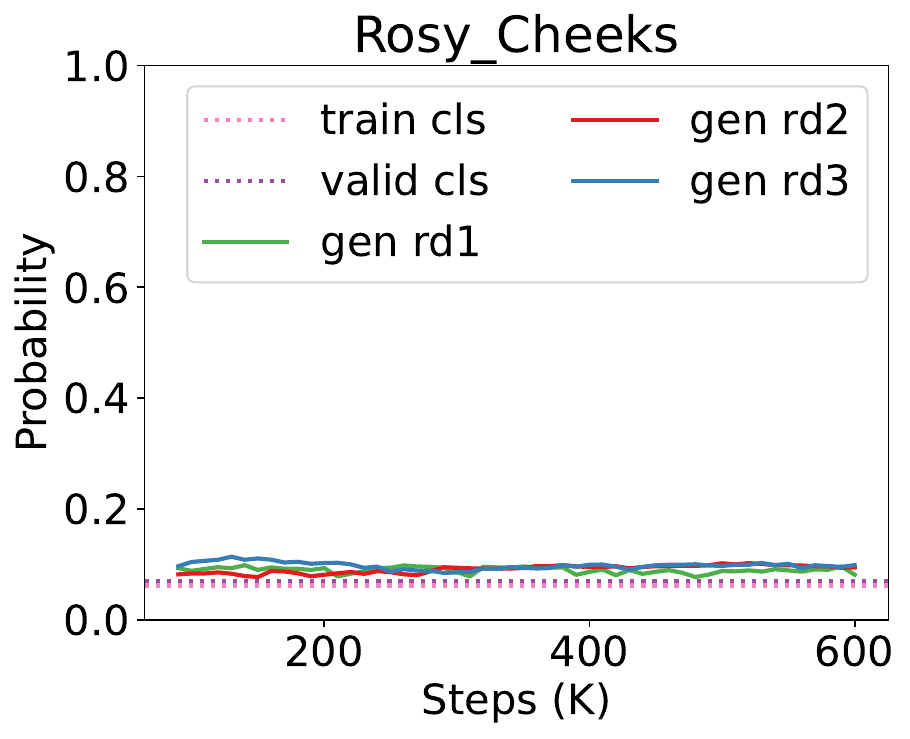}
		\caption{Rosy Cheeks}
	    \end{subfigure}
    \begin{subfigure}{0.19\textwidth}
		\includegraphics[width=\textwidth]{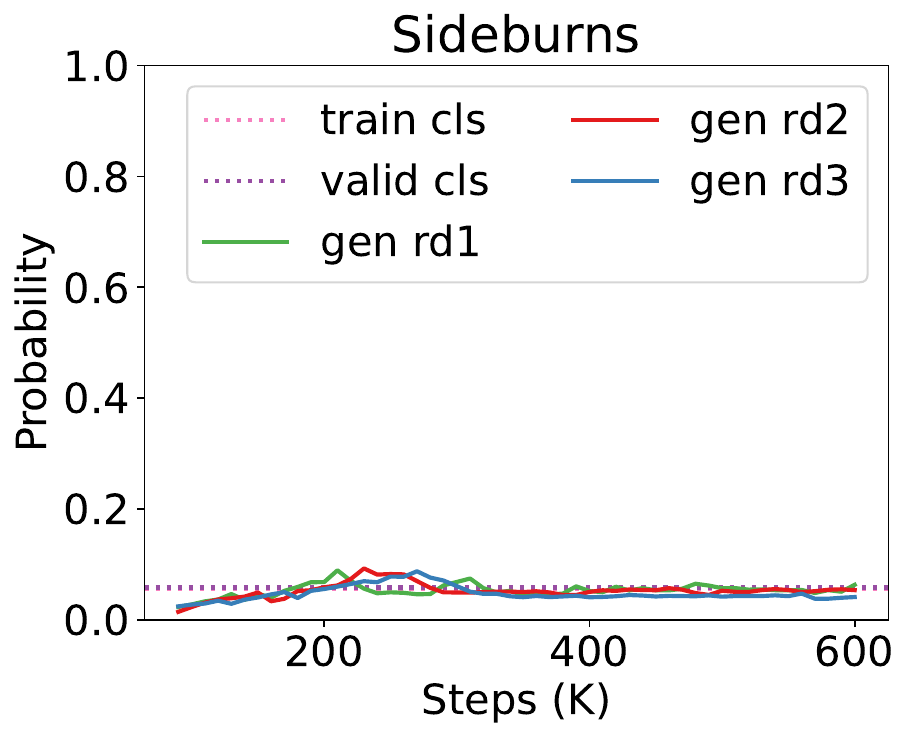}
		\caption{Sideburns}
    \end{subfigure}
    \vfill
    \begin{subfigure}{0.19\textwidth}
		\includegraphics[width=\textwidth]{Figs/marginal_bias_curves/Smiling.pdf}
		\caption{Smiling}
    \end{subfigure}
    \begin{subfigure}{0.19\textwidth}
		\includegraphics[width=\textwidth]{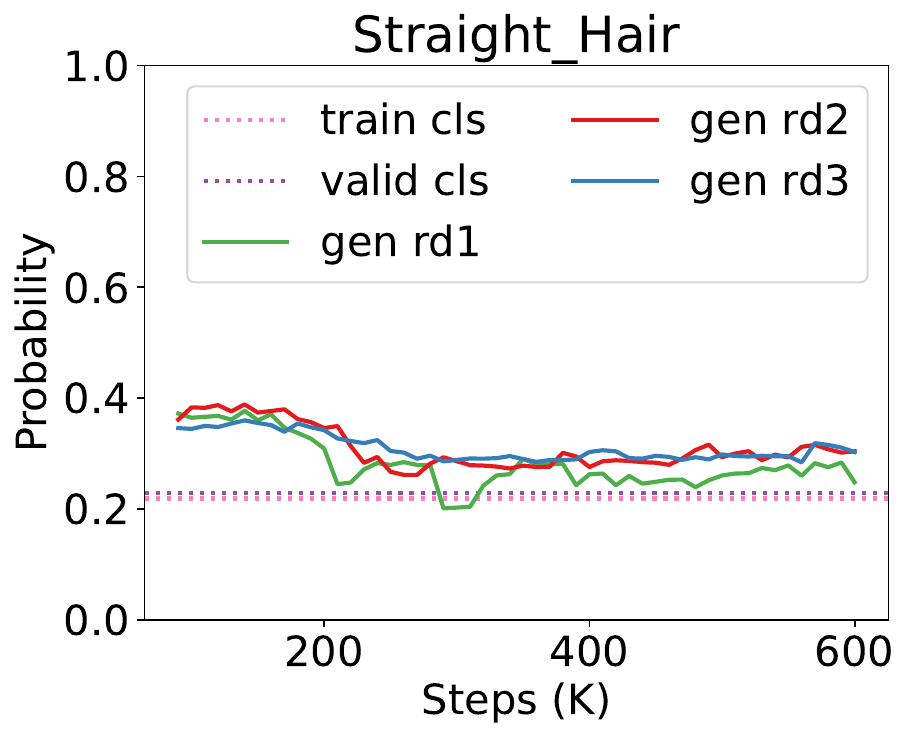}
		\caption{Straight hair}
    \end{subfigure}
     \begin{subfigure}{0.19\textwidth}
		\includegraphics[width=\textwidth]{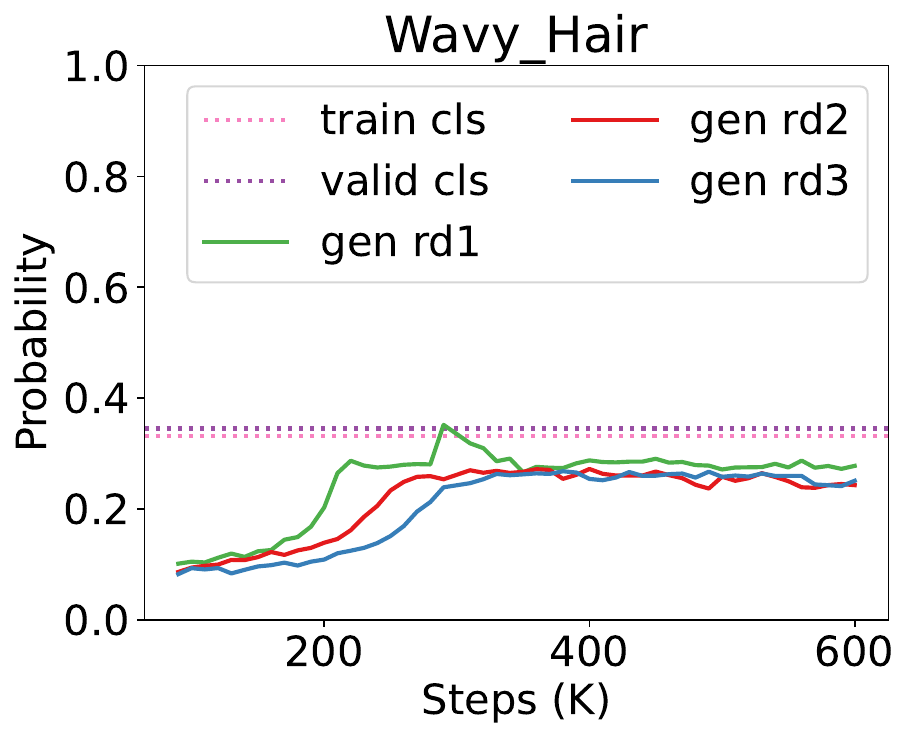}
		\caption{Wavy Hair}
	    \end{subfigure}
    \begin{subfigure}{0.19\textwidth}
		\includegraphics[width=\textwidth]{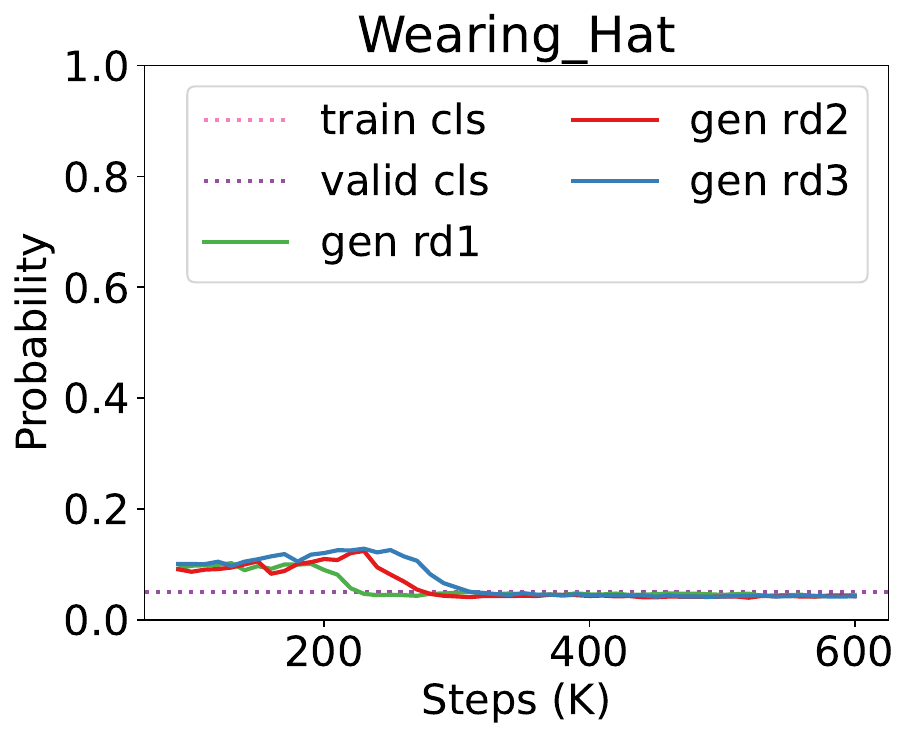}
		\caption{Wearing Hat}
    \end{subfigure}
    \begin{subfigure}{0.19\textwidth}
		\includegraphics[width=\textwidth]{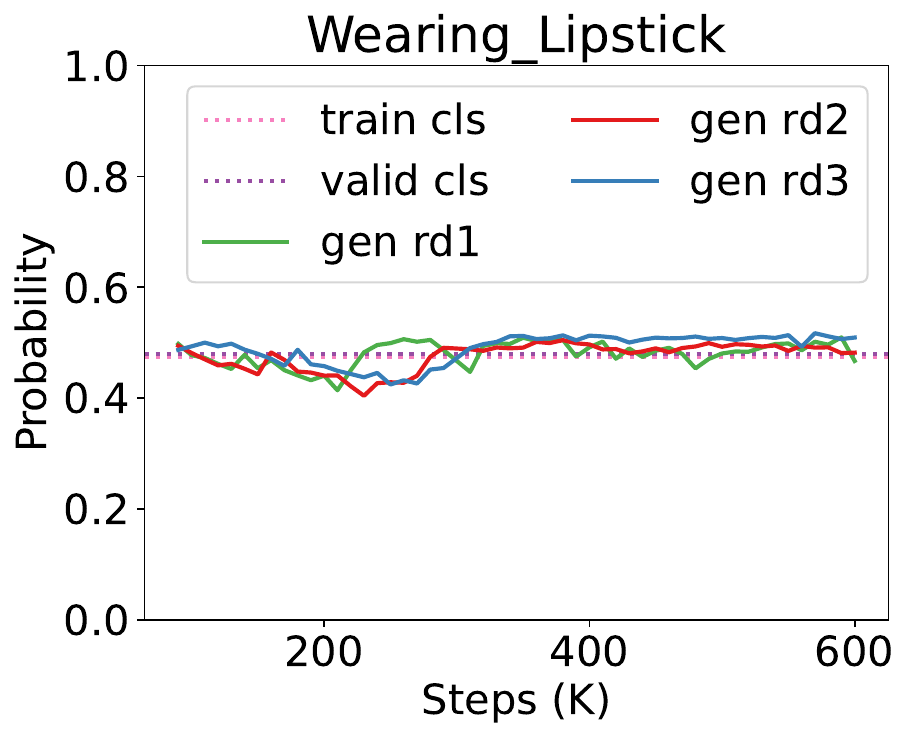}
		\caption{\begin{tiny}
		    Wearing Lipstick
		\end{tiny}}
    \end{subfigure}
    \vfill
    \begin{subfigure}{0.19\textwidth}
		\includegraphics[width=\textwidth]{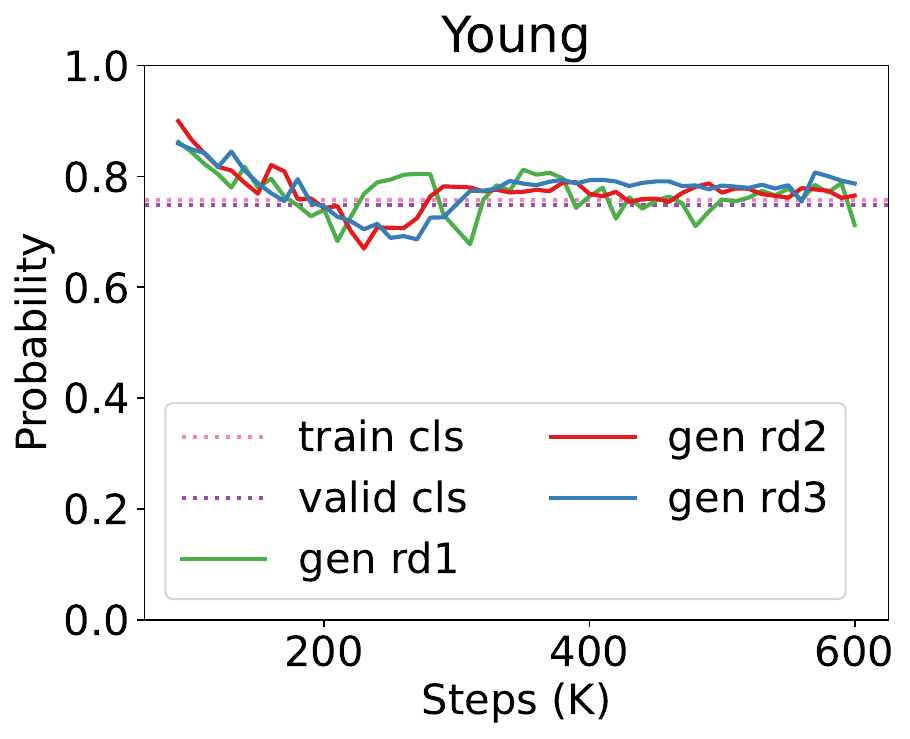}
		\caption{Young}
    \end{subfigure}
	\caption{The probabilities of attributes in CelebA during training (ResNext-based classifier).}
	\label{fig:attr_bias_celeba_all}
\end{figure}

\begin{figure}[!tbp]
    \centering
    \begin{subfigure}{0.19\textwidth}
		\includegraphics[width=\textwidth]{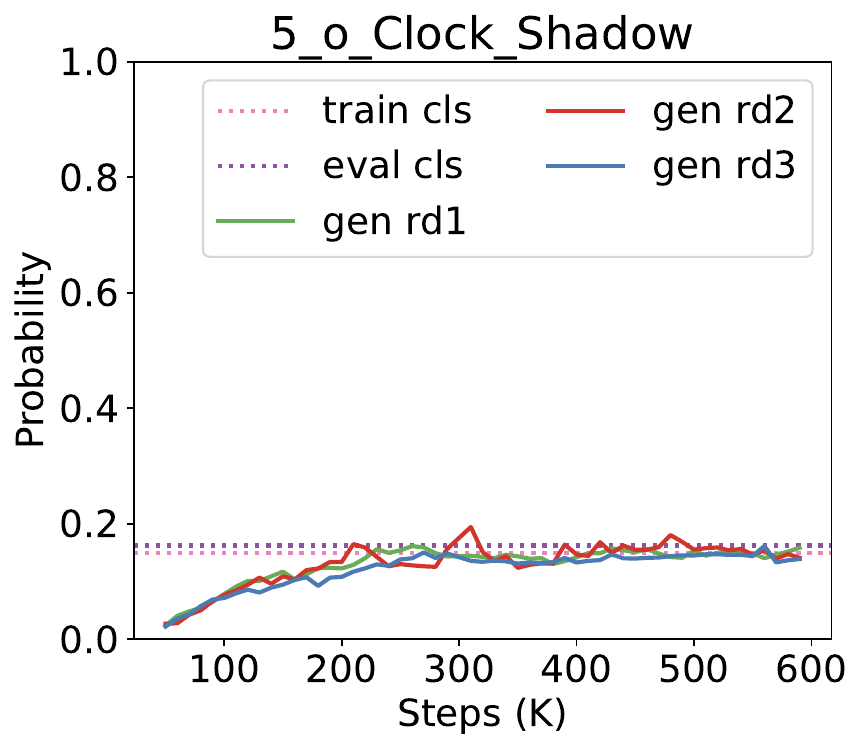}
		\caption{5o Clock Shadow}
    \end{subfigure}
    \begin{subfigure}{0.19\textwidth}
		\includegraphics[width=\textwidth]{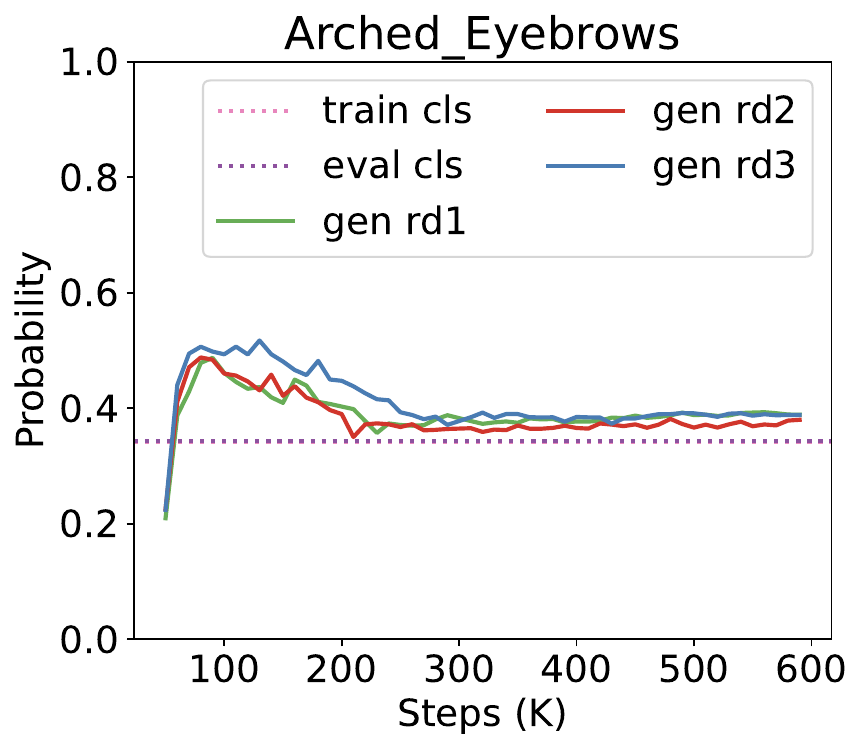}
		\caption{\begin{tiny}
		    Arched Eyebrows
		\end{tiny}}
        \label{fig:arched_eyebrows_swint}
    \end{subfigure}
     \begin{subfigure}{0.19\textwidth}
		\includegraphics[width=\textwidth]{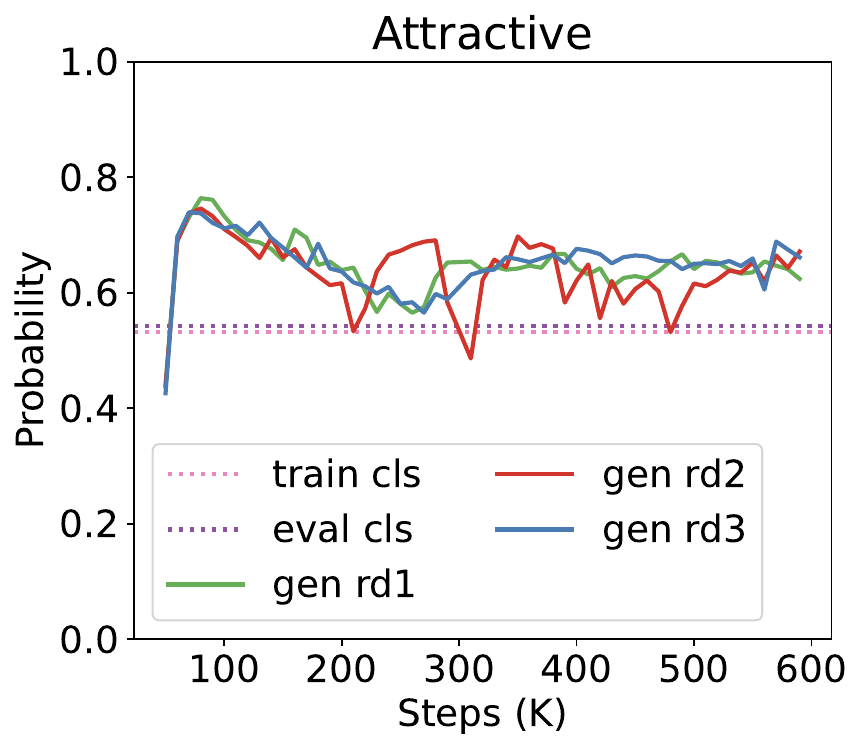}
		\caption{Attractive}
	    \end{subfigure}
    \begin{subfigure}{0.19\textwidth}
		\includegraphics[width=\textwidth]{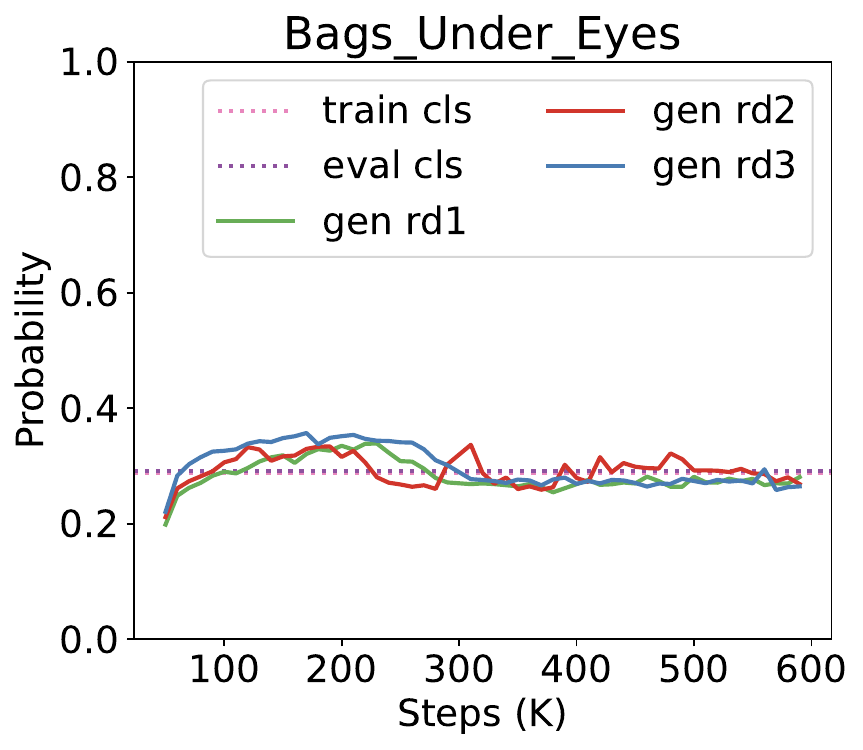}
		\caption{Bags Under Eyes}
	    \end{subfigure}
    \begin{subfigure}{0.19\textwidth}
		\includegraphics[width=\textwidth]{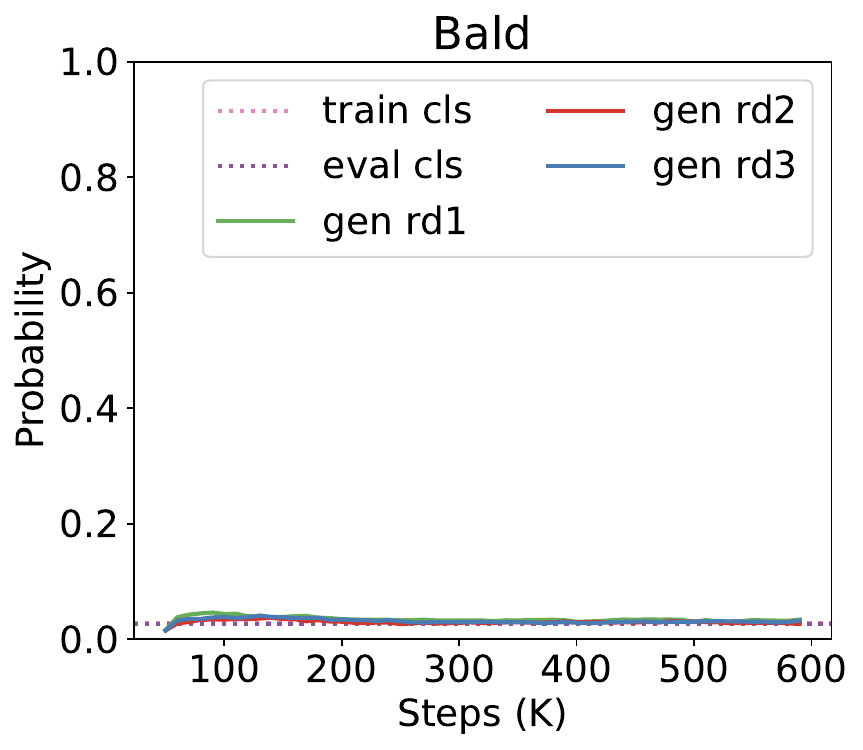}
		\caption{Bald}
    \end{subfigure}
     \vfill
    \begin{subfigure}{0.19\textwidth}
		\includegraphics[width=\textwidth]{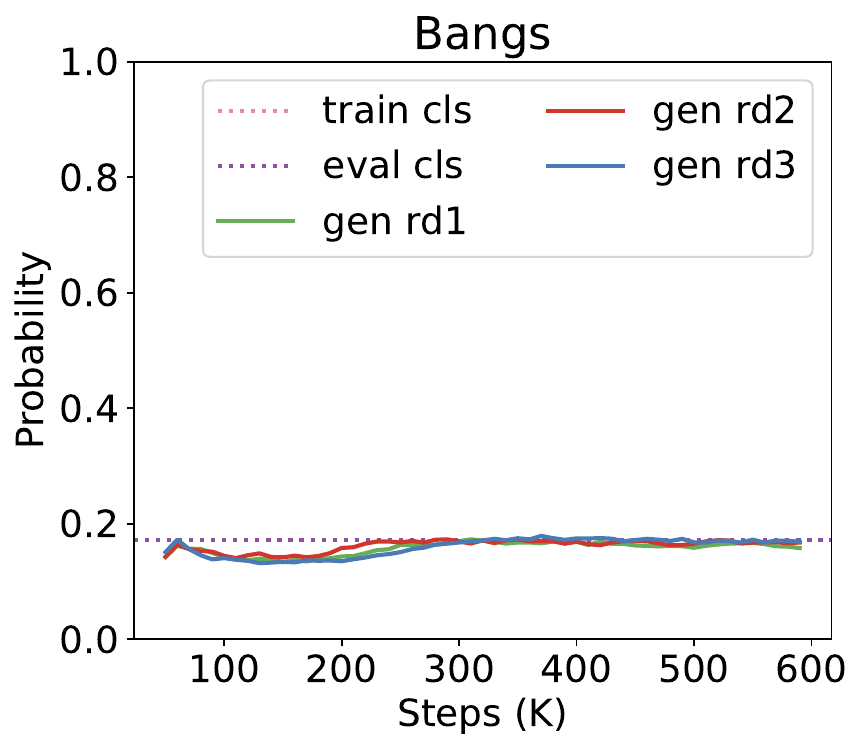}
		\caption{Bangs}
    \end{subfigure}
    \begin{subfigure}{0.19\textwidth}
		\includegraphics[width=\textwidth]{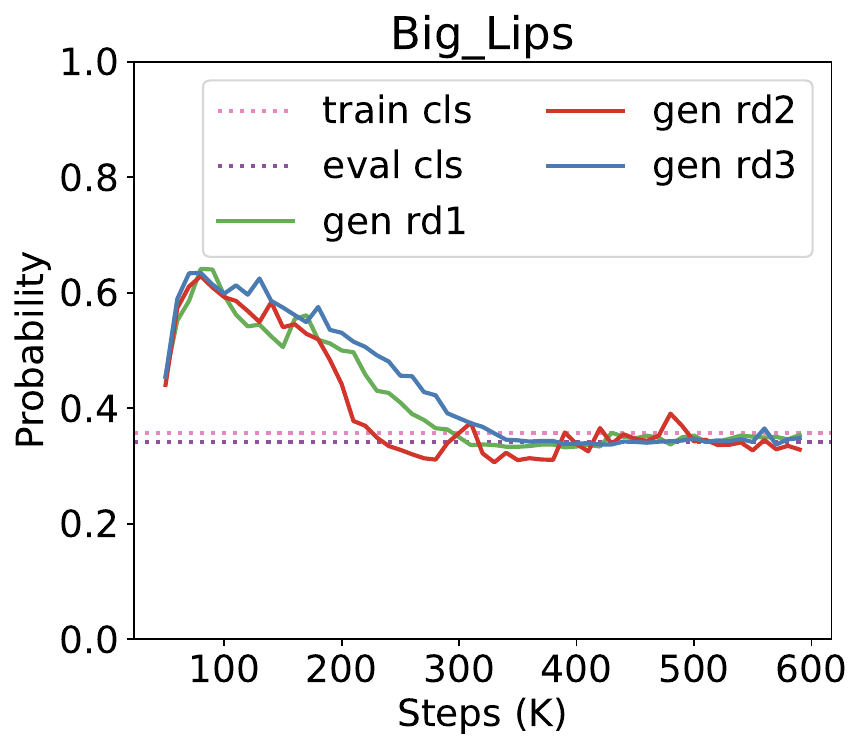}
		\caption{Big Lips}
    \end{subfigure}
     \begin{subfigure}{0.19\textwidth}
		\includegraphics[width=\textwidth]{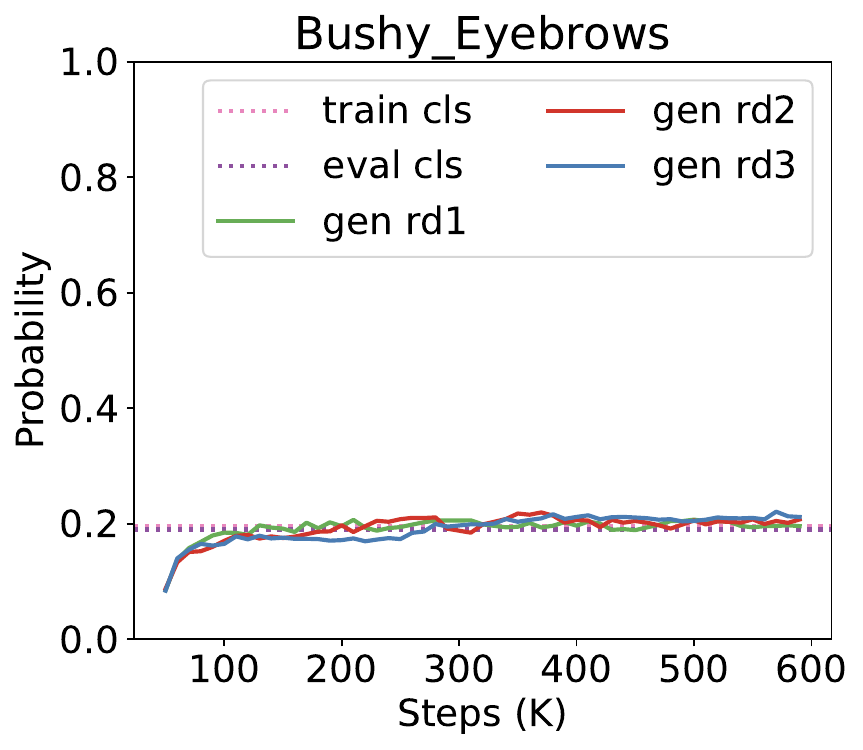}
		\caption{Big Nose}
	    \end{subfigure}
    \begin{subfigure}{0.19\textwidth}
		\includegraphics[width=\textwidth]{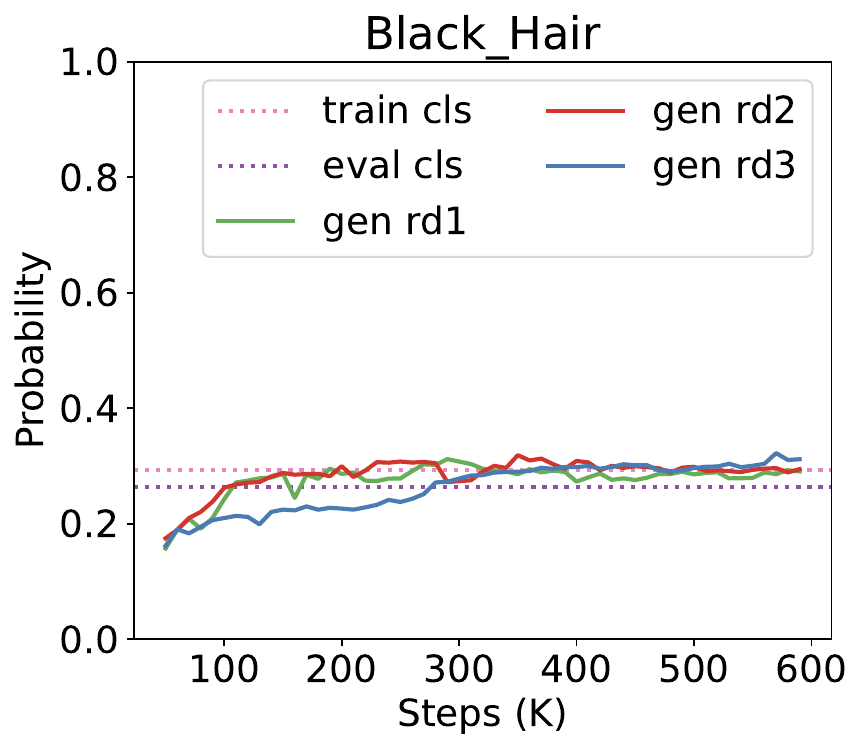}
		\caption{Black Hair}
	    \end{subfigure}
    \begin{subfigure}{0.19\textwidth}
		\includegraphics[width=\textwidth]{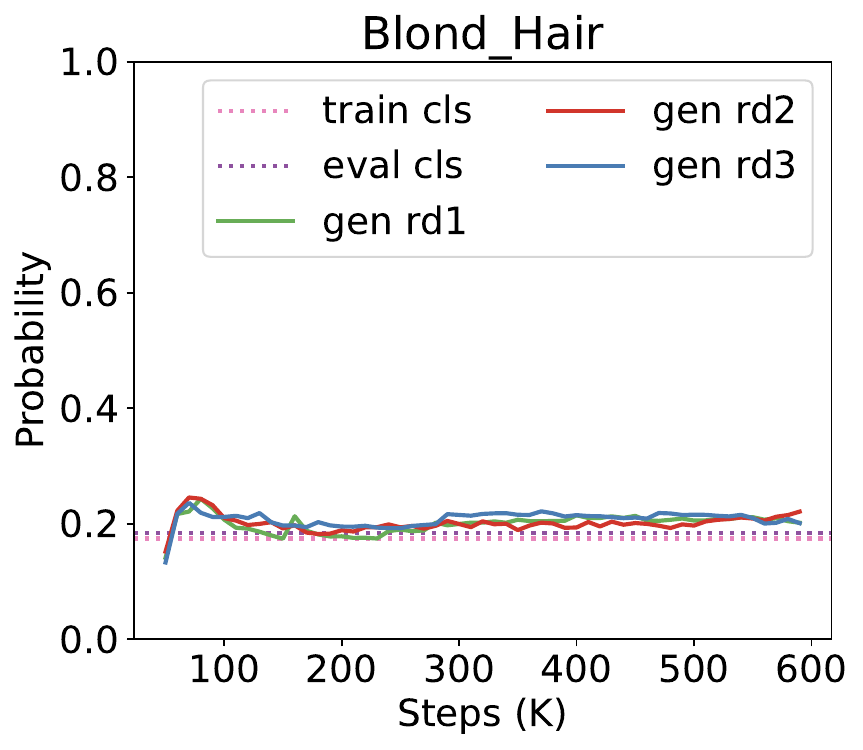}
		\caption{Blond Hair}
    \end{subfigure}
    \vfill
    \begin{subfigure}{0.19\textwidth}
		\includegraphics[width=\textwidth]{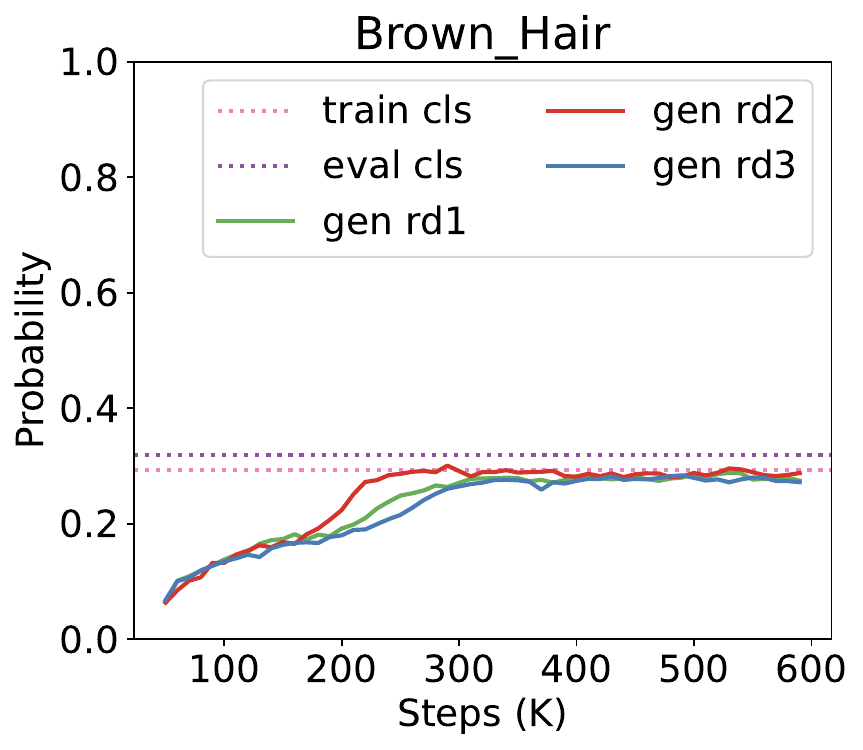}
		\caption{Brown Hair}
    \end{subfigure}
    \begin{subfigure}{0.19\textwidth}
		\includegraphics[width=\textwidth]{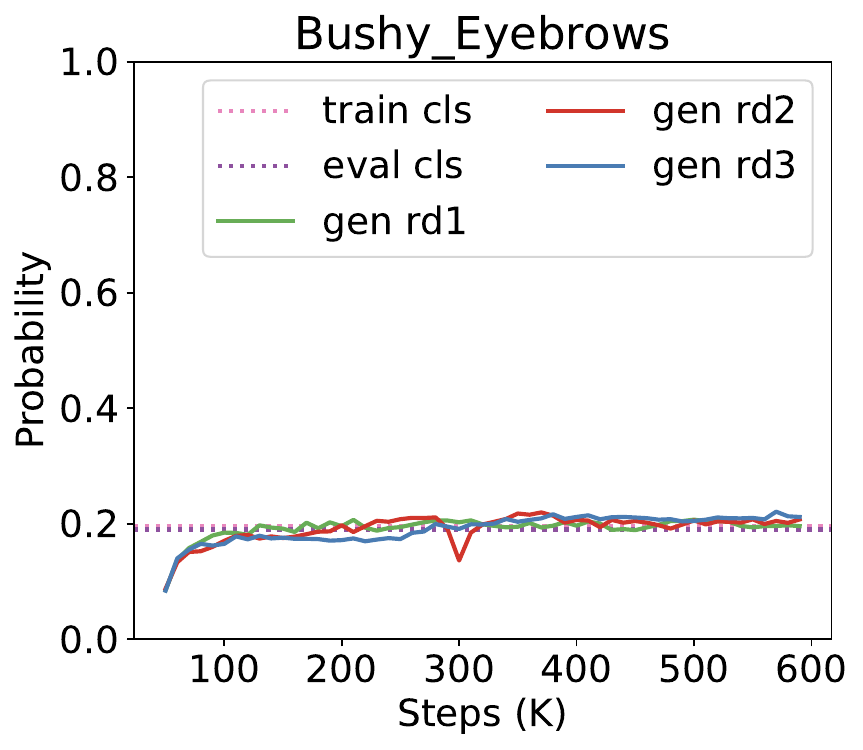}
		\caption{Bushy Eyebrows}
    \end{subfigure}
     \begin{subfigure}{0.19\textwidth}
		\includegraphics[width=\textwidth]{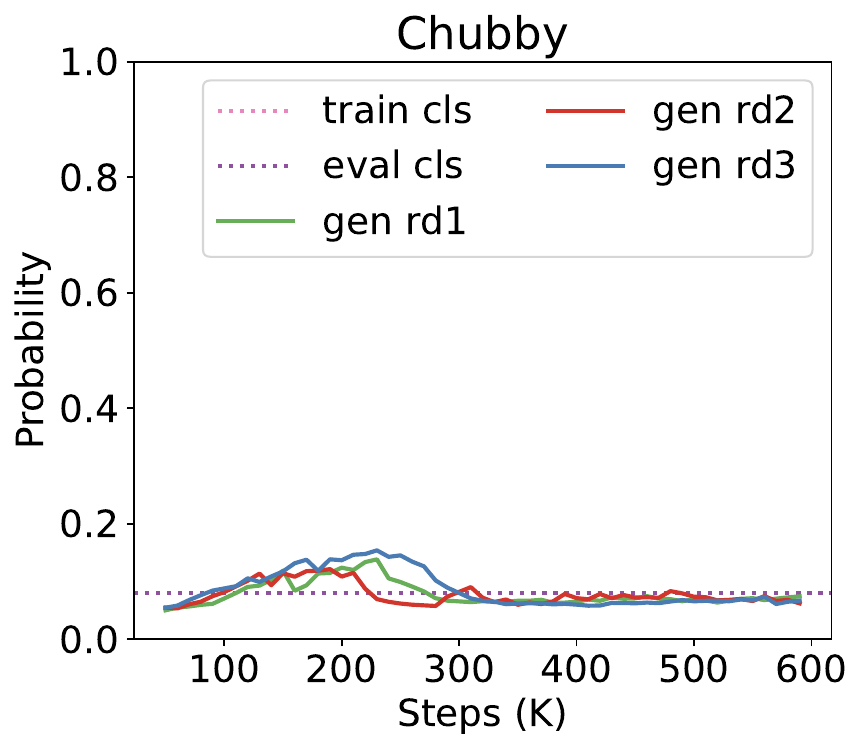}
		\caption{Chubby}
	    \end{subfigure}
    \begin{subfigure}{0.19\textwidth}
		\includegraphics[width=\textwidth]{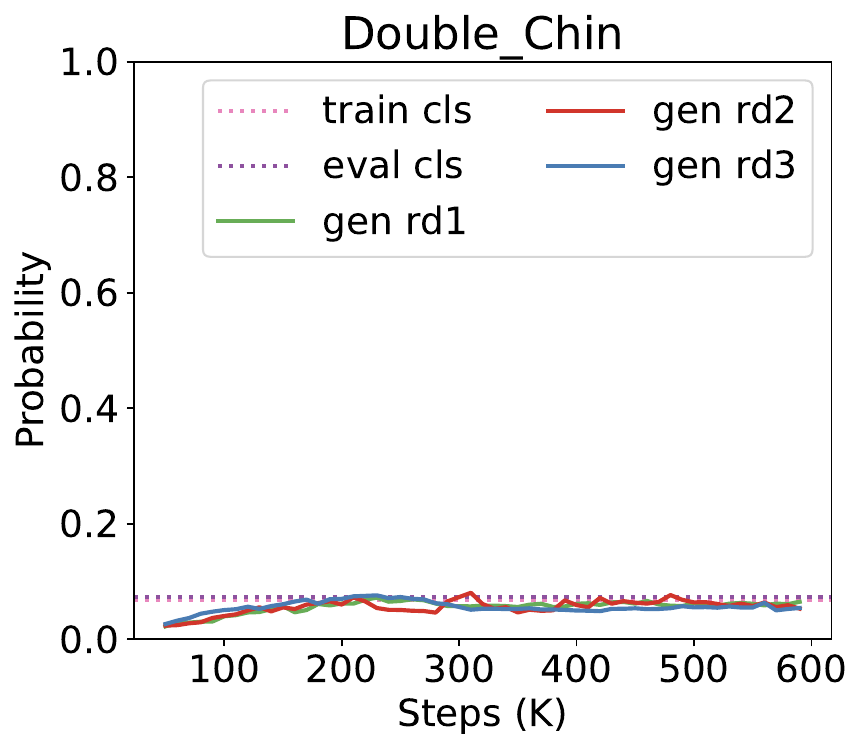}
		\caption{Double Chin}
	    \end{subfigure}
    \begin{subfigure}{0.19\textwidth}
		\includegraphics[width=\textwidth]{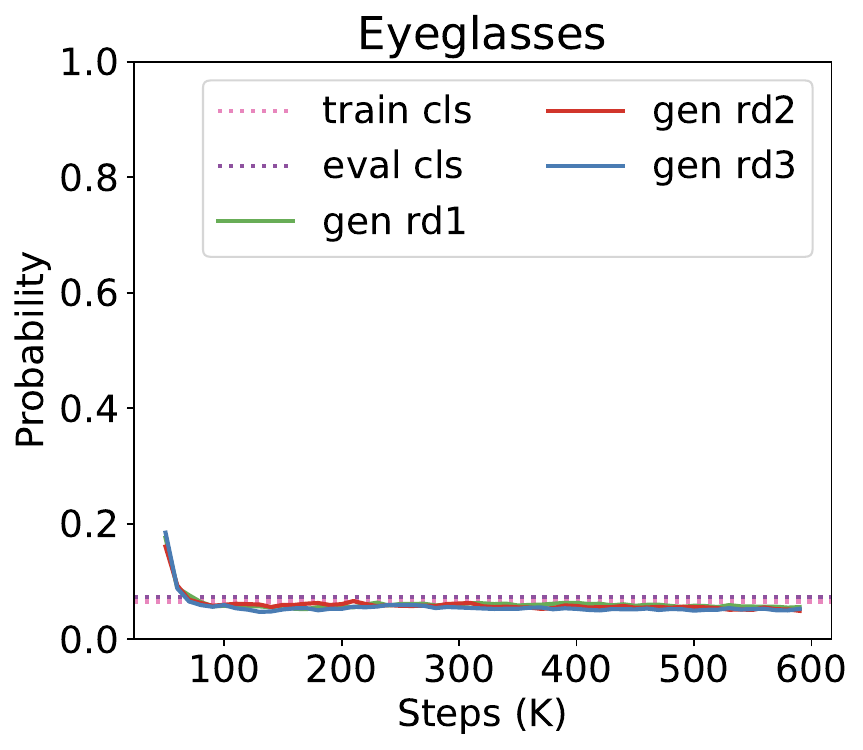}
		\caption{Eyeglasses}
    \end{subfigure}
    \vfill
    \begin{subfigure}{0.19\textwidth}
		\includegraphics[width=\textwidth]{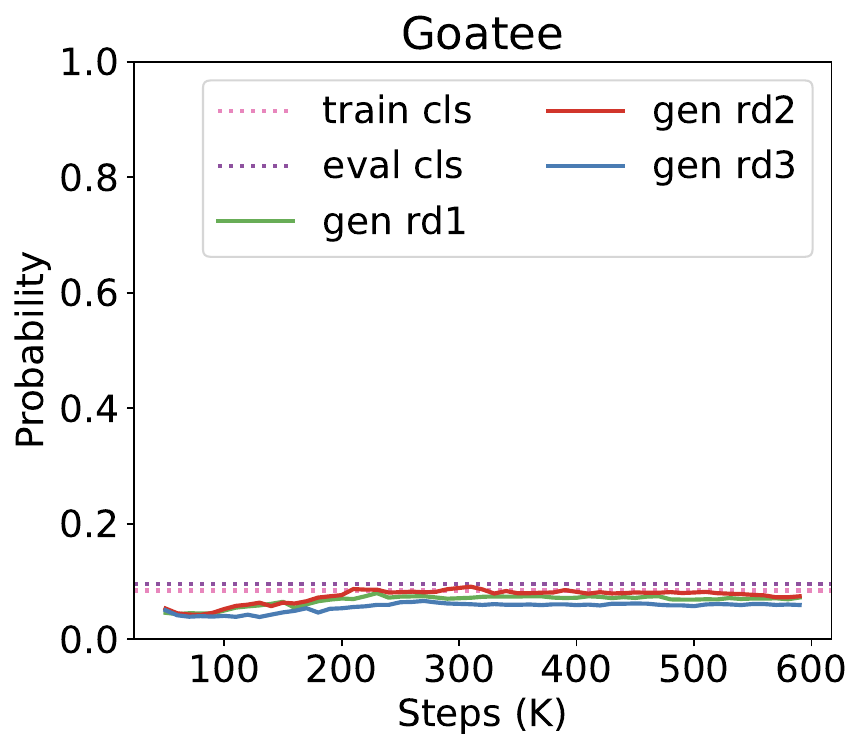}
		\caption{Goatee}
    \end{subfigure}
    \begin{subfigure}{0.19\textwidth}
		\includegraphics[width=\textwidth]{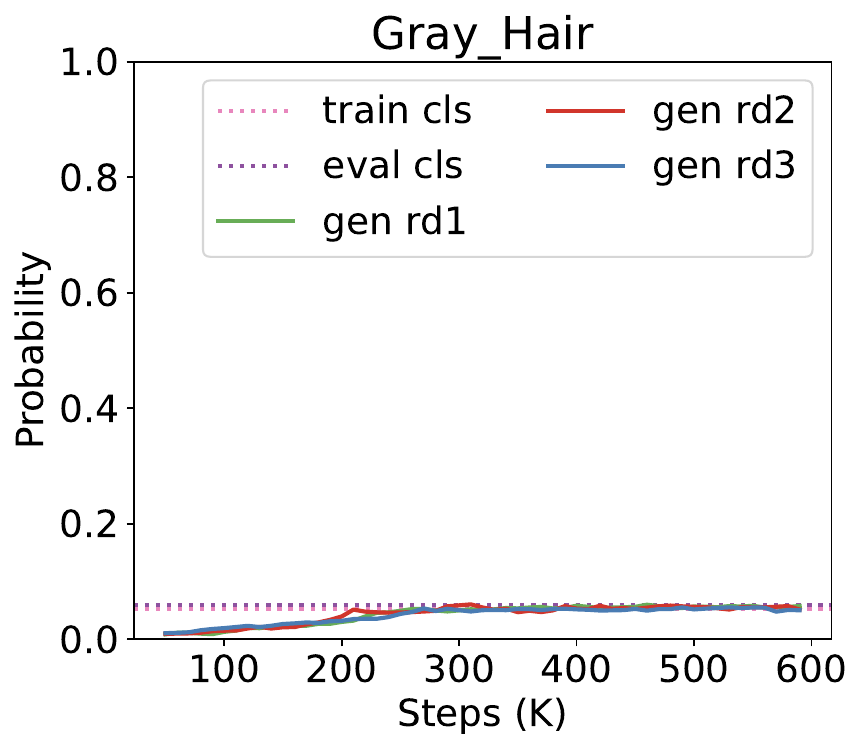}
		\caption{Gray Hair}
    \end{subfigure}
     \begin{subfigure}{0.19\textwidth}
		\includegraphics[width=\textwidth]{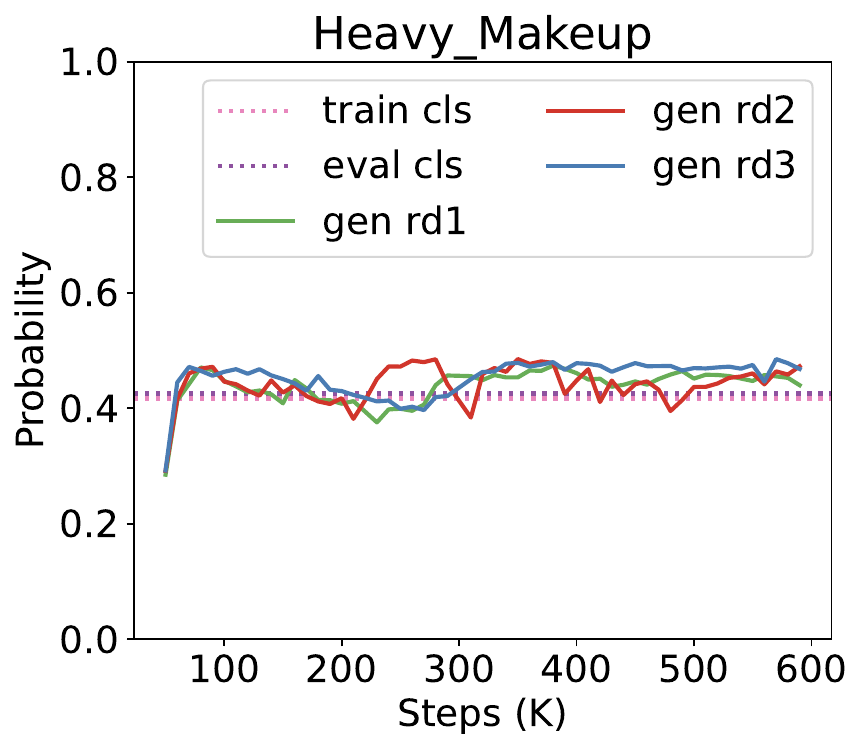}
		\caption{Heavy Makeup}
	    \end{subfigure}
    \begin{subfigure}{0.19\textwidth}
		\includegraphics[width=\textwidth]{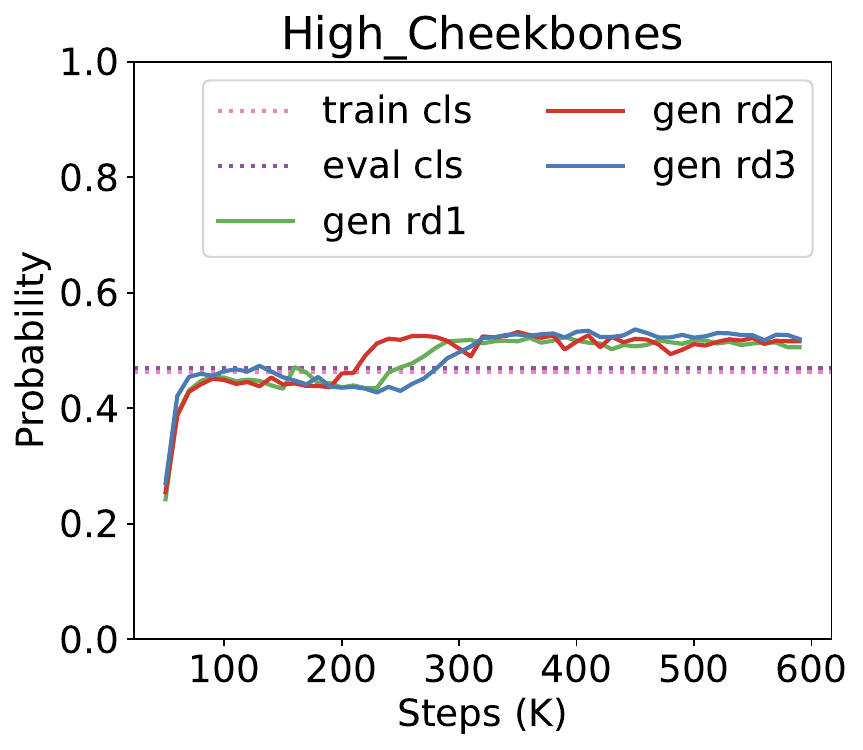}
		\caption{High Cheekbones}
	    \end{subfigure}
    \begin{subfigure}{0.19\textwidth}
		\includegraphics[width=\textwidth]{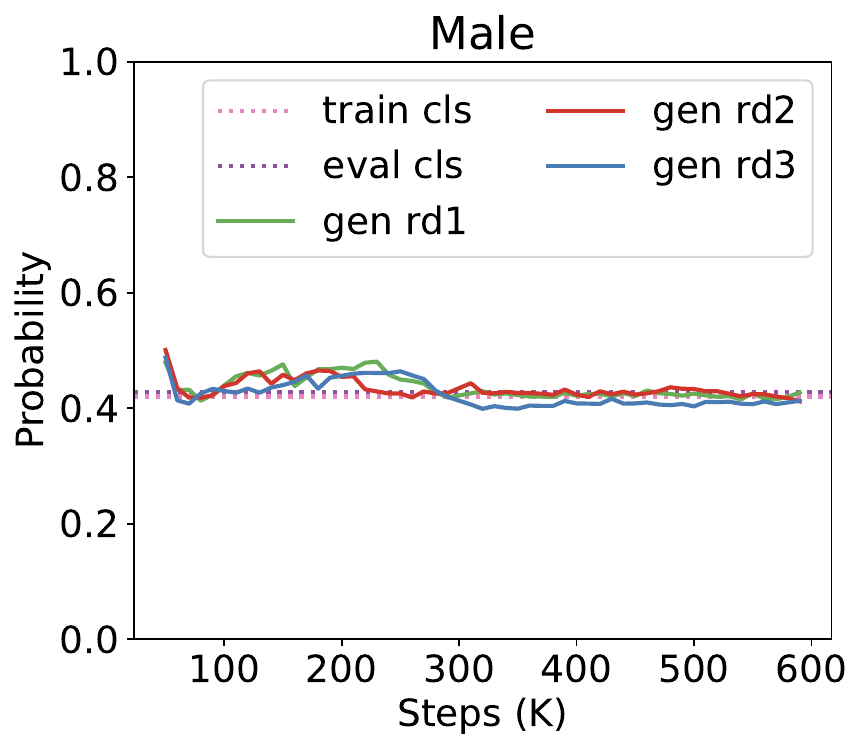}
		\caption{Male}
    \end{subfigure}
    \vfill
    \begin{subfigure}{0.19\textwidth}
		\includegraphics[width=\textwidth]{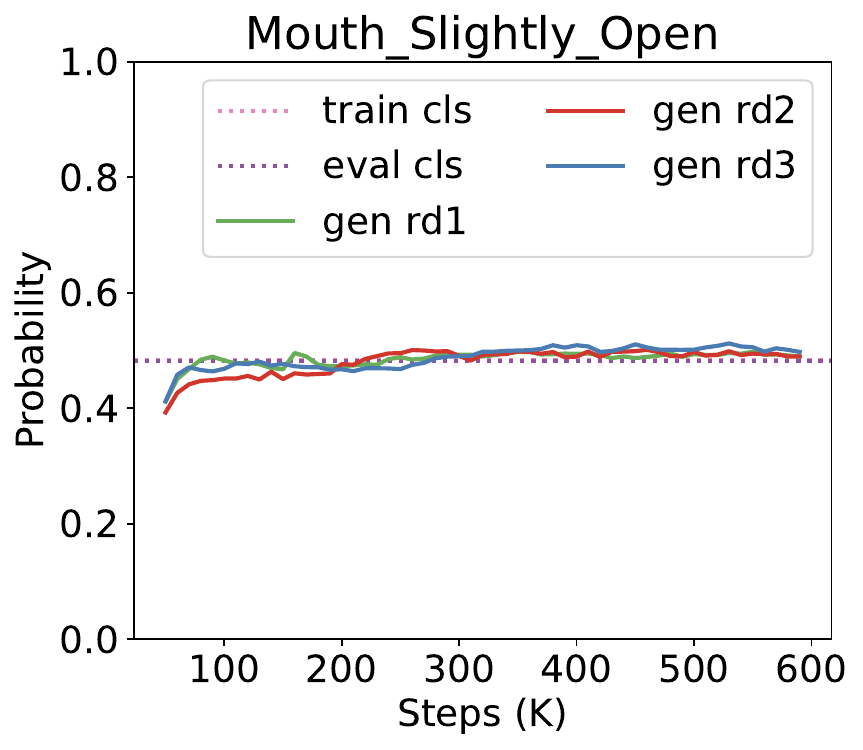}
		\caption{\begin{tiny}
		    Mouth Slightly Open
		\end{tiny}}
    \end{subfigure}
    \begin{subfigure}{0.19\textwidth}
		\includegraphics[width=\textwidth]{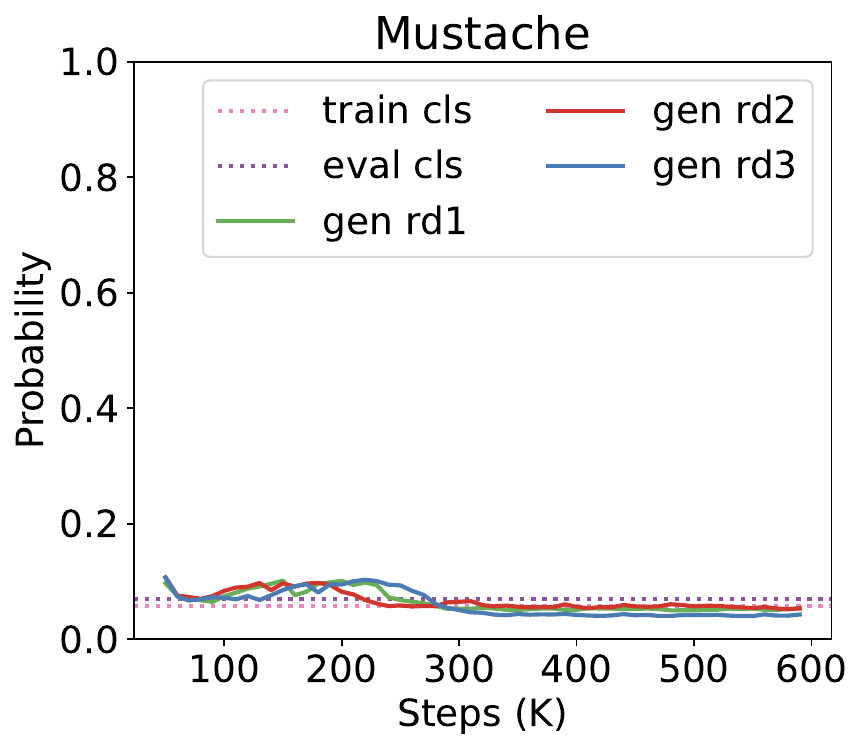}
		\caption{Mustache}
    \end{subfigure}
     \begin{subfigure}{0.19\textwidth}
		\includegraphics[width=\textwidth]{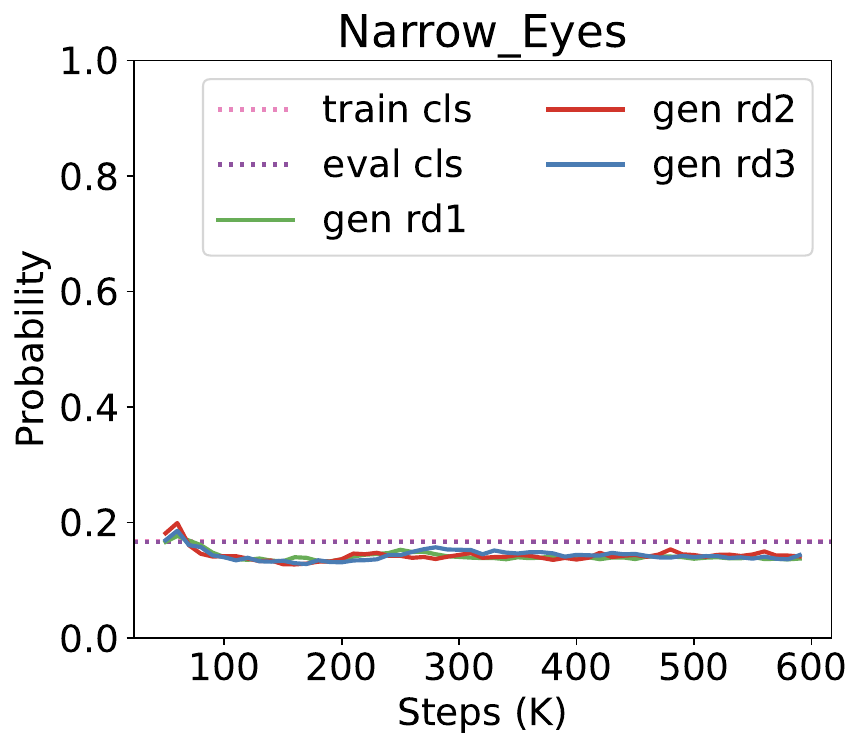}
		\caption{Narrow Eyes}
	    \end{subfigure}
    \begin{subfigure}{0.19\textwidth}
		\includegraphics[width=\textwidth]{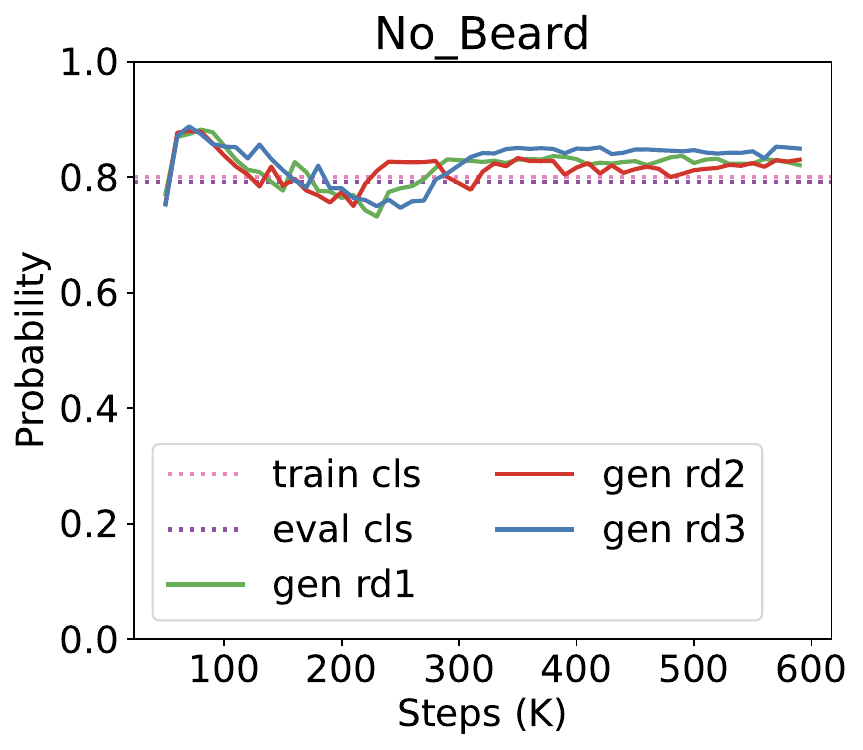}
		\caption{No Beard}
	    \end{subfigure}
    \begin{subfigure}{0.19\textwidth}
		\includegraphics[width=\textwidth]{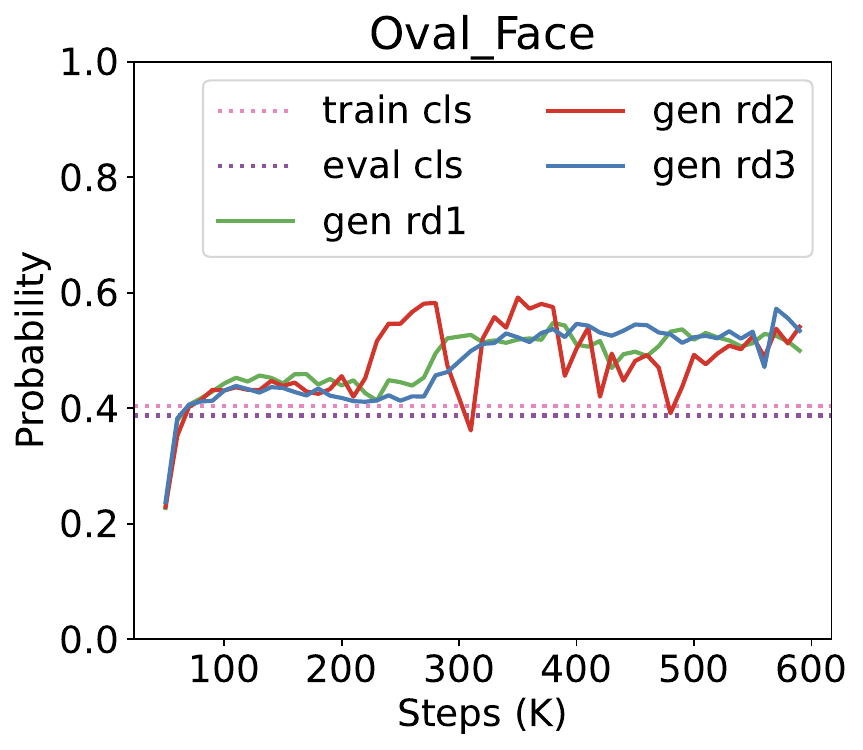}
		\caption{Oval Face}
    \end{subfigure}
    \vfill
    \begin{subfigure}{0.19\textwidth}
		\includegraphics[width=\textwidth]{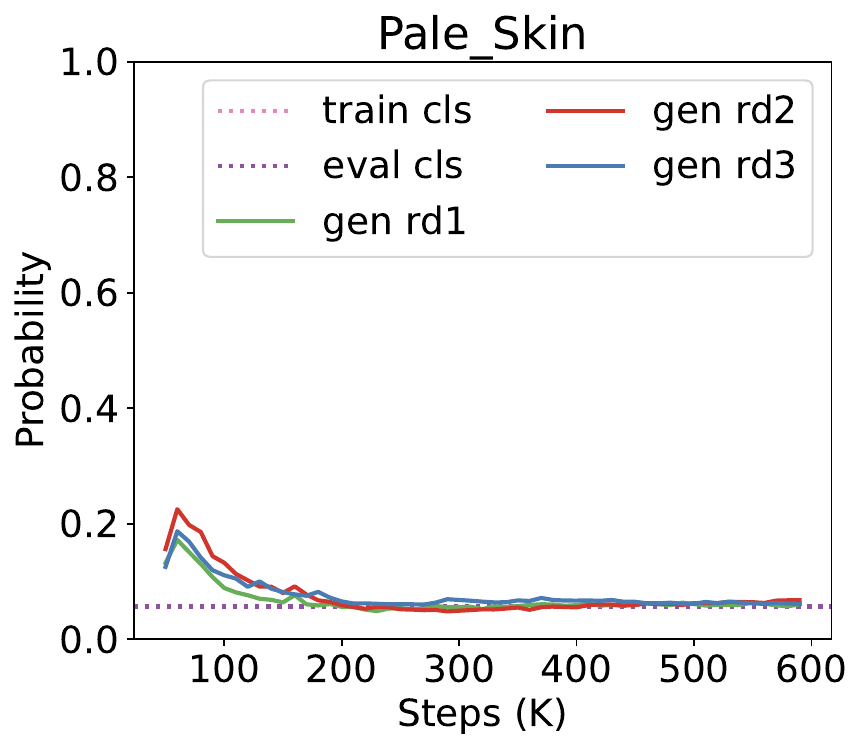}
		\caption{Pale Skin}
    \end{subfigure}
    \begin{subfigure}{0.19\textwidth}
		\includegraphics[width=\textwidth]{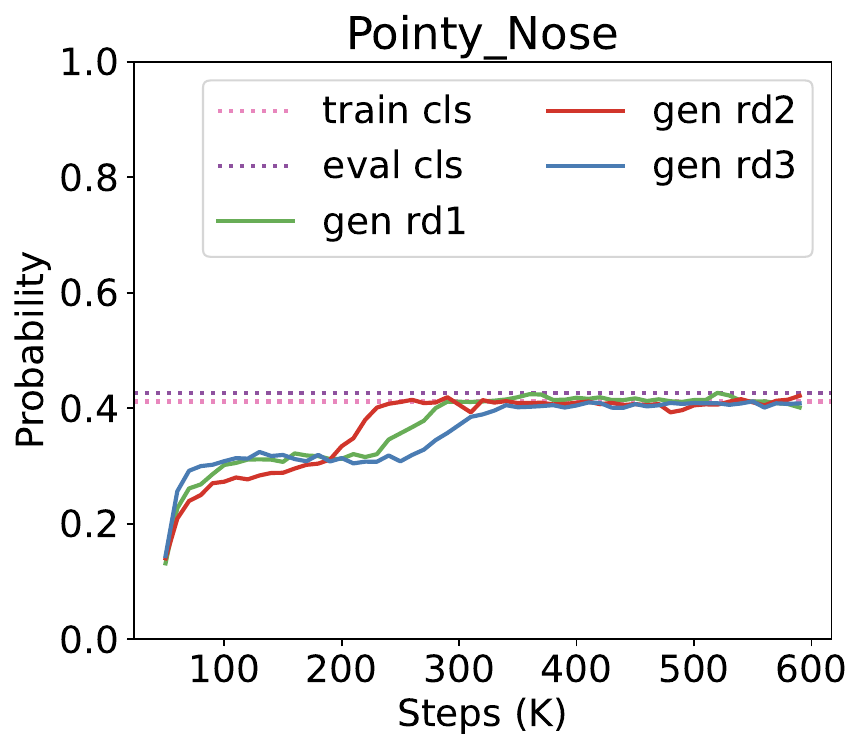}
		\caption{Pointy Nose}
    \end{subfigure}
     \begin{subfigure}{0.19\textwidth}
		\includegraphics[width=\textwidth]{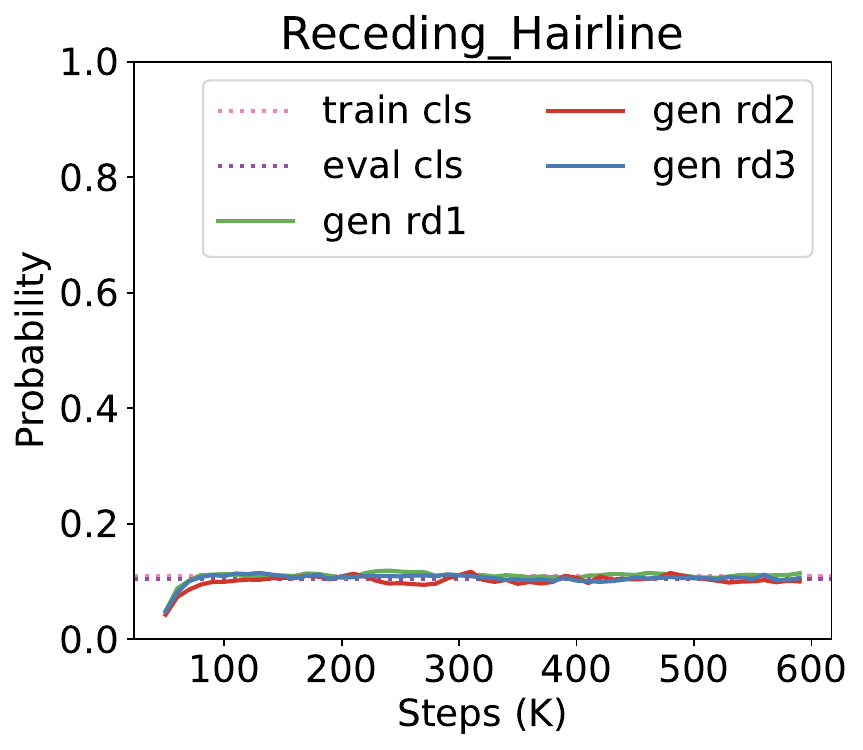}
		\caption{\begin{tiny}
		    Receding Hairline
		\end{tiny}}
	    \end{subfigure}
    \begin{subfigure}{0.19\textwidth}
		\includegraphics[width=\textwidth]{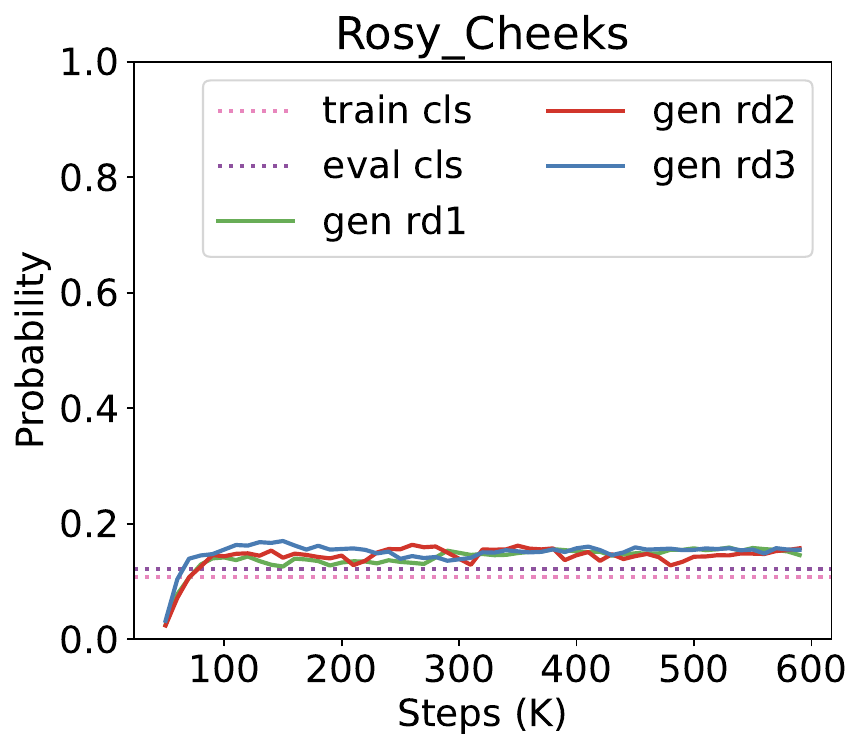}
		\caption{Rosy Cheeks}
	    \end{subfigure}
    \begin{subfigure}{0.19\textwidth}
		\includegraphics[width=\textwidth]{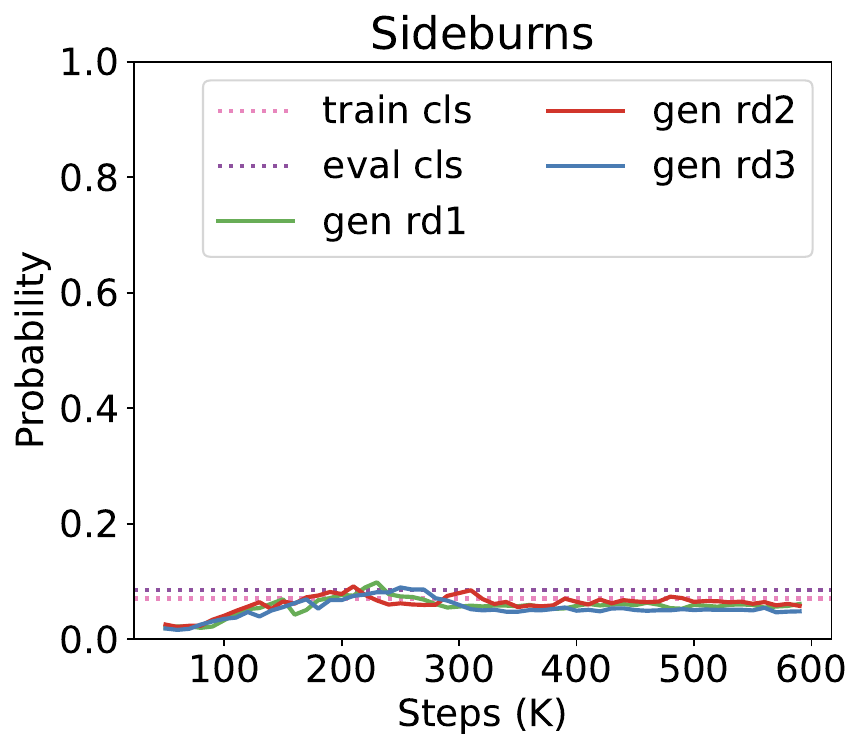}
		\caption{Sideburns}
    \end{subfigure}
    \vfill
    \begin{subfigure}{0.19\textwidth}
		\includegraphics[width=\textwidth]{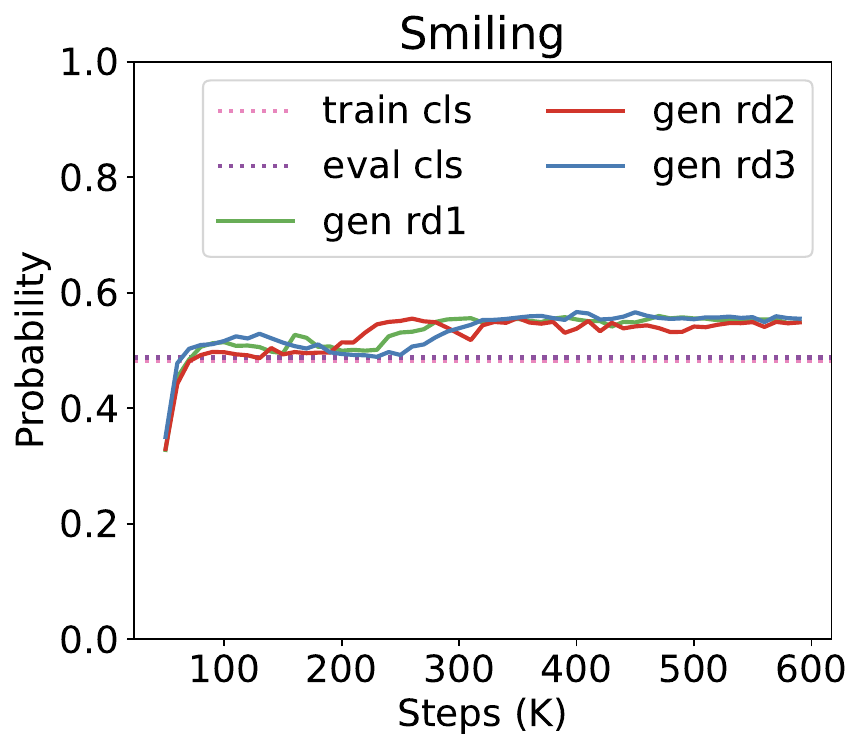}
		\caption{Smiling}
    \end{subfigure}
    \begin{subfigure}{0.19\textwidth}
		\includegraphics[width=\textwidth]{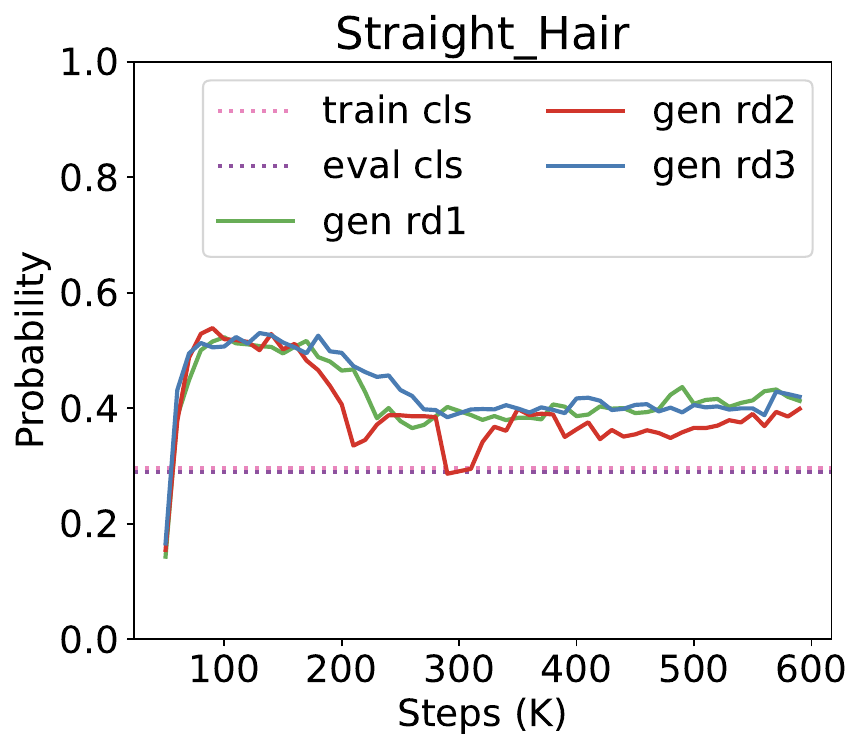}
		\caption{Straight hair}
    \end{subfigure}
     \begin{subfigure}{0.19\textwidth}
		\includegraphics[width=\textwidth]{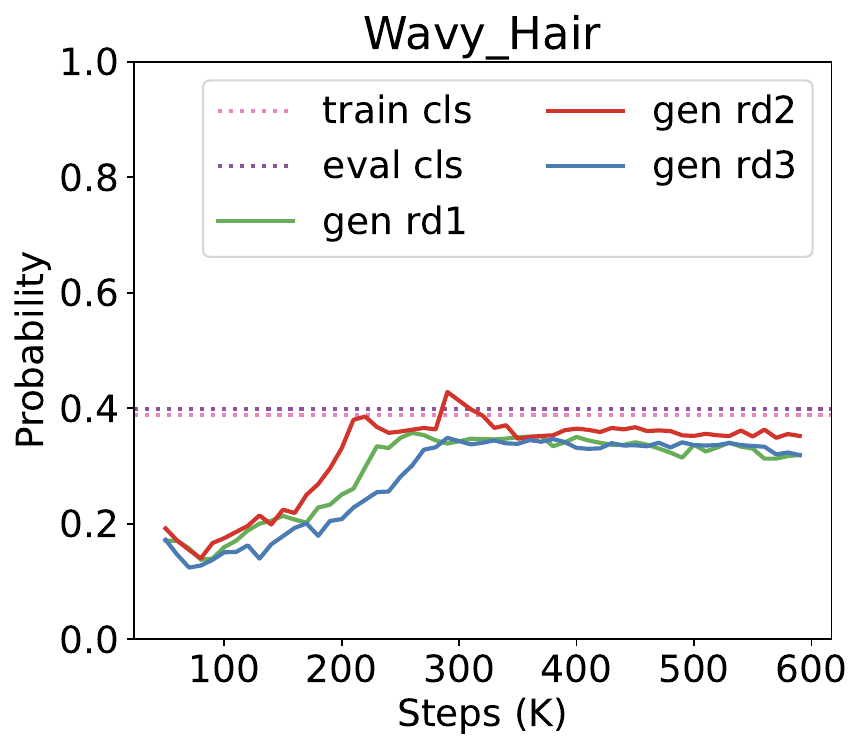}
		\caption{Wavy Hair}
	    \end{subfigure}
    \begin{subfigure}{0.19\textwidth}
		\includegraphics[width=\textwidth]{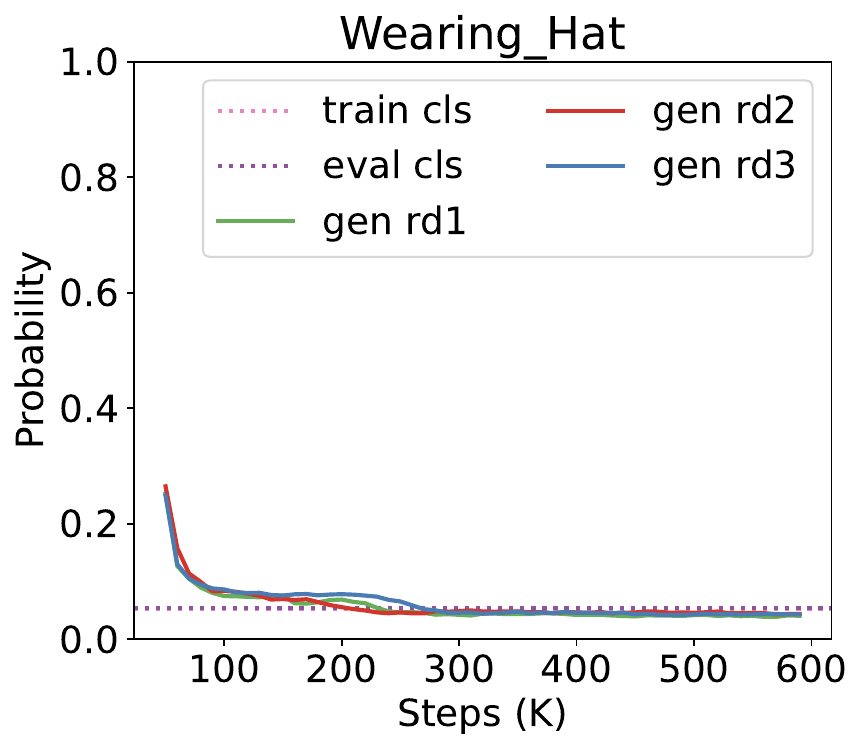}
		\caption{Wearing Hat}
    \end{subfigure}
    \begin{subfigure}{0.19\textwidth}
		\includegraphics[width=\textwidth]{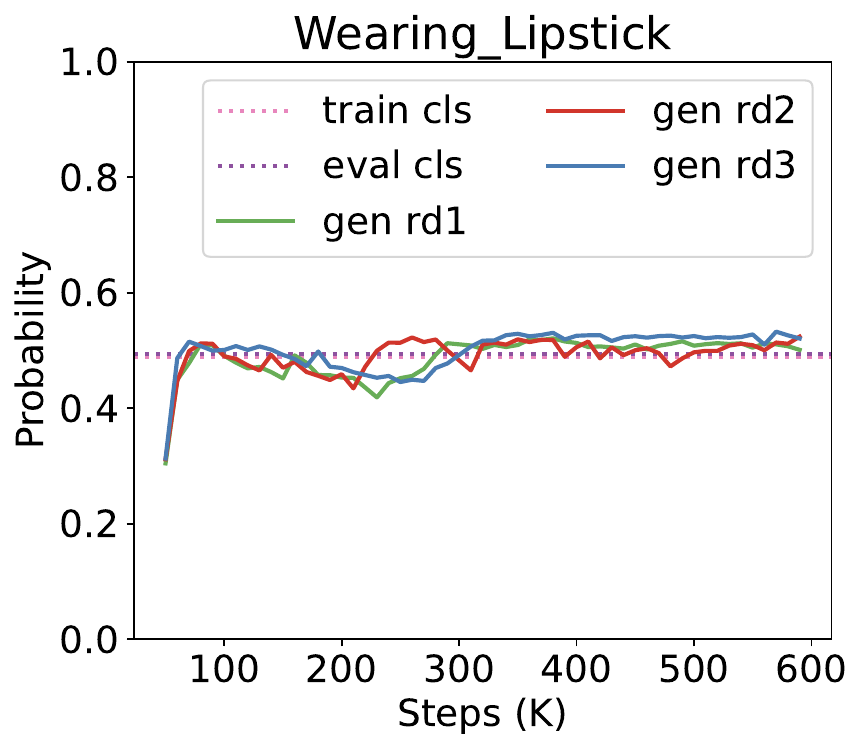}
		\caption{\begin{tiny}
		    Wearing Lipstick
		\end{tiny}}
    \end{subfigure}
    \vfill
    \begin{subfigure}{0.19\textwidth}
		\includegraphics[width=\textwidth]{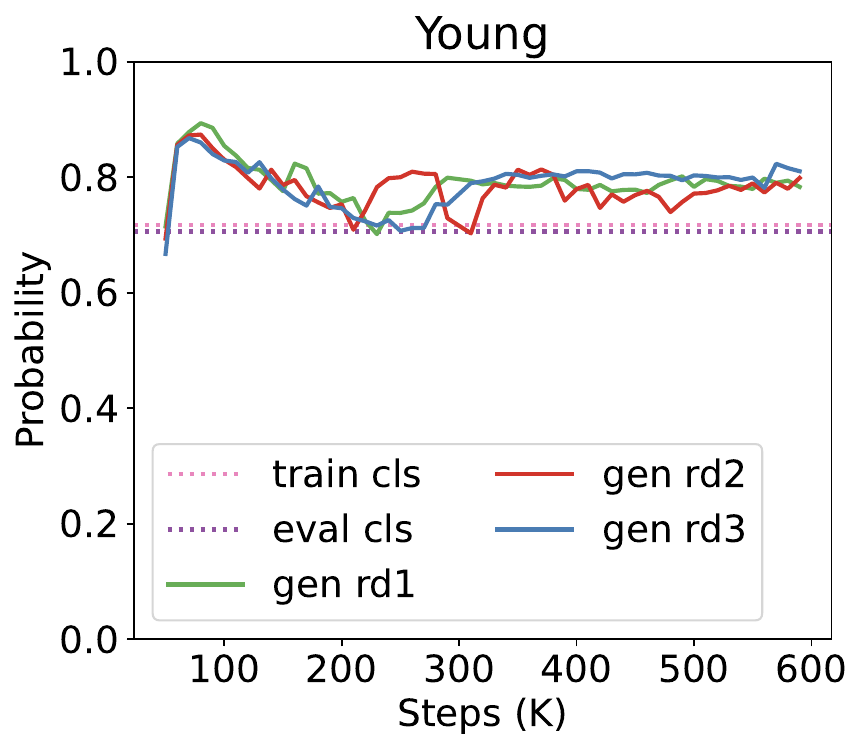}
		\caption{Young}
    \end{subfigure}
	\caption{The probabilities of attributes in CelebA during training (SwinTransformer-based classifier).}
	\label{fig:attr_bias_celeba_all_swint}
\end{figure}

Fig.~\ref{fig:attr_bias_dp_all} show the bias probability for each attribute in DeepFashion dataset. Figs.~\ref{fig:attr_bias_celeba_all} and \ref{fig:attr_bias_celeba_all_swint} show the bias probability for each attribute in CelebA dataset using ResNeXt-based and SwinTransformer-based classifiers respectively. 

\clearpage

\section{Samples of generated images}
\label{app:gen_samples}
\begin{figure}[!htbp]
    \centering
    \includegraphics[width=0.95\linewidth]{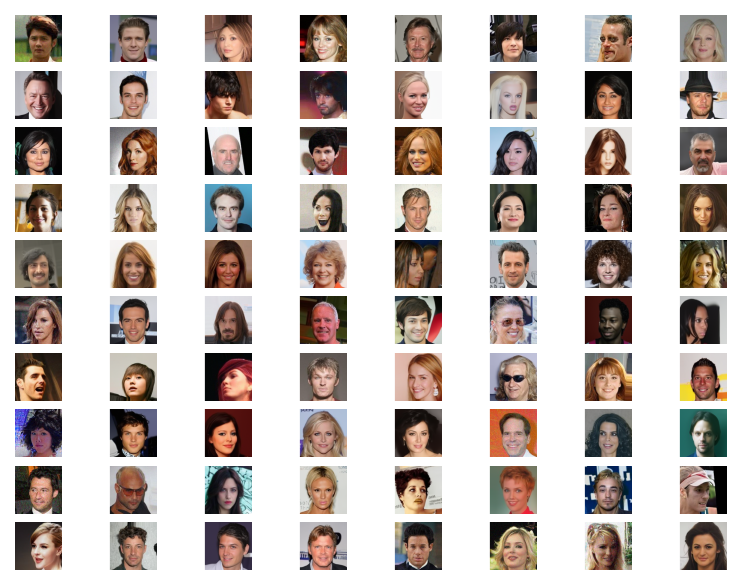}
    \caption{Image samples from large diffusion model generations on CelebA dataset.}
    \label{fig:sample_celeba_large_diffusion}
\end{figure}

\begin{figure}[!tbp]
    \centering
    \includegraphics[width=0.95\linewidth]{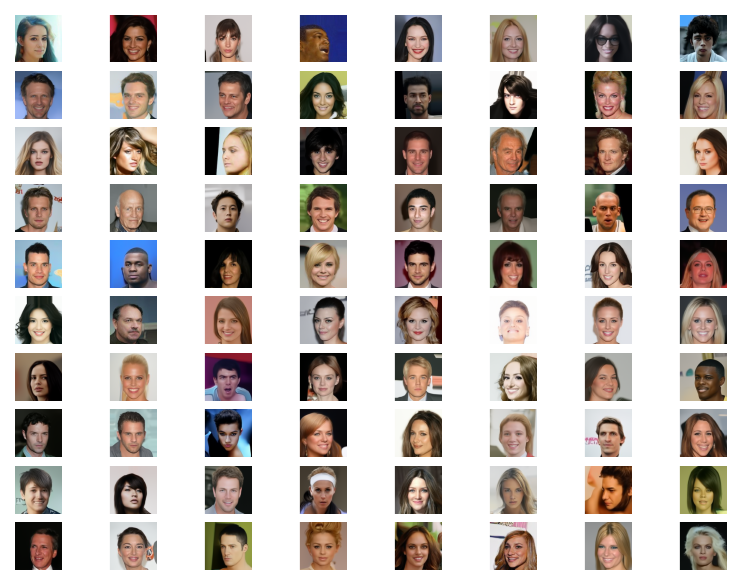}
    \caption{Image samples from the small diffusion model trained on CelebA dataset.}
    \label{fig:sample_celeba_small_diffusion}
\end{figure}

\begin{figure}[!tbp]
    \centering
    \includegraphics[width=0.95\linewidth]{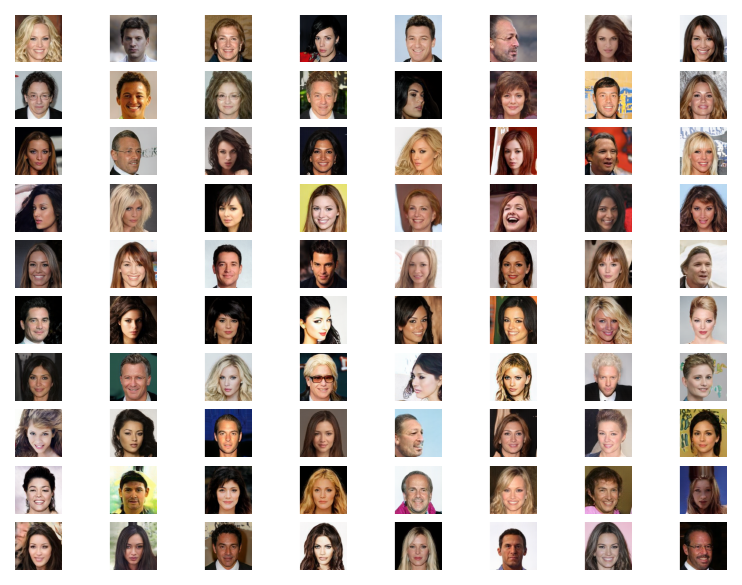}
    \caption{Image samples from the BigGAN model trained on CelebA dataset.}
    \label{fig:sample_celeba_gan}
\end{figure}

\begin{figure}[!tbp]
    \centering
    \includegraphics[width=0.95\linewidth]{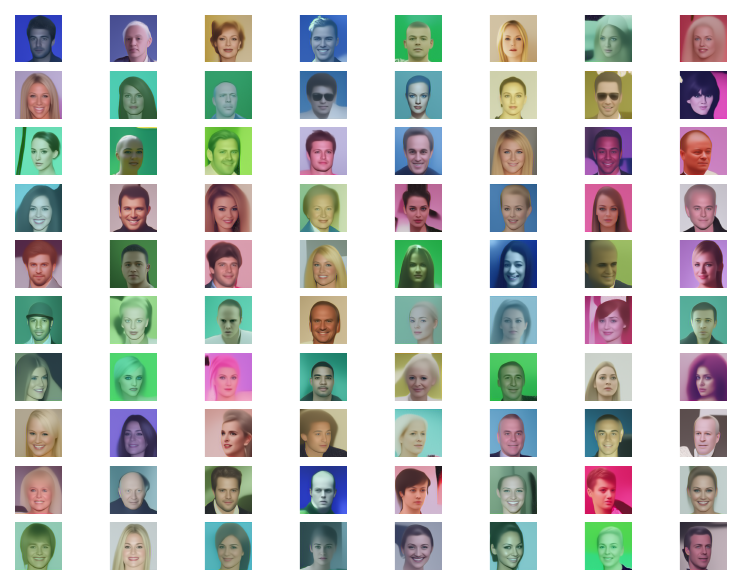}
    \caption{Image samples from the tiny diffusion model trained on CelebA dataset.}
    \label{fig:sample_celeba_tiny_diffusion}
\end{figure}

\begin{figure}[!tbp]
    \centering
    \includegraphics[width=0.95\linewidth]{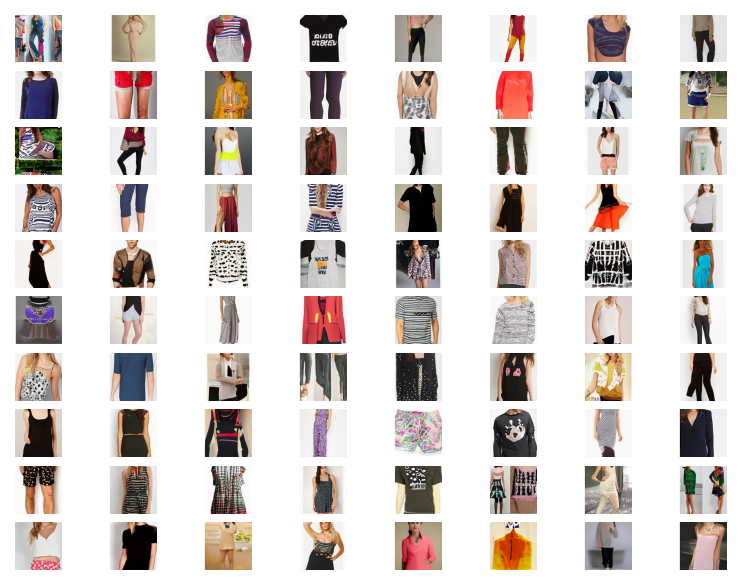}
    \caption{Image samples from the large diffusion model trained on DeepFashion dataset.}
    \label{fig:sample_deepfashion_large_diffusion}
\end{figure}

For different models and different dataset, we sample 80 images from the generation set and present them in Figs.~\ref{fig:sample_celeba_large_diffusion}, \ref{fig:sample_celeba_small_diffusion}, \ref{fig:sample_celeba_gan}, \ref{fig:sample_celeba_tiny_diffusion} and \ref{fig:sample_deepfashion_large_diffusion}.

\end{document}